\documentclass[10pt]{article}
\usepackage[utf8]{inputenc}
\usepackage{geometry}
\usepackage{bbm}
\usepackage{amsmath,amssymb,amsthm}
\newtheorem{theorem}{Theorem}
\newtheorem{proposition}{Proposition}

\newtheorem{lemma}{Lemma}
\newtheorem{assumption}{Assumption}

\theoremstyle{definition}
\newtheorem{definition}{Definition}
\newtheorem{remark}{Remark}
\newtheorem{example}{Example}
\newtheorem{problem}{Problem}

\numberwithin{equation}{section}
\numberwithin{figure}{section}
\numberwithin{table}{section}
\numberwithin{theorem}{section}
\numberwithin{proposition}{section}
\numberwithin{corollary}{section}
\numberwithin{lemma}{section}
\numberwithin{definition}{section}
\numberwithin{remark}{section}
\numberwithin{example}{section}

\usepackage{graphicx}
\usepackage{subfig}
\usepackage{float}
\usepackage[toc,page]{appendix}
\usepackage{xcolor}
\usepackage{makecell}
\usepackage{booktabs}
\usepackage{longtable}
\usepackage{algorithm}
\usepackage{algpseudocode}
\newcommand{\R}{\mathbb{R}}
\newcommand{\E}{\mathbb{E}}
\newcommand{\F}{\mathcal{F}}
\newcommand{\X}{\mathcal{X}}
\newcommand{\Z}{\mathcal{Z}}
\newcommand{\PS}{\mathcal{P}}
\newcommand{\prob}{\mathbb{P}}
\newcommand{\N}{\mathbb{N}}
\newcommand{\HS}{\mathcal{H}}

\newcommand{\Q}{\mathcal{Q}}
\newcommand{\C}{\mathcal{C}}

\newcommand{\x}{\mathbf{x}}

\newcommand{\w}{\mathbf{w}}

\newcommand{\vb}{\mathbf{v}}
\newcommand{\ep}{\epsilon}

\DeclareMathOperator*{\argmax}{arg\,max}
\DeclareMathOperator*{\argmin}{arg\,min}

\DeclareMathOperator{\TV}{TV}
\DeclareMathOperator{\law}{law}

\DeclareMathOperator{\KL}{KL}
\DeclareMathOperator{\dom}{Dom}
\DeclareMathOperator{\sprt}{sprt}
\DeclareMathOperator{\proj}{Proj}
\DeclareMathOperator{\iid}{i.i.d.}
\DeclareMathOperator{\iidsample}{\stackrel{\iid}\sim}
\DeclareMathOperator{\AC}{\mathbf{AC}}
\DeclareMathOperator{\NC}{\mathbf{NC}}
\DeclareMathOperator{\TC}{\mathbf{TC}}
\DeclareMathOperator{\TF}{\textsf{TF}}
\DeclareMathOperator{\poly}{\textsf{poly}}
\newcommand{\fip}{\proj^{-1}_{<\omega}}
\newcommand{\nsm}{\Phi}
\DeclareMathOperator{\id}{id}
\newcommand{\eye}{\text{\faEye}}
\newcommand{\see}{\text{\faHandLizardO}}
\newcommand{\fasmile}{\text{\faSmileO}}
\newcommand{\detach}{\text{\faHandStopO}}
\newcommand{\bos}{\langle\textsf{bos}\rangle}
\newcommand{\eos}{\langle\textsf{eos}\rangle}
\newcommand{\sot}{\langle \textsf{s} \rangle}
\newcommand{\eot}{\langle \textsf{/s} \rangle}
\newcommand{\sor}{\langle \textsf{r} \rangle}
\newcommand{\eor}{\langle \textsf{/r} \rangle}

\newcommand{\blank}{\langle \textsf{blank} \rangle}

\newcommand{\vs}{\vspace{0.5em}}
\usepackage{hyperref}
\allowdisplaybreaks
\usepackage{authblk}
\usepackage{marvosym}
\usepackage{fontawesome}
\usepackage[most]{tcolorbox}

\newcommand\silentfootnote[1]{%
    \bgroup
    \renewcommand\thefootnote{\fnsymbol{footnote}}%
    \renewcommand\thempfootnote{\fnsymbol{mpfootnote}}%
    \footnotetext[0]{#1}%
    \egroup
}

\title{A First-Principles Theory of Slow Thinking\\and Active Perception}
\date{}

\author[1,2]{Hongkang Yang}
\author[3]{Zhi-Qin John Xu}
\author[1,2]{Feiyu Xiong}
\author[4,2,\Letter]{Weinan E}

\affil[1]{MemTensor (Shanghai) Technology Co., Ltd.}
\affil[2]{Institute for Advanced Algorithms Research, Shanghai}
\affil[3]{Shanghai JiaoTong University}
\affil[4]{Peking University}

\usepackage{titling}
\begin{document}

\maketitle
\silentfootnote{\Letter ~Corresponding author: weinan@math.pku.edu.cn}

\vspace{-3.5em}

\begin{abstract}
As part of a series on first-principles modeling of cognitive functions, this paper attempts to provide a mathematical formulation of thinking and perception.
It formally derives slow thinking or more generally, active perception, and encompasses the design, training and inference of slow thinking large language models.
Our starting point is the lifting and projection of probability distributions on the observable and latent spaces, with the objective of representing complex data distributions by simple function families such as neural networks.
A theory called ``active lifting" is proposed, based on the sampling of latent sequences and an intrinsic drive to reduce uncertainty with maximum rate.
It derives a large design space, containing the slow thinking models in a subspace that we call the static theory.
These models are positioned on the representation hierarchy and sampler hierarchy induced by the static theory, and can be upgraded by climbing the two hierarchies.
Active lifting further derives an inference process with an internal time axis, and a training objective that resembles minimum-length coding as well as the invention of languages.
Thus, it characterizes the agency of perception, including the emergence of the slow thinking formats.
Technical by-products of this theory include a three-stage pathway for improving slow thinking models, a unified approach to constructing encoders and generative models for all data modalities, \textit{a priori} formation of human-like visual representations, and a possible solution to policy collapse.
\end{abstract}

\begin{figure}[ht]
\centering
\includegraphics[width=0.9\linewidth]{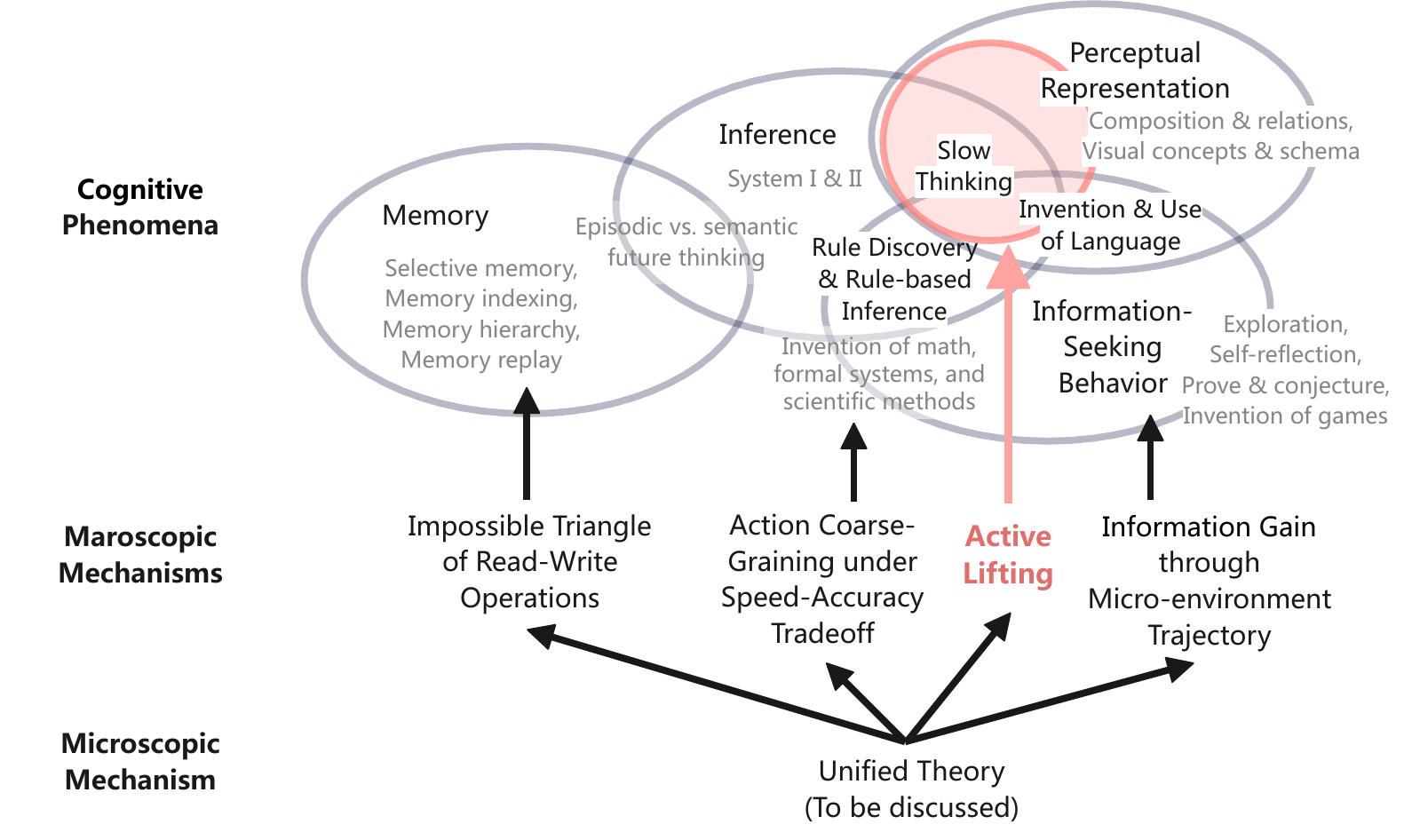}
\caption{The modeling hierarchy of cognitive functions.
The coverage of this paper is marked in red.
The other topics are left for our paper series and this figure is subject to constant update.
Examples of cognitive phenomena (in gray) are included to illustrate their diversity.}
\label{fig: cognitive modeling}
\end{figure}

\newpage

\begin{figure}[ht]
\centering
\includegraphics[width=0.7\linewidth]{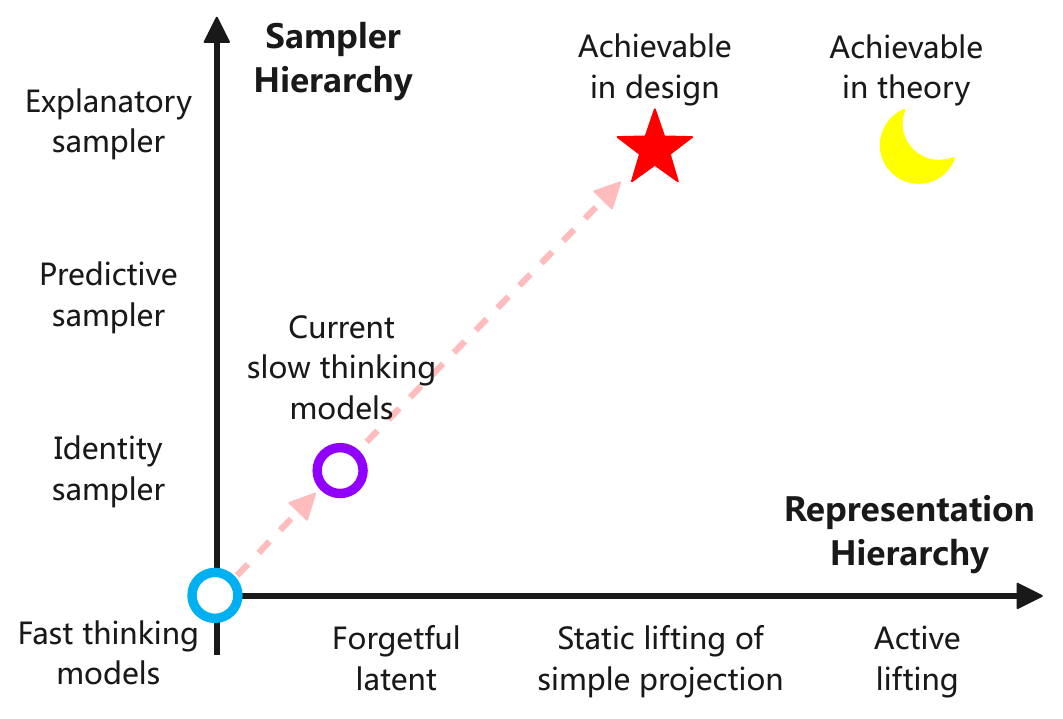}
\caption{Road-map for improving slow thinking models, as a by-product of our theory.
This figure summarizes the improvements proposed in Sections \ref{sec. quick improvement}-\ref{sec. long-term improvement} and \ref{sec. perceptual}.
The two axes are the representation hierarchy (illustrated in Figures \ref{fig: representation hierarchy} and \ref{fig: forgetful in hierarchy}) and the sampler hierarchy (Figure \ref{fig: sampler hierarchy}).}
\label{fig: climb hierarchy}
\end{figure}

\tableofcontents

\section{Introduction}

This paper is the first in a series on the first-principles modeling of cognitive functions including thinking, perception, memory, the abstraction of rules and concepts, and spontaneous self-directed learning, etc.
By first-principle, we refer to deriving phenomena of interest from a fundamental mathematical formulation, without relying on the empirical knowledge of these phenomena.
The ultimate goal of this series is to formulate a first-principles approach to artificial intelligence and provide guidance to the design of fully self-directed and self-evolving AI models.
By self-directed, we mean that the models acquire data by themselves, not requiring human-prepared data.
By self-evolving, we mean that their cognitive functions emerge spontaneously, not requiring human-designed architectures or workflows.


As a scientific discipline, the field of artificial intelligence is quite unique in how it has evolved.
On one hand, it has managed to achieve astonishing success by trial and error, and  pushing  engineering limits.
On the other hand, it has experienced serious ups and downs in a way not seen in other disciplines \cite{mitchell2021ai,crevier1993ai,hendler2008avoiding,agar2020science}.
It is fair to say that in the absence of first principles, we are prone to making unrealistic predictions, wasting resources and even leading the field to wrong directions.

Since artificial intelligence has always drawn inspiration from human intelligence, to seek the first principles of AI, it is natural that we first turn to neuroscience.
Unfortunately, searching for first principles of neuroscience is likely a much harder task.
To resolve this dilemma, let us go back to the story of aviation.
Humans got the idea of flight from birds. But to learn how to fly and fly well, there is no need to understand the detailed anatomy of birds, we only need to understand the important physical issues, namely lift and drag, and this is the subject of gas dynamics.
Indeed, building upon gas dynamics, scientists like von Karman have succeeded in formulating the first principles of aviation, which has made aerospace engineering a principled subject.
One lesson we learn from this story is that to formulate the first principles of artificial intelligence, we should first build a mathematical theory of cognitive functions.

For centuries, cognitive functions have been studied at a phenomenological level in cognitive science.
A well-known example is the phenomenon of fast and slow thinking,
also known as  the ``system 1" and ``system 2" mental processes,
the former being unconscious, instinctive thinking and the latter being conscious, deliberative thinking \cite{kahneman2011fastslow,kahneman2003maps,stanovich2000individual}.
Our aim is to first build mathematical theories for the major cognitive functions,
as intermediate steps,
and then formulate a microscopic theory that unifies them all.
A tentative plan of this paper series is given by Figure \ref{fig: cognitive modeling}.
It depicts a three-level modeling hierarchy, such that the cognitive functions can be derived in a first-principles fashion from the macroscopic theories, which all arise from a unified microscopic theory.
The hierarchy simplifies the intractably diverse phenomena to elementary mechanisms that could be simpler to implement.
We are satisfied as long as our theories are sufficient to induce cognitive functions, regardless of whether they adhere to the actual mechanisms of human cognition.

The subject of this paper is slow thinking, or more generally, active perception.
In machine learning, slow thinking refers to the generation of hidden texts (often called ``chain-of-thoughts") during the encoding and decoding of large language models (LLMs), which enhances their performance \cite{openai2024o1card,guo2025deepseek,zelikman2022star,wei2022chain,ling2017program}.
Meanwhile, by active perception, we mean converting observations to ``understandable descriptions".
Its agency comes from the freedom of inventing such descriptions and the freedom of searching for the right description for each observation.
Thus, slow thinking can be seen as a special case of active perception, e.g.\@ a difficult reading material becomes more understandable when annotated with the deep thoughts of the reader.
A more complete characterization of active perception would involve the movements of the observer in the environment, but this paper focuses on the internal states and leave the modeling of movements to future work.

The general problem can be formulated as follows.
\begin{problem}[informal]
\label{problem. intro}
Let $P_*$ be a probability distribution over an arbitrary data space $\X$.
Let $\PS_M$ be the space of probability distributions that can be modeled under some computational constraints (e.g.\@ can be parametrized by neural networks, or more generally, bounded-depth polynomial-size Boolean circuits).
Construct a ``lifted" version of $P_*$ that belongs to $\PS_M$.
\end{problem}
By lifting, we mean converting $P_*$ to some distribution $P_{\text{lift}}$ over some ``larger" space (the latent space)  $\Z$.
Since $\Z$ is large, the complicated structures of $P_*$ can be unfolded, and thus $P_{\text{lift}}$ becomes easier to model and may belong to $\PS_M$.
A representative example is the hidden Markov models (HMMs);
while the observable process of an HMM can have arbitrarily long dependency, its latent process is simply a Markov chain,
so the lifting from the observable distribution to the latent distribution (or the joint latent-observable distribution) greatly simplifies its modeling.
Since the lifted distributions are generally non-unique, some maximum principle is needed to determine the optimal lifting, which is similar to determining the optimal transport among all probabilistic couplings \cite{villani2003topics,kantorovich1960mathematical}.
The theme of this paper is to formalize and solve this problem.

We propose a theoretical framework called ``active lifting".
In analogy to humans, the theory gives rise to ``languages" that are both efficient and regular (in the sense to be described below), and describes perceptual data with those languages.
It derives the intuitive notions of perceptual efficiency and linguistic regularity, and thus one motivation for inventing languages.
Unsurprisingly, the slow thinking of LLMs  is mathematically a special case of the theory,
as slow thinking can be understood as the LLMs inventing a mental language to unravel the complicated reasoning of the input texts.
The specialness is that the $P_{\text{lift}}$ of slow thinking models have to follow certain ``formats".
This paper focuses more on the sub-theory of slow thinking than active perception, and a thorough treatment of the latter is left for the rest of the series.

The presentation of our theory proceeds from the particular to the general.
First, we develop the static theory with a prescribed and fixed lifting function, which is already a broad setup that includes the existing slow thinking models.
The static theory derives not only the theoretical rationale for performing slow thinking (climbing the representation hierarchy), but also the optimal policies of slow thinking (the sampler hierarchy).
Technically, slow thinking is shown to be the result of maximizing both the approximation ability and sampling efficiency of distribution families that are parametrized by liftings (or equivalently, projections).
In some sense, these two objectives set the upper and lower bounds of a slow thinking model.
As depicted in Figure \ref{fig: climb hierarchy}, one can ascend the representation hierarchy by moving from forgetful latents (one feature of the existing models, to be defined later) to persistent and ubiquitous thinking (i.e.\@ general models under a static lifting),
and ascend the sampler hierarchy by moving from identity samplers to explanatory samplers.

Next, to prepare for active lifting, we present a unified objective, the maximization of the rate of uncertainty reduction.
Using this objective function, one can qualitatively derive the entire static theory.
It also leads to a natural loss that balances fast and slow thinking, when the LLMs are allowed to perform slow thinking anytime and anywhere during pretraining.

Finally, we establish the general framework of active lifting, a model based on unconstrained sampling of latent sequences.
It characterizes the two aspects of the agency of perception:
For inference, perception is endowed with an internal time axis and the model actively searches for possible ``understandings" of each observation;
while for training, the model tries to reduce the remaining uncertainty in its observations as quickly as possible across the observation steps.
The unified objective in this setup resembles the minimum-length coding problem, plus a constraint on the approximation ability.
So if we interpret each sampled latent sequence as a text description of an observation, then a well-trained model forms a language that efficiently describes the observations, while being learnable by neural nets or bounded-depth circuits.
The optimizing solution will be shown to contain implicitly a lifting (or projection) function, so the static theory with a prescribed lifting is a special case, and thus we call this framework active lifting.


We will label the theory developed here a ``macroscopic" theory for thinking and perception. In subsequent work we will develop similar macroscopic theories for other cognitive functions such as memory, spontaneous and self-directed learning, etc, leading up to the  unified theory which we label as ``microscopic". In analogy with physics, the macroscopic theories play the role of Newtonian dynamics, gas dynamics and solid mechanics for particles, gases and solids, and the microscopic theory plays the role of quantum mechanics.

Although these macroscopic theories are only intermediate steps in our road-map, some technical by-products may be of independent, practical interest.
The by-products of this paper include:
\begin{enumerate}
\item Three stages of improvements for slow thinking models (Figure \ref{fig: climb hierarchy}).
The first stage (Section \ref{sec. quick improvement}) improves sampling efficiency with explanatory samplers.
The second stage (Section \ref{sec. short-term improvement}) improves approximation ability with persistent and ubiquitous thinking, along with an intrinsic balance of fast and slow thinking.
The third stage (Sections \ref{sec. long-term improvement} and \ref{sec. adapt to text}) enables the model to develop a free form of slow thinking, unconstrained by any prescribed format.

\item A unified approach to constructing encoders for all data modalities (cf.\@ Section \ref{sec. active lifting} for training and Section \ref{sec. time axis} for inference).
In particular, for image data, the encoder might spontaneously develop a human-like multiscale compositional representation (Section \ref{sec. minimum-length coding}), even though its training and architecture do not involve any such prior knowledge.
The learned representation might be able to unify three common representations in computer vision: patch auto-regression, diffusion, and part-based modeling (Section \ref{sec. visual field})

\item Also, a unified approach to building generative models for all data modalities (Section \ref{sec. generative}).
Theoretically, it helps to solve the non-uniqueness problem of generative modeling,
offering the new option of ``linguistic coupling" besides the mainstream options of free coupling (e.g.\@ generative adversarial network \cite{goodfellow2014generative} and variational autoencoder \cite{kingma2013auto}) and product coupling (e.g.\@ diffusion \cite{song2019generative} and flow matching \cite{albergo2022building}).

\item A possible cause and solution for the policy collapse problem of slow thinking models (Sections \ref{sec. inquisitive sampler} and \ref{sec. sampler training}).
It is a common issue during the reinforcement learning (RL) training of these models, such that the entropy of their sampling distributions quickly diminishes and their performance no longer improves.
In our modeling, to minimize the estimation error of the training gradient, the model needs to approximate two samplers, the posterior sampler and inquisitive sampler.
The omission of the latter may cause the model to favor exploitation over exploration in the latent space.
\end{enumerate}


\subsection{Motivating Examples}

We begin by discussing three intuitive examples for learning, inference, and visual encoding.
Although the derivations of our theory do not rely on empirical knowledge, these examples make the theory easier to understand.

\begin{figure}[!ht]
\centering
\includegraphics[width=1\linewidth]{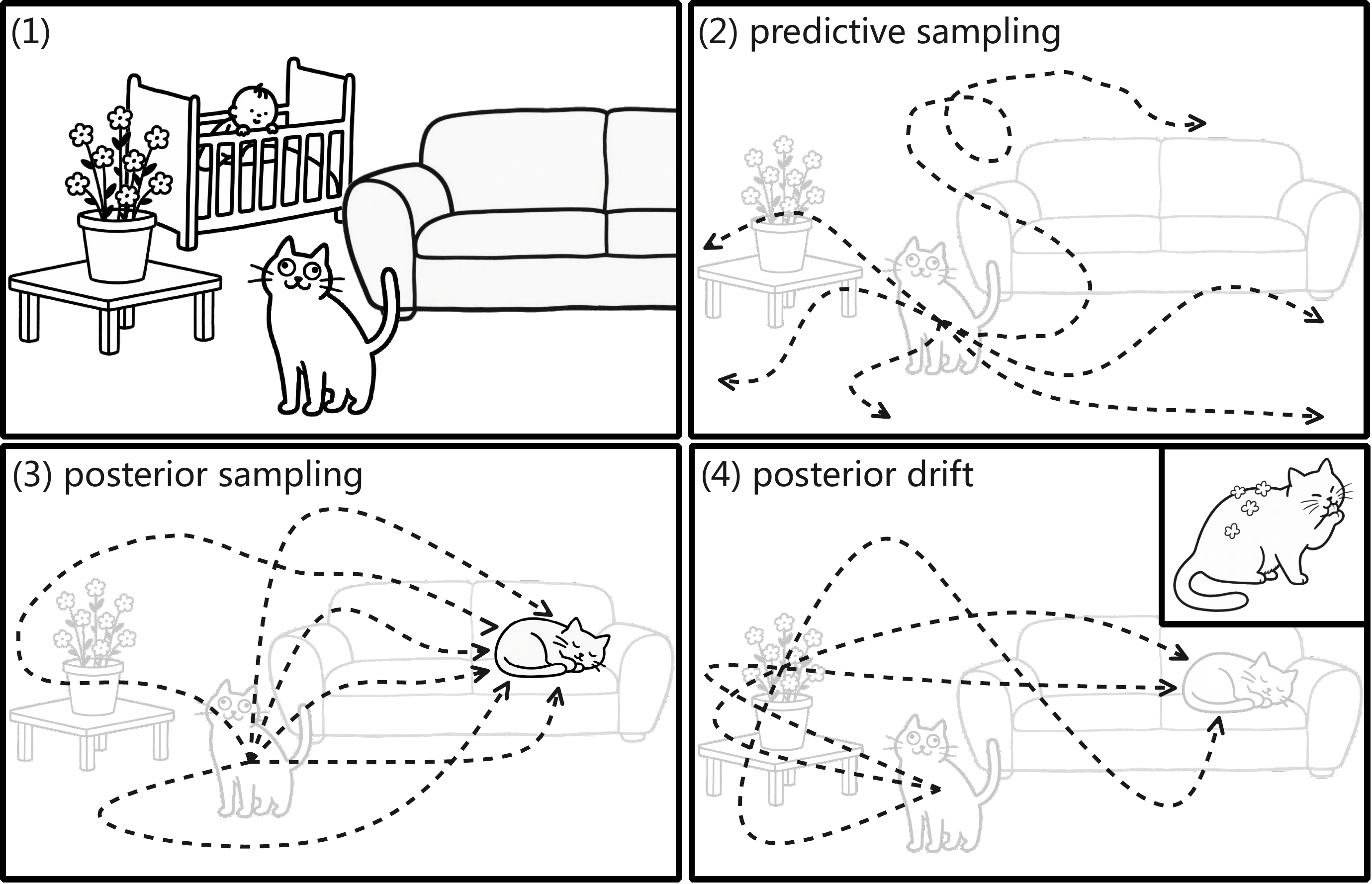}
\caption{Illustration of Example \ref{ex. cat trajectory}.
The four panels are: your room, the predictive sampling distribution at time $t_0$,
the posterior sampling distribution given an observation at time $t_1$,
and the posterior sampling distribution for $[t_0,t_1]$ given a new observation at time $t_2$.}
\label{fig: cat sampler}
\end{figure}

\begin{example}
\label{ex. cat trajectory}
Suppose you are a baby.
You peek out of the crib and look around the room, which appears like Panel 1 of Figure \ref{fig: cat sampler}.
At time $t_0$, a cat shows up.
You decide to study its movement.
As cats are mysteriously stochastic creatures, the possible trajectories of the cat starting from $t_0$ have high variability, as depicted in Panel 2.
At time $t_1 > t_0$, you peek out again, and the cat now sleeps on the sofa.
The possible trajectories between $t_0$ and $t_1$ collapse to those that end with a particular terminal state, as depicted in Panel 3.
Despite that these posterior trajectories still have considerable variability, they provide some knowledge of the movement of the cat,
e.g.\@ that it can jump as high as the sofa, and can travel at a speed of $\text{the distance}/(t_1-t_0)$.
In comparison, the predicted trajectories in Panel 2 are unhelpful for your learning, since it is very unlikely that such a trajectory can end with the right terminal state.
After that, at time $t_2 > t_1$, you notice that there are some flower petals stuck on the cat.
You realize that the cat may have interacted with the flowerpot.
So the possible trajectories further concentrate around those that pass by the flowerpot, as depicted in Panel 4.
This change indicates that the posterior distribution of latent variables at earlier times may be influenced by observations from later times, a phenomenon that we later refer to as ``posterior drift".
These sampled trajectories bring you new knowledge, e.g.\@ cats are fond of rubbing against certain surfaces, cats' fur can easily attract some objects, and cats might travel at a faster speed than previously estimated.
\end{example}

\begin{figure}[!ht]
\centering
\includegraphics[width=1\linewidth]{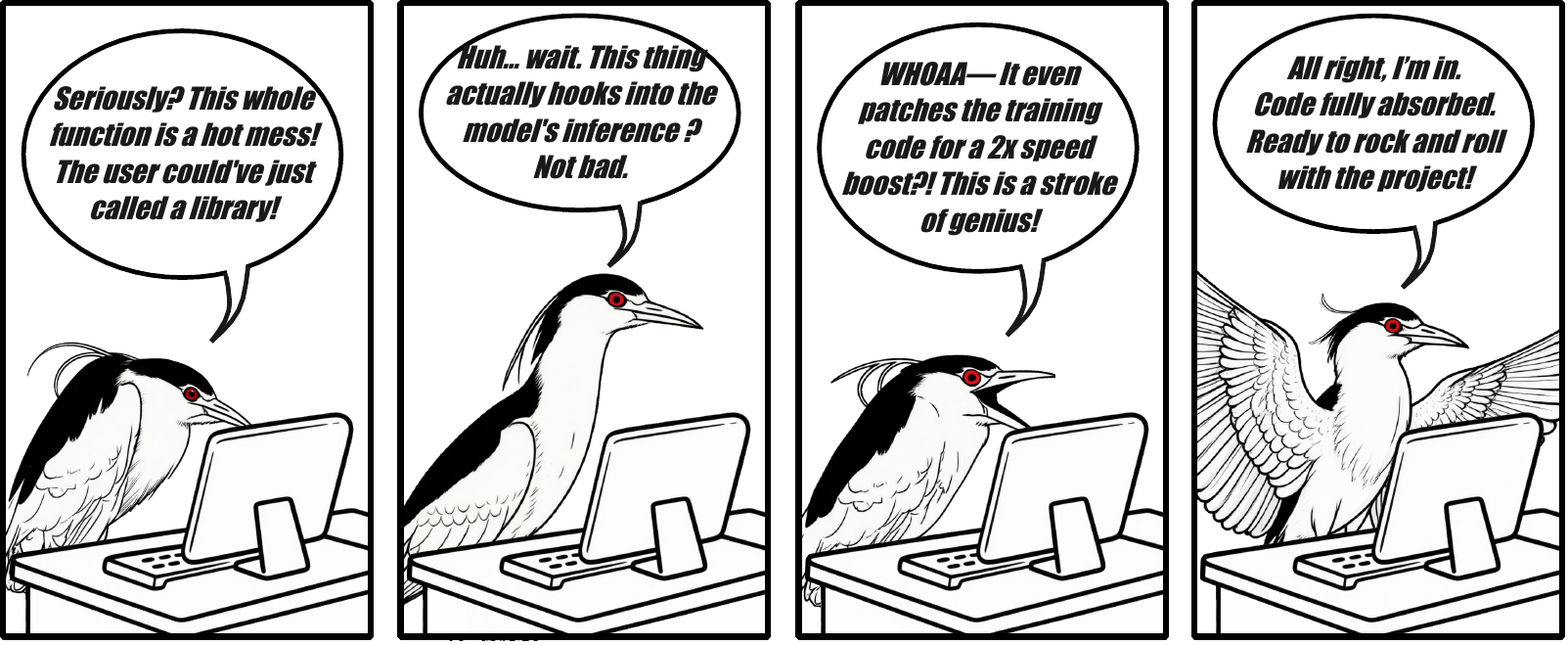}
\caption{Illustration of Example \ref{ex. night heron}.
The well-functioning of an AI agent depends on a proper encoding of the provided context, which depends on the generation of a comprehensive understanding in the case of slow thinking.
In general, the latter must be performed in a non-causal manner.}
\label{fig: night heron}
\end{figure}

\begin{example}
\label{ex. night heron}
Suppose you ask your AI agent to join a coding project that you have been working on.
The agent needs to read and understand your existing code before generating further code.
Figure \ref{fig: night heron} illustrates how its understanding may evolve during reading.
To generate a rationale for a text segment, sometimes both the preceding and subsequent contexts are needed.
Misunderstanding can happen if the agent ponders on a code snippet while having read only the preceding code, as depicted in Panel 1.
As the agent sees how this snippet is applied in subsequent code, it develops a more complete rationale and thus deeper understanding, as depicted in Panel 2.
Panel 3 further illustrates that new understanding of the code snippet can be obtained from the very late parts of the project.
This is similar to the posterior drift in Example \ref{ex. cat trajectory}.
Finally in Panel 4, as the agent finishes reading everything, it can generate a complete rationale for every part of the code and has fully understood your project.
Technically, this task demands more than bi-directional attention \cite{devlin2019bert,raffel2020T5}, since a proper understanding requires sequential computation and cannot be handled by bounded-depth circuits,
an intuition that will be justified in Section \ref{sec. separation I}.
\end{example}

\begin{figure}[!ht]
\centering
\includegraphics[width=1\linewidth]{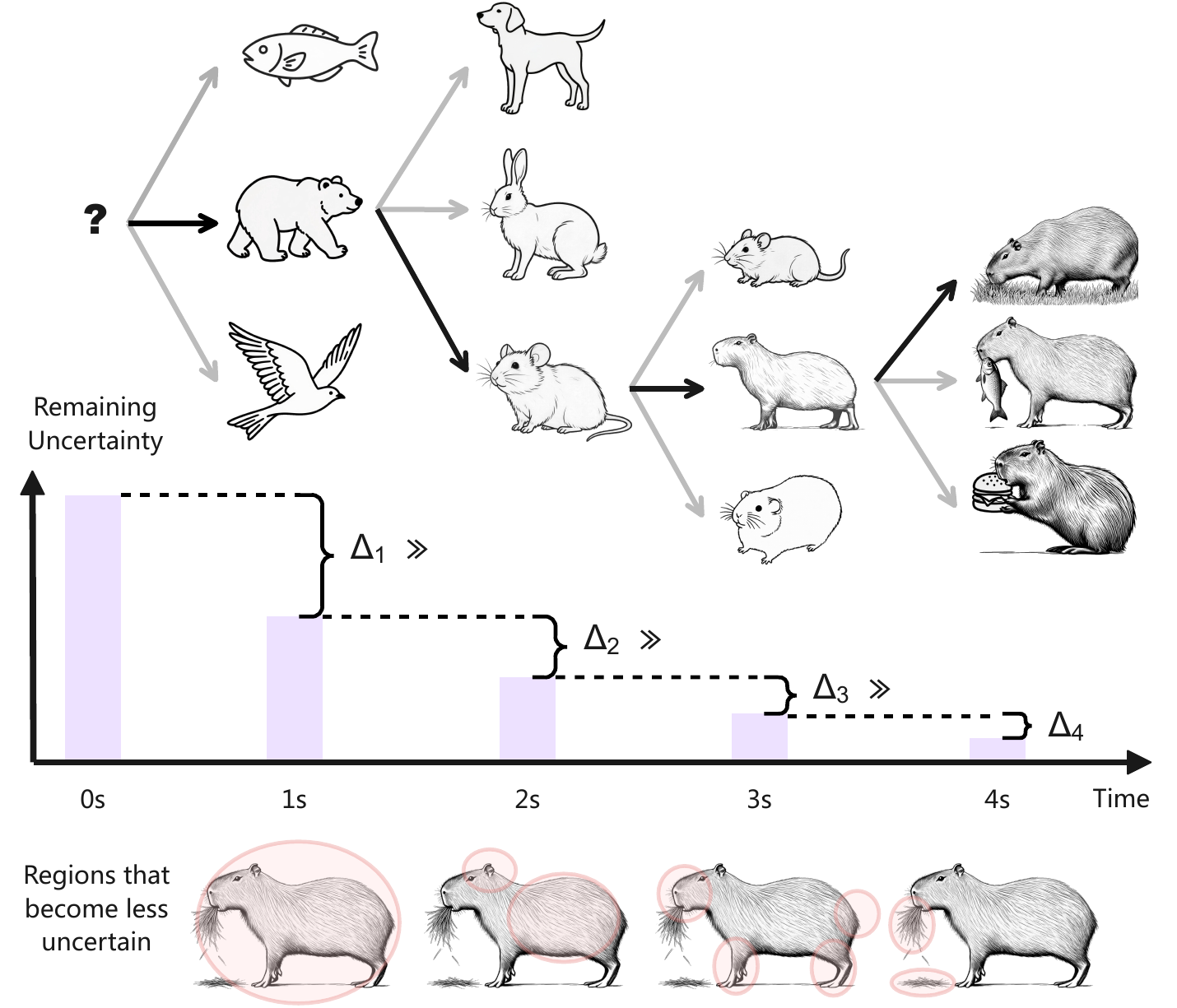}
\caption{Illustration of Example \ref{ex. capybara}.
One possible way to model the agency of human vision is to maximize the rate of uncertainty reduction.
The bottom illustrates the information extracted by an optimal policy over time.
The top depicts the posterior distributions given the partial information at each second, which collapse gradually to the actual scene.
The middle plot indicates that the remaining uncertainty decreases over time, and since its rate is maximized, the greatest reductions tend to happen at earlier steps.}
\label{fig: capybara}
\end{figure}

\begin{example}
\label{ex. capybara}
Suppose you are a kid and your parents take you to the zoo.
You are facing an animal called ``capybara" that you have never seen before.
Your curiosity compels you to ``load" its appearance into your mind as quickly as possible.
So what is going on in your mind?
Figure \ref{fig: capybara} depicts one possibility.
In the 1st second, you perceive that the capybara is a fluffy creature, unlike the birds and fishes (Later you will learn that this is called ``mammals").
In the 2nd second, you perceive that the capybara has a chubby body and round ears, similar to mice (or rodents), unlike dogs and rabbits.
In the 3rd second, you perceive that the capybara has a square face and no tail, distinct from the rats and hamsters.
Finally, in the 4th second, you perceive that the capybara is eating grass.
This is an illustration of the ``agency" and ``urgency" of your visual perception.
Agency means that you can decide which information gets loaded first,
and urgency means that this loading order tries to eliminate uncertainty with maximum speed.
Specifically, urgency may be modeled by the minimization of
$$\sum_{t=0}^{\infty} H_t = \sum_{t=1}^{\infty} t ~ (H_{t-1} - H_t)$$
where $H_t$ is some non-negative measure of the uncertainty at step $t$ (assuming that $H_{\infty}=0$).
The right-hand side can be seen as decomposing the prior uncertainty $H_0$ into progressive reductions $H_{t-1} - H_t$ and allocating these reductions to the observation steps $t$,
and minimization requires allocating larger reductions to earlier steps.
If discriminative features are defined as perceptions effective for reducing the posterior uncertainty $H_t$, then we are motivated to perceive such features at early time, thus determining the ``loading order".
In this example, uncertainty is reduced in the following manner:
In the beginning, the prior distribution of the possible appearances of capybara has high uncertainty,
and then in later seconds, the posterior distribution narrows down to roughly the distribution of mammals,
then the distribution of rodents,
and the distribution of capybara-like creatures,
and finally concentrates onto the image of capybara with the right living environment.
So not only is uncertainty being reduced, but the rate of reduction is also maximized.

Of course, one also needs to take into account the comprehensibility of perceptions.
If one's visual focus jumps around abruptly, then the sequence of perceived visual features would be difficult to understand.
One extreme example would be reading books, such that one's visual focus usually follows the printed words in the linear order, which is the most understandable order.
In general, we balance urgency and comprehensibility in a flexible manner; for instance, when reading a research paper, urgency compels us to first look for the keywords, main ideas, novel methods, etc.,
while comprehensibility compels us to read the sentences, tables and plots in some logical order, no matter how complex the formatting of this paper is.
We will see how this balance can be achieved in Section \ref{sec. active lifting}.
\end{example}





\subsection{Overview}

First, we establish the static theory in Sections \ref{sec. separation theorems} and \ref{sec. latent sampling}, where we derive the representation hierarchy and sampler hierarchy.
Then, Section \ref{sec. unified objective} proposes a maximum principle that can qualitatively derive the static theory in a unified fashion.
For concreteness, Section \ref{sec. to existing models} derives the existing slow thinking models (in particular, DeepSeek-R1 \cite{guo2025deepseek}), and proposes three stages of improvements for those models.
Section \ref{sec. perceptual} establishes the framework of active lifting, generalizing the static theory.
Section \ref{sec. experiment} describes a preliminary experiment of our stage one improvement of slow thinking.
Section \ref{sec. related work} lists the remaining related works.
Section \ref{sec. conclusion} concludes this paper.
All proofs are listed in the Appendices.

\section*{List of Symbols}

Most of these notations will be introduced later in this paper,
and this table is for reference only.

\begin{longtable}{@{}p{0.19\textwidth} p{0.78\textwidth}@{}}
\toprule
\textbf{Symbol} & \textbf{Description} \\
\midrule
\endfirsthead
\multicolumn{2}{c}{{\tablename\ \thetable{} -- continued from previous page}} \\
\toprule
\textbf{Symbol} & \textbf{Description} \\
\midrule
\endhead
\midrule
\multicolumn{2}{r}{{Continued on next page}} \\
\endfoot
\bottomrule
\endlastfoot


$\X, \Z$ & The data space and the latent space \\
$\PS(\X)$ & The space of Borel probability measures (if $\X$ is a topological space) or the space of probability measures (if $\X$ is just a measureable space) \\
$P_*$ & The target distribution $\in\PS(\X)$ \\
$P_f$ & A parametrized probability measure with function parameter $f$ \\
$f\#P$ & Pushforward or ``transport" of a measure $P$ by function $f$ \\
$Q\colon \X\to\PS(\Z)$ & A conditional probability measure \\
$\law(X)$ & The probability law of a random variable $X$ \\
$\mathbbm{1}_S$ & The indicator function of a set $S$ \\
$\delta_x$ & Dirac measure at an element $x\in\X$ \\
$\Sigma, \Omega$ & The observable vocabulary and the latent vocabulary \\
$\N, \omega$ & The natural numbers $\{0,1,2,\dots\}$ and the countable infinity \\
$\Omega^*$ & The space of finite sequences \\
$\varnothing$ & Often, the empty sequence $\in \Omega^*$ \\
$\Omega^{\omega}$ & The space of infinite sequences (with product topology) \\
$d_{\omega}$ & Metric that induces the product topology of $\Omega^{\omega}$ \\
$\Omega^{\leq T}, \Omega^{\geq T}$ & The subspace of sequences with lengths bounded above or below by $T \in \N$ \\
$\sqcup$ & Disjoint union \\
$\Omega^{\leq\omega}$ & The union $\Omega^* \sqcup \Omega^{\omega}$ \\
$|x|$ & The length $\in \N\cup\{\omega\}$ of a sequence $x\in\Omega^{\leq \omega}$ \\
$x_t$ & The $t$-th entry of a sequence $x\in\Omega^{\geq t}$ \\
$\{x^{(i)}\}_{i=1}^n$ & A set of sequences indexed by $i$ \\
$x \sqsubseteq y$ & Sequence $x \in \Omega^{\leq\omega}$ is a prefix of $y \in \Omega^{\leq \omega}$, or simply $y$ starts with $x$ \\
$[x]$ & The cylinder set $\subseteq \Omega^{\omega}$ of a sequence $x\in\Omega^*$ \\
$\preceq$ & Partial order (\ref{eq. partial order}) on the subsets of $\Omega^*$ \\
$xy$ & Concatenation of a finite sequence $x$ and a sequence $y$ \\
$x^n, x^{\omega}$ & Repetition of a finite sequence $x$ for $n$ times or infinite times \\
$\text{gcd}(x,y)$ & Longest common prefix of two sequences \\
$P$ & Often, a latent distribution over the latent space $\Z$ \\
$P^{\leq |x|}(x) = P([x])$ & Probability of sequences that start with $x \in \Omega^{\leq\omega}$, given $P \in \PS(\Omega^{\omega})$ \\
$P(\cdot|[x])$ & Conditional probability measure on the event $[x]$, given that $P([x])>0$ \\
$P^{\leq |x|+|y|}(y|x)$ & Conditional probability of suffix $y \in \Omega^{\leq\omega}$ on prefix $x\in\Omega^*$ \\
$P(y \mid x\to S)$ & Probability conditioned on the suffix being in some prefix-free subset $S \subseteq \Omega^{\leq\omega}$ \\
$P_{A,B,\lambda}$ & A hidden Markov model with stochastic matrices $A,B$ and probability vector $\lambda$ \\
$\TF$ & The space of Transformer functions \\
$\TC^0$ & Bounded-depth polynomial-size threshold circuits \\
$\textsf{DLOGTIME}$ & Computable in logarithmic time with a deterministic Turing machine \\
$\proj\colon\Omega^{\omega}\rightharpoonup\Sigma^{\omega}$ & The projection as a partial function from $\Omega^{\omega}$ to $\Sigma^{\omega}$ \\
$\fip$ & The lifting (static theory) \\
$\dom(\proj)$ & The domain of a (partial) function \\
$\proj^{<\omega} $ & The prefix map \\
$\nsm, \nsm^{\leq 1}$ & The next-segment map and the next-token set \\
$\Q_{\proj},\Q_{+1}$ & The spaces of latent samplers for the static and active theories \\
$Q_*,Q_{\eye}$ & The posterior sampler and the inquisitive sampler \\
$Q,Q^{\see}$ & The (inference) sampler and the train sampler \\
$Q_{\id},\Q_{\proj}^{\text{\faForward}},\Q_{\proj}^{\text{\faPlay}}$ & The identity sampler, the predictive samplers, the causal samplers \\
$\chi^2$ & The chi-square divergence \\
$\KL,\TV$ & KL-divergence and total variation \\
$\detach$ & The stop-gradient or ``detach" operation \\
$\bos,\eos$ & The beginning/end-of-text symbols \\
$\sot,\eot$ & The start/end-of-thought symbols \\
$\mathcal{T}$ & The space of ``thought" sequences \\
$P\otimes Q$ & The product measure if both $P,Q$ are measures, or the joint measure if either one is a conditional measure \\
$Q^{\leq\omega}$ & The infinite sequence sampler induced by a sampler $Q \in \Q_{+1}$ \\
$Q^{\leq\omega}_{<\omega}$ &  The infinite sequence sampler with text prefixes as inputs \\
$P_{\text{lift}} = P_*\otimes Q^{\leq\omega}$ & The lifted distribution \\
$Q^{\leq \omega}_* \otimes P_*^{\leq\omega} = P_{\text{lift}}$ & The marginal distribution and the posterior data distribution \\
$d P / d Q$ & Radon-Nikodym derivative \\
$\Sigma^{\omega}_{\text{\faCode}},\Omega^{\omega}_{\text{\faCode}}$ & Observable/latent sequences in the conversation format \\
$\Sigma^*_{\square}$ & The set of admissible communications in the conversation format \\
$\Delta_d$ & The probability simplex in $\R^d$
\end{longtable}

\clearpage

\section{Separation of Approximation Ability}
\label{sec. separation theorems}


The static theory, when viewed as a standalone framework, features two optimization problems,
the maximizations of approximation ability and sampling efficiency.
The former concerns the hierarchy of parametrized families of probability measures, such that higher families can approximate more sophisticated target distributions.
The latter is about the hierarchy of subspaces of latent samplers, such that larger subspaces can achieve higher sampling efficiency.
Later, a unified derivation using one objective will be discussed in Section \ref{sec. unified objective}.

\begin{figure}[!ht]
\centering
\includegraphics[width=0.6\linewidth]{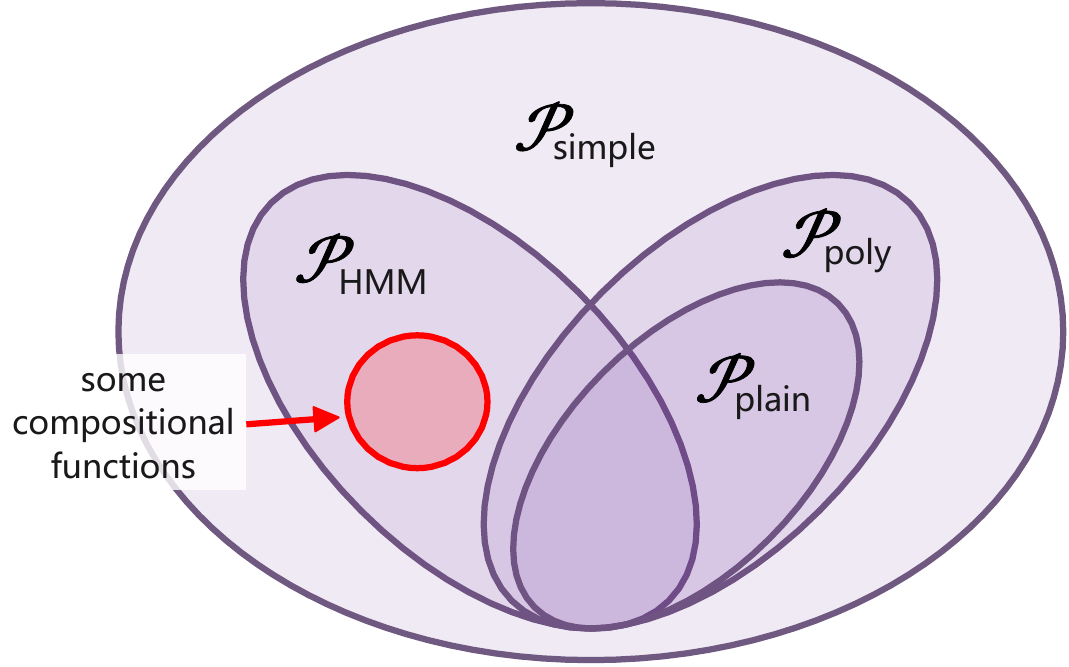}
\caption{The representation hierarchy.
It illustrates the various approaches to modeling distributions over sequences and compares their expressivity.
The simple family of hidden Markov models ($\PS_{\text{HMM}}$) is included as a reference, separating the fast-thinking models ($\PS_{\text{plain}}$) from a more expressive and yet achievable class of models ($\PS_{\text{simple}}$).
A detailed description is provided in Section \ref{sec. approximation summary}.
The existing slow-thinking models also lack expressivity and their position in this hierarchy will be depicted in Figure \ref{fig: forgetful in hierarchy}.}
\label{fig: representation hierarchy}
\end{figure}

This section studies the first problem of approximation ability.
A summary of the key results is provided by Figure \ref{fig: representation hierarchy}.
Although the specific function family of Transformers is used, the derivation works for constant-depth Boolean circuits in general and thus can be regarded as a first-principles approach.

\subsection{From Learning Ability to Approximation Ability}
\label{sec. learnability}

To begin with, this section introduces approximation theory and its significance to the learning ability of a model.
Specifically, approximation theory not only determines whether a target distribution can be approximated by a model,
but also how fast it can approximate this distribution during training.

As a general setup, the data space $\X$ is a measureable space, $\PS$ is the space of probability measures over $\X$, and $P_* \in \PS$ denotes the target distribution.
Let $\F$ be a space of measureable functions with a norm, or more generally a complexity measure, $\|\cdot\|_{\F}$.
A distribution parametrization $f\mapsto P_f$ is a mapping from $\F$ to $\PS$ and we denote its image by $\PS_{\F}$.
Meanwhile, if $\F$ is the result of some parametrization $\theta \mapsto f_{\theta}$, then we call it function parametrization.
This subspace $\PS_{\F}$ is what we mean by a model or parametrized family.
The two most common distribution parametrizations are potential model
$$P_f=\text{softmax}(f):= \frac{e^f}{\int_{\X} e^f}$$
where $\F$ is a subspace of measureable functions $\X\to\R\cup\{-\infty\}$,
and transport model
$$P_f=f\#P_{\text{latent}}:=\law(f(X)), ~ X\sim P_{\text{latent}}$$
where $P_{\text{latent}}$ is a base distribution on some latent space $\Z$ and $\F$ is a subspace of measureable functions $\Z \to \X$.
A complexity measure can be defined for all probability measures:
For any $P \in \PS$,
\begin{equation}
\label{eq. distribution norm}
\|P\|_{\F} := \inf_{f\in\F} \bigl\{ \|f\|_{\F} \bigm| P=P_f \bigr\}
\end{equation}
By definition, $\|P\|_{\F} < \infty$ if and only if $P \in \PS_{\F}$.
Intuitively, (\ref{eq. distribution norm}) measures the difficulty of fitting $P$ with this model.
If for instance the training objective is the KL-divergence $\KL(P_*\|P_f)$, then it cannot reach $0$ unless $\|P_*\|_{\F} < \infty$.
Thus, one purpose of approximation theory is to design appropriate distribution parametrizations and function parametrizations such that $\|P_*\|_{\F} < \infty$.

\begin{remark}
\label{remark. logic}
Besides practicality, defining distribution parametrizations with functions as parameters might be necessary from a logical point of view.
Since functions belong to second-order logic (i.e.\@ functions are special sets) while probability measures belong to third-order logic (i.e.\@ they take sets as input), it seems more tractable to model measures indirectly through functions than to model them directly.
\end{remark}

For concreteness, here are some examples of the complexity measure $\|\cdot\|_{\F}$.
If $\F$ consists of neural networks, then one can define $\|\cdot\|_{\F}$ as the parameter norm of the network or any other measure of the magnitude of its parameters \cite{e2022barron,e2019residual}.
In case multiple parameters lead to the same function, the infimum over those instantiations is taken.
Several theorems cited below adopt two-layer networks whose first-layer weights are frozen, commonly known as random feature functions \cite{rahimi2008uniform}.
Their norm $\|\cdot\|_{\F}$ is the well-known RKHS norm \cite{aronszajn1950theory,berlinet2011reproducing}, which equals the parameter norm of the trainable second-layer weights \cite{rahimi2008uniform}.

Besides reachability, the complexity measure $\|P_*\|_{\F}$ determines the convergence rate of training.
Suppose for simplicity that training is performed with continuous-time gradient flow, the random feature functions for $\F$, and the population loss instead of mini-batch loss.
Denote the trained model at time $t$ by $P_t$.
Then, the training loss has the following estimate for various models
\begin{equation}
\label{eq. train loss upper bound}
\KL(P_*\|P_t) \lesssim \frac{\|P_*\|_{\F}}{t^a}
\end{equation}
where $a>0$ is a constant.
Specifically, this is Theorem 6.2.1 of \cite{yang2023thesis}, which considers the potential model (or softmax distribution).
Several results of \cite{yang2023thesis} also take similar forms, including Proposition 6.2.3 for Flow Matching, and Proposition 5.4.5 and Lemma 5.4.9 for a simplified GAN model.
This estimate suggests that, if a good parametrized family $\PS_{\F}$ is chosen such that the complexity $\|P_*\|_{\F}$ is small, then training will likely proceed more rapidly.
The dependency of training loss on $\|\cdot\|_{\F}$ is expected, as lower/upper bounds in the form of (\ref{eq. train loss upper bound}) are common in the study of supervised learning, e.g.\@ \cite[Section 6]{dieuleveut2017harder}, \cite[Theorem 4.1]{e2020comparative}, \cite[Corollary 5.6]{e2020multilayer} and \cite[Section 6.2]{yang2023thesis}.

\begin{remark}
\label{remark. misspecified approximation}
The training loss estimate (\ref{eq. train loss upper bound}) can be generalized to the case when $\|P_*\|_{\F} = \infty$.
For instance, by Remark 4 of \cite{yang2023thesis}, the upper bound (\ref{eq. train loss upper bound}) may hold with a smaller exponent $a' \in (0,a)$ that depends on the ``density" of $\PS_{\F}$ around $P_*$.
This paper does not go into such fine-grained analysis of approximation theory, and will primarily focus on reducing $\|P_*\|_{\F}$.
\end{remark}

In addition, we mention that approximation ability not only determines the training loss but also indirectly affects the test loss (or generalization error).
In practice, the target distribution $P_*$ is hidden from us, accessible only through a finite sample set $\{X_i\}_{i=1}^n \iidsample P_*$ with size $n$,
and the model needs to recover $P_*$ so that we can generate new samples.
Denote the empirical distribution by $P_*^{(n)}=\frac{1}{n}\sum_{i=1}^n\delta_{X_i}$, and the trained model at time $t$ by $P^{(n)}_t$.
%
Similar to (\ref{eq. train loss upper bound}), the test loss has the following estimate
\begin{equation}
\label{eq. test loss implicit}
\KL(P_*\|P^{(n)}_t) \lesssim \frac{\|P_*\|_{\F}}{t^a} + \frac{t^b}{\sqrt{n}}
\end{equation}
for some constant $b>0$.
Specifically, this is Theorem 5.2.1 of \cite{yang2023thesis} for softmax distributions.
Theorem 5.3.6 (for Flow Matching) and Theorems 5.4.1 and 5.4.7 (for simplified GAN) also take similar forms.
With early-stopping $t=\Theta(n^{\frac{1}{2(a+b)}}\|P_*\|_{\F}^{\frac{1}{a+b}})$, the test loss becomes
\begin{equation}
\label{eq. test loss early stop}
\KL(P_*\|P^{(n)}_t) \lesssim \|P_*\|_{\F}^{\frac{b}{a+b}}n^{-\frac{a}{2(a+b)}}
\end{equation}
The exponent $-\frac{a}{2(a+b)}$ over the sample size $n$ does not involve the dimension or size of $\X$, so the model can escape from the curse of dimensionality if a good parametrization is chosen,
e.g.\@ if $\|P_*\|_{\F}$ is not on the order of $10^{\text{dim}(\X)}$.






Hence, the ability of a model to represent and learn a target distribution can be characterized by its approximation ability.
Designing a more expressive model means parametrizing a larger subspace $\PS_{\F} \subseteq \PS$ and reducing the complexity $\|P_*\|_{\F}$ for any target $P_*$ of interest.
The common approach is to design a function parametrization that better suits $P_*$, e.g.\@ a specific neural network architecture that incorporates our prior knowledge of the data characteristics of $P_*$.
However, our separation theorems will show that this is inadequate for distributions over sequences such as texts, so an stronger distribution parametrization is needed.

\subsection{Sequence Data}
\label{sec. sequence}

This section defines the sequence spaces to prepare for our technical analysis.
Sequences are central to this paper not only because we focus on the text modality, but also because our modeling of all data modalities relies on a latent space that consists of infinite sequences.

Given any finite set $\Sigma$, let $\Sigma^{\omega}$ be the space of infinite sequences;
$\Sigma$ is referred to as its vocabulary, and each $a \in \Sigma$ as a token.
Each $x \in \Sigma^{\omega}$ can be written as a list of tokens $(x_t)_{t=1}^{\infty}$.
We usually use $t$ to index the sequences for reasons that will become clear in Section \ref{sec. perceptual}.
We always equip $\Sigma^{\omega}$ with the product topology, which seems to be the most natural choice.
The product topology is compact by Tychonoff's theorem and is equivalent to the topology induced by the metric
\begin{equation}
\label{eq. metric product topology}
d_{\omega}(x,y) = 2^{-\inf\{t\mid x_t\neq y_t\}}
\end{equation}
Denote the space of finite sequences by $\Sigma^*=\bigcup_{t=0}^{\infty}\Sigma^t$.
Denote the empty string by $\varnothing \in \Sigma^*$.
Denote $\Sigma^{\leq\omega}=\Sigma^*\sqcup\Sigma^{\omega}$, where $\sqcup$ means disjoint union.
Denote the length of $x\in\Sigma^{\leq\omega}$ by $|x|$.
For any $T\in\N$, denote $\Sigma^{\leq T}=\bigcup_{t=0}^T \Sigma^t$ and $\Sigma^{\geq T} = \bigcup_{t=T}^{\infty} \Sigma^t$.
For any $x \in \Sigma^{\omega} \cup \Sigma^{\geq T}$, denote its first $T$ entries by $x_{\leq T} \in \Sigma^T$, and similarly define $x_{\geq T}$.
For any $x,y\in\Sigma^{\leq\omega}$, if $|x|\leq|y|$ and $x = y_{\leq |x|}$, then we say that $x$ is a prefix of $y$ and denote $x \sqsubseteq y$.
The cylinder of any $x\in\Sigma^{\leq\omega}$ is defined as
\begin{equation*}
[x] := \{x' \in \Sigma^{\omega} \mid x \sqsubseteq x'\}
\end{equation*}
For any $x,y\in\Sigma^{\leq\omega}$, denote their longest common prefix by $\text{gcd}(x,y)\in\Sigma^{\leq\omega}$,
\begin{equation*}
\text{gcd}(x,y) = \argmax_{z\in\Sigma^{\leq\omega}} \big\{|z| \bigm| z \sqsubseteq x, ~ z \sqsubseteq y \big\}
\end{equation*}
We say that a subset $S \subseteq \Sigma^{\leq\omega}$ is prefix-free, if there does not exist $z,z'\in S$ such that $z \sqsubseteq z'$.
A partial order $\preceq$ can be defined on the subsets of $\Sigma^*$:
For any $S,S' \subseteq \Sigma^*$,
\begin{equation}
\label{eq. partial order}
S \preceq S' ~\longleftrightarrow~ \forall z' \in S', ~\exists z\in S, ~ z \sqsubseteq z'
\end{equation}
For any sequences $x\in\Sigma^*$ and $y\in\Sigma^{\leq\omega}$, denote their concatenation by $xy$.
For any subsets $A,B \subseteq \Sigma^*$, define their concatenation $AB\subseteq \Sigma^*$ as
\begin{equation}
\label{eq. concatenation}
AB = \{ ab \mid a\in A, b\in B \}
\end{equation}


Given any measureable space $\X$, any probability measure $P$ over $\X$, and any event (measureable subset) $E \subseteq \X$ such that $P(E)>0$,
define the conditional probability measure $P(\cdot|E)$ by
\begin{equation*}
P(\cdot|E) = \frac{\mathbbm{1}_E}{P(E)}P
\end{equation*}
Thus, given any Borel probability measure $P$ over $\Sigma^{\omega}$ and any $x\in\Sigma^*$ such that $P([x])>0$, we can define the conditional probability measure $P(\cdot|[x])$.
The sample $X\sim P(\cdot|[x])$ satisfies $x \sqsubseteq X$ almost surely.
Denote by $P^{\leq T}$ (and $P^{< T}$) the marginal distribution of $P$ over the first $T$ (or $T-1$) entries.
Thus, $P^{\leq |x|}(x) = P([x])$ for any $x\in\Sigma^*$.
We often use the following notation for any $x,y\in\Sigma^*$,
\begin{equation*}
P^{\leq |x|+|y|}(y|x) := P\big([xy]\bigm|[x]\big)
\end{equation*}



\begin{remark}
Our derivations often start from objects defined on infinite sequences $\Sigma^{\omega}$ and end with objects on finite sequences $\Sigma^*$.
This may seem a detour, as one might want to define and study these objects directly in the realistic setting of finite sequences.
However, it is common practice in math to take such detours, by first going into an extended and more ``complete" space, whose completeness allows us to solve for our desired objects more easily, and then translating the solutions back to the original space.
For instance, we go to complex numbers when solving linear ODEs even if the solutions are real functions, and go to Sobolev spaces when solving PDEs even if the solutions are smooth.
In our case, the set of limit points that complete $\Sigma^*$ is exactly $\Sigma^{\omega}$ (when $\Sigma^*$ is equipped with the metric $d_*(x,y)=2^{-|\text{gcd}(x,y)|}$).
We will see that several desirable properties such as causality (Section \ref{sec. continuous projection}) and determinism (Section \ref{sec. active lifting}) take on simple forms when expressed in terms of infinite sequences.
\end{remark}

\subsection{Probability Space}
\label{sec. loss}

This section characterizes the space of probability measures.
A semi-divergence is introduced that will help to build the representation hierarchy.

Let $\PS(\Sigma^{\omega})$ be the space of Borel probability measures over $\Sigma^{\omega}$.
Its weak topology is used by default,
namely, a sequence $\{P_n\}$ converges to a limit $P_*$ if and only if $\E_{P_n}[f]\to\E_{P_*}[f]$ for any bounded continuous function $f\colon \Sigma^{\omega}\to\R$.

Approximation ability will be studied with respect to the weak topology,
so let us define a convenient divergence over $\PS(\Sigma^{\omega})$ that captures the weak topology.
Recall that the KL divergence can be expressed as a sum of entry-wise KL divergences:
For any $T\in\N$ and $P,P_*\in\PS(\Sigma^T)$,
\begin{equation*}
\KL(P_*\|P) = \sum_{t=1}^T \E_{X \sim P_*^{<t}} \Big[ \KL\big(P_*^{\leq t}(\cdot|X) \big\| P^{\leq t}(\cdot|X) \big) \Big]
\end{equation*}
The topology induced by KL divergence is too strong when $T$ becomes infinity, so we weaken it as follows:
For any $P,P_*\in\PS(\Sigma^{\omega})$,
\begin{equation}
\label{eq. infinite divergence}
D_0(P_*,P) := \sum_{t=1}^{\infty} 2^{-t} \min\Big(1,~ \E_{X\sim P_*^{<t}} \Big[ \KL\big(P_*^{\leq t}(\cdot|X) \big\| P^{\leq t}(\cdot|X) \big) \Big] \Big)
\end{equation}
One can check that this is a well-defined divergence that ranges in $[0,1]$, and that it induces the right topology.

\begin{proposition}
\label{prop. weak topology}
A sequence $\{P_n\}$ converges to $P_*$ in the weak topology of $\PS(\Sigma^{\omega})$ if and only if $D_0(P_*,P_n)\to 0$.
\end{proposition}



One detail to consider, however, is to ignore numerical error.
Intuitively, for any model that tries to fit a target distribution $P_*\in\PS(\Sigma^{\omega})$, its error $D_0$ has two sources, the logical errors and numerical errors.
For instance, if a language model $P$ cannot solve a math problem, then it incurs a logical error, e.g.\@ $P^{\leq T}(\text{``37"}\mid\text{``18+19="})$ is close to 0, so the KL divergence conditioned on the problem would be large.
Here, $T$ is the number of tokens of the text ``18+19=37".
Meanwhile, even if the model answers correctly, there could still be some numerical error, e.g.\@ $P_*^{\leq T}(\text{``37"}\mid\text{``18+19="})=0.9591635\dots$ whereas $P^{\leq T}(\text{``37"}\mid\text{``18+19="})=0.96$ due to limited floating-point precision for example.

We want $D_0$ to ignore the numerical errors for a technical reason.
Our theorems consider Transformer networks with floating-point numbers, because this is not only the case for all practical scenarios, but also needed when modeling these networks as Boolean circuits \cite{merrill2023parallelism,chiang2025TC0}.
For illustration, the number $1+\ep$ is rounded to $1$ whenever $\ep\in[-\frac{1}{4096},\frac{1}{2048}]$ when using float16, $\ep\in[-\frac{1}{256},\frac{1}{128}]$ with bfloat16, and $\ep\in[-\frac{1}{32},\frac{1}{16}]$ with float8\_e4m3.
Thus, the integrand $\log P_*/P$ in (\ref{eq. infinite divergence}) will in general include an irreducible error $\log(1+\ep)\approx \ep$.
Intuitively, what matters for both humans and LLMs is whether they can solve problems correctly, rather than computing the exact real-number probabilities.
So we modify $D_0$ to allow for small pointwise errors:
\begin{align}
\label{eq. semi-divergence}
D(P_*,P) &:= \sum_{t=1}^{\infty} 2^{-t} \min\big(1,~ D_t(P_*, P) \big) \\
\nonumber
D_t(P_*,P) &:= \E_{X\sim P_*^{<t}} \Big[ \KL_{0.01}\big(P_*^{\leq t}(\cdot|X) \big\| P^{\leq t}(\cdot|X) \big) \Big] \\
\label{eq. KL 0.01}
\KL_{0.01}(p_*\|p) &:= \inf_{q} \Big\{ \KL(p_*\|q) \Bigm| q\geq 0, ~\int q=1 ,~ \|\log(q/p)\|_{\infty} \leq 0.01 \Big\}
\end{align}
where $\|\cdot\|_{\infty}$ denotes the supremum norm of finite-dimensional vectors,
and we require $q=0$ for entries where $p=0$.
This $\KL_{0.01}$ minimizes over all surrogate distributions $q$ whose pointwise difference is small but sufficient to absorb the numerical error of $p$.
This $D$ is non-negative and can be seen as a ``semi-divergence".


This $D$ is purely theoretical;
It is used only for defining the representation hierarchy, and does not participate in training or evaluation.
Specifically, given two subsets (or topological subspaces) $\PS_1,\PS_2 \subseteq \PS(\Sigma^{\omega})$,
we say that $\PS_1$ can approximate $\PS_2$ if
\begin{equation}
\label{eq. semidivergence approximate}
\sup_{P_* \in \PS_2} \inf_{P \in \PS_1} D(P_*,P) = 0
\end{equation}
and that
$\PS_1$ contains $\PS_2$ if
\begin{equation}
\label{eq. semidivergence contain}
\sup_{P_* \in \PS_2} \min_{P \in \PS_1} D(P_*,P) = 0
\end{equation}
So what matters is only whether $D$ can approximate or reach $0$, instead of its exact value.
(Technically, these relations do not seem to be transitive, though this property is not needed for this paper).

\subsection{Parametrization}
\label{sec. parametrization}

Next, this section specifies the distribution parametrization and function parametrization.

Following Section \ref{sec. learnability}, we want to use the potential model parametrization $P_f = e^f/\int_{\Sigma^{\omega}} e^f$, but the denominator may not be well-defined.
So a slight detour is taken based on the following proposition, which shows that there is a correspondence between $\PS(\Sigma^{\omega})$ and the next-token distributions.

\begin{proposition}
\label{prop. from next-token to infinity}
Denote by $\PS_{+1}$ the set of functions $P_{+1}\colon \Sigma^*\to\PS(\Sigma)$, and denote the function values by $P_{+1}(\cdot|x)$ for all $x\in\Sigma^*$.
There exists a surjective map $\mathfrak{I}\colon \PS_{+1}\to\PS(\Sigma^{\omega})$ such that for any $P\in\PS(\Sigma^{\omega})$ and any $P_{+1} \in \mathfrak{I}^{-1}(P)$,
\begin{equation*}
P^{\leq t+1}(\cdot|x) = P_{+1}(\cdot|x)
\end{equation*}
for any $t\in\N$ and $x\in\Sigma^t$ such that $P^{\leq t}(x)>0$.
\end{proposition}

Proposition \ref{prop. from next-token to infinity} indicates that it suffices to define the next-token distributions $P^{\leq t+1}(\cdot|x)$ in order to define any $P \in \PS(\Sigma^{\omega})$.
So we parametrize the former by potential model.
Denote $\R_-=\R\cup\{-\infty\}$.
Given any function $f\colon \Sigma^*\to\R_-^{|\Sigma|}$ whose output vectors are not uniformly $-\infty$, we set
\begin{align}
\label{eq. softmax conditional}
\begin{split}
\forall x \in \Sigma^*, \quad P_f^{+1}(\cdot|x) &= \text{softmax}(f(x)) \\
P_f &= \mathfrak{I}(P_f^{+1}) \in \PS(\Sigma^{\omega})
\end{split}
\end{align}
This settles the distribution parametrization.
Typically, $f(x)$ is called the logit vector.


While the proofs in later sections should be applicable to bounded-depth polynomial-size circuits in general,
for concreteness the function parametrization is set to the Transformer networks \cite{vaswani2017attention}.

\begin{definition}
\label{def. TF}
Denote by $\TF(L,H,d_h,p)$ the set of Transformer networks with $L$ blocks, $H$ heads, head dimension $d_h$, and floating-point precision of $p$ bits.
\end{definition}

\noindent
Regarding the details,
\begin{itemize}

\item The need to compute $f(\varnothing)$ in (\ref{eq. softmax conditional}) gives rise to the ``beginning-of-sentence" token $\bos$ and the input sequence becomes $(\bos,x_1,\dots)$.
Typically we do not write out $\bos$ and assume that the Transformer inserts it automatically.

\item The vocabulary of the Transformer is set to $\Sigma \cup \{\bos\}$.
So the embedding size is $|\Sigma|+1$, and output size is $|\Sigma|$.
Later, when we switch to latent sequences over some finite set $\Omega$, the vocabulary will be assumed to be $\Omega \cup \{\bos\}$.

\item Denote by $\R_p$ the set of $p$-bits numbers and $\text{cast}_p \colon \R\cup\{\pm\infty\} \to \R_p$ the nearest casting.
Any floating-point scheme is acceptable, as long as $\text{cast}_p$ converges pointwise to the identity function on $\R\cup\{\pm\infty\}$ as $p\to\infty$.

\item In practice, the self-attention of Transformers is usually causal (or at least block-wise causal \cite{nie2025llada}), so that the computation of (\ref{eq. softmax conditional}) over increasing $t$ can be reused.
Nevertheless, our separation theorems hold regardless of whether the attention is causal.



\item The remaining hyperparameters do not affect our proofs.
For instance, the MLP layers can be gated or not \cite{shazeer2020glu}, use any activation function, and have any fixed width ratio (usually $2\sim4$).
The position encoding can be any popular choice \cite{raffel2020exploring,su2024roformer,press2021train} or none.
The layernorms can be prenorm, postnorm or none \cite{nguyen2019prenorm}.
\end{itemize}
A notational nuance is that in machine learning papers $f(x)$ is usually the $|x|\times|\Sigma|$ matrix that includes the logit vector at each token of $x$,
whereas in this paper $f(x)$ corresponds to the logit vector at the last token.

\begin{remark}
\label{remark. precision}
There are some interesting results regarding the floating-point precision $p$.
For instance, if $p$ is constant, as in Definition \ref{def. TF}, then by \cite[Theorem 3.1]{li2024chain} every Transformer $f \in \TF(L,H,d_h,p)$ can be modeled by an $\AC^0$ circuit family (cf.\@ Appendix \ref{appendix. circuit complexity} for the definitions), which has very limited expressivity and cannot solve simple problems such as \textsf{MAJORITY} (deciding if a sequence $\in\{0,1\}^*$ has more $1$'s than $0$'s) \cite[Corollary 3.34]{vollmer1999introduction}.
Meanwhile, if $p$ is allowed to grow with the input length, specifically $p=O(\poly(|x|))$, then by \cite[Theorem 13]{chiang2025TC0} Transformers can be modeled by $\TC^0$ circuit families (cf.\@ Appendix \ref{appendix. circuit complexity}), which are slightly more expressive than $\AC^0$ and can solve \textsf{MAJORITY}.
We will treat Transformers as $\TC^0$ circuits when upper bounding their expressivity (Theorems \ref{thm. separation I} and \ref{thm. separation III}), despite that we are using constant precision $p$ for simplicity.
The motivation for this overkill is that, as discussed in Section \ref{sec. loss}, we consider only the logical errors instead of numerical errors, in the hope that our separation theorems can reveal the intrinsic limitations of Transformers and fast thinking.
\end{remark}

\subsection{Separation, Part I}
\label{sec. separation I}

This section establishes the first part of the separation theorem, that models with bounded circuit depth cannot approximate many distributions that are apparently simple.

The target distributions are set to be hidden Markov models (HMM).
One reason is that HMMs are a simple family, so if a model cannot fit HMMs, then it must be very poor.
Another reason is that HMMs are not too far from realistic scenarios, as classical physics is modeled by memory-less differential equations, and one can imagine an HMM whose hidden states represent the mental states of human speakers and whose observable states are speech segments.

\begin{definition}[HMM]
\label{def. HMM}
Given two finite sets $\Omega$ and $\Sigma$,
a hidden Markov model (HMM) is a tuple $(A,B,\lambda)$, where $A\in\R^{|\Omega|\times |\Omega|}$ and $B\in\R^{|\Omega|\times |\Sigma|}$ are stochastic matrices (i.e.\@ each row $A_i$ or $B_j$ is a probability vector) and $\lambda \in \R^{|\Omega|}$ is a probability vector.
For $t=1,2,\dots$, define the random variables $Z_t$ that range in $\Omega$, and $X_t$ that range in $\Sigma$.
All of them are drawn independently:
$Z_1\sim \lambda$, $Z_{t+1}\sim A_{Z_t}$, $X_t \sim B_{Z_t}$.
Denote by $P_{A,B,\lambda} \in \PS(\Sigma^{\omega})$ the distribution of the observable sequence
\begin{equation*}
P_{A,B,\lambda} = \law\big((X_t)_{t=1}^{\infty}\big)
\end{equation*}
\end{definition}

Despite that the latent variables of HMMs are memory-less, the observable variables are not, and their dependency can have arbitrarily long range, which leads to the following theorem.
One technicality is that the theorem assumes that $\TC^0 \subsetneq \NC^1$, the two sets being ``problems" that can be decided by two classes of Boolean circuits (cf.\@ Appendix \ref{appendix. circuit complexity} for definitions).
Similar to the famous conjecture $\mathbf{P} \subsetneq \mathbf{NP}$, this conjecture is widely believed to be true in the theory of computing \cite{yao1989circuits,allender2010amplifying,strobl2024formal}.

\begin{theorem}
\label{thm. separation I}
Assume that $\TC^0 \subsetneq \NC^1$.
There exists an HMM $(A,B,\lambda)$ that no Transformer can approximate.
Specially, for any configuration $(L,H,d_h,p)$, the semi-divergence (\ref{eq. semi-divergence}) cannot be reduced to zero
\begin{equation}
\label{eq. separation 1}
\inf_{f\in \TF(L,H,d_h,p)} D(P_{A,B,\lambda}, P_f) > 0
\end{equation}
\end{theorem}


\begin{figure}[!ht]
\centering
\includegraphics[width=0.85\linewidth]{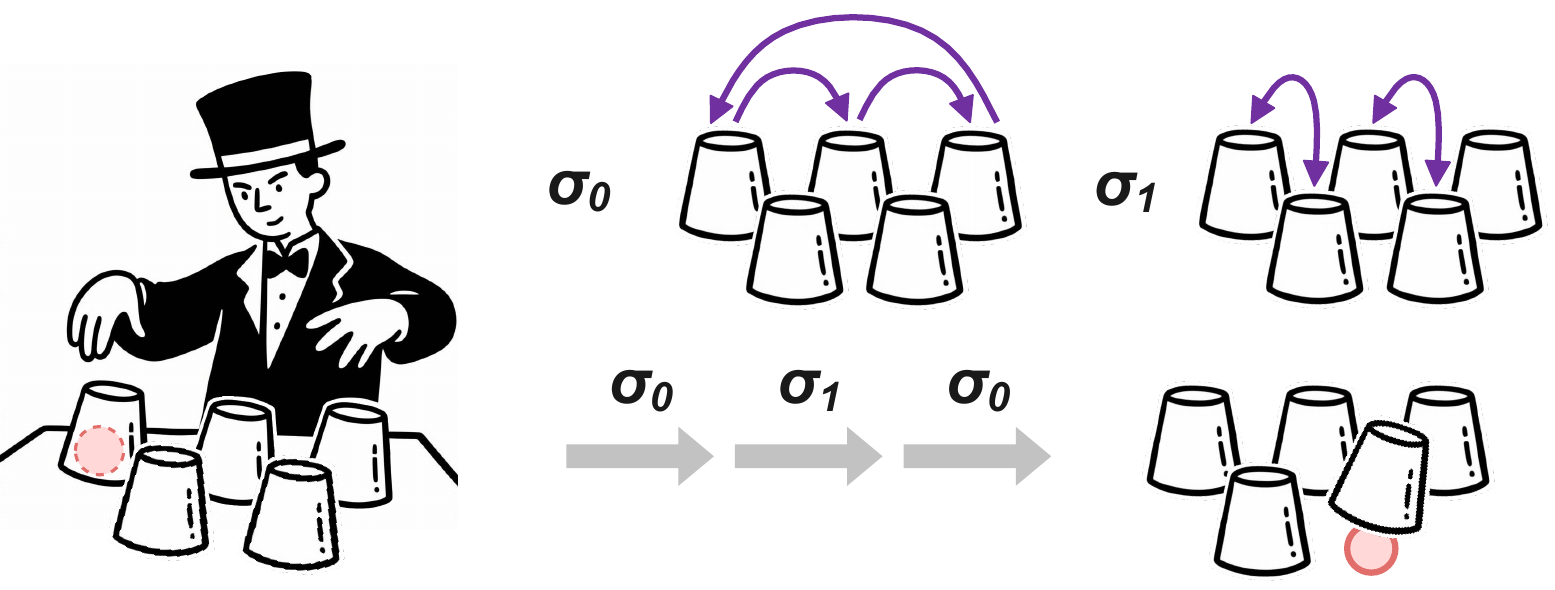}
\caption{Illustration of the toy function (\ref{eq. A5 composition function}) constructed in Appendix \ref{append. separation I}, which can be compared to the ``cups and ball" game.
A ball is put under the first cup in the beginning, and we try to locate it after some arbitrary swappings are applied.
By Lemma \ref{lemma. A5 is NC1}, if $\TC^0 \subsetneq \NC^1$, then there is no way to compute this function ``fully in parallel".
The proof of Theorem \ref{thm. separation I} is to embed this function into an HMM.}
\label{fig: cups and ball}
\end{figure}

Our proof is based on a toy function (\ref{eq. A5 composition function}) that resembles the ``cups and ball" game illustrated in Figure \ref{fig: cups and ball}.
Informally, the assumption $\TC^0 \subsetneq \NC^1$ (and Lemma \ref{lemma. A5 is NC1}) means that the number of steps required for solving the game must grow with the amount of swappings,
if any realistic algorithm is used.
This is intuitive as the most human-like approach is to keep track of the location of the ball one step after another, thus taking $O(t)$ steps.
(Even if using divide-and-conquer, we still need $O(\log t)$ steps).
Such latent processing is common in real life, as one maintains (and keeps updating) an estimation of the others' mental states during a conversation,
and maintains an understanding of a book while reading it.
It seems impossible that these latent processes can always be performed within a constant number of steps, since they appear inherently sequential;
when studying ``Real Analysis" for instance, one needs to first understand ``measureable sets" and then learn the ``measureable functions", ``Lebesgue integral" and so on, and there is no way to skip steps.
(Divide-and-conquer does not seem to work for these examples either).

More concretely, one can try to plot the computations performed by an algorithm on an input sequence,
with all variables depicted as nodes and their dependencies as (directed) edges, thus forming a (directed acyclic) graph.
For instance, the graph of ``tracking the ball step by step" is a chain of length $t$, and the graph of ``divide-and-conquer" is a tree of depth $\lceil \log_2 t \rceil$.
With longer input sequences, the graphs become bigger, which can be measured by depth (the length of the longest directed path in the graph) and size (the number of nodes).
In some sense, depth concerns sequential computation, while width (size divided by depth) concerns parallel computation.
So one promising way to say that a task is inherently sequential is that, no matter what algorithm is used, the depths of its graphs are always unbounded as the input grows.
A technical issue, though, is that there is a degenerate way to achieve bounded depth, by keeping a lookup table that maps every possible input to the correct output,
e.g.\@ recording all $2^t$ input-output pairs of the toy function (\ref{eq. A5 composition function}),
and thus reducing the depth to 3 (input-lookup-output).
This is degenerate since these exponential-size  graphs are unrealistic for implementation.
Thus, the correct characterization of sequentiality is that, no matter what algorithm is used, the depths of its graphs are always unbounded, as long as their sizes are polynomial in $t$.
Equivalently, if any algorithm achieves bounded depths, the sizes must be superpolynomial.
Basically, this is the intuition behind the assumption $\TC^0 \subsetneq \NC^1$ and Appendix \ref{appendix. circuit complexity}.

Meanwhile, if we plot the computations inside a Transformer with each hidden activation being a node, then the resulting graphs have depths bounded by $O(L)$ (constant in $t$) and sizes bounded by $O(L H d_h t^2)$ (polynomial in $t$).
In particular, the attention mechanism that enables token interactions is merely parallel computation.
As a result, Transformers cannot model inherently sequential functions.
This is the intuition behind Theorem \ref{thm. separation I}.
The formal statement is that every Transformer $f \in \TF(L,H,d_h,p)$ can be modeled as a \textsf{DLOGTIME}-uniform $\TC^0$ circuit family $\{C_t\}_{t=0}^{\infty}$ \cite{chiang2025TC0} (cf.\@ Appendix \ref{appendix. circuit complexity}).
Each $C_t$ is a Boolean circuit that implements $f$ over length-$t$ inputs (in binary), producing the same output logits (in binary).
$\TC^0$ captures bounded depth, while \textsf{DLOGTIME}-uniformity is a condition stronger than polynomial size.
One detail is that, based on Definition \ref{def. TF}, this limitation applies to Transformers of various architectures, e.g.\@ regardless of whether the attention is causal or bi-directional, which resolves the last remark in Example \ref{ex. night heron}.

\begin{remark}
The limited expressivity of Transformers has been extensively studied in various works \cite{merrill2022saturated,liu2022transformers,strobl2023averagehard,strobl2024formal,li2024chain,chiang2025TC0} (also cf.\@ Remark \ref{remark. precision}),
and Theorem \ref{thm. separation I} applies these results to show that Transformers are inadequate for modeling general HMMs.
The toy function (\ref{eq. A5 composition function}) is based on the symmetry group $S_5$, which is used in both the classical Barrington's theorem that characterizes $\NC^1$ \cite{barrington1986bounded} and recent analyses of Transformers that adopt this theorem \cite{li2024chain,strobl2024formal}.
\end{remark}


\begin{remark}
A more general heuristic underlying our construction is function composition, as the toy function (\ref{eq. A5 composition function}) is simply a composition of permutations.
An interesting conflict is that while compositionality is an important feature of natural languages \cite{donatelli2023compositionality,peng2024limitations}, plain Transformers are known to have difficulty in solving compositional tasks \cite{dziri2023faith}.
It seems plausible that function compositions in general must be modeled by sequential instead of parallel computation, such that the depth complexity of composed functions grows linearly in the number of compositions;
however, this is currently an open conjecture in the theory of computation \cite{karchmer1995super,gavinsky2017toward}.
\end{remark}


\subsection{Projection}
\label{sec. projection}

To prepare for our separation theorems, this section defines projections, which help to reduce the norm $\|P_*\|_{\F}$ and increase the learnability of $P_*$.

\subsubsection{Motivation}

Generally speaking, suppose there are some arbitrary measurable spaces $\X$ and $\Z$ (the spaces of observable and latent data) and a measureable function $\proj\colon \Z\to\X$.
We can consider an alternative parametrization
\begin{equation}
\label{eq. projection parametrization}
f\mapsto \proj\#P_f
\end{equation}
and a new norm can be defined:
\begin{align}
\label{eq. projection norm}
\|P\|_{\F,\proj} &:= \inf_{f\in\F} \big\{ \|f\|_{\F} \bigm| P=\proj\#P_f \big\}\\
\nonumber
&= \inf_{Q\in\PS(\Z)} \bigl\{ \|Q\|_{\F} \bigm| P=\proj\#Q \big\}
\end{align}

The parametrization (\ref{eq. projection parametrization}) could be much more expressive than the simple $f\mapsto P_f$, if $\Z$ is ``larger" than $\X$.
In that case, one could get $\|P_*\|_{\F,\proj} \ll \|P_*\|_{\F}$, which greatly improves the estimates (\ref{eq. train loss upper bound}, \ref{eq. test loss early stop}).
As an informal example, consider the random variables $(X_p,X_s,X_a)$ such that $X_p$ ranges among math competition problems, $X_a|X_p$ is the final answer, and $X_s|(X_p,X_a)$ is a well-written detailed solution.
Define the conditional distributions $Q=\law(X_s,X_a|X_p), P_*=\law(X_a|X_p)$, and the projection $\proj\colon (x_p,x_s,x_s)\mapsto(x_p,x_s)$.
Intuitively, it is much easier for LLMs, as well as humans, to model $(X_s,X_a)|X_p$ than $X_a|X_p$, so we have
\begin{equation}
\label{eq. norm separation}
\|P_*\|_{\F,\proj} \leq \|Q\|_{\F} \ll \|P_*\|_{\F}
\end{equation}
Later sections will provide more rigorous and stronger results that separate the expressivity of $P_f$ and $\proj\#P_f$.

Separation theorems have been a popular topic in the studies of the approximation ability of neural networks \cite{daniely2017depth,safran2019depth,bach2017breaking,wu2022spectral}.
It is often established that there exist some target functions $f_*$ such that deep networks can approximate them using polynomially many neurons ($\poly(d,1/\ep)$ with dimension $d$ and error $\ep$), whereas shallow networks cannot.
Informally, if we define $\F_l$ as the space of depth-$l$ neural networks, and the norm $\|f_*\|_{\F_l}$ as the minimum number of neurons sufficient for a $\F_l$ function to approximate $f_*$ up to error $\ep$,
then $\|f_*\|_{\F_m} \ll \|f_*\|_{\F_l}$ if $m>l$.
Thus, (\ref{eq. norm separation}) can be seen as a separation result.
In fact, our result will be stronger than depth separation.
With the help of $\proj$, even shallow networks can surpass networks with arbitrary depth.
It indicates that the parametrization $\proj\#P_f$ is more effective for our goal of reducing the norm of $P_*$.

\subsubsection{Continuous Projections}
\label{sec. continuous projection}

To begin with, we characterize the appropriate projections for sequence data.

The minimum requirement is that the projection is a Borel measureable partial function $\proj\colon \Omega^{\omega}\rightharpoonup\Sigma^{\omega}$.
Borel measureability is needed to make the pushforward $\proj\#P$ well-defined.
Meanwhile, by partial function we mean that the domain $\dom(\proj)$ is not necessarily all of the latent space $\Z$,
e.g.\@ if $\Z$ consists of sequences, then $\dom(\proj)$ could be sequences that obey some syntactical rules.
For simplicity, we write $\proj(S)=\proj\big(S\cap\dom(\proj)\big)$ for any subset $S \subseteq \Z$.

One basic requirement for $\proj$ is surjectivity, that it can map to anywhere in $\X$ and thus can model any distribution in $\PS(\X)$.
Specifically, Proposition \ref{prop. surjective projection} indicates that in some sense, a surjective $\proj$ is necessary and sufficient for ensuring that $\PS(\Sigma^{\omega}) = \{ \proj\#P \mid P \in \PS(\Omega^{\omega})\}$.

Since the observable data are infinite sequences  $\X=\Sigma^{\omega}$,
it is natural to define the latent space also by infinite sequences, $\Z=\Omega^{\omega}$ for any finite set $\Omega$.
Recall from Section \ref{sec. loss} that the likelihoods of any finite-length sequence $x \in \Sigma^*$ can be expressed in terms of the cylinders $[x]$
\begin{equation*}
(\proj\#P)^{\leq |x|}(x) = (\proj\#P)([x]) = P\big(\proj^{-1}([x])\big)
\end{equation*}
To simplify computation, we want the subset $\proj^{-1}([x])$ to have a simple form, perhaps by imposing some regularity on $\proj$.

It turns out that the simple property of continuity is sufficient for this purpose, as indicated by the following proposition.
We say that the partial function $\proj$ is continuous, if for any open set $O \subseteq \Sigma^{\omega}$, its preimage $\proj^{-1}(O)$ is relatively open in $\dom(\proj)$.
As usual, the product topologies of $\Omega^{\omega}$ and $\Sigma^{\omega}$ are used.

\begin{proposition}
\label{prop. inverse projection}
Let $\proj\colon\Omega^{\omega}\rightharpoonup\Sigma^{\omega}$ be any continuous surjective partial function whose domain $\dom(\proj)$ is closed.
For any $x\in\Sigma^*$, there exists at least one subset $I_x \subseteq \Omega^*$ such that
\begin{enumerate}
\item $I_x$ is prefix-free, non-empty and finite
\item $[z] \cap \dom(\proj) \neq \varnothing$ for every $z\in I_x$
\item the inverse $\proj^{-1}([x])$ can be expressed as a finite and disjoint union of cylinders
\begin{equation}
\label{eq. disjoint union of cylinders}
\proj^{-1}([x]) = \bigsqcup_{z\in I_x} [z] \cap \dom(\proj)
\end{equation}
\end{enumerate}
Moreover, among all $I_x$ that satisfies these properties, there exists a unique minimal set $I_x^*$ with respect to the partial order $\preceq$ defined by (\ref{eq. partial order}).
Denote the mapping $x\mapsto I_x^*$ by the set-valued function $\fip\colon\Sigma^*\to2^{\Omega^*}$.
Then, $\fip$ is monotone in the sense that
\begin{equation}
\label{eq. monotone}
\forall x,x'\in \Sigma^*, ~ x \sqsubseteq x' \to \fip(x) \preceq \fip(x')
\end{equation}
and $\fip$ converges pointwise to $\proj^{-1}$ in the sense of
\begin{equation*}
\forall x\in \Sigma^{\omega}, \quad \bigcap_{t=0}^{\infty} \bigsqcup_{z\in \fip(x_{\leq t})} [z] \cap \dom(\proj) = \proj^{-1}(x)
\end{equation*}
\end{proposition}

\begin{definition}[Lifting]
\label{def. lifting}
For any continuous surjective partial function $\proj\colon\Omega^{\omega}\rightharpoonup\Sigma^{\omega}$ with closed domain $\dom(\proj)$,
we refer to the function $\fip$ constructed in Proposition \ref{prop. inverse projection} as the ``lifting" defined by $\proj$.
\end{definition}

\begin{remark}[Causality]
\label{remark. causal}
Informally speaking, for any (partial) function $\proj\colon\Omega^{\omega}\rightharpoonup\Sigma^{\omega}$, being continuous is equivalent to being causal, in the sense that tokens ``at the back" of the input sequence do not affect tokens ``at the front" of the output sequence.
Since $\Omega^{\omega}$ is compact, a continuous $\proj$ is uniformly continuous, so for any $\ep>0$, there exists some $\delta>0$ such that any $z,z'\in\dom(\proj)$ with $d_{\omega}(z,z')<\delta$ satisfies $d_{\omega}\big(\proj(z),\proj(z')\big)<\ep$,
where $d_{\omega}$ is the metric (\ref{eq. metric product topology}).
Thus, for any $t$, there exists some $T$ such that if two input sequences differ only after position $T$, the output sequences must be the same before position $t$.
From this perspective, Proposition \ref{prop. inverse projection} means that a causal projection must have a tractable inverse.
\end{remark}

\begin{remark}[Injectivity]
\label{remark. injective}
The lifting $\fip$ is injective in the following sense:
For any inputs $x,x' \in \Sigma^*$ that are prefix-free (neither is the prefix of the other), the sets $\fip(x), \fip(x')$ are disjoint.
In particular, for any prefix $x\in\Sigma^*$, the sets $\fip(xa), a\in\Sigma$ are all disjoint, a property helpful for computing next-token probabilities.
However, if the inputs are not prefix-free, then there are counterexamples.
For instance, define
\begin{equation*}
\proj\colon\{a,b\}^{\omega}\to\{a,b\}^{\omega}, \quad \proj(z) = \begin{cases}
a^{\omega} ~\text{ if }~ z_1=a\\
z_{>1} ~\text{ else}
\end{cases}
\end{equation*}
where $a^{\omega}$ denotes the infinite sequence with all $a$, and $z_{>1}$ is $z$ with the first entry removed.
$\proj$ is a continuous surjective function, and $\fip(a^k) = \{ a, ba^k\}$, where $a^k$ denotes the all-$a$ sequence with length $k$.
So it always holds that $a \in \fip(a^k) \cap \fip(a^j) \neq \varnothing$.
As a result, in general $\fip$ is not the inverse of any (partial) function $\Omega^*\rightharpoonup\Sigma^*$.
\end{remark}

To facilitate the study of conditional probabilities, we make the following definition.
For any $x,y\in\Sigma^t$, monotonicity (\ref{eq. monotone}) implies that $\fip(x) \preceq \fip(xy)$, so for any $z \in \fip(xy)$, there exists a $z' \in \fip(x)$ such that $z' \sqsubseteq z$, and $z'$ is unique since $\fip(x)$ is prefix-free.
Thus, we can define the following set
\begin{equation}
\label{eq. next-segment set}
\nsm_x(z,y) := \big\{ z^{-1}z' \bigm| z' \in \proj^{-1}_{\omega}(xy), ~ z \sqsubseteq z' \big\}
\end{equation}
where $z^{-1}z'$ means stripping the prefix $z$ from $z'$.
It follows that
\begin{equation*}
\fip(xy) = \bigsqcup_{z\in\fip(x)} \big\{ zs \bigm| s \in \nsm_x(z,y) \big\}
\end{equation*}
Note that this set $\nsm_x(z,y)$ is prefix-free and could be empty.
It also includes $\fip$ as a special case, as $\nsm_{\varnothing}(\varnothing,x) = \fip(x)$.

\begin{definition}[Next-segment map $\nsm$]
\label{def. next-segment map}
For any continuous surjective partial function $\proj\colon\Omega^{\omega}\rightharpoonup\Sigma^{\omega}$ with closed domain,
we refer to the set-valued mapping $\nsm_x\colon\fip(x)\times\Sigma^*\to2^{\Omega^*}$ constructed in (\ref{eq. next-segment set}) as the next-segment map.
For simplicity, we write $\nsm$ when the sequence $x$ is clear from context.
\end{definition}

Examples of $\nsm$ will be given in the next section.
The following lemma summarizes the above discussion.

\begin{lemma}
\label{lemma. segments prefix-free}
Given any continuous surjective partial function $\proj\colon\Omega^{\omega}\rightharpoonup\Sigma^{\omega}$ with closed domain,
any $x\in\Sigma^*$ and any $z\in\fip(x)$,
\begin{enumerate}
\item For any $y\in\Sigma^*$, the set $\nsm_x(z,y)$ is finite and prefix-free.
    
\item For any prefix-free subset $S \subseteq \Sigma^*$,
the sets $\{\nsm_x(z,y) \mid y\in S\}$ are disjoint,
and their union $\bigsqcup_{y\in S}\nsm_x(z,y)$ is prefix-free.

\item For any decomposition of $x$ into segments, $x = x^{(1)}\dots x^{(k)}$ with $x^{(i)} \in \Sigma^*$, the sequence $z$ can be uniquely decomposed into
\begin{equation*}
z = z^{(1)} \dots z^{(k)}, \quad z^{(i)} \in \nsm_{x^{(<i)}}\big(z^{(<i)},x^{(i)}\big)
\end{equation*}
In particular, for the token-wise decomposition $x=x_1\dots x_{|x|}$, we have the unique decomposition
\begin{equation*}
z = z^{(1)} \dots z^{(|x|)}, \quad z^{(t)} \in \nsm_{x_{<t}}\big(z^{(<t)},x_t\big)
\end{equation*}

\item For any $w \in [z] \cap \dom(\proj)$, there exists a unique $s \in \bigsqcup_{a\in\Sigma}\nsm_x(z,a)$ such that $zs \sqsubseteq w$.
\end{enumerate}
\end{lemma}

The following definition is helpful for describing conditional distributions.

\begin{definition}[Next-segment distribution]
\label{def. next-segment distribution}
Given any $P \in \PS(\Omega^{\omega})$, any $z \in \Sigma^*$ such that $P([z])>0$, and any prefix-free subset $S \subseteq \Omega^*$, 
denote
\begin{equation*}
P(S|z) := P\Big(\bigsqcup_{s\in S} [zs] \Bigm| [z]\Big) = \sum_{s \in S} P^{\leq |z|+|s|}(s|z)
\end{equation*}
Furthermore, if given the sample $Z \sim P(\cdot|[z])$ there almost surely exists a unique $s \in S$ such that $zs \sqsubseteq Z$, then define the distribution $P(\cdot|z\to S) \in \PS(S)$ by
\begin{equation*}
\forall s\in S, \quad P(s \mid z\to S) := P(\{s\}|z)
\end{equation*}
\end{definition}

Given the condition of Lemma \ref{lemma. segments prefix-free}, its property 4 implies that, for any $P \in \PS(\dom(\proj))$ such that $P([z])>0$,
the next-segment distribution
\begin{equation*}
P\Big(\cdot\Big|z\to\bigsqcup_{a\in\Sigma}\nsm_x(z,a)\Big)
\end{equation*}
is always well-defined.
If we denote by $\proj_{x,z}(s) = a$ the mapping from each latent segment $s \in \bigsqcup_{a\in\Sigma}\nsm_x(z,a)$ to its corresponding token $a$, then the next-token distribution conditioned on the latent sequence $z$ can be expressed in two ways
\begin{equation}
\label{eq. next token condition on latent}
\forall a \in \Sigma, \quad P\big( \nsm_x(z,a) \big| z \big) = \Big( \proj_{x,z}\#P\big(\cdot\big|z\to\bigsqcup_{a'\in\Sigma}\nsm_x(z,a')\big) \Big)(a)
\end{equation}

As the inverse map $\proj^{-1}$ can be converted to its finite-length counterpart $\fip$,
we remark that $\proj$ itself can also be converted to a map $\proj^{<\omega}$ on finite-length sequences $\Omega^*$.
Denote $\Sigma^{\leq \omega} = \Sigma^* \cup \Sigma^{\omega}$,
and denote by $\text{gcd}(S)$ the longest common prefix of any nonempty subset $S\subseteq \Sigma^{\leq \omega}$.

\begin{definition}[Prefix projection]
\label{def. prefix projection}
Given any nonempty partial function $\proj\colon\Omega^{\omega}\rightharpoonup\Sigma^{\omega}$, define the partial function $\proj^{<\omega}\colon\Omega^*\to\Sigma^{\leq\omega}$ by
\begin{align*}
\dom(\proj^{<\omega}) &= \big\{ z \bigm| z \in \Omega^*, ~ [z] \cap \dom(\proj) \neq \varnothing \big\}\\
\proj^{<\omega}(z) &:= \text{gcd}\big(\proj([z])\big)
\end{align*}
\end{definition}


Note that the condition ``$[z] \cap \dom(\proj) \neq \varnothing$ for every $z\in I_x$" of Proposition \ref{prop. inverse projection} can be written more concisely as $I_x \subseteq \dom(\proj^{<\omega})$.
As expected, this prefix projection converges to $\proj$, as indicated by the following result.

\begin{lemma}
\label{lemma. prefix projection}
For any continuous partial function $\proj\colon\Omega^{\omega}\rightharpoonup\Sigma^{\omega}$ and any $z\in\dom(\proj)$,
the sequence $\bigcup_{t=1}^{\infty} \proj^{<\omega}(z_{\leq t}) = \proj(z)$.
\end{lemma}

Despite what the notations may suggest, the lifting $\fip$ is generally not the inverse of $\proj^{<\omega}$ (or the inverse of any function, as indicated by Remark \ref{remark. injective}).
Based on Remark \ref{remark. injective}, one can show that $\fip(x)$ is generally not contained in $(\proj^{<\omega})^{-1}(x)$.
Based on Example \ref{ex. think} in the next section, one can show that $(\proj^{<\omega})^{-1}(x)$ is generally not contained in $\fip(x)$.

\subsubsection{Projected Distributions}

Now we can describe the marginal and conditional distributions of the observable distribution $\proj\#P$ in terms of the latent distribution $P$.

The following theorem implies that the next-token (more generally next-segment) probabilities of $\proj\#P$ are weighted averages of the next-segment probabilities of $P$.
The weights $Q_*$ is the posterior distribution of latent sequences given a prefix $x$.

\begin{theorem}
\label{thm. conditional latent distribution}
Let $\proj\colon\Omega^{\omega}\rightharpoonup\Sigma^{\omega}$ be any continuous surjective partial function with closed domain and $P\in\PS(\dom(\proj))$ be any distribution.
Let $\fip$ be the lifting and $\nsm$ be the next-segment map.
Then, for any $x \in \Sigma^*$,
\begin{equation}
\label{eq. latent marginal}
(\proj\#P)^{\leq |x|}(x) = \sum_{z\in \fip(x)} P^{\leq |z|}(z)\\
\end{equation}
and for any $x,y\in\Sigma^*$ such that $(\proj\#P)^{\leq |x|}(x)>0$,
\begin{equation}
\label{eq. latent conditional}
(\proj\#P)^{\leq |x|+|y|}(y|x) = \int P\big(\nsm(z,y)\big|z\big) \text{d}Q_*(z|x)
\end{equation}
where $Q_*(\cdot|x)$ is the restriction of $P$ to $\fip(x)$
\begin{equation}
\label{eq. latent posterior}
\forall z \in \fip(x), \quad Q_*(z|x) := \frac{P^{\leq |z|}(z)}{(\proj\#P)^{\leq |x|}(x)}
\end{equation}
\end{theorem}

Recall that the classical autoregressive decomposition is computed as follows:
For any $P\in\PS(\Sigma^{\omega})$ and any $x \in \Sigma^*$,
\begin{equation}
\label{eq. classical autoregression}
P^{\leq |x|}(x) = P^{<|x|}(x_{<|x|}) P^{\leq |x|}(x_{|x|} \mid x_{<|x|}) = \prod_{t=1}^{|x|} P^{\leq t}(x_t|x_{<t})
\end{equation}
With the parametrization (\ref{eq. softmax conditional}) and causal Transformers (Definition \ref{def. TF}), the computation of $f(x_1, \dots x_t)$ can be performed incrementally over $t=1,\dots,|x|$ to save computation (i.e.\@ key-value cache reuse).
Similarly, Theorem \ref{thm. conditional latent distribution} implies that, the distribution $\proj\#P$ can be computed incrementally through the segments $s \in \nsm_{x_{< t}}(z,x_t)$ for $t=1,\dots T$ and $z \in \fip(x_{< t})$.
Namely, we can identify each latent sequence $z \in \fip(x)$ with an array of $|x|$ segments
\begin{align*}
z = z^{(1)}\dots z^{(|x|)}, \quad z^{(t)} \in \nsm_{x_{<t}}(z^{(<t)}, x_t)
\end{align*}
and compute the likelihood autoregressively
\begin{equation*}
(\proj\#P)^{\leq|x|}(x) = \sum_{z\in\fip(x)} \prod_{t=1}^{|x|} P \Big( z^{(t)} \Bigm| z^{(<t)} \to \bigsqcup_{a\in\Sigma} \nsm_{x_{<t}}(z^{(<t)},a) \Big)
\end{equation*}

Here are some examples of projections (that are continuous surjective partial functions with closed domains).

\begin{example}[Identity]
\label{ex. identity}
One trivial projection is the identity function $\id\colon\Sigma^{\omega}\to\Sigma^{\omega}$.
In this case, $\fip$ is simply the identity $x\mapsto \{x\}$,
and the next-segment map is $\nsm_x(z,a) \equiv \{a\}$.
Then, the next-token probability $(\proj\#P)^{\leq t}(x_t|x_{<t})$ simplifies to the classical autoregression (\ref{eq. classical autoregression}).
\end{example}

\begin{example}[Product space]
\label{ex. product}
Let the latent vocabulary be $\Omega = \Sigma \times \Gamma$ for some arbitrary finite set $\Gamma$.
Denote each $z \in \Omega^{\omega}$ by $z=((x_t,y_t))_{t=1}^{\infty}$.
Define the projection by $\proj\colon ((x_t,y_t))_{t=1}^{\infty} \mapsto (x_t)_{t=1}^{\infty}$.
Its domain $\dom(\proj)=\Omega^{\omega}$ is closed.
The lifting ranges over all of $\Gamma^{|x|}$
\begin{align*}
\forall x\in\Sigma^*, \quad &\fip(x) = \big\{((x_t,y_t))_{t=1}^{|x|} \bigm| y_1,\dots y_{|x|} \in \Gamma \big\}\\
\forall z \in \fip(x), ~ \forall a\in\Sigma, \quad &\nsm(z, a) = \big\{ (a,y) \bigm| y \in \Gamma \big\}
\end{align*}
The probabilities (\ref{eq. latent conditional}) can be computed recursively
\begin{align*}
(\proj\#P)^{\leq 1}(x_1) &= \sum_{y_1 \in \Gamma} P^{\leq 1}((x_1,y_1)) \\
(\proj\#P)^{\leq t}(x_t|x_{<t}) &= \sum_{y_{<t} \in \Gamma^{t-1}}  P^{<t}(y_{<t}|x_{<t}) \sum_{y_t \in \Gamma} P^{\leq t}\big((x_t,y_t)\big|((x_i,y_i))_{i<t}\big) \\
\text{where} ~ P^{<t}(y_{<t}|x_{<t}) &= \frac{P^{<t}\big(((x_i,y_i))_{i<t}\big)}{\sum_{y_{<t}' \in \Gamma^{t-1}} P^{<t}\big(((x_i,y_i'))_{i<t}\big)} \\
\text{where} ~ P^{\leq t}\big(((x_i,y_i))_{i\leq t}\big) &= P^{\leq t}\big( (x_t,y_t) \big| ((x_i,y_i))_{i<t}\big) P^{<t}\big(((x_i,y_i))_{i< t}\big)
\end{align*}
such that the terms $P^{\leq t}\big((x_t,y_t)\big|((x_i,y_i))_{i<t}\big)$ are produced and the terms $P^{\leq t}\big(((x_i,y_i))_{i\leq t}\big)$ are cached at each step.
\end{example}

\begin{figure}[ht]
\centering
\includegraphics[width=0.7\linewidth]{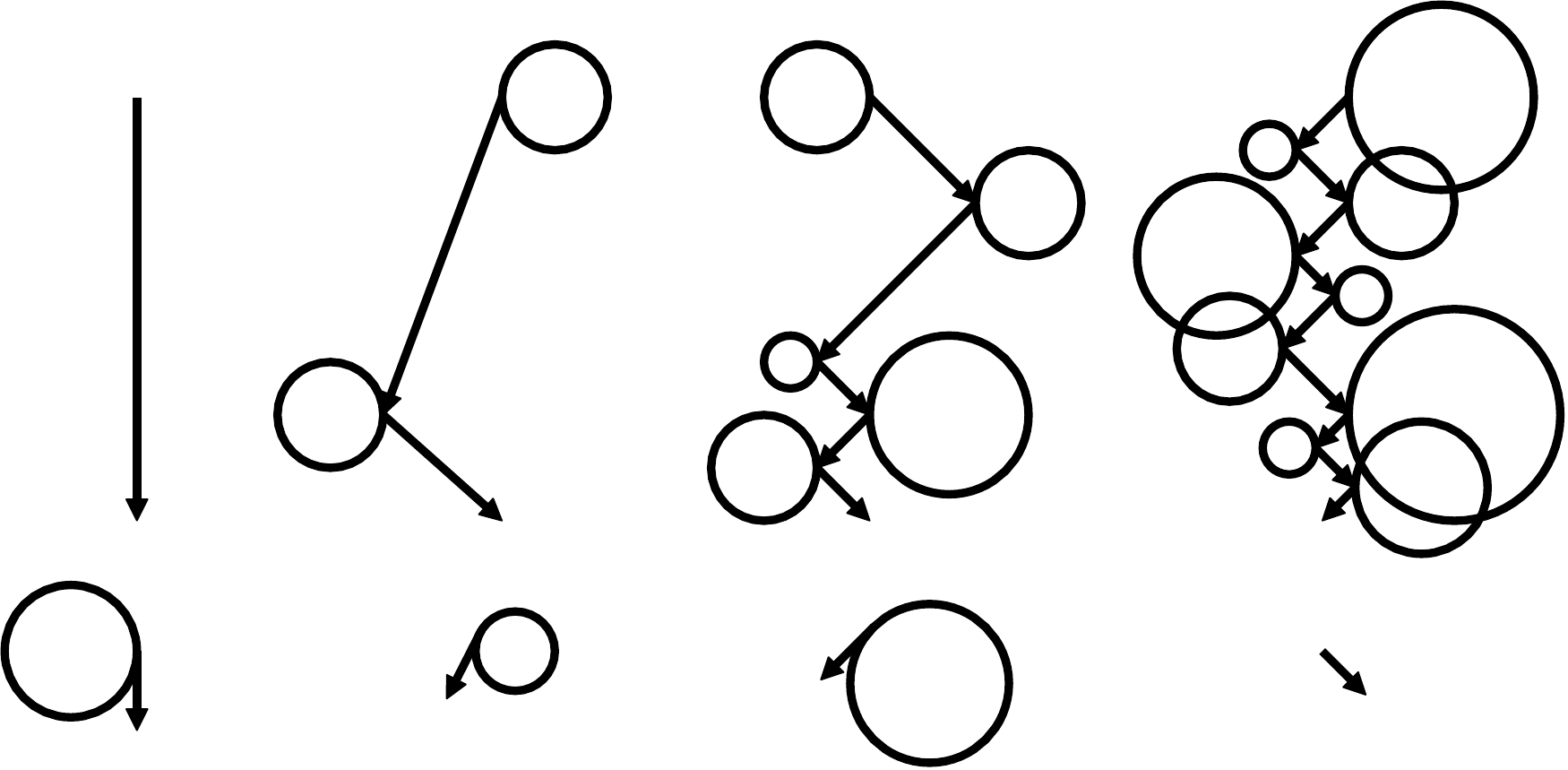}
\caption{Illustration of the lifting $\fip$ and next-segment map $\nsm$ of Example \ref{ex. think}.
Top: The latent sequences $z=(y^{(t)} x_t)_{t=1}^{|x|}\in\fip(x)$ of an observable sequence $x$ may contain ``thoughts" $y^{(t)}$ with variable lengths.
The bubbles represent nonempty thoughts $y^{(t)}$, while the straight lines are segments of $x$.
Bottom: Each segment $s \in \nsm(z,a)$ may also contain a variable-length thought.}
\label{fig: path integral}
\end{figure}

\begin{example}[Pause to think]
\label{ex. think}
Let $\sot$ and $\eot$ be the ``start-of-thought" and ``end-of-thought" symbols.
Set $\Omega = \Sigma \cup \{\sot,\eot\}$.
Define
\begin{align*}
\dom(\proj) &= \big\{ (y^{(t)} x_t)_{t=1}^{\infty} \bigm| x_t\in\Sigma ~~\text{and}~~ y^{(t)} \in \mathcal{T}_{c_t} \big\} \\
\mathcal{T}_c &= \{\varnothing\} \cup \{ \sot y \eot \mid y \in\Sigma^*, ~1\leq |y|\leq c\}
\end{align*}
where $\{c_t\}_{t=1}^{\infty}$ are arbitrary positive constants that upper bound the length of the ``thoughts" $y^{(t)}$.
Define the projection by stripping the ``thoughts"
\begin{equation*}
\proj\colon (y^{(t)} x_t)_{t=1}^{\infty} \mapsto (x_t)_{t=1}^{\infty}
\end{equation*}
Thus, $\proj$ is a surjective partial function.
Proposition \ref{prop. segment-wise partial function} below indicates that $\dom(\proj)$ is closed and $\proj$ is continuous.
Its lifting is simply
\begin{equation*}
\forall x\in\Sigma^*, \quad \fip(x) = \big\{(y^{(t)}x_t)_{t=1}^{|x|} \bigm| y^{(t)} \in \mathcal{T}_{c_t} \big\}
\end{equation*}
such that a nonempty ``thought" may be inserted before each token.
The next segments happen to be independent of $z \in \fip(x)$,
\begin{equation*}
\forall a \in \Sigma, \quad \nsm_x\big( z, a\big) = \big\{ y a \bigm| y \in \mathcal{T}_{c_{|x|+1}} \big\}
\end{equation*}

The following result is a general criterion for the continuity of $\proj$ and closedness of $\dom(\proj)$.
Recall from Remark \ref{remark. causal} that continuity is equivalent to ``causality".
So intuitively if each entry of $\proj(z)$ depends only on ``local" information, then $\proj$ should be continuous.

\begin{proposition}
\label{prop. segment-wise partial function}
Suppose there are nonempty subsets $\{S_t\}_{t=1}^{\infty}$ of $\Omega^*$ and finite constants $\{l_t\}_{t=1}^{\infty} \subseteq \N$,
such that for each $t$, each $s\in S_t$ satisfies $1\leq |s| \leq l_t$.
Then, $S := \{(s_t)_{t=1}^{\infty} \mid s_t \in S_t \}$ is a closed subset of $\Omega^{\omega}$ with respect to the product topology.
Moreover, for each $t$, let $f_t\colon S_t\to\Sigma^{\geq 1}$ be any mapping.
Then, the following is a continuous partial function $\Omega^{\omega}\rightharpoonup\Sigma^{\omega}$
\begin{equation*}
\dom(\proj) = S, \quad \proj\colon (s_t)_{t=1}^{\infty} \mapsto \big(f_t(s_t)\big)_{t=1}^{\infty}
\end{equation*}
\end{proposition}

An illustration of the functions $\fip$ and $\nsm$ of Example \ref{ex. think} is provided by Figure \ref{fig: path integral}.
This $\proj$ includes the format used by reasoning models such as DeepSeek-R1 \cite{guo2025deepseek} and Quiet-STaR \cite{zelikman2024quietstar}.
Specifically, Quiet-STaR fixes the thought lengths $|y^{(t)}|$ to 16,
while DeepSeek-R1 restricts $\sprt P$ to latent sequences whose thoughts $y^{(t)}$ are non-empty only when $x_t$ is the last token of each user query (more details are given in Section \ref{sec. deepseek representation}).
\end{example}

\subsubsection{Simple Projections}

Finally, we restrict the complexity of the projection.
The motivation is that the parametrization (\ref{eq. projection parametrization}) may admit degenerate solutions, e.g.\@ $P_f$ could be the uniform distribution over $\Omega^{\omega}$ that merely serves as the source of randomness for a sophisticated random variable $\proj$ that produces natural language texts.
Instead, we want to show that even with very simple projections, the parametrization $\proj\#P_f$ can easily surpass the plain parametrization $P_f$ in terms of approximation ability.

One natural option is to require that the lifting $\fip$ be circuits with low complexity.

\begin{definition}[Simple projection]
\label{def. simple projection}
Given a continuous surjective partial function $\proj\colon\Omega^{\omega}\rightharpoonup\Sigma^{\omega}$ with closed domain,
we call $\proj$ a \textit{simple projection} if its lifting $\fip$ is a \textsf{DLOGTIME}-uniform $\textbf{TC}^0$ nondeterministic circuit family with variable-length output,
and if the domain of its prefix map $\dom(\proj^{<\omega}) \subseteq \Omega^*$ can be decided by a \textsf{DLOGTIME}-uniform $\textbf{TC}^0$ circuit family.
\end{definition}

The definition of these concepts are provided in Appendix \ref{appendix. circuit complexity}.
Informally speaking, nondeterminism allows set-valued outputs, necessary for modeling $\fip\colon\Sigma^*\to 2^{\Omega^*}$.
Since the latent sequences $z\in \fip(x)$ are of variable lengths (both within the set $\fip(x)$ and for different inputs $x \in \Sigma^t$), the nondeterministic circuits for $\fip$ need to support variable-length outputs.

Meanwhile, the decidability of $\dom(\proj^{<\omega})$ means that we can easily check if any $z \in \Omega^*$ belongs to $\dom(\proj^{<\omega})$.
By Definition \ref{def. prefix projection}, this is equivalent to deciding whether $[z] \cap \dom(\proj) = \varnothing$.
This property allows us to compute the set of admissible next-tokens for $z$
\begin{align}
\label{eq. admissible next-token}
\nsm^{\leq 1}(z) &:= \{a \mid a\in\Omega, ~[za] \cap \dom(\proj) \neq \varnothing \} \\
\nonumber
&= \{a \mid a\in\Omega, ~za \in \dom(\proj^{<\omega}) \}
\end{align}
Specifically, if $\proj$ is a simple projection, then there exists a a \textsf{DLOGTIME}-uniform $\textbf{TC}^0$ circuit family $\{C_t\}_{t=1}^{\infty}$ such that $C_t(z,a) = 1$ if $a \in \nsm^{\leq 1}(z)$ and $0$ otherwise.
The next-token sets are useful for building autoregressive models over $\dom(\proj)$.

It is straightforward to check that the projections of Examples \ref{ex. identity} and \ref{ex. product} are simple projections.
Regarding Example \ref{ex. think}, if the length bounds $c_t$ are given by a polynomial, then $\proj$ is also a simple projection.
One possible implementation of $\fip$ is that the circuit $C_T$ places a thought $\sot y^{(t)} \eot$ before each input token $x_t$, for each $t=1,\dots T$.
These $y^{(t)}$ are specified by the auxiliary input bits and could range over all of $\big(\Sigma\cup\{\blank\}\big)^{c_t}$.
Then, $C_T$ identifies the $y^{(t)}$'s that are all $\blank$ and sets their $\sot,\eot$ to $\blank$.
Similarly, an implementation of circuits that decide $\dom(\proj^{<\omega})$ is to check,
given any $z\in\Omega^*$, whether its $\sot,\eot$ symbols appear alternately,
and if each nonempty thought $y^{(t)}$ has length $\leq c_t$.

Definition \ref{def. simple projection} seems to provide just the right amount of expressivity.
On one hand, \textsf{DLOGTIME}-uniform $\textbf{TC}^0$ circuits are very weak, e.g.\@ simple operations such as integer multiplication and iterated addition are $\textbf{TC}^0$-complete \cite{chandra1984constant,jerabek2012root}, and thus can be implemented in realistic settings.
On the other hand, we will see in Theorem \ref{thm. separation II} that $\proj\#P_f$ with simple projections is strictly more expressive than plain $P_f$.


Finally, we consider a technical issue.
Recall that our distribution parametrization is defined by conditional softmax (\ref{eq. softmax conditional}).
If the parameter function $f$ always outputs real value (instead of $-\infty$), then $\sprt P_f = \Omega^{\omega}$.
Instead, if we want to enforce that $\sprt P_f \subseteq \dom(\proj)$ when $\dom(\proj)$ is a proper subset of $\Omega^{\omega}$, then some entries of the output logits $f(x) \in \R^{|\Omega|}$ need to be masked (set to $-\infty$).

\begin{definition}[Distribution parametrization with prescribed support]
\label{def. softmax with masking}
Given any function $f\colon \Omega^*\to\R$ and partial function $\proj\colon \Omega^{\omega}\rightharpoonup\Sigma^{\omega}$ with nonempty closed domain $\dom(\proj)$,
define the distribution $P_f \in \PS(\dom(\proj))$ by
\begin{align*}
P_f = \mathfrak{I}(P_f^{+1}), \quad \forall z \in \Omega^*, ~ P_f^{+1}(\cdot|z) = \text{softmax}(\tilde{f}(z)), \quad \forall a \in \Omega, ~ \tilde{f}(z)_a = \begin{cases}
f(z)_a, ~ \text{if} ~ a \in \nsm^{\leq 1}(z) \\
-\infty, ~\text{else}
\end{cases}
\end{align*}
It follows that $\sprt P_f \subseteq \dom(\proj)$.
Whenever we write $P_f \in \dom(\proj)$ or $\proj\#P_f$, we assume that this parametrization is used.
\end{definition}

With a simple projection, this masking can be easily performed.
Such modification of the output logits during LLM inference is common in practice \cite{huggingface2026generationconfig,hewitt2022truncation,meister2023locally}, not to mention the more sophisticated operations \cite{keskar2019ctrl,shi2024thorough}.

Hence, in the static setting with a prescribed lifting function, Problem \ref{problem. intro} can be formalized as follows.
A more general formulation will be provided in Section \ref{sec. incremental modeling}.

\begin{problem}[static]
\label{problem. static}
Given a target distribution $P_* \in \PS(\Sigma^{\omega})$ and a simple projection $\proj\colon\Omega^{\omega}\rightharpoonup\Sigma^{\omega}$,
find a latent distribution $P\in\PS(\Omega^{\omega})$ such that $\proj\#P=P_*$ and $P=P_f$ for some finite numbers $L,H,d_h,p$ and some $f \in \TF(L,H,d_h,p)$.
\end{problem}

\subsection{Separation, Part II}
\label{sec. separation II}

This section establishes the second half of our separation theorem, that HMMs can be easily modeled with the help of projections.

\begin{theorem}
\label{thm. separation II}
For any HMM $(A,B,\lambda)$, there exist a finite set $\Omega$, a simple projection $\proj$ (cf. Definition \ref{def. simple projection}) and constants $d_h^*, p^*$ such that, for any $d_h \geq d_h^*, p\geq p_*$,
\begin{equation*}
\min_{f \in \TF(0,1,d_h,p)}  D(P_{A,B,\lambda}, \proj\#P_f)  = 0
\end{equation*}
In particular, the minimizer $f$ exists.
\end{theorem}

\begin{remark}[Context dependency vs.\@ representation size]
\label{remark. context dependency representation size}
Theorem \ref{thm. separation II} is expected as its proof simply models the latent distribution of an HMM, which is memory-less, and thus no Transformer layer is needed ($L=0$).
So what is the purpose of using Transformers in practice?
One possible explanation is cost optimization.
Any HMM that models natural language well should have complicated hidden states that capture the real world, and the hidden space size $|\Omega|$ would be astronomical.
Instead of setting the hidden dimension to $|\Omega|$ as in the proof, one can rely on a lazy representation that is distributed over time.

Specifically, suppose the HMM is built on a 2D grid world of size $10^2$ and each grid point has $10$ states, so the hidden space has size $|\Omega|=10^{100}$.
The state transition follows some rule that obeys locality, e.g.\@ the rule of the traffic, weather, or social interaction.
An agent moves around this grid, and the observable state $X_t=(I_t,O_t)$ consists of the agent's location $I_t$ and the current state $S_t$ of this location, so the observable state space is small, $|\Sigma|=100\times 10$.
A compromise needs to be made between state size and context length, i.e.\@ between a memoryless posterior over $10^{100}$ hidden states, and a posterior over 1000 observable states whose context dependency may be intractable as in Theorem \ref{thm. separation I}.

A desirable solution is that the model keeps an intermediate state size and a context sequence with controlled length.
The past information is half-processed and stored as a bounded sequence of feature vectors with moderate dimension, and next-token prediction requires only bounded amount of computation, thus implementable with bounded-depth circuits.


\end{remark}

\begin{remark}[Emergence of memory]
Remark \ref{remark. context dependency representation size} tries to derive in a first-principles manner a bounded-length latent context to replace the unbounded-length text context of Transformers and the unbounded-dimensional feature vector of RNNs.
This representation resembles a memory mechanism, in that it is moderately compact and continuously updated.
In an earlier paper \cite{yang2024memory}, we modeled the emergence of memory mechanisms as the result of minimizing total read-write cost.
Remark \ref{remark. context dependency representation size} points towards an alternative theory for the genesis of this cognitive capacity from the need of approximation ability.
\end{remark}

\subsection{Separation, Part III}
\label{sec. separation III}

This section provides a supplementary result to our separation theorems, that there needs to be sufficiently many latent sequences to ensure the expressivity of $\proj\#P_f$.

\begin{definition}
\label{def. polynomially simple projection}
A polynomially simple projection is a simple projection (cf.\@ Definition \ref{def. simple projection}) such that each nondeterministic circuit $C_t$ has only $O(\log t)$ auxiliary bits.
As a result, the set $\fip(x)$ has size $O(\poly(|x|))$ for any $x\in\Sigma^*$.
\end{definition}

Recall that Example \ref{ex. product} has $|\fip(x)| = |\Gamma|^{|x|}$ and Example \ref{ex. think} has $|\fip(x)| \geq |\Sigma|^{c|x|}$ (assuming that the upper bounds are constant $c_t \equiv c$).
Since they have exponentially many latent sequences, these projections are not polynomially simple.

\begin{example}
Similar to Example \ref{ex. think}, set $\Omega = \Sigma \cup \{\sot,\eot\}$.
However, the ``thoughts" are inserted very sparsely:
For some constant $c$,
\begin{align*}
\dom(&\proj) = \big\{ \big( \sot y^{(t)} \eot x^{(t)} \big)_{t=1}^{\infty} \bigm| x^{(t)}\in\Sigma^{2^t}~\text{and}~ y^{(t)} \in \Sigma^c \big\}\\
&\proj\colon  (y^{(t)} x^{(t)})_{t=1}^{\infty} \mapsto (x^{(t)})_{t=1}^{\infty}
\end{align*}
Namely, we have fixed-length thoughts separated by exponentially growing text segments.
It is straightforward to check that $\proj$ is a polynomially simple projection.
\end{example}

The following theorem indicates that polynomially simple projections cannot help to model HMMs, strengthening the result of Theorem \ref{thm. separation I}.
Thus, at least superpolynomially many latent sequences are required for ensure the expressivity of the parametrization $\proj\#P_f$.

\begin{theorem}
\label{thm. separation III}
Asume that $\TC^0 \subsetneq \NC^1$.
There exists an HMM $(A,B,\lambda)$ such that for any finite set $\Omega$,
any polynomially simple projection $\proj$,
and any configuration $(L,H,d_h,p)$, the semi-divergence (\ref{eq. semi-divergence}) cannot be reduced to zero
\begin{equation}
\label{eq. separation III}
\inf_{f\in \TF(L,H,d_h,p)}  D(P_{A,B,\lambda}, \proj\#P_f)  > 0
\end{equation}
\end{theorem}

This result is expected since a ``meta" $\TC^0$ circuit family whose gates are $\TC^0$ circuits (with uniform depth bound and uniform polynomial size bound) remains a $\TC^0$ circuit family.
A similar result is provided by \cite[Theorem 3.2]{li2024chain}, which shows that Transformers whose chain-of-thoughts have at most logarithmic lengths can only recognize $\TC^0$ languages.

\subsection{Summary}
\label{sec. approximation summary}

This section has introduced simple projections $\proj$ and the distribution parametrization $\proj\#P_f$, and demonstrated through Theorems \ref{thm. separation I} and \ref{thm. separation II} that $\proj\#P_f$, compared to the usual parametrization $P_f$, is capable of modeling a lot more distributions over sequences.
As a result, it is expected that simple projections can greatly reduce the norm of the target distribution in general
\begin{equation*}
\|P_*\|_{\F,\proj} \ll \|P_*\|_{\F} \leq \infty
\end{equation*}
and thus models with projections enjoy smaller training error (\ref{eq. train loss upper bound}) and generalization error (\ref{eq. test loss early stop}).

An illustration of our separation theorems is provided in Figure \ref{fig: representation hierarchy}, showing the hierarchy of the four families (or subspaces) of $\PS(\Sigma^{\omega})$.
Technically, these four families are defined as follows
\begin{align*}
\PS_{\text{HMM}} &= \big\{P_{A,B,\lambda} \bigm| \text{finite set} ~ \Omega, ~ A\in\Delta_{|\Omega|}^{|\Omega|}, ~ B\in\Delta_{|\Sigma|}^{|\Omega|}, ~ \lambda \in \Delta_{|\Omega|}\big\} \\
\PS_{\text{plain}} &= \big\{P_f \bigm| f \in \TF_{<\omega}(\Sigma) \big\} \\
\PS_{\text{poly}} &= \big\{ \proj\#P_f \bigm| \text{finite set} ~ \Omega, ~ f \in \TF_{<\omega}(\Omega), ~ \text{polynomially-simple projection} ~ \proj \big\}\\
\PS_{\text{simple}} &= \big\{ \proj\#P_f \bigm| \text{finite set} ~ \Omega, ~ f \in \TF_{<\omega}(\Omega), ~ \text{simple projection} ~ \proj \big\}
\end{align*}
where $\Delta_d$ and $\Delta_d^k$ denote the spaces of probability vectors and stochastic matrices,
and $\TF_{<\omega}(\Omega)$ denotes the space of Transformers with vocabulary set to $\Omega \cup \{\bos\}$ and with finite sizes, namely
\begin{equation*}
\TF_{<\omega}(\Omega) := \bigcup_{L=1}^{\infty} \bigcup_{H=1}^{\infty} \bigcup_{d_h=1}^{\infty} \bigcup_{p=1}^{\infty} \TF(L,H,d_h,p)
\end{equation*}
With subspace containment defined by (\ref{eq. semidivergence contain}),
Theorems \ref{thm. separation I} and \ref{thm. separation III} indicate that neither $\PS_{\text{plain}}$ nor $\PS_{\text{poly}}$ contains $\PS_{\text{HMM}}$.
Some of those difficult HMMs are defined by compositional functions such as (\ref{eq. A5 composition function}).
Meanwhile, Theorem \ref{thm. separation II} indicate that $\PS_{\text{simple}}$ contains $\PS_{\text{HMM}}$, and thus $\PS_{\text{simple}}$ is strictly larger than $\PS_{\text{plain}},\PS_{\text{poly}}$.
Hence, Figure \ref{fig: representation hierarchy} is established.

The challenge of using $\proj\#P_f$ is that its computations (\ref{eq. latent marginal}, \ref{eq. latent conditional}) involve summation over the set of latent sequences, $\fip(x)$.
Theorem \ref{thm. separation III} indicates that the size of this set must be superpolynomial in $|x|$, and Examples \ref{ex. product} and \ref{ex. think} further imply that the size is in general exponential in $|x|$.
Thus, exact computation is infeasible, and the following section designs efficient algorithms for training and inference.

\section{Latent Sampling and Optimal Samplers}
\label{sec. latent sampling}

\begin{figure}[!ht]
\centering
\includegraphics[width=0.8\linewidth]{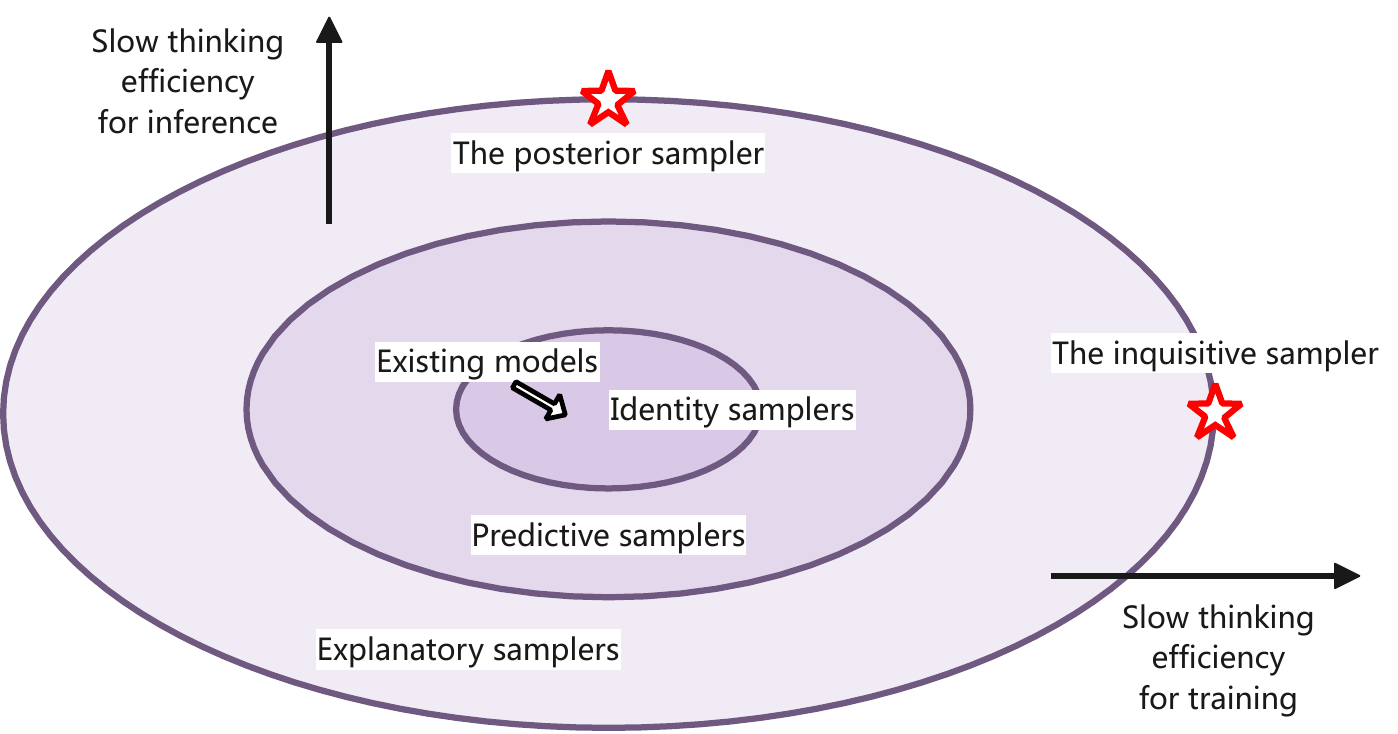}
\caption{The sampler hierarchy.
The slow thinking scaling laws derived in this section largely depends on the efficiency of latent sampling, and thus on how well the implemented samplers can approximate the optimal samplers.
Thus, a hierarchy in terms of expressivity is established, that includes the identity sampler, the predictive samplers, and the explanatory samplers.
In general, this hierarchy is strict, with each containment being proper.
We prove that the optimal samplers for inference and training (marked by stars) are generally not predictive samplers.
Since the existing models rely on the identity sampler, their slow thinking scaling laws are suboptimal.}
\label{fig: sampler hierarchy}
\end{figure}

With the significance of the parametrization $\proj\#P_f$ established, this section studies how to perform training and inference with $\proj\#P_f$.
Our methods will be based on latent sampling, which leads to the sampler hierarchy (Figure \ref{fig: sampler hierarchy}) and slow thinking scaling laws (Table \ref{table: slow thinking scaling laws}).

Specifically, training means, given any sample $x \in \Sigma^*$,
updating the function parameter $f$ to increase the likelihood $(\proj\#P_f)^{\leq |x|}(x)$.
This probability is given by equation (\ref{eq. latent marginal}) of Theorem \ref{thm. conditional latent distribution}.
Alternatively, given any pair, $x_p, x_s \in \Sigma^*$ (e.g.\@ a problem and a solution), increase the conditional likelihood $(\proj\#P_f)^{\leq |x_p|+|x_s|}(x_s|x_p)$.
Meanwhile, inference means, given any prefix $x \in \Sigma^*$ (also known as prompt), producing a sample from $(\proj\#P_f)(\cdot|[x])$, the conditional distribution over sequences that starts with $x$.

Two operations, encoding and decoding, can be defined in the context of $\proj\#P_f$.
Let $x \in \Sigma^*$ be any observable sequence, which can be either a training sample or the prompt for inference.
Encoding means inferring the posterior distribution of the latent sequences $z\in\fip(x)$.
Decoding means the continuation of $x$ conditioned on a latent sequence $z \in \fip(x)$.
It follows that training involves only encoding, while inference requires first encoding and then decoding.
The encoding stage of inference is often called prefill.

One major difficulty for encoding is the intractable size of the set $\fip(x)$.
Theorem \ref{thm. separation III} indicates that the size must be superpolynomial in $|x|$, and Examples \ref{ex. product} and \ref{ex. think} further imply that the size is in general exponential in $|x|$.
Thus, plain enumeration is infeasible.
We tackle $\fip(x)$ with sampling, and solve for the optimal samplers.

\begin{assumption}
\label{assume. sec sampling}
Throughout this section, we assume that the projection $\proj$ and the latent distribution $P$ satisfy the condition of Theorem \ref{thm. conditional latent distribution}.
\end{assumption}

\subsection{Unconditional Encoding}
\label{sec. encoding}

First, we study how to compute $(\proj\#P)^{\leq |x|}(x)$ efficiently.
This probability can be expressed as a summation (\ref{eq. latent marginal}) over all possible latent sequences $\fip(x)$.

\subsubsection{Efficient Estimation}

There are three common approaches to efficiently estimating the sum of many non-negative numbers.
The first approach is the forward algorithm.
Consider Example \ref{ex. product} for instance,
\begin{align*}
(\proj\#P)^{\leq |x|}(x) &= \sum_{y_1,\dots y_{|x|} \in \Gamma} P^{\leq t}((x_t,y_t)_{t\leq |x|})\\
&= \sum_{y_1\in\Gamma} P^{\leq 1}(y_1|x_1) \Big( \sum_{y_2 \in \Gamma} P^{\leq 2}\big(y_2\big|x_{\leq 2}, y_1\big) \Big( \dots \sum_{y_{|x|}\in\Gamma}P^{\leq |x|}\big(y_{|x|}\big|x,y_{<|x|}\big) \Big)\Big)\\
\text{where} &~ P^{\leq t}\big(y_t \big| x_{\leq t}, y_{<t} \big) = \frac{P^{\leq t}\big((x_t,y_t)\big|(x_i,y_i)_{i<t}\big)}{\sum_x P^{\leq t}\big((x,y_t)\big|(x_i,y_i)_{i<t}\big)}
\end{align*}
Suppose $P$ is a Markov chain, such that $P^{\leq t+1}(\cdot|((x_i,y_i))_{i\leq t}) \equiv P^{\leq 2}(\cdot|(x_t,y_t))$ for all $t\geq 1$.
Then, the summation simplifies to a series of matrix-vector multiplications
\begin{align}
\label{eq. Markov forward algorithm}
(\proj\#P)^{\leq |x|}(x) &= P^o_{x_1} \Big( \prod_{t=2}^{|x|} P^{\to}_{x_{t-1},x_t} \Big) \mathbf{1}\\
\nonumber
P^o_{x_1,y_1} &:= P^{\leq 1}(y_1|x_1), ~ P^{\to}_{x_1,x_2,y_1,y_2} := P^{\leq 2}\big(y_2\big|(x_1,y_1),x_2\big)
\end{align}
where $\mathbf{1}\in\R^{|\Gamma|}$ is a column vector, $P^o$ is a $|\Sigma|\times|\Gamma|$ matrix, and $P^{\to}$ is a $|\Sigma|\times|\Sigma|\times|\Gamma|\times|\Gamma|$ tensor.
Thus, computing $(\proj\#P)^{\leq |x|}(x)$ takes $O(|x|\cdot |\Gamma|^2)$ operations.
In particular, given any HMM $(A,B,\lambda)$, if we set
\begin{equation*}
P^o_{x_1,y_1} = \frac{\lambda_{y_1} B_{y_1,x_1}}{(\lambda B)_{x_1}}, \quad P^{\to}_{x_1,x_2,y_1,y_2} = \frac{A_{y_1,y_2}B_{y_2,x_2}}{(AB)_{y_1,x_2}}
\end{equation*}
then (\ref{eq. Markov forward algorithm}) becomes the forward algorithm of HMMs.
Nevertheless, the Markov assumption, or approximating the latent distribution $P$ by bigram distribution, seems too crude;
whereas if $k$-gram approximations are considered, then a modified version of (\ref{eq. Markov forward algorithm}) would require $O(|x|\cdot |\Gamma|^k)$ operations, which is too costly.

The second approach is maximum \textit{a posteriori} estimation.
Specifically, it assumes that the summation (\ref{eq. latent marginal}) is dominated by a very small subset $\{z^{(1)},\dots z^{(k)}\} \subseteq \fip(x)$, such that the left side would be very close to the right side:
\begin{equation*}
\sum_{i=1}^k P^{\leq |z^{(i)}|}(z^{(i)}) \leq \sum_{z\in\fip(x)}P^{\leq |z|}(z)
\end{equation*}
Then, what remains is to locate these maximizers $z^{(i)}$ by various search methods such as evolutionary algorithms.
However, this approach does not seem suitable for (\ref{eq. latent marginal}), because conceptually $P$ could be insensitive to token-wise mutations,
e.g.\@ if $z^{(i)}$ is a natural language solution for a math problem, then it should not matter if we swap ``First" with ``First of all" and ``Thus" with ``Therefore" etc.
Even if only a fraction of $z$ can be modified, say $10\%$, there are still $\geq 2^{0.1|z|}$ candidates.
So, in general, the summation (\ref{eq. latent marginal}) may contain exponentially many non-negligible terms.

The third approach is Monte-Carlo estimation.
Given any conditional distribution $Q(\cdot|x) \in \PS(\Omega^*)$ whose support equals $\fip(x)$ for all $x\in\Sigma^*$, we have
\begin{align*}
(\proj\#P)^{\leq |x|}(x) &= \sum_{z\in \fip(x)} P^{\leq |z|}(z) = \int \frac{P^{\leq |z|}(z)}{Q(z|x)} dQ(z|x)
\end{align*}
Denoting the above value by $p_*$, an estimator $p_n$ can be constructed by
\begin{equation}
\label{eq. Monte-Carlo estimator}
p_n := \frac{1}{n}\sum_{i=1}^n \frac{P^{\leq |Z^{(i)}|}(Z^{(i)})}{Q(Z^{(i)}|x)}, \quad \{Z^{(i)}\}_{i=1}^n \iidsample Q(\cdot|x)
\end{equation}
for any sampling size $n$.
One can check that this estimator is strongly consistent (i.e.\@ $\lim_{n\to\infty} p_n = p_*$ almost surely) and unbiased (i.e.\@ $\E[p_n]=p_*$).
To avoid numerical overflow or underflow, (\ref{eq. Monte-Carlo estimator}) can be computed with log probabilities
\begin{align*}
\log p_n &= \text{logsumexp} \Big( \big( \log P^{\leq |Z^{(i)}|}(Z^{(i)}) - \log Q(Z^{(i)}|x) \big)_{i=1}^n \Big) - \log n\\
\text{logsumexp}\big((c_i)_{i=1}^n\big) &= \max_i c_i + \log \sum_{j=1}^n \exp(c_j - \max_i c_i)
\end{align*}

Hence, this paper adopts Monte-Carlo estimation to compute (\ref{eq. latent marginal}).
One reason is that it is theoretically correct without additional assumptions, unlike the other approaches.
The other reason is that its convergence rate, discussed below, has no explicit dependence on the size of $\fip(x)$, and thus may scale well with long sequences $x$.

Regarding the feasibility of sampling from a distribution $Q(\cdot|x)$ whose support is exactly $\fip(x)$, in theory this is easy if $\proj$ is a simple projection (Definition \ref{def. simple projection}).
Specifically, let $C_t$ be a non-deterministic circuit from Definition \ref{def. simple projection} that models $\fip$ on length-$t$ inputs, and let $\{0,1\}^k$ be the auxiliary inputs for this nondeterministic circuit (cf.\@ Definition \ref{def. non-deterministic circuit}) such that $C_t(x,\{0,1\}^k) = \fip(x)$ for all $x\in\Sigma^t$.
Then, let $Y$ be a random variable uniformly distributed in $\{0,1\}^k$, the random variable $C_t(x,Y)$ ranges over all of $\fip(x)$ and hits each latent sequence $z \in \fip(x)$ with probability $\geq 2^{-k}$.

\begin{remark}[Sequential sampling]
\label{remark. sequential sampling}
While the samples are produced in parallel in the Monte-Carlo estimator (\ref{eq. Monte-Carlo estimator}), there are sequential sampling methods that can generate dependent samples to achieve certain effects.
For instance, stochastic gradient descent produces ``samples" $\{\theta_t\}_{t=1}^{\infty}$ that approximate the local minima of a landscape,
while Quasi-Monte Carlo produces ``uniformly placed" samples to reduce estimation error \cite{morokoff1995quasi}.
Sequential sampling can be implemented implicitly over $\fip(x)$.
As an example, consider the projection of Example \ref{ex. think};
its thought space can be equivalently written as
\begin{equation*}
\mathcal{T}_c = \bigcup_{K=0}^{\infty} \mathcal{T}^K, \quad \mathcal{T}^0 = \{\varnothing\}, \quad \mathcal{T}^K = \big\{ \sot y^{(1)} \dots y^{(K)} \eot \bigm| y^{(1)}, \dots y^{(K)} \in \Sigma^*, ~ \sum_{k=1}^K |y^{(k)}| \leq c \big\}
\end{equation*}
where each $y^{(k)}$ ($k<K$) ends with some recognizable delimiter such as multiple line breaks.
So now we view each thought as a trace of multiple thoughts.
If the sampling distribution $Q(\cdot|x)$ decodes the latent samples autoregressively, then within a decoded thought trace $Y \in \mathcal{T}_c$, each thought $Y^{(K)}$ can be dependent on the earlier thoughts $\{Y^{(k)}\}_{k=1}^{K-1}$.
This dependency can model sophisticated thinking strategies of humans, e.g.\@ trying novel approaches to a math problem if the usual approaches all fail.
We have observed such implicit sequential sampling in the thinking process of DeepSeek-R1, as it often transitions among multiple proof strategies, and such transitions are named as the ``reconstruction" stage in the phenomenological studies of these models \cite{marjanovic2025thoughtology}.
Nevertheless, this paper does not go into the semantic meanings of the latent samples, and focuses only on the formal modeling of latent sampling.
The philosophy is that we avoid any artificial specification of the functioning of the model, and only provide the soil for its cognitive function to grow on its own.
\end{remark}

\subsubsection{Samplers and Variance Reduction}
\label{sec. sampler}

We define samplers as follows.
\begin{definition}
\label{def. sampler}
A (latent) sampler $Q$ is a partial function $\Sigma^* \to \PS(\Omega^*)$
such that $\dom(Q)$ is prefix-closed (i.e.\@ if $x \in \dom(Q)$, then $x' \in \dom(Q)$ for all $x' \sqsubseteq x$),
and that for each $x\in\dom(Q)$, the conditional distribution $Q(\cdot|x)$ is supported on $\fip(x)$.
Denote the set of samplers by $\Q_{\proj}$.
\end{definition}
Note that this definition only depends on $\Sigma,\Omega$ and $\proj$, but not on the latent distribution $P$.
By definition, the posterior $Q_*$ defined in (\ref{eq. latent posterior}) is a sampler with
\begin{equation}
\label{eq. domain of posterior}
\dom(Q_*) = \big\{ x \in \Sigma^* \bigm|(\proj\#P)^{\leq |x|}(x)>0 \big\}
\end{equation}

For algorithmic design, one may additionally require that the sampler be a generative model that can produce samples $Z\sim Q(\cdot|x)$ and also evaluate the probabilities $Q(z|x)$,
but this is always assumed doable in our mathematical analysis.

Considering function parametrization, $P$ and $Q$ may be implemented with different functions, the same function but with different prompts, or the same function and prompt.
The last case, however, is quite restrictive and suboptimal, as will be discussed in Section \ref{sec. prediction vs explanation}.

In order to accelerate the convergence of the Monte-Carlo estimator $p_n$ to the true value $p_*$, the variance of $p_n$ should be minimized.
The variance is simply
\begin{align}
\label{eq. Monte-Carlo var}
\begin{split}
var(p_n) &= \frac{1}{n} \int \Big( \frac{P^{\leq |z|}(z)}{Q(z|x)} - (\proj\#P)^{\leq |x|}(x) \Big)^2 dQ(z|x)\\
&= (\proj\#P)^{\leq |x|}(x)^2 \cdot \frac{\chi^2\big(Q_*(\cdot|x) \big\| Q(\cdot|x) \big)}{n}
\end{split}
\end{align}
where $\chi^2$ is the chi-square divergence, $\chi^2(p\|q) = \int (p/q)^2 dq - 1$.
So the optimal sampler, which makes $var(p_n)=0$, is exactly the posterior distribution $Q_*$.

To approximate $Q_*$, one can train the sampler $Q$ by minimizing one of the following losses:
either $var(p_n)$,
or the reverse KL divergence
\begin{align}
\label{eq. reverse KL sampler}
\KL\big(Q(\cdot|x)\big\|Q_*(\cdot|x)\big) &= \int \log \frac{Q(z|x)}{P^{\leq |z|}(z)} dQ(z|x) + C
\end{align}
or the forward KL divergence
\begin{align*}
\KL\big(Q_*(\cdot|x)\big\|Q(\cdot|x)\big) &= c \int \frac{P^{\leq |z|}(z)}{Q(z|x)}\log \frac{P^{\leq |z|}(z)}{Q(z|x)} dQ(z|x) + C
\end{align*}
where $c>0,C$ are constants that do not depend on $Q$.
The forward KL is presented in the above form, since we can only sample from $Q$ instead of $Q_*$, and thus the integration must be with respect to $Q$.
The reverse KL seems most favorable:
All losses involve the term $P^{\leq |z|}(z) / Q(z|x)$, which could incur high variance when $z$ is sampled from $Q(\cdot|x)$, but reverse KL only contains the logarithm of this term, which is much milder than the others.

Hence, this paper adopts the reverse KL (\ref{eq. reverse KL sampler}) for sampler training.

\begin{remark}
The reverse KL divergence can be seen as a special case of a larger family of losses.
For any $n \geq 1$, consider
\begin{align}
\label{eq. sampler NLL loss}
L\big(Q(\cdot|x)\big) &= \E[-\log p_n] = - \int \dots \int \log\Big( \frac{1}{n} \sum_{i=1}^n \frac{P^{\leq|z^{(i)}|}(z^{(i)})}{Q(z^{(i)}|x)} \Big) \prod_{i=1}^n dQ(z^{(i)}|x)
\end{align}
with $p_n$ defined by (\ref{eq. Monte-Carlo estimator}).
By Proposition \ref{prop. sampler loss minimizer}, the unique minimizer of (\ref{eq. sampler NLL loss}) is exactly the posterior distribution $Q_*(\cdot|x)$.
Also, with $n=1$, the loss simplifies to the reverse KL
\begin{equation*}
\E[-\log p_1] = \KL\big( Q(\cdot|x) \big\| Q_*(\cdot|x) \big)
\end{equation*}
The advantage of this loss is that, since $\E[-\log p_n]$ can be used as the loss function of the latent distribution $P$, the training of the latent distribution and the sampler can be unified into a single objective
\begin{equation*}
\min_f \min_g \E[-\log p_n], \quad p_n = \frac{1}{n}\sum_{i=1}^n q_i, \quad q_i = \frac{P_f^{\leq |Z^{(i)}|}(Z^{(i)})}{Q_g(Z^{(i)}|x)}, \quad \{Z^{(i)}\}_{i=1}^n \iidsample Q_g(\cdot|x)
\end{equation*}
where $f,g$ are the parameter functions of $P,Q$.
However, the disadvantage is that inside the gradient
\begin{equation*}
\nabla_g \E[-\log p_n] = \E\Big[ \sum_{i=1}^n c_i \nabla_g \log Q_g(Z^{(i)}|x) \Big], \quad c_i = \frac{q_i}{\sum_{j=1}^n q_j} - \log\Big( \frac{1}{n} \sum_{j=1}^n q_j \Big)
\end{equation*}
the differences between the coefficients $c_i$ are small, $0\leq |c_i-c_j| \leq 1$.
In comparison, for the reverse KL divergence, the differences between the coefficients can be unbounded, $|\log q_i - \log q_j| \in [0,+\infty]$, as will be calculated in Section \ref{sec. sampler training}.
So the reverse KL could be more discriminative and thus more efficient for gradient-based training.
\end{remark}


\begin{remark}
It is worth mentioning that there are more general strategies than using $Q_*$ directly as the training target.
Abstractly speaking, the task is to build a generative model given an unnormalized density function, e.g.\@ our density $P$ restricted to the subset $\fip(x)$.
One alternative loss is the forward KL divergence $\KL(\tilde{Q}\|Q)$ from \cite{gabrie2022adaptive}, where the surrogate target $\tilde{Q}$ is made by ``pushing" $Q$ slightly towards $P$ by $\nabla \log P$.
\end{remark}

\begin{remark}
The variance of the estimator (\ref{eq. Monte-Carlo estimator}) due to its summand $P^{\leq |x|}(z) / Q(z|x)$ can be further reduced,
if the summation is over log-probabilities.
This is in fact feasible if we assume that the summand follows some log-normal distribution, namely
\begin{equation*}
\law\big( S \big) = \mathcal{N}(\mu,\sigma^2), \quad S = \log P^{\leq |Z|}(Z) - \log Q(Z|x), \quad Z \sim Q(\cdot|x)
\end{equation*}
for some $\mu$ and $\sigma^2$.
Since the mean of log-normal distribution is $\exp(\mu+\sigma^2/2)$, the following is an unbiased estimator for log probabilities
\begin{align*}
\log (\proj\#P)^{\leq |x|}(x) &\approx \frac{1}{n}\sum_{i=1}^n S_i + \frac{1}{2(n-1)}\sum_{i=1}^n \Big( S_i - \frac{1}{n}\sum_{j=1}^n S_j \Big)^2\\
S_i &= \log P^{\leq |Z^{(i)}|}(Z^{(i)}) - \log Q(Z^{(i)}|x), \quad \{Z^{(i)}\}_{i=1}^n \iidsample Q(\cdot|x)
\end{align*}
and one can show that the optimal sampling distribution that makes $var=0$ is again (\ref{eq. latent posterior}).
Nevertheless, the log-normal assumption is not guaranteed to hold in general.
\end{remark}

\subsection{Conditional Encoding}
\label{sec. conditional encoding}

This section discusses the subtlety of estimating the conditional probability $(\proj\#P)^{\leq |x|+|y|}(y|x)$.

A biased estimator can be obtained by simply dividing the Monte-Carlo estimators (\ref{eq. Monte-Carlo estimator}) of unconditional probabilities
\begin{align}
\label{eq. conditional estimator}
& (\proj\#P)^{\leq |x|+|y|}(y|x) = \frac{p_*}{q_*} \approx \frac{p_n}{q_n} = \frac{\frac{1}{n}\sum_{i=1}^n p^{(i)}}
{\frac{1}{n}\sum_{i=1}^n q^{(i)}} \\
\nonumber
& p_* := (\proj\#P)^{\leq |x|+|y|}(xy) = \E[p^{(i)}], \quad p^{(i)} := \frac{P^{\leq |W^{(i)}|}(W^{(i)})}{Q(W^{(i)} | xy)}, \quad \{W^{(i)}\}_{i=1}^n \iidsample Q(\cdot| xy) \\
\nonumber
& q_* := (\proj\#P)^{\leq |x|}(x) = \E[q^{(i)}], \quad q^{(i)} := \frac{P^{\leq |Z^{(i)}|}(Z^{(i)})}{Q(Z^{(i)} | x)}, \quad \{Z^{(i)}\} \iidsample Q(\cdot| x)
\end{align}
where all samples $W^{(i)}, Z^{(i)}$ are independent.
Despite that $p_n,q_n$ are unbiased, their quotient is in general overestimated with a relative error of $O(n^{-1})$.
Specifically, using formal calculation (based on Lemma \ref{lemma. quotient estimator}) and the variance formula (\ref{eq. Monte-Carlo var}), one can compute the relative error
\begin{align}
\label{eq. bias conditional estimator}
\begin{split}
\Big( \E\Big[\frac{p_n}{q_n}\Big] - \frac{p_*}{q_*} \Big) \Big/ \frac{p_*}{q_*} &= \E\Big[ \frac{q_*}{q_n} \Big] - 1 \\
&= \sum_{k=2}^{\infty} (-1)^k ~ \E \Big[ \Big(\frac{1}{n}\sum_i \frac{q^{(i)}}{q_*} - 1 \Big)^k \Big] = \frac{\chi^2\big( Q_*(\cdot|x) \big\| Q(\cdot|x) \big)}{n} + O(n^{-2})
\end{split}
\end{align}
The first line implies that the relative error does not depend on $y$, so the estimator is unbiased if one only needs an unnormalized density that is proportional to $(\proj\#P)^{\leq |x|+t}(\cdot|x)$, e.g.\@ when explicitly estimating the next-token distribution, the output would eventually be normalized to a probability vector.
Similarly, the mean squared error (MSE) of the estimator can be computed formally by
\begin{align}
\label{eq. mse conditional estimator}
\begin{split}
\E\Big[\Big(\frac{p_n}{q_n} - \frac{p_*}{q_*}\Big)^2\Big] &= var(p_n) ~\E[q_n^{-2}] + p_*^2 ~\E\big[ \big( q_n^{-1} - q_*^{-1} \big)^2 \big] \\
&= \frac{var(p_n)}{q_*^2} \big( 1 + O(n^{-1}) \big) + \frac{p_*^2}{q_*^2} \Big(\frac{var(q_n)}{q_*^2} + O(n^{-2})\Big) \\
&= \Big(\frac{p_*}{q_*}\Big)^2 \frac{\chi^2\big( Q_*(\cdot|xy) \big\| Q(\cdot|xy) \big) + \chi^2\big( Q_*(\cdot|x) \big\| Q(\cdot|x) \big)}{n} + O(n^{-2})
\end{split}
\end{align}
One nice property of this estimator is that its bias and MSE can be arbitrarily small.
One way to see this is that if $Q(\cdot|x) = Q_*(\cdot|x)$ and $Q(\cdot|xy) = Q_*(\cdot|xy)$, then $p_n = p_*$ and $q_n = q_*$ almost surely.
Another way is that the above estimates are proportional to the $\chi^2$ errors of $Q$, which goes to 0 as $Q$ converges to $Q_*$ (so do the terms in $O(n^{-2})$ that measure the higher order differences between $Q$ and $Q_*$).

\begin{remark}
\label{remark. reduce bias of quotient}
A simple way to reduce the bias (\ref{eq. bias conditional estimator}) is to compute $q_n$ by reusing the samples of $p_n$.
Assume that, instead of using fully independent samples $\{W^{(i)}\}_{i=1}^n$ and $\{Z^{(i)}\}_{i=1}^n$, we only require the pairs $\{(W^{(i)},Z^{(i)})\}_{i=1}^n$ to be i.i.d., allowing some dependency between $W^{(i)}$ and $Z^{(i)}$.
Then, the formal computation (\ref{eq. bias conditional estimator}) becomes
\begin{align*}
\Big( \E\Big[\frac{p_n}{q_n}\Big] - \frac{p_*}{q_*} \Big) \Big/ \frac{p_*}{q_*} &= \sum_{k=1}^{\infty} (-1)^k c_k, \quad c_k = \E\Big[\frac{p_n}{p_*}\Big(\frac{q_n}{q_*}-1\Big)^k\Big]
\end{align*}
The first few $c_k$ are
{\small
\begin{align*}
c_1 &= \frac{1}{n} \E \Big[ \Big(\frac{p^{(1)}}{p_*}-1\Big) \Big(\frac{q^{(1)}}{q_*} - 1 \Big) \Big] \\
c_2 &= \frac{1}{n} \E\Big[ \Big(\frac{q^{(1)}}{q_*} - 1 \Big)^2 \Big] + \frac{1}{n^2} \E\Big[ \Big(\frac{p^{(1)}}{p_*} - 1 \Big) \Big(\frac{q^{(1)}}{q_*} - 1 \Big)^2 \Big] \\
c_3 &= \frac{1}{n^2} \E\Big[ \Big(\frac{q^{(1)}}{q_*} - 1 \Big)^3 \Big] + \frac{1}{n^2} \E \Big[ \Big(\frac{p^{(1)}}{p_*}-1\Big) \Big(\frac{q^{(1)}}{q_*} - 1 \Big) \Big] \E\Big[ \Big(\frac{q^{(1)}}{q_*} - 1 \Big)^2 \Big] + \frac{1}{n^3} \E\Big[ \Big(\frac{p^{(1)}}{p_*}-1\Big) \Big(\frac{q^{(1)}}{q_*} - 1 \Big)^3 \Big]
\end{align*}
}%
The sum of the rest of the $c_k$ are $O(n^{-2})$ (cf. Lemma \ref{lemma. quotient estimator}).
The first term of $c_2$ is the source of the $\chi^2/n$ bias in (\ref{eq. bias conditional estimator}).
Ideally, if the random variables $p^{(1)}/p_*$ and $q^{(1)}/q_*$ are positively correlated, then the bias would be reduced
\begin{align*}
\sum_{k=1}^{\infty} (-1)^k c_k &= c_2 - c_1 + O(n^{-2}) \\
&= \frac{1}{n}\E\Big[ \Big(\frac{q^{(1)}}{q_*} - 1 \Big)^2 \Big] - \frac{1}{n}\E\Big[ \Big(\frac{q^{(1)}}{q_*} - 1 \Big) \Big(\frac{p^{(1)}}{p_*} - 1 \Big) \Big] + O(n^{-2}) \\
&< \frac{\chi^2\big( Q_*(\cdot|x) \big\| Q(\cdot|x) \big)}{n}
\end{align*}
Here is one way to achieve positive correlation:
Given $W \sim Q(\cdot|xy)$, Lemma \ref{lemma. segments prefix-free} implies that $W$ can be uniquely decomposed into $W=ZS$ for $Z\in\fip(x)$ and $S\in\nsm_x(Z,y)$,
and Lemma \ref{lemma. prefix-free projection} allows us to denote $Z=\Pi_{\fip(x)}(W)$.
Redefine the estimator (\ref{eq. conditional estimator}) by setting
\begin{equation*}
Z^{(i)} = \Pi_{\fip(x)} W^{(i)}, \quad Q(\cdot|x) = \int Q(\cdot s|xy) ds
\end{equation*}
So the samples for $q_n$ are the part of the samples of $p_n$ that belong to $\fip(x)$, and the density function $Q(\cdot|x)$ is adjusted accordingly.
If the sampler $Q(\cdot|xy)$ is implemented by the usual parametrization (Definition \ref{def. softmax with masking}), which allows access to the tokenwise conditional probabilities, then $\int Q(z s|xy) ds$ can be obtained by simply restricting to the token probabilities over $z$.
Denote the mapping from $W$ to $p_n/p_*$ and $q_n/q_*$ by $f(W)$ and $g(W)$.
Informally speaking, since the functional forms of $f$ and $g$ are very similar, one can expect the random variables $f(W)$ and $g(W)$ to be positively correlated.
\end{remark}

One may try to derive an alternative estimator based on the expectation (\ref{eq. latent conditional}), but there would be two issues.
Suppose we define the following estimator
\begin{align}
\nonumber
(\proj\#P)^{\leq |x|+|y|}(y|x) &= \int P\big(\nsm(z,y)\big|z\big) \text{d}Q_*(z|x)\\
\label{eq. conditional estimator 2}
&\approx \frac{1}{n}\sum_{i=1}^n \frac{P^{\leq |Z^{(i)}|+|S^{(i)}|}(S^{(i)}|Z^{(i)})}{Q(S^{(i)}|x,Z^{(i)},y)} \frac{\frac{P^{\leq |Z^{(i)}|}(Z^{(i)})}{(\proj\#P)^{\leq |x|}(x)}}{Q(Z^{(i)}|x)}\\
\nonumber
Z^{(i)} &\iidsample Q(\cdot|x), \quad S^{(i)} \stackrel{\text{ind}}\sim Q(\cdot|x,Z^{(i)},y)
\end{align}
where $Q(\cdot|x,z,y)$ is any distribution with support $\nsm_x(z,y)$, and we set $P^{\leq |Z^{(i)}|+|S^{(i)}|}(S^{(i)}|Z^{(i)}) = 0$ if $\nsm_x(Z^{(i)},y) = \varnothing$.
The obvious issue is the need to estimate the term $(\proj\#P)^{\leq |x|}(x)$, which makes this estimator biased, similar to (\ref{eq. conditional estimator}).
Yet, a greater issue is the variability of the posterior distribution $Q_*$.
Assuming for simplicity that $(\proj\#P)^{\leq |x|}(x)$ is known, the variance of this estimator is given by
\begin{equation*}
\frac{(\proj\#P)^{\leq |x|+|y|}(y|x)^2}{n} \chi^2\big( Q_*(\cdot|xy) \big\| Q(\cdot|xy) \big), \quad Q(zs|xy) := Q(s|x,z,y)Q(z|x)
\end{equation*}
where $Q(zs|xy)$ is well-defined thanks to the unique decomposition of $zs$ (cf.\@ Lemma \ref{lemma. segments prefix-free}).
Then by Lemma \ref{lemma. chi-square}, we have
\begin{equation*}
\chi^2\big( Q_*(\cdot|xy) \bigm\| Q(\cdot|xy) \big) \geq \chi^2\Big( \int Q_*(\cdot s|xy) ds \Bigm\| Q(\cdot|x) \Big)
\end{equation*}
where the target is the marginal distribution of $Q_*(\cdot|xy)$ over $\fip(x)$
\begin{equation}
\label{eq. posterior early marginal}
\int Q_*(z s|xy) ds = \big(\Pi_{\fip(x)} \# Q_*(\cdot|xy) \big)(z) \propto P\big(\nsm_x(z,y)|z\big) Q_*(z|x)
\end{equation}
In general $z\mapsto P\big(\nsm_x(z,y)|z\big)$ is a non-constant function over $\fip(x)$ and could vary with $y$,
so one can expect that
\begin{equation*}
\inf_{Q \in \PS(\fip(x))} ~ \sup_{y\in\Sigma^*} ~ \chi^2\Big( \int Q_*(\cdot s|xy) ds \Bigm\| Q \Big) > 0
\end{equation*}
In particular, we generally have
\begin{equation*}
\chi^2\Big( \int Q_*(\cdot s|xy) ds \Bigm\| Q(\cdot|x) \Big) > 0 \quad \text{and} \quad \int Q_*(\cdot s|xy) ds \not\equiv Q_*(\cdot|x)
\end{equation*}
It follows that the estimator (\ref{eq. conditional estimator 2}) suffers from an irreducible variance, as long as the sampler $Q(\cdot|x)$ does not depend on $y$.
This issue will be studied more thoroughly in Section \ref{sec. prediction vs explanation}.

If we instead use some sampler $Q(\cdot|x,y) \in \PS(\fip(x))$, then one can check that the estimator (\ref{eq. conditional estimator 2}) becomes equivalent to the estimator (\ref{eq. conditional estimator}) up to the parametrization of $Q$.

Hence, this paper uses (\ref{eq. conditional estimator}) to estimate conditional distributions.





\subsection{Decoding}
\label{sec. decoding}

Decoding is to sample from the conditional distribution $(\proj\#P)(\cdot|[x])$ given any prompt $x\in\Sigma^*$.
We discuss three sampling methods: a straightforward method with fixed sampling error,
a simple method with arbitrarily small error,
and an unbiased but indirect method.

The first method requires a sampler $Q$ that approximates the posterior $Q_*$,
e.g.\@ trained to have small loss (\ref{eq. reverse KL sampler}).
The decoding $X^{(\omega)}$ is produced by
\begin{equation}
\label{eq. decode once}
X^{(\omega)} = \proj\big(Z^{(\omega)}\big), \quad Z^{(\omega)} \sim P\big(\cdot\big|[Z^{(0)}]\big), \quad Z^{(0)} \sim Q(\cdot|x)
\end{equation}
This random sample enjoys a bounded error.

\begin{proposition}
\label{prop. decode error bound}
For any sequence $x\in\Sigma^*$ with $(\proj\#P)^{\leq |x|}(x)>0$ and any sampler $Q$ such that $\sprt Q(\cdot|x) \subseteq \sprt Q_*(\cdot|x)$,
the random variable $X^{\omega}$ is well-defined and its sampling error is bounded by
\begin{align*}
\KL\big((\proj\#P)(\cdot|[x]) \bigm\| \law(X^{\omega}) \big) &\leq \KL\big(Q_*(\cdot|x) \bigm\| Q(\cdot|x)\big)\\
\KL\big( \law(X^{\omega}) \bigm\| (\proj\#P)(\cdot|[x]) \big) &\leq \KL\big(Q(\cdot|x) \bigm\| Q_*(\cdot|x)\big)
\end{align*}
\end{proposition}
Note that the sampling error is measured by the plain KL divergence over $\PS(\Sigma^{\omega})$,
which is much finer than the loss $D_0$ or the weak topology of $\PS(\Sigma^{\omega})$ and in general equals $\infty$.
Thus, the upper bound of Proposition \ref{prop. decode error bound} is a bit non-trivial.

$X^{(\omega)}$ can be generated incrementally, which is helpful for real-time streaming,
such that users can watch each token $X_t$ gets printed as soon as it is generated, instead of after the entire sequence $X^{(\omega)}$ has been decoded.
Let $\proj^{<\omega}$ be the prefix projection (Definition \ref{def. prefix projection}) of $\proj$.
The following sequence $X^{(t)} \in \Sigma^*$ keeps growing and converges pointwise to $X^{(\omega)}$ as $t\to\infty$:
\begin{equation}
\label{eq. decode stream}
X^{(t)} = \proj^{<\omega}(Z^{(t)}), \quad Z^{(t)} = Z^{(t-1)} Z_t, \quad Z_t \sim P^{\leq t + |Z^{(0)}|}(\cdot|Z^{(t-1)}), \quad Z^{(0)} \sim Q(\cdot|x)
\end{equation}
The validity of this streaming decoding is established by the following proposition.

\begin{proposition}
\label{prop. stream decode equivalence}
For any sequence $x\in\Sigma^*$ with $(\proj\#P)^{\leq |x|}(x)>0$ and any sampler $Q$ such that $\sprt Q(\cdot|x) \subseteq \sprt Q_*(\cdot|x)$,
the infinite sequence $\bigcup_{t=1}^{\infty} X^{(t)}$ is a well-defined random variable on $\Sigma^{\omega}$, and has the same distribution as $X^{(\omega)}$.
\end{proposition}

For convenience, the streaming process (\ref{eq. decode stream}) can be written equivalently as follows, in terms of the next-segment distribution (Definition \ref{def. next-segment distribution})
\begin{align*}
X^{(t)} &= X^{(t-1)}X_t, \quad X_t = \proj_{X^{(t-1)},Z^{(t-1)}}(S_t), \quad Z^{(t)} = Z^{(t-1)} S_t \\
S_t &\sim P\big(\cdot \big| Z^{(t-1)}\to\bigsqcup_{a\in\Sigma} \nsm_{X^{(t-1)}}(Z^{(t-1)},a) \big)\\
X^{(0)} &= \varnothing, \quad Z^{(0)} \sim Q(\cdot|x)
\end{align*}

The second method is error-tolerant and only requires a sampler $Q$ with $\sprt Q(\cdot|x) = \sprt Q_*(\cdot|x)$.
The initial sequence $Z^{(0)}$ is produced by resampling:
\begin{align}
\label{eq. resampling posterior}
\begin{split}
Z^{(0)} &= Z^{(I,0)}, \quad I \sim \text{Categorical}(p_1, \dots p_n), \quad \{ Z^{(i,0)}\}_{i=1}^n \iidsample Q(\cdot|x) \\
& \text{where} \quad p_i = \frac{P^{\leq |Z^{(i,0)}|}(Z^{(i,0)}) \big/ Q(Z^{(i,0)}|x)}{\sum_{j=1}^n P^{\leq |Z^{(j,0)}|}(Z^{(j,0)}) \big/ Q(Z^{(j,0)}|x)}
\end{split}
\end{align}
Then, one follows the rest of either (\ref{eq. decode once}) or (\ref{eq. decode stream}) to produce the decoding $X^{(\omega)}$.
This method enjoys an arbitrarily small error:

\begin{proposition}
\label{prop. TV decoding error}
For any $x\in\Sigma^*$ with $(\proj\#P)^{\leq |x|}(x)>0$ and any sampler $Q$ with $\sprt Q(\cdot|x) = \sprt Q_*(\cdot|x)$,
the total variation distance $\TV$ between $(\proj\#P)(\cdot|[x])$ and the distribution of $X^{(\omega)}$ defined by (\ref{eq. resampling posterior}) is bounded in expectation by
\begin{align*}
\E_{\{ Z^{(i,0)} \}_{i=1}^n } \big[ \TV\big( (\proj\#P)(\cdot|[x]), \law(X^{(\omega)}) \big) \big] & \leq \E\big[ \TV\big( Q_*(\cdot|x), \law(Z^{(0)}) \big) \big] \\
&\leq \sqrt{\frac{6\chi^2\big( Q_*(\cdot|x) \big\| Q(\cdot|x)\big) + 2}{n}}
\end{align*}
\end{proposition}

The third method explicitly estimates the next-token distributions $(\proj\#P)^{\leq |x|+t+1}(\cdot|X^{(t)})$
and is discussed in Appendix \ref{appendix. unbiased decode}.
One potential advantage is that its estimation, based on Monte-Carlo, is unbiased,
but the algorithm is more complicated and costly than the previous two.

\subsection{Training}
\label{sec. training}

With inference solved, this section derives the training of the latent distribution $P$ and the sampler $Q$.
Interestingly, it turns out that the optimal sampler for training is different from the optimal sampler for inference (namely $Q_*$), and we name the former as the inquisitive sampler.

Suppose the latent distribution is parametrized by $P=P_f$, with parameter function parametrized by $f=f_{\theta}$.
The variable $\theta$ ranges in some Hilbert space (or Euclidean space) with norm $\|\cdot\|$.
For concreteness, one can use the conditional softmax (Definition \ref{def. softmax with masking}) for $P_f$ and Transformers (Definition \ref{def. TF}) for $f_{\theta}$.

\subsubsection{Training the Latent Distribution}
\label{sec. training latent}

Given any data distribution (over finite-length sequences) $P_* \in \PS(\Sigma^*)$,
the most common objective is to minimize the cross entropy
\begin{equation}
\label{eq. cross entropy}
\min_{\theta} L(P_f), \quad L(P) = -\int \log (\proj\#P)^{\leq |x|}(x) dP_*(x)
\end{equation}
It may be of theoretical interest to note that this loss $L$ can be interpreted as a KL divergence on $P$, not just on $\proj\#P$ (more details are provided in Appendix \ref{appendix. KL finite to infinite}).
As discussed in previous sections, the exact computation of $(\proj\#P)^{\leq |x|}(x)$ is intractable, so some estimation is needed.
For gradient-based training, what matters is not to estimate the loss value but the loss gradient
\begin{align*}
\vb_* := \nabla_{\theta}L(P_f) &= - \int \frac{\sum_{z\in\fip(x)} \nabla_{\theta} P_f^{\leq |z|}(z)}{(\proj\#P_f)^{\leq |x|}(x)} dP_*(x) \\
&= - \iint \nabla_{\theta} \log P_f^{\leq |z|}(z) dQ_*(z|x) dP_*(x)
\end{align*}
where $Q_*$ is the posterior sampler of $P_f$.
A Monte-Carlo estimator for $\vb_*$ can be defined as follows:
First, replace $P_*$ by text samples and expand $Q_*$ to get
\begin{equation*}
\vb^{(B)} := - \frac{1}{B} \sum_{b=1}^B \frac{\sum_{z\in\fip(X^{(b)})}P_f^{\leq |z|}(z) ~ \nabla_{\theta} \log P_f^{\leq |z|}(z)}{\sum_{z\in\fip(X^{(b)})}P_f^{\leq |z|}(z)}, \quad \{X^{(b)}\}_{b=1}^B \iidsample P_*
\end{equation*}
where $B$ is the amount of text samples per batch.
Then, estimate both the numerator and denominator by sampling:
\begin{align*}
\vb_n^{(B)} &:= - \frac{1}{B} \sum_{b=1}^B \frac{\frac{1}{n} \sum_{i=1}^n \tilde{q}_i^{(b)} ~ \nabla_{\theta} \log P_f^{\leq |\tilde{Z}^{(b,i)}|}(\tilde{Z}^{(b,i)})}{\frac{1}{n} \sum_{i=1}^n q_i^{(b)}} \\
q_i^{(b)} &= \frac{P_f^{\leq |Z^{(b,i)}|}(Z^{(b,i)}) }{ Q(Z^{(b,i)}|X^{(b)})}, \quad \tilde{q}_i^{(b)}= \frac{P_f^{\leq |\tilde{Z}^{(b,i)}|}(\tilde{Z}^{(b,i)}) }{ \tilde{Q}(\tilde{Z}^{(b,i)}|X^{(b)})} \\
\{X^{(b)}\}_{b=1}^B &\iidsample P_*, \quad \{Z^{(b,i)}\}_{i=1}^n \iidsample Q(\cdot|X^{(b)}), \quad \{\tilde{Z}^{(b,i)}\}_{i=1}^n \iidsample \tilde{Q}(\cdot|X^{(b)})
\end{align*}
Note that in general the samplers $Q,\tilde{Q}$ for the denominator and numerator could be different.
In practice, in order to obtain a desired mini-batch gradient, one would define a mini-batch loss, and then automatic differentiation programs such as PyTorch can produce the gradient.
The mini-batch loss for $\vb_n^{(B)}$ can be defined as
\begin{equation}
\label{eq. minibatch loss two sampler}
L_n^{(B)}(\theta) := -\frac{1}{B}\sum_{b=1}^B \frac{\frac{1}{n} \sum_{i=1}^n \detach(\tilde{q}_i^{(b)}) ~ \log P_f^{\leq |\tilde{Z}^{(b,i)}|}(\tilde{Z}^{(b,i)})}{\frac{1}{n} \sum_{i=1}^n \detach(q_i^{(b)})}
\end{equation}
where \faHandStopO~detaches a term from differentiation, a common practice in PyTorch.
Thus,
\begin{equation*}
\vb_n^{(B)} = \nabla_{\theta} L_n^{(B)}(P_f)
\end{equation*}

One special case is to assume that the numerator and denominator share the same sampler ($Q=\tilde{Q}$) and samples ($Z^{(b,i)}=\tilde{Z}^{(b,i)}$).
The mini-batch loss can have a simpler form
\begin{align}
\label{eq. minibatch loss one sampler}
L_n^{(B)}(\theta) &= -\frac{1}{B} \sum_{b=1}^B \log\Big( \frac{1}{n} \sum_{i=1}^n q_i^{(b)} \Big) \\
\nonumber
\nabla_{\theta} L_n^{(B)}(P_f) &= - \frac{1}{B} \sum_{b=1}^B \frac{\frac{1}{n} \sum_{i=1}^n q_i^{(b)} ~ \nabla_{\theta} \log P_f^{\leq |Z^{(b,i)}|}(Z^{(b,i)})}{\frac{1}{n} \sum_{i=1}^n q_i^{(b)}}
\end{align}
The ``original loss" that corresponds to this mini-batch loss would be
\begin{align}
\label{eq. finite-sample loss}
L_n^{(\infty)} &= - \int \E \big[\log p_n(x) \big] dP_*(x) \\
\nonumber
p_n(x) &= \frac{1}{n} \sum_{i=1}^n \frac{P_f^{\leq |Z^{(i)}(x)|}(Z^{(i)}(x))}{Q(Z^{(i)}(x)|x)} , \quad \{Z^{(i)}(x)\}_{i=1}^n \iidsample Q(\cdot|x)
\end{align}
This formulation can be interpreted as a slightly different way to make the loss (\ref{eq. cross entropy}) tractable, such that one estimates the loss value instead of the loss gradient.
The next section, however, will show that this is suboptimal.
Also, we will see later that the case $n=1$ is of special interest, such that many existing slow thinking models can be derived from the loss $L_1^{(\infty)}$.

Hence, in this paper, we use either (\ref{eq. minibatch loss two sampler}) or (\ref{eq. minibatch loss one sampler}) to train the latent distribution $P_f$.

While the above discussion is about unconditional encoding, one can also consider supervised finetuning.
Let $P_* \in \PS(\Sigma^* \times \Sigma^*)$ be a distribution of question-answer pairs $(x_q,x_a)$.
The finetuning loss can be defined based on
\begin{equation*}
L(P) = - \int \log (\proj\#P)^{\leq |x_q|+|x_a|}(x_a|x_q) dP_*(x_q,x_a)
\end{equation*}
and the conditional estimators from Section \ref{sec. conditional encoding}.

As a remark, note that if the trivial projection, namely the identity function of Example \ref{ex. identity}, is used, then both losses (\ref{eq. minibatch loss two sampler}, \ref{eq. minibatch loss one sampler}) simplify to the training loss of fast-thinking LLMs.
Since $\fip(x) \equiv x$, all samplers reduce to $Q(\cdot|x)=\delta_x$ and the samples $Z^{(b,i)},\tilde{Z}^{(b,i)}$ reduce to $X^{(b)}$, and thus
\begin{equation*}
L^{(B)}_n(\theta) = - \frac{1}{B} \sum_{b=1}^B \log P_f^{\leq |X^{(b)}|}(X^{(b)})
\end{equation*}
which is the mini-batch version of $\KL(P_*\|P_f)$.

\subsubsection{The Inquisitive Sampler}
\label{sec. inquisitive sampler}

Next, we specify the samplers $Q,\tilde{Q}$ in the mini-batch loss (\ref{eq. minibatch loss two sampler}).
Since the estimated gradient $\vb_n^{(B)}$ should approximate the true gradient $\vb_*$,
we want the samplers to minimize the expected error:
\begin{equation}
\label{eq. min grad mse batch}
\min_{Q,\tilde{Q} \in \Q_{\proj}} \E\big[\|\vb_*-\vb_n^{(B)}\|^2\big]
\end{equation}
under the trivial constraint that both $\dom(Q)$ and $\dom(\tilde{Q})$ contain $\dom(Q_*)$,
namely, the samplers $Q,\tilde{Q}$ are defined on all prefixes $x \in \Sigma^*$ that can be generated by $P_f$.

Since $\vb^{(B)}-\vb_n^{(B)}$ is an average of $B$ i.i.d.\@ random vectors, Jensen's inequality implies that
\begin{align*}
\E\big[\|\vb_*-\vb_n^{(B)}\|^2\big] &= \E\big[\|\vb_*-\vb^{(B)}\|^2\big] + \E\big[\|\vb^{(B)}-\vb_n^{(B)}\|^2\big] \\
&\leq \frac{\E\big[\|\vb_*-\vb^{(1)}\|^2\big]}{B} + \E\big[\|\vb^{(1)}-\vb_n^{(1)}\|^2\big]
\end{align*}
and equality holds only if $\vb^{(1)} - \vb_n^{(1)}$ is constant almost surely.
Since the first term does not depend on the samplers, the objective (\ref{eq. min grad mse batch}) can be replaced with the slightly stronger condition
\begin{equation}
\label{eq. min grad mse}
\forall x\in\dom(Q_*), \quad Q(\cdot|x), \tilde{Q}(\cdot|x) = \argmin_{Q',\tilde{Q}'\in\PS(\fip(x))} \E\big[\|\vb_*(x)-\vb_n(x)\|^2\big]
\end{equation}
with $\vb_*(x)$ and $\vb_n(x)$ defined by
\begin{align}
\label{eq. target gradient}
\vb_*(x) & := - \nabla_{\theta} \log (\proj\#P_f)^{\leq |x|}(x) = - \frac{\sum_{z\in\fip(x)}P_f^{\leq |z|}(z) ~ \nabla_{\theta} \log P_f^{\leq |z|}(z)}{\sum_{z\in\fip(x)}P_f^{\leq |z|}(z)} \\
\nonumber
\vb_n(x) & := - \frac{\frac{1}{n} \sum_{i=1}^n \tilde{q}_i(x) ~ \nabla_{\theta} \log P_f^{\leq |\tilde{Z}^{(i)}(x)|}(\tilde{Z}^{(i)}(x))}{\frac{1}{n} \sum_{i=1}^n q_i(x)} \\
\nonumber
&q_i(x) = \frac{P_f^{\leq |Z^{(i)}(x)|}(Z^{(i)}(x)) }{ Q'(Z^{(i)}(x))}, \quad \{Z^{(i)}(x)\}_{i=1}^n \iidsample Q' \\
\nonumber
&\tilde{q}_i(x) = \frac{P_f^{\leq |\tilde{Z}^{(i)}(x)|}(\tilde{Z}^{(i)}(x))}{\tilde{Q}'(\tilde{Z}^{(i)}(x))}, \quad \{\tilde{Z}^{(i)}(x)\}_{i=1}^n \iidsample \tilde{Q}'
\end{align}
Proposition \ref{prop. min grad mse} indicates that (\ref{eq. min grad mse}) has a unique solution, such that $Q$ is the posterior sampler $Q_*$ as usual, whereas $\tilde{Q}$ is a distinct sampler $Q_{\eye}$ defined by
\begin{align}
\label{eq. inquisitive sampler}
Q_{\eye}(z|x) &:= \frac{ P_f^{\leq |z|}(z) ~ \big\| \nabla_{\theta} \log P_f^{\leq |z|}(z) \big\|}{\sum_{z'\in\fip(x)} P_f^{\leq |z'|}(z') ~ \big\| \nabla_{\theta} \log P_f^{\leq |z'|}(z') \big\|}
\end{align}
for all $x\in \dom(Q_*)$.

Besides the proof of Proposition \ref{prop. min grad mse}, there is a quick way to derive (\ref{eq. inquisitive sampler}).
Suppose that the denominator of $\vb_n(x)$ has zero variance.
Then,
\begin{align*}
\vb_1(x) = - \frac{Q_*(Z|x)}{\tilde{Q}'(Z)} \nabla_{\theta}\log P_f^{\leq |Z|}(Z), \quad Z \sim \tilde{Q}'
\end{align*}
and $\E[\vb_1(x)] = \vb_*(x)$.
So
\begin{align*}
\E\big[ \|\vb_*(x)-\vb_1(x)\|^2 \big] &= c \int \frac{P_f^{\leq |z|}(z|x)^2}{\tilde{Q}'(z)^2} \big\| \nabla_{\theta}\log P_f^{\leq |z|}(z) \big\|^2 d\tilde{Q}'(z) + C \\
&= c ~ \chi^2\big( Q_{\eye}(\cdot|x) \bigm\| \tilde{Q}' \big) + C
\end{align*}
and thus the unique minimizer is $Q_{\eye}$.
We give it a name below.

\begin{definition}
\label{def. inquisitive sampler}
The inquisitive sampler $Q_{\eye}$ is the sampler defined by (\ref{eq. inquisitive sampler}), given any differentiably parametrized log-likelihood function $\theta\mapsto \log P_f$.
\end{definition}

The name ``inquisition" is inspired by the ``exploration vs.\@ exploitation" problem in reinforcement learning.
Specifically, the distinction between the inquisitive sampler $Q_{\eye}$ and the posterior sampler $Q_*$ can be compared to that of exploration and exploitation.
Informally, we can imagine $z,z'\in\fip(x)$ to be two proofs of a math statement $x$, such that $z$ follows a routine that one's mind $P_f$ is familiar with, while $z'$ exhibits a very novel approach.
It may happen for instance that $P_f^{\leq |z'|}(z') \approx P_f^{\leq |z|}(z)/2$ and thus $Q_*(z'|x) < Q_*(z|x)$,
whereas $\|\nabla_{\theta} \log P_f^{\leq |z'|}(z')\| \geq 4 \|\nabla_{\theta} \log P_f^{\leq |z|}(z)\|$ and thus $Q_{\eye}(z'|x) > Q_{\eye}(z|x)$.
So $Q_{\eye}$ tends to explore new ideas and $Q_*$ tends to exploit established methods.

\begin{remark}[Data curriculum]
\label{remark. inquisitive}
As a slightly more formal way to express ``exploration vs.\@ exploitation", the distribution $Q_{\eye}(\cdot|x)$ can be interpreted as the ``uncertain" part of $Q_*(\cdot|x)$.
Note that for any differentiable map $\theta \mapsto p$ from a Hilbert space to $(0,1)$,
\begin{equation*}
\nabla_{\theta} p = p \nabla_{\theta} \log p = (1-p) \big(-\nabla_{\theta} \log (1-p)\big)
\end{equation*}
So $\|\nabla_{\theta} p\|$ is roughly proportional to $p(1-p)$, which is further proportional to the entropy $H(p) = -p\log p - (1-p)\log(1-p)$.
One may conclude that
\begin{equation*}
Q_{\eye}(\cdot|x) \propto H\big(Q_*(\cdot|x)\big)
\end{equation*}
In some sense, $Q_{\eye}(\cdot|x)$ induces a data curriculum for $\proj\#P_f$,
that throughout training, the samples $Z \sim Q_{\eye}(\cdot|x)$ are always right on the edge of the ability of $P_f$,
encouraging $P_f$ to move out of its comfort zone.
\end{remark}

\begin{remark}[Translation invariance]
Intuitively, the inquisitive sampler should be invariant with respect to any uniform translation of the vectors $\nabla_{\theta} \log P_f^{\leq |z|}(z)$ and depend only on the distances $\| \nabla_{\theta} \log P_f^{\leq |z|}(z) - \vb_*(x) \|$ instead of the norms $\|\nabla_{\theta} \log P_f^{\leq |z|}(z)\|$.
This would be the case if we modify the estimator $\vb_n(x)$ to a weighted average
\begin{equation*}
\tilde{\vb}_n(x) = - \frac{\sum_{i=1}^n \tilde{q}_i(x) ~ \nabla_{\theta} \log P_f^{\leq |\tilde{Z}^{(i)}(x)|}(\tilde{Z}^{(i)}(x))}{\sum_{i=1}^n \tilde{q}_i(x)}
\end{equation*}
Then, one can show that as $n\to\infty$, a solution that achieves the approximately minimum estimation error
$$\min_{\tilde{Q}(\cdot|x)} \E\big[ \| \vb_*(x) - \tilde{\vb}_n(\x) \|^2\big] + O(n^{-2})$$
is given by
\begin{equation*}
\tilde{Q}(z|x) = \frac{P_f^{\leq |z|}(z) ~ \big\| \nabla_{\theta} \log P_f^{\leq |z|}(z) - \vb_*(x) \big\|}{\sum_{z'\in\fip(x)} P_f^{\leq |z'|}(z') ~ \big\| \nabla_{\theta} \log P_f^{\leq |z'|}(z') - \vb_*(x) \big\|}
\end{equation*}
It can be an alterative implementation of the inquisitive sampler.
\end{remark}


So far we have shown that, while the optimal sampler for inference is simply the posterior $Q_*$, optimal training requires two samplers, $Q_*$ and $Q_{\eye}$.
Despite that (\ref{eq. inquisitive sampler}) resembles (\ref{eq. latent posterior}),
the inquisitive sampler is parametrization-dependent, unlike the posterior sampler.
They are different as long as the norm $\|\nabla_{\theta} \log P_f^{\leq |z|}(z)\|$ is a non-constant function in $z\in\fip(x)$.
It follows that the mini-batch loss (\ref{eq. minibatch loss two sampler}) has higher potential than (\ref{eq. minibatch loss one sampler}) in terms of the accuracy of gradient estimation, and may lead to better performance when trained on the same data.

\begin{definition}
\label{def. two sampler training}
When training is performed with the loss (\ref{eq. minibatch loss two sampler}),
we refer to the sampler in the numerator as the ``train sampler", and denote it by $Q^{\see}$.
We refer to the sampler in the denominator (\ref{eq. minibatch loss two sampler}) (or more generally, any likelihood estimator (\ref{eq. Monte-Carlo estimator})) as the ``inference sampler", and denote it by $Q$.
\end{definition}

Thus, the mini-batch loss (\ref{eq. minibatch loss two sampler}) (for any sample $x \in \dom(Q_*)$) becomes
\begin{align}
\label{eq. minibatch loss inquisitive}
L^{\see}_n (\theta) &= - \frac{\frac{1}{n} \sum_{i=1}^n \log P_f^{\leq |Z^{(i)}_{\eye}|}(Z^{(i)}_{\eye}) ~\detach(q^{(i)}_{\eye})}{\frac{1}{n} \sum_{i=1}^n \detach(q^{(i)})}\\
\nonumber
&q^{(i)} = \frac{P_f^{\leq |Z^{(i)}|}(Z^{(i)})}{Q(Z^{(i)}|x)}, \quad \{Z^{(i)}\}_{i=1}^n \iidsample Q(\cdot|x) \\
\nonumber
&q^{(i)}_{\eye} = \frac{P_f^{\leq |Z^{(i)}_{\eye}|}(Z^{(i)}_{\eye})}{Q^{\see}(Z^{(i)}_{\eye}|x)}, \quad \{Z^{(i)}_{\eye}\}_{i=1}^n \iidsample Q^{\see}(\cdot|x)
\end{align}
The gradient estimator $\vb_n^{\see}(x) = \nabla_{\theta}L^{\see}_n(P_f)$ has an expected error of
\begin{align}
\label{eq. grad mse one sample}
\E\big[ \| \vb_*(x) - \vb_n^{\see}(x) \|^2 \big] = \frac{1}{n} \Big[ &C_{\downarrow}^2 ~ (\proj\#P_f)^{\leq|x|}(x)^2 ~ \chi^2\big( Q_*(\cdot|x) \big\| Q(\cdot|x) \big) \\
\nonumber
+ &C_{\uparrow}^2 ~\chi^2\big( Q_{\eye}(\cdot|x) \big\| Q^{\see}(\cdot|x) \big) + (C_{\uparrow}^2 - C_{\downarrow}^2) \Big] + O(n^{-2}) \\
\nonumber
\text{where} \quad & C_{\uparrow} := \int \big\| \nabla_{\theta} \log P_f^{\leq |z|}(z) \big\| dQ_*(z|x) \\
\nonumber
& C_{\downarrow} := \|\vb_*(x)\| = \Big\| \int \nabla_{\theta} \log P_f^{\leq |z|}(z) dQ_*(z|x) \Big\| \leq C_{\uparrow}
\end{align}
A formal derivation is provided in Appendix \ref{appendix. gradient estimation error}.
Similar to the previous error bounds, (\ref{eq. grad mse one sample}) has the nice form $O(\chi^2/n)$, this time with two divergence terms.
The additional term $C_{\uparrow}^2 - C_{\downarrow}^2$ indicates that the sampling error persists even when using the optimal samplers, unless all constituent vectors lie in the same one-dimensional subspace.

\begin{remark}[Policy collapse]
\label{remark. policy collapse}
A common pitfall of training slow thinking models is ``policy collapse" \cite{yu2025DAPO,cui2025entropy,huang2025spark}, such that the entropy of the sampling distribution of the chain-of-thoughts quickly drops to near zero early in training, hampering the improvement of the models.
Our calculation indicates a possible cause of this issue.
As will be analyzed in Section \ref{sec. deepseek training}, the existing slow thinking models
implement only the inference sampler $Q$
and are trained to fit only the posterior sampler $Q_*$, neglecting the inquisitive sampler $Q_{\eye}$.
This could lead to a positive feedback cycle during training:
Since $Q$ fits $Q_*$, which is proportional to the latent model $P$ by definition,
if $P$ has some mode $z_* \in \fip(x)$, then $Q$ would tend to concentrate at $z_*$,
so the samples $Z^{(i)} \sim Q(\cdot|x)$ tend to resemble $z_*$,
and thus $P$ being trained on $Z^{(i)}$ become further concentrated at $z_*$,
and so on, until total collapse.
In terms of Remark \ref{remark. inquisitive}, the model is trained to focus only on exploitation instead of exploration.
Hence, to ensure stable and sustainable training, it might be necessary to implement the train sampler $Q^{\see}$.
\end{remark}

\subsubsection{Training the Samplers}
\label{sec. sampler training}

Having derived the training loss of the latent distribution $P_f$, we now derive the losses for the samplers $Q$ and $Q^{\see}$, which can be trained simultaneously with $P_f$.

Similar to $P_f$, we assume that $Q,Q^{\see}$ are parametrized by some variables $\psi,\psi^{\see}$ in a Hilbert space, such that their log-likelihood functions are differentiable in $\psi,\psi^{\see}$ for any $x\in\dom(Q_*)$ and $z\in\fip(x)$.
For concreteness, one can use conditional softmax (Definition \ref{def. softmax with masking}) plus Transformers (Definition \ref{def. TF}) as usual.
To decode $z$ conditioned on $x$, one can simply include $x$ in the prompt.
The two samplers can be implemented with either separate models or the same model with different prompts,
e.g.\@ the prompt for $Q$ can start with ``$\langle$inference mode$\rangle$" and the prompt for $Q^{\see}$ with ``$\langle$training mode$\rangle$".
(One can even consider the special case when $P_f$ and the samplers share the same model/parameter.
This will be studied in Sections \ref{sec. prediction vs explanation} and \ref{sec. deepseek sampling}.
For now, let $P_f$ and the samplers be distinct models.)

As indicated by the gradient estimation error (\ref{eq. grad mse one sample}) as well as Proposition \ref{prop. min grad mse}, the inference sampler $Q$ should approximate the posterior sampler $Q_*$,
and the training sampler $Q^{\see}$ should approximate the inquisitive sampler $Q_{\eye}$.
Recall that Section \ref{sec. encoding} has derived the reverse KL loss (\ref{eq. reverse KL sampler}).
So the inference sampler should minimize the following
\begin{equation*}
\int \KL\big(Q(\cdot|x)\big\|Q_*(\cdot|x)\big) dP_*(x) = \iint \log \frac{Q(z|x)}{P^{\leq |z|}(z)} dQ(z|x) dP_*(x) + C
\end{equation*}
Since the loss involves an expectation with $Q(\cdot|x)$, the trick of policy gradient can help to compute the gradient
\begin{align*}
\nabla_{\psi} \KL\big( Q(\cdot|x) \big\| Q_*(\cdot|x) \big) &= \sum_{z\in\fip(x)} \log \frac{Q(z|x)}{P^{\leq |z|}(z)} \nabla_{\psi} Q(z|x) + \int \nabla_{\psi} \log Q(z|x) dQ(z|x) \\
&= \int \log \frac{Q(z|x)}{P^{\leq |z|}(z)} \nabla_{\psi} \log Q(z|x) ~dQ(z|x) + 0
\end{align*}
So it is equivalent to use the following implementation-friendly loss
\begin{align*}
L_{\text{sampler}}(Q) := - \iint \detach \Big( \log \frac{P^{\leq |z|}(z)}{ Q(z|x)} \Big) \log Q(z|x) ~ d \detach \big( Q (z|x) \big) dP_*(x)
\end{align*}
where the detached terms are ignored by differentiation, and $\nabla_{\psi} L_{\text{sampler}}$ produces the same gradient.
Then, a mini-batch loss can be defined by
\begin{align}
\label{eq. minibatch sampler}
L^{(B,n)}_{\text{sampler}}(Q) &= - \frac{1}{B(n-1)} \sum_{b=1}^B \sum_{i=1}^n \detach( A^{(b,i)} ) \log Q(Z^{(b,i)}|X^{(b)}) \\
\nonumber
A^{(b,i)} &= R^{(b,i)} - \frac{1}{n} \sum_{j=1}^n R^{(b,j)}, \quad R^{(b,i)} = \log \frac{P^{\leq |Z^{(b,i)}|}(Z^{(b,i)})}{Q(Z^{(b,i)}|X^{(b)})}\\
\nonumber
&\{X^{(b)}\}_{b=1}^B \iidsample P_*, \quad \{Z^{(b,i)}\}_{i=1}^n \iidsample Q(\cdot|X^{(b)})
\end{align}
The normalized terms $A^{(b,i)}$ are used to reduce the variance of the gradient $\nabla_{\theta} L_{\text{sampler}}^{(B,n)}$,
and the coefficient $(n-1)^{-1}$ makes sure that the gradient is unbiased, $\E[\nabla_{\theta}L_{\text{sampler}}^{(B,n)}] = \nabla_{\theta}L_{\text{sampler}}$.
In terms of reinforcement learning, $Z^{(b,i)}$, $R^{(b,i)}$ and $A^{(b,i)}$ are known as the episode, reward and advantage.

Similarly, $Q^{\see}$ can be trained by (\ref{eq. minibatch sampler}) with reward modified to
\begin{equation}
\label{eq. train sampler reward}
R^{(b,i)}_{\eye} = \log \frac{P_f^{\leq |Z^{(b,i)}_{\eye}|}(Z^{(b,i)}_{\eye}) ~ \big\| \nabla_{\theta} \log P_f^{\leq |Z^{(b,i)}_{\eye}|}(Z^{(b,i)}_{\eye}) \big\|}{Q^{\see}(Z^{(b,i)}_{\eye}|X^{(b)})}, \quad \{Z^{(b,i)}_{\eye}\}_{i=1}^n \iidsample Q^{\see}(\cdot|X^{(b)})
\end{equation}
In practice, the gradient norm $\|\log\nabla_{\theta} P_f\|$ can be easily obtained with little cost.
For instance, in PyTorch the gradient norm can be returned after backpropagation (at least when the per-device micro-batch size is 1) \cite{torch2025clipgrad}.
Since the reward is detached in (\ref{eq. minibatch sampler}), one never needs to perform the costly differentiation of $\|\log\nabla_{\theta} P_f\|$.

Note that there is much overlap between the terms of the losses (\ref{eq. minibatch loss inquisitive}) and (\ref{eq. minibatch sampler}), which can save much cost when training $P_f$ and $Q$ together.
For each training step, one only needs to sample $X,Z$ once and compute the log-likelihoods of $P_f,Q$ once, and then the two losses are ready.
The sampler $Q^{\see}$ can be similarly treated.




\begin{remark}[Batch size]
Language model training typically adopts a large batch size, and one main reason is to increase hardware utilization
(specifically, to increase arithmetic intensity and also scale with more machines and processors).
From this perspective, the counterpart of batch size could be $B \times n$ for the training of $\proj\#P_f$,
since the degree of parallelization of both (\ref{eq. minibatch loss inquisitive}) and (\ref{eq. minibatch sampler}) is $B\times n$.
So if the best batch size for training a baseline model is 3200,
an appropriate choice of $(B,n)$ could be $(200,16)$.
This is especially helpful if one is short of data for some particular task or domain.
\end{remark}

\begin{remark}[Policy gradient]
As a follow-up to Remark \ref{remark. policy collapse}, note that the inquisitive sampler may be a general variance-reduction technique applicable to all policy gradient models, not limited to slow thinking LLMs.
For instance, consider a toy problem with a reward function $R$ defined on fixed-length episodes $\Omega^T$ for some $T$,
and a policy model $\pi_{\theta}$, which is a parametrized distribution on $\Omega^T$.
The objective is $\max_{\theta} -L(\theta)$, $L(\theta)=-\E_{Z \sim \pi_{\theta}}\big[R(Z)\big]$,
and one typically adopts some variant of the following loss
\begin{equation*}
L^{(n)}(\theta) = - \frac{1}{n-1} \sum_{i=1}^n A_i \log \pi_{\theta}(Z^{(i)}), \quad \{Z^{(i)}\}_{i=1}^n \iidsample \pi_{\theta}, \quad A_i = R(Z^{(i)}) - \frac{1}{n} \sum_{j=1}^n R(Z^{(j)})
\end{equation*}
whose gradient is unbiased, $\E[\nabla_{\theta}L^{(n)}] = \nabla_{\theta}L$.
As discussed in Remark \ref{remark. policy collapse}, sampling directly from $\pi_{\theta}$ may lead to policy collapse.
So instead, one may implement a separate policy $\pi^{\see}$ and train $\pi$ on the more general loss
\begin{align*}
L^{(n)}(\theta) &= - \frac{1}{n-1} \sum_{i=1}^n A_i \log \pi_{\theta}(Z^{(i)}) ~ \detach\Big(\frac{\pi_{\theta}(Z^{(i)})}{\pi^{\see}(Z^{(i)})}\Big) \\
& \{Z^{(i)}\}_{i=1}^n \iidsample \pi^{\see}, \quad A_i = R(Z^{(i)}) - \frac{1}{n} \sum_{j=1}^n R(Z^{(j)})
\end{align*}
which remains unbiased.
The variance-minimizing policy is given by
\begin{align*}
\pi_{\eye} &:= \argmin_{\pi'\in\PS(\Omega^T)} \E\Big[ \big\| \nabla_{\theta}L - \nabla_{\theta}L^{(n)} \big\|^2 \Big] \\
&= \argmin_{\pi'\in\PS(\Omega^T)} \int \frac{\big\|\nabla_{\theta}\log\pi_{\theta}(z) \big\|^2 \pi_{\theta}(z)^2}{\pi'(z)^2} \Big[ \big( R(z) - \E_{\pi'}[R] \big)^2 + \frac{\E_{\pi'}[R^2] - \E_{\pi'}[R]^2}{n-1} \Big] d\pi'(z)
\end{align*}
which can be approximated by
\begin{equation*}
\pi_{\eye} \approx C \big| R - \E_{\pi^{\see}}[R] \big| ~ \big\| \nabla_{\theta} \log \pi_{\theta} \big\| ~ \pi_{\theta}
\end{equation*}
Thus, to approximate $\pi_{\eye}$, the policy $\pi^{\see}$ can be trained on the loss
\begin{align*}
L_{\text{sampler}}^{(n)}(\pi^{\see}) &= - \frac{1}{n} \sum_{i=1}^n A^{\see}_i \log \pi^{\see}(Z^{(i)}) \\
A^{\see}_i &= R^{\see}_i - \frac{1}{n} \sum_{j=1}^n R^{\see}_i, \quad R^{\see}_i = \detach\Big( \log \frac{ | A_i | \big\| \nabla_{\theta} \log \pi_{\theta}(Z^{(i)}) \big\| \pi_{\theta}(Z^{(i)})}{\pi^{\see}(Z^{(i)}) } \Big)
\end{align*}
where $Z^{(i)}$ and $A_i$ are shared with $L^{(n)}$.
\end{remark}

\subsection{Slow Thinking}
\label{sec. slow thinking}

This section provides a characterization of slow thinking and its scaling laws, in terms of the samplers $Q$ and $Q^{\see}$.

The motivation is that the previous sections indicate that the performance of inference and training depends heavily on the two divergences,
$$\chi^2\big(Q_*(\cdot|x) \big\| Q(\cdot|x) \big) \quad \text{and} \quad \chi^2\big( Q_{\eye}(\cdot|x) \big\| Q^{\see}(\cdot|x) \big)$$
namely, the alignment between our implemented samplers and the theoretically optimal samplers.
Specifically, this dependency is illustrated by the variance of unconditional likelihood estimation (\ref{eq. Monte-Carlo var}),
the bias and MSE of conditional likelihood estimation (\ref{eq. bias conditional estimator}, \ref{eq. mse conditional estimator}),
the bias of the two direct decoding algorithms (Propositions \ref{prop. decode error bound} and \ref{prop. TV decoding error}),
the variance of the indirect decoding algorithm (Proposition \ref{prop. estimate conditional}), and the MSE of gradient estimation (\ref{eq. grad mse one sample}).
So it should be generally feasible to translate the approximation errors of these samplers to their efficacy during inference and training.

For concreteness, consider the following two informal examples.

\begin{example}[Inference]
\label{ex. inference efficiency}
In a math exam, one tries to solve a multiple-choice question $x_q$ with four answers $a^{(1)}, \dots a^{(4)} \in \Sigma^*$, with $a^{(1)}$ being the correct answer.
Let $\proj$ be the projection from ``thoughts" $\Omega^{\omega}$ to ``texts" $\Sigma^{\omega}$, and $P$ be a distribution of thoughts generated by one's mind.
Correctly solving the question amounts to
\begin{equation*}
\text{argmax}_{i\in\{1,\dots 4\}} (\proj\# P)^{\leq |x^{(i)}|}(x^{(i)}) = 1, \quad x^{(i)} := x_q a^{(i)} \eos
\end{equation*}
namely, $a^{(1)}$ gets the highest likelihood.
An end-of-sentence symbol $\eos$ is appended to exclude special cases, e.g.\@ an incorrect answer is the prefix of the correct answer.
Suppose we estimate each of the four likelihoods with the estimator (\ref{eq. Monte-Carlo estimator}) and each $i$ is assigned with a budget of $n_i$ samples (or thoughts).
Then, the chance of answering correctly becomes $\prob[\text{argmax}_i ~ q_i  = 1]$, where $q_i$ are the estimators for the probabilities $p_i$
\begin{align*}
p_i &:= (\proj\# P)^{\leq |x^{(i)}|}(x^{(i)}), \quad q_i := \frac{1}{n_i}\sum_{j=1}^{n_i} \frac{P^{\leq |Z^{(i,j)}|}(Z^{(i,j)})}{Q(Z^{(i,j)} | x^{(i)})}, \quad \{ Z^{(i,j)}\}_{j=1}^{n_i} \iidsample Q(\cdot | x^{(i)})
\end{align*}
If each $Z^{(i,j)}$ is interpreted as a candidate derivation of answer $a_i$,
then $Q(\cdot|x)$ is one's solution strategy that produces candidate derivations,
and $P$ is a discriminator that judges whether a derivation obeys mathematical logic.
By Chebyshev's inequality, the error probability can be bounded by
\begin{equation*}
\prob[\text{argmax}_i~q_i \neq 1] \leq \sum_{i=2}^4 \prob[q_1 \leq q_i] \leq \sum_{i=2}^4 \frac{\frac{p_1^2\chi^2_1}{n_1} + \frac{p_i^2 \chi^2_i}{n_i}}{(p_1-p_i)^2}, \quad \chi^2_i := \chi^2\big(Q_*(\cdot|x^{(i)})\big\|Q(\cdot|x^{(i)})\big)
\end{equation*}
Given a total budget $n \geq \sum_i n_i$, the optimal error bound becomes
\begin{equation}
\label{eq. inference error bound}
\text{error}_n = \min_{\sum_i n_i \leq n} \prob[\text{argmax}_i~q_i \neq 1] \leq \frac{4}{n} \sum_{i=2}^4 \frac{ p_1^2 \chi^2_1 + p_i^2 \chi^2_i}{(p_1-p_i)^2} + O(n^{-2})
\end{equation}
where $O(n^{-2})$ is the discretization error.
If we interpret $\chi^2\big(Q_*(\cdot|x^{(i)})\big\|Q(\cdot|x^{(i)})\big)$ as the sub-optimality of one's solution strategy compared to that of a perfect problem solver,
then (\ref{eq. inference error bound}) implies the intuitive relation that one's error rate is proportional to the sub-optimality of one's strategy, and inversely proportional to the amount of trials $n$.
\end{example}

\begin{example}[Learning]
\label{ex. learning efficiency}
Suppose one is reading a passage $x$ from a math textbook.
Assume that one's mind is parametrized by $P=P_f$ and $f=f_{\theta}$ as in Section \ref{sec. training},
and that ``knowledge gain" can be modeled by the gradient $\vb_*(x)$ (\ref{eq. target gradient}).
Since it is intractable to iterate over all possible ``interpretations" $z \in \fip(x)$ of this passage, for practical cost one can only use the estimator
\begin{equation*}
\vb_n(x) = \nabla_{\theta} L_n^{\see}(P_f) = \frac{\frac{1}{n} \sum_{i=1}^n \nabla_{\theta} \log P_f^{\leq |Z^{(i)}_{\eye}|}(Z^{(i)}_{\eye}) \frac{P_f^{\leq |Z^{(i)}_{\eye}|}(Z^{(i)}_{\eye})}{Q^{\see}(Z^{(i)}_{\eye}|X^{(i)})}} {\frac{1}{n} \sum_{i=1}^n \frac{P_f^{\leq |Z^{(i)}|}(Z^{(i)})}{Q(Z^{(i)}|X^{(i)})}}, \quad Z^{(i)} \sim Q(\cdot|x), ~ Z^{(i)}_{\eye} \sim Q^{\see}(\cdot|x)
\end{equation*}
with $L_n^{\see}$ defined by (\ref{eq. minibatch loss inquisitive}).
According to (\ref{eq. grad mse one sample}), the amount of ``misunderstanding" scales as
\begin{equation}
\label{eq. learning grad asymp}
\E\big[ \|\vb_*(x) - \vb_n(x) \|^2 \big] \asymp \frac{1}{n} \Big[ \chi^2\big( Q_*(\cdot|x) \big\| Q(\cdot|x) \big) + \chi^2\big(Q_{\eye}(\cdot|x) \big\| Q^{\see}(\cdot|x) \big) + 1 \Big]
\end{equation}
To appreciate this error, consider for instance that one way for the inference sampler $Q(\cdot|x)$ to be suboptimal is to miss the modes of the posterior $Q_*(\cdot|x)$.
Then, most of the samples $Z^{(i)} \sim Q(\cdot|x)$ tend to have small likelihood $P^{\leq |Z^{(i)}|}(Z^{(i)})$ and thus are not ``reasonable interpretations" of $x$.
It follows that the denominator of $\vb_n$ is underestimated, one becomes overly surprised by $x$, and the amount of update $\|\vb_n\|$ is excessive.
Similarly, if $Q^{\see}(\cdot|x)$ misses the modes of $Q_{\eye}(\cdot|x)$, then one misses the valuable ideas in the passage and learns little, and the norm of the numerator of $\vb_n$ is underestimated.
Worse still, if both occur, then $\vb_n$ would be a moderately long vector that points in some wrong direction
(e.g.\@ towards simply memorizing $x$ and giving up any logical effort, or relying too much on one's familiar proof routines) and quickly drives one to misunderstanding.
Hence, the suboptimality of $Q(\cdot|x)$ and $Q_{\see}(\cdot|x)$ demands one to sample more to avoid any possible failure, and makes learning inefficient.
The error estimate (\ref{eq. learning grad asymp}) captures both sources of sampling error and thus characterizes one's learning efficiency.
\end{example}

Examples \ref{ex. inference efficiency} and \ref{ex. learning efficiency} indicate that the performance and efficiency of both inference and training are determined by the quality of the samplers and the amount of sampling.
Hence, in the setting of the representation $\proj\#P_f$, we informally define slow thinking as follows.

\begin{definition}[Slow thinking]
\label{eq. slow thinking}
Define the target $V_*(p,x,\theta)$ as any operator that takes the following function $p$ as input
\begin{equation*}
\forall x \in \Sigma^*, \quad p(x|\theta) := (\proj\#P_f)^{\leq |x|}(x), \quad f=f_{\theta}
\end{equation*}
A \textit{slow thinking} algorithm is a consistent estimator of $V_*(p,x,\theta)$, within which any summation over $\fip(x)$ is approximated by sampling.
Specifically, given a fixed $\proj$, denote by $V_n(x,\theta)$ the stochastic output of a slow thinking algorithm, given total sampling size $n$.
Then,
\begin{equation*}
\forall \theta, \forall x \in \Sigma^*, \quad \lim_{n\to\infty} \text{error}\big( V_*(p,x,\theta), V_n(x,\theta) \big) = 0 \quad \text{almost surely}
\end{equation*}
where error denotes any difference function.
The \textit{slow thinking scaling law} of this algorithm is the function
\begin{equation*}
n \mapsto \E\big[\text{error}\big(V_*(p,x,\theta), V_n(x,\theta) \big) \big]
\end{equation*}
where expectation is over the randomness of $V_n$ due to sampling.
Alternatively, the scaling law may use other input variables, such as the amount of computations (FLOPs) or the actual time of implementation.
\end{definition}

In contrast, a fast thinking algorithm directly models a distribution in the observable data space instead of using any latent space, e.g.\@ the distributions $\PS_{\text{plain}}$.

Results from the previous sections can be expressed in terms of Definition \ref{eq. slow thinking} and are summarized in Table \ref{table: slow thinking scaling laws}.
It turns out that all scaling laws are proportional to terms of the form $\chi^2/n$,
and thus all errors go to zero as the sampling size $n\to \infty$.
One can extend the sample $x$ in Definition \ref{eq. slow thinking} to a pair $(x_p,x_a)$ to allow for conditional encoding or to a data distribution $P_*$.

\begin{table}[!ht]
\small
\centering
\begin{tabular}{cccc}
 \toprule 
 Task & Target $V_*$ & Error Function & Scaling Law (or Its Upper Bound) \\
 \midrule\midrule
 \makecell{Encoding} & \makecell{Likelihood\\$(\proj\#P_f)^{\leq |x|}(x)$} & Squared error & $(\proj\#P)^{\leq |x|}(x)^2 \cdot \frac{\chi^2 (Q_*(\cdot|x) \| Q(\cdot|x) )}{n}$ \\ 
 \midrule
 \makecell{Decoding\\(direct)} & \makecell{Conditional distribution\\$(\proj\#P_f)(\cdot|[x])$} & Total variation & $\sqrt{\frac{6\chi^2 ( Q_*(\cdot|x) \| Q(\cdot|x)) + 2}{n}}$ \\
 \midrule
 \makecell{Decoding\\(indirect)} & \makecell{Next-token distribution\\$(\proj\#P_f)^{\leq|x|+t+1}(\cdot|X^{(t)})$} & $L^2$ distance & $\frac{32 \chi^2 ( Q_*(\cdot|x) \| Q(\cdot|x) ) + 12}{n}$ \\
 \midrule
 Training & \makecell{Gradient\\$\nabla_{\theta}(\proj\#P_f)^{\leq |x|}(x)$} & \makecell{Difference\\norm} & 
 \makecell{$\frac{C_{\downarrow}^2 (\proj\#P_f)^{\leq|x|}(x)^2 \chi^2( Q_*(\cdot|x) \| Q(\cdot|x) )}{n}$\\$+ \frac{C_{\uparrow}^2 \chi^2( Q_{\eye}(\cdot|x) \| Q^{\see}(\cdot|x) ) + (C_{\uparrow}^2 - C_{\downarrow}^2)}{n} + O(n^{-2})$}
  \\
 \midrule
 \makecell{Multi-Choice\\Question} & \makecell{Correct choice\\$a_1$} & \makecell{Error\\probability} & $\frac{4}{n} \sum_{i=2}^4 \frac{ p_1^2 \chi^2_1 + p_i^2 \chi^2_i}{(p_1-p_i)^2}+O(n^{-2})$ \\
 \bottomrule
\end{tabular}
\caption{Slow thinking scaling laws derived in this section.
The $\chi^2$ divergences measure the inefficiency of the implemented samplers compared to the optimal samplers.
The second decoding algorithm from Section \ref{sec. decoding} is denoted by direct decoding, and the third algorithm by indirect decoding.
The first decoding algorithm is not included since its sampling size is fixed $n=1$.}
\label{table: slow thinking scaling laws}
\end{table}

\subsection{Prediction vs. Explanation}
\label{sec. prediction vs explanation}

This section studies the non-causality of the posterior distribution $Q_*$ (as well as $Q_{\eye}$) and its implication to the samplers.
A hierarchy of sampler classes is obtained, which leverage increasingly more information from the input data and eventually can reach $Q_*$ and $Q_{\eye}$.

To begin with, let us consider the simplest sampler.
Ideally, one would like to directly decode a latent sequence $z$ from the latent distribution $P$, without using any separate model or prompt
(even a separate prompt would require a separate encoding operation and thus greater cost).
Yet, the challenge is to ensure that the sampled $z$ corresponds to the provided data $x\in\Sigma^*$, namely $z\in\fip(x)$.
In general this is not guaranteed, but we can consider a special case, when $\fip$ has an easy-to-handle structure.


\begin{definition}
We say that a projection $\proj$ is ``disentangled" if its next-segment map $\nsm_x(z,a)$ depends only on the observable context $x$.
Specifically, there exists a set-valued function $\mathcal{S}$ on $\Sigma^*$
such that there is a canonical isomorphism
\begin{equation}
\label{eq. disentangled nsm}
\nsm_x(z,a) \simeq \mathcal{S}(x) \times \{a\}
\end{equation}
for any $x \in \Sigma^*$, $z\in\fip(x)$ and $a\in\Sigma$.
As a result, the set of latent sequences $\fip(x)$ can be identified with $\{x\} \times \prod_{t=1}^{|x|} \mathcal{S}(x_{<t})$.
\end{definition}

That $\{a\}$ on the right of (\ref{eq. disentangled nsm}) is indispensable, since the latent segments in $\nsm_x(z,a)$ must map to $a$, and thus the information of $a$ must get involved in some way.
One can check that the projections of both Examples \ref{ex. product} and \ref{ex. think} are disentangled.
For instance, Example \ref{ex. think} can set $\mathcal{S}(x)$ to be the thoughts $\mathcal{T}_{c_{|x|+1}}$.
Meanwhile, the projection from Remark \ref{remark. injective} is not disentangled, since $\nsm_a(a,a)=\{\varnothing\} \neq \{a\} = \nsm_a(ba,a)$.

Given a disentangled projection, we can denote each $z \in \fip(x)$ by $((s_t,x_t))_{t=1}^{|x|}$ with $s_t\in \mathcal{S}(x_{<t})$.
As an abuse of notation, we denote the next-segment distribution by
\begin{align*}
P^{\leq |x|} \big( s_{|x|+1} \bigm| ((s_t, x_t))_{t=1}^{|x|} \big) &= \frac{ \sum_{a \in \Sigma} P\big(\big[ ((s_t, x_t))_{t=1}^{|x|} (s_{|x|+1},a) \big]\big)}{\sum_{s \in \mathcal{S}(x)} \sum_{a \in\Sigma} P\big(\big[ (s_t, x_t)_{t=1}^{|x|} (s, a) \big]\big)} \\
\text{where} ~~ P\big(\big[ (s_t,x_t)_{t=1}^{|x|} \big]\big) &:= P([z]) = P^{\leq |z|}(z)
\end{align*}
Thus, a sample $Z \in \fip(x)$ can be obtained by autoregressively sampling $P$
\begin{equation*}
Z=((S_t,x_t))_{t=1}^{|x|}, \quad S^{(t)} \sim P^{\leq t} \big(\cdot\big|((S_{\tau}, x_{\tau}))_{\tau=1}^{t-1}\big) ~~\text{for}~ t=1,\dots |x|
\end{equation*}
Then, Assumption \ref{assume. sec sampling} implies that $Z\in\fip(x)$ almost surely.
It follows that the following conditional distribution defines a sampler, $Q_{\id} \in \Q_{\proj}$.
\begin{equation}
\label{eq. identity sampler}
Q_{\id}(\cdot|x) := \law(Z), \quad Q_{\id}(z|x) = \prod_{t=1}^{|x|} P^{\leq t} \big( s_t \bigm| ((s_{\tau}, x_{\tau}))_{\tau=1}^{t-1} \big)
\end{equation}
In fact, $Q_{\id}(\cdot|x)$ is supported on all of $\fip(x)$ if $\sprt P = \dom(\proj)$.
We will see in Section \ref{sec. deepseek sampling} that the samplers of existing reasoning models such as DeepSeek-R1 \cite{guo2025deepseek} and Quiet-STaR \cite{zelikman2024quietstar} are all constructed in this way.

\begin{definition}
\label{def. identity sampler}
Given a disentangled projection, we call the sampler $Q_{\id}$ the \textit{identity sampler}.
\end{definition}

The name identity emphasizes that $Q_{\id}$ does not involve any model other than $P$ or any initial prompt other than the empty string $\varnothing$, i.e.\@ at $t=1$ in (\ref{eq. identity sampler}), the condition is empty.
Note that the identity sampler is ``strictly causal" with respect to its input $x$: Consider the random sequence $S$
\begin{equation}
\label{eq. random disentangled sequence}
S_1 \dots S_{|x|} = S, \quad (S,x) \simeq Z, \quad Z \sim Q_{\id}(\cdot|x), \quad S_t \in \mathcal{S}(x_{<t})
\end{equation}
By (\ref{eq. identity sampler}), the random variable $S_{\leq t}$ does not depend on the ``present and future" information $x_{\geq t}$.
We generalize the identity sampler to capture this causality.

\begin{definition}
\label{def. predictive sampler}
Given a disentangled projection, a sampler $Q\in\Q_{\proj}$ is called a \textit{predictive sampler} if for any $x\in\Sigma^*$ and $t=1,\dots |x|$, the random latent sequence $S_{\leq t}$ defined by (\ref{eq. random disentangled sequence}) and $Q$ does not depend on $x_{\geq t}$.
Denoted the set of predictive samplers by $\Q_{\proj}^{\text{\faForward}}$.
\end{definition}

To simplify the math and make our arguments more general, we study a larger class of causal samplers that may depend on the ``present" information $x_t$.
This is fair, since if we can show that this larger class is inadequate for some goal, then the predictive samplers cannot achieve this goal either.

Recall from Lemma \ref{lemma. segments prefix-free} that any latent sequence $z\in\fip(x)$ can be uniquely decomposed into a list of segments $z^{(1)}\dots z^{(|x|)}$ such that each segment $z^{(t)}$ is responsible for producing the token $x_t$, namely $z^{(t)} \in \nsm_{x_{<t}}(z^{(<t)},x_t)$.
Thus, we use the notation $Z^{(\leq t)}$ whenever $Z$ is a random variable that ranges in $\fip(x)$.

The following definition does not assume that $\proj$ is disentangled.

\begin{definition}
\label{def. causal sampler}
A sampler $Q \in \Q_{\proj}$ is called a \textit{causal sampler} if for any $x\in\Sigma^*$ and $1\leq t < |x|$,
given the random variable $Z\sim Q(\cdot|x)$,
the prefix $Z^{(\leq t)}$ does not depend on $x_{>t}$.
Denoted the set of causal samplers by $\Q_{\proj}^{\text{\faPlay}}$.
\end{definition}

A characterization of causal samplers is provided by the following proposition.
The projected distribution in (\ref{eq. causal sampler identity}) can be equivalently written as a marginal distribution as in (\ref{eq. posterior early marginal}).
\begin{proposition}
\label{prop. causal sampler condition}
Given a sampler $Q\in \Q_{\proj}$, the following conditions are equivalent
\begin{enumerate}
\item $Q$ is a causal sampler.
\item For any $x,x'\in \dom(Q)$ such that $x \sqsubseteq x'$,
\begin{equation}
\label{eq. causal sampler identity}
Q(\cdot|x) = \Pi_{\fip(x)}\# Q(\cdot|x')
\end{equation}
where $\Pi_{\fip(x)}$ is the partial function
\begin{equation*}
\Pi_{\fip(x)} = \big\{ (z,z') \bigm| z\in\Omega^*, ~ z' \in \fip(x), ~z' \sqsubseteq z \big\}
\end{equation*}
\item There exist a conditional distribution $\tilde{Q}$ (a partial function from $\Sigma^*\times\Omega^*$ to $\PS(\Omega^*)$) such that for any $x\in\Sigma^*$ and $z\in\fip(x)$,
\begin{equation}
\label{eq. causal factorization}
Q(z|x) = \prod_{t=1}^{|x|} \tilde{Q}\big(z^{(t)} \bigm| x_{\leq t}, z^{(<t)}\big)
\end{equation}
\end{enumerate}
\end{proposition}

The major drawback of causal samplers is that in general they are far from the variance-minimizing sampler, namely the posterior $Q_*$.
This is demonstrated by the following two results.

\begin{theorem}
\label{thm. posterior not causal}
Consider the subset of probability measures $P \in \PS(\Omega^{\omega})$ whose posterior sampler $Q_*$ is a causal sampler,
\begin{equation*}
\PS_{\proj}^{\text{\faPlay}} := \{P\in\PS(\Omega^{\omega}) \mid Q_* \in \Q_{\proj}^{\text{\faPlay}} \}
\end{equation*}
Suppose $\proj$ is non-trivial such that $|\fip(x_0)|>1$ for some $x_0\in\Sigma^*$.
For every $t$ greater than some finite threshold $t_0$,
the set of marginal distributions
\begin{equation*}
\PS_{\leq t}^{\text{\faPlay}} = \{P^{\leq t} \mid P \in \PS_{\proj}^{\text{\faPlay}} \} \subseteq \PS(\Sigma^t)
\end{equation*}
has measure zero with respect to the $(|\Sigma|^t-1)$-dimensional Lebesgue measure of the probability simplex $\PS(\Sigma^t)$.
\end{theorem}

\begin{proposition}
\label{prop. approximable by causal}
For any sampler $Q \in \Q_{\proj}$,
if $Q \notin \Q_{\proj}^{\text{\faPlay}}$, then
\begin{equation*}
\inf \big\{ \sup_{x\in\dom(Q)} \chi^2\big(Q(\cdot|x)\big\|Q'(\cdot|x)\big) \bigm| Q' \in \Q_{\proj}^{\text{\faPlay}}, ~\dom(Q) \subseteq \dom(Q')\big\} > 0
\end{equation*}
\end{proposition}

Basically, Theorem \ref{thm. posterior not causal} indicates that the posterior $Q_*$ is ``almost surely" not a causal sampler,
and then Proposition \ref{prop. approximable by causal} shows that $Q_*$ cannot even be approximated by causal samplers with respect to the $\chi^2$ divergence.
Since the inquisitive sampler $Q_{\eye}$ (\ref{eq. inquisitive sampler}) has similar form as the posterior $Q_*$, one can expect that the optimal sampler for training is ``almost surely" non-causal as well.
Since Table \ref{table: slow thinking scaling laws} indicates that the slow thinking scaling laws all depend on $\chi^2(Q_*(\cdot|x)\| Q(\cdot|x))$,
one may conclude that causal samplers (and thus predictive samplers) are inadequate for slow thinking in terms of sample efficiency.

\begin{definition}
We will often refer to $\Q_{\proj}$ as \textit{explanatory samplers}, to emphasize that the causal samplers $\Q_{\proj}^{\text{\faPlay}}$ is only a proper subset of $\Q_{\proj}$.
\end{definition}


Hence, we have established the following hierarchy
\begin{equation*}
\begin{array}{c}
\text{The identity sampler} \\
\{Q_{\id}\}
\end{array} \subseteq
\begin{array}{c}
\text{Predictive samplers} \\
\Q_{\proj}^{\text{\faForward}}
\end{array} \subseteq
\begin{array}{c}
\text{Causal samplers} \\
\Q_{\proj}^{\text{\faPlay}}
\end{array} \subseteq
\begin{array}{c}
\text{Explanatory samplers} \\
\Q_{\proj}
\end{array}
\end{equation*}
In general, each containment is proper, and in particular, $Q_*,Q_{\eye} \in \Q_{\proj} \backslash \Q_{\proj}^{\text{\faPlay}}$.

Besides the rigorous proofs, the difference between predictive and explanatory samplers can be illustrated by intuitive examples.
Example \ref{ex. cat trajectory} implies that only explanatory samplers have the sufficient information for approximating the posterior sampler $Q_*$.
An analog of Example \ref{ex. cat trajectory} in the text modality will be provided in Figure \ref{fig: compute pi}.
Meanwhile, below is a simpler example for texts.

\begin{example}
\label{ex. predictive vs explanatory}
Consider the following two texts, which are identical from the perspective of a human reader.
\begin{align*}
x_1=& \text{``}\dots \text{by Tychonoff's theorem,} ~ \Omega^{\omega} ~ \text{is compact} \dots \text{"} \\
x_2=& \text{``}\dots \Omega^{\omega} ~ \text{is compact, by Tychonoff's theorem} \dots \text{"}
\end{align*}
Suppose the projection of Example \ref{ex. think} is adopted, so the latent sequences $\fip(x)$ of each text $x$ is $x$ inserted with thoughts $\sot y \eot$, $y \in \Sigma^*$.
Suppose the latent distribution $P$ is an autoregressive model (\ref{eq. softmax conditional}), so the latent sequences $z \in \fip(x)$ with high probability $P^{\leq |z|}(z)$ are the ones that can be easily understood while being read uni-directionally.
In other words, these $z$ are the modes of $Q_*(\cdot|x)$.
Since the logic of the text $x_1$ is presented linearly, the latent sequence $z=x_1$ is uni-directionally understandable, so it does not matter whichever sampler we use.
However, for $x_2$, a latent sequence of the following form might be necessary
\begin{align*}
z= \text{``} \dots \textcolor{purple}{\sot\text{by Tychonoff's theorem}\eot} ~ \Omega^{\omega} ~ \text{is compact, by Tychonoff's theorem} \dots \text{"}
\end{align*}
For explanatory samplers, generating the thought in $z$ is easy, as they can see the entire text $x_2$ while doing this.
For predictive samplers, however, what is visible to them is only the preceding context ``\dots", so they have to come up with the logic on their own, which may have a much smaller chance.
\end{example}

As a result, the explanatory samplers are generally more efficient than the handicapped predictive samplers, and this advantage can manifest in all aspects, from better understanding the users' prompts to extracting more knowledge from training data.

Still, one interesting subtlety is that humans solve Example \ref{ex. predictive vs explanatory} in a slightly different manner.
When reading $x_2$, we can first attend to ``by Tychonoff's theorem" and then turn to ``$\Omega^{\omega}$ is compact".
So the human agency makes our perception more flexible than autoregressive models.
The modeling of this agency will be discussed in Section \ref{sec. perceptual}.

\begin{remark}
The names ``predictive/explanatory samplers" come from the notions of ``forward predictive inference" and ``backward explanatory inference" in cognitive psychology \cite{van2001effects,yeari2017online} and ``inference to the best explanation" in philosophy \cite{lipton2004inference,davey2024inferring}.
A human reader may perform forward prediction to infer the possible subsequent sentences of the currently read sentence, or backward explanation to try to understand a sentence based on the entire article as context.
One related notion is ``deductive vs.\@ abductive reasoning" \cite{josephson1996abductive,campos2011distinction},
such that abduction is the logical derivation of the probable causes of some observations.
We do not use these names since the sampling $Z\sim Q(\cdot|x)$, as a general stochastic procedure,
is not limited to the special case of logical inference.
\end{remark}

The non-causal behavior of the posterior $Q_*$ might be of independent interest, so we give it a name below.

\begin{definition}[Posterior drift]
\label{def. posterior drift}
By Theorem \ref{thm. posterior not causal}, given a latent distribution $P\in\PS(\Omega^{\omega})$, an observation sequence $x\in\Sigma^*$ and an earlier time $1\leq t < |x|$,
the marginal of the posterior distribution $Q_*(\cdot|x)$ over the earlier latent sequences $\fip(x_{\leq t})$ is generally different from the posterior $Q_*(\cdot|x_{\leq t})$ given only the ``incomplete information" $x_{\leq t}$.
\end{definition}

\begin{example}
This posterior drift can be highly non-causal in the sense that for any $t_0$ and arbitrarily large $t_2 > t_1 \gg t_0$, there exist $P$ and $x\in\Sigma^{\geq t_2}$ such that
\begin{equation*}
\Pi_{\fip(x_{\leq t_0})} \# Q_*(\cdot|x_{\leq t_1}) \neq \Pi_{\fip(x_{\leq t_0})} \# Q_*(\cdot|x_{\leq t_2})
\end{equation*}
As an informal example, if a novel ends with an incredible twist, readers may reasonably suspect that the author has already planned for this in the beginning.
Then, one can set $t_0,t_1,t_2$ to be the opening, the paragraph before the twist, and the ending.
So the posterior distribution of the author's early intention at $t_0$ is updated by a very late observation between $t_1$ and $t_2$.
This intuition is in accordance with Example \ref{ex. night heron} and Panel 4 of Example \ref{ex. cat trajectory}.
(A formal example can be constructed based on the projection of Remark \ref{remark. injective}.)
\end{example}

\begin{remark}
There may be a misunderstanding that the explanatory samplers are not applicable to inference,
as these non-causal samplers seem to require the follow-up text that has not been decoded yet.
This is not the case as demonstrated by the decoding methods in Section \ref{sec. decoding}.
The sampler $Q$ is only involved in the ``prefill stage" when the prompt $x$ is encoded, while the decoding stage simply uses the latent model $P_f$ to decode a latent sequence $Z^{(t)}$.
Thereby, the scaling law for decoding (Proposition \ref{prop. decode error bound}) only involves $\chi^2(Q_*(\cdot|x)\|Q(\cdot|x))$, the divergence over $x$, and non-causality takes place only within $x$.
Intuitively, we are modeling a reader who first reads the text $x$ thoroughly (possibly in a nonlinear order), and then starts writing uni-directionally, as in Example \ref{ex. night heron}.
\end{remark}

\subsection{Summary}
\label{sec. sampling summary}

This section has demonstrated that the basic operations of the model $\proj\#P_f$, namely encoding, decoding and training, can be and perhaps must be implemented through the sampling of the latent sequences $\fip(x)$.
Thus, a slow thinking model is determined by the tuple $(\Sigma,\Omega,\proj,P,Q)$,
namely the observable vocabulary, the latent vocabulary, the projection $\proj\colon \Omega^{\omega}\rightharpoonup\Sigma^{\omega}$, the latent distribution, and the (inference) sampler.
Optionally, one can include the train sampler $Q^{\see}$, and also specify the parametrization of $P$ and the samplers.
Then, encoding can be performed by (\ref{eq. Monte-Carlo estimator}), conditional encoding by (\ref{eq. conditional estimator}) and Remark \ref{remark. reduce bias of quotient},
decoding by either (\ref{eq. decode stream}) or (\ref{eq. resampling posterior}),
training by either (\ref{eq. minibatch loss inquisitive}, \ref{eq. minibatch sampler}, \ref{eq. train sampler reward}) in the two-sampler setting or (\ref{eq. minibatch loss one sampler}, \ref{eq. minibatch sampler}) in the one-sampler setting.

The efficiency of these operations is determined by the alignment between the implemented samplers and the optimal samplers, and this is characterized by a family of slow thinking scaling laws (Table \ref{table: slow thinking scaling laws}).
The optimal training sampler is distinct from the optimal inference sampler, which mirrors the exploration vs.\@ exploitation problem.
These optimal samplers are generally non-causal, so current slow thinking models, which rely on predictive samplers, are suboptimal in terms of sampling variance, and can be improved by using explanatory samplers for both training and prefill.
The hierarchy of the samplers, in particular the gap between current practice and optimality, is illustrated in Figure \ref{fig: sampler hierarchy}.


\section{Unified Objective}
\label{sec. unified objective}

So far we have constructed the static theory for slow thinking based on two objectives, the optimization of both approximation ability and sampling efficiency.
It is conceptually possible, however, to derive the entire static theory from one objective, the maximization of the rate of uncertainty reduction in real time.
This objective will also pave the way to the general theory of active lifting.
Different from the other sections, this section mainly uses informal arguments.



Let $P_* \in \PS(\Sigma^*)$ be any target distribution;
for convenience, assume that it has finite entropy and prefix-free support.
Let $P_{\theta}$ be an abstract model, parametrized by an abstract parameter $\theta$.
Assume that $P_{\theta}$ ranges in some abstract subspace of $\PS(\Sigma^{\omega})$ that includes the plain parametrization $P_f$, the projection parametrization $\proj\#P_f$ for various simple projections, and potentially many other possible parametrizations.
As usual, the training objective is cross entropy
\begin{equation}
\label{eq. reducible uncertainty}
C(\theta) = -\int \log P_{\theta}^{\leq |x|}(x) dP_*(x) - H(P_*)
\end{equation}
The entropy $H(P_*)$ is included so that the minimum value of $C$ becomes $0$ by Proposition \ref{prop. cross entropy to KL}.
Thus, $C$ can be interpreted as the amount of ``reducible uncertainty" of this learning task, apart from the irreducible uncertainty $H(P_*)$.

Denote the update steps by $s$, and the parameter at each step by $\theta_s$.
Assume that the update trajectory $(\theta_s)_{s=0}^{\infty}$ ranges in some admissible set $\Theta$
(e.g.\@ with fixed initialization $\theta_0$, and transitions $\theta_s\to\theta_{s+1}$ constrained by some update rules).
Denote the actual elapsed time by $t$.
Given any update trajectory $(\theta_s)_{s=0}^{\infty}$, denote the time interval of each step by $(t_{s-1},t_s]$ (with $t_0=0$)
and the step function by $\text{step}(t) = \sum_{s=1}^{\infty} \mathbbm{1}_{t > t_s}$.

Given some time limit $0<T<\infty$, a more realistic objective can be
\begin{equation*}
\min_{(\theta_s)_{s=0}^{\infty} \in \Theta} C\big( \theta_{\text{step}(T)} \big)
\end{equation*}
So the objective is to reduce the loss as much as possible within some time limit.
If $T$ is unknown \textit{a priori}, we can consider the following loss whose integral $\int_0^{\infty} dt$ represents the uninformative prior
\begin{align}
\label{eq. unified objective}
\int_0^{\infty} C( \theta_{\text{step}(t)} ) dt &= \sum_{s=0}^{\infty} C( \theta_s ) \cdot (t_{s+1}-t_s) \\
\nonumber
&= \sum_{s=1}^{\infty} t_s \cdot \big( C( \theta_{s-1} ) - C(\theta_s)\big) + \begin{cases}
0, ~~ \text{if} ~ \lim\limits_{s\to\infty}C(\theta_s)=0 \\
\infty, ~~ \text{else}
\end{cases}
\end{align}
where we assume that $\lim_{s\to\infty}t_s=\infty$.
This objective urges the model to reduce $C$ as rapidly as possible with respect to the real time.

\begin{figure}[!ht]
\centering
\includegraphics[width=0.85\linewidth]{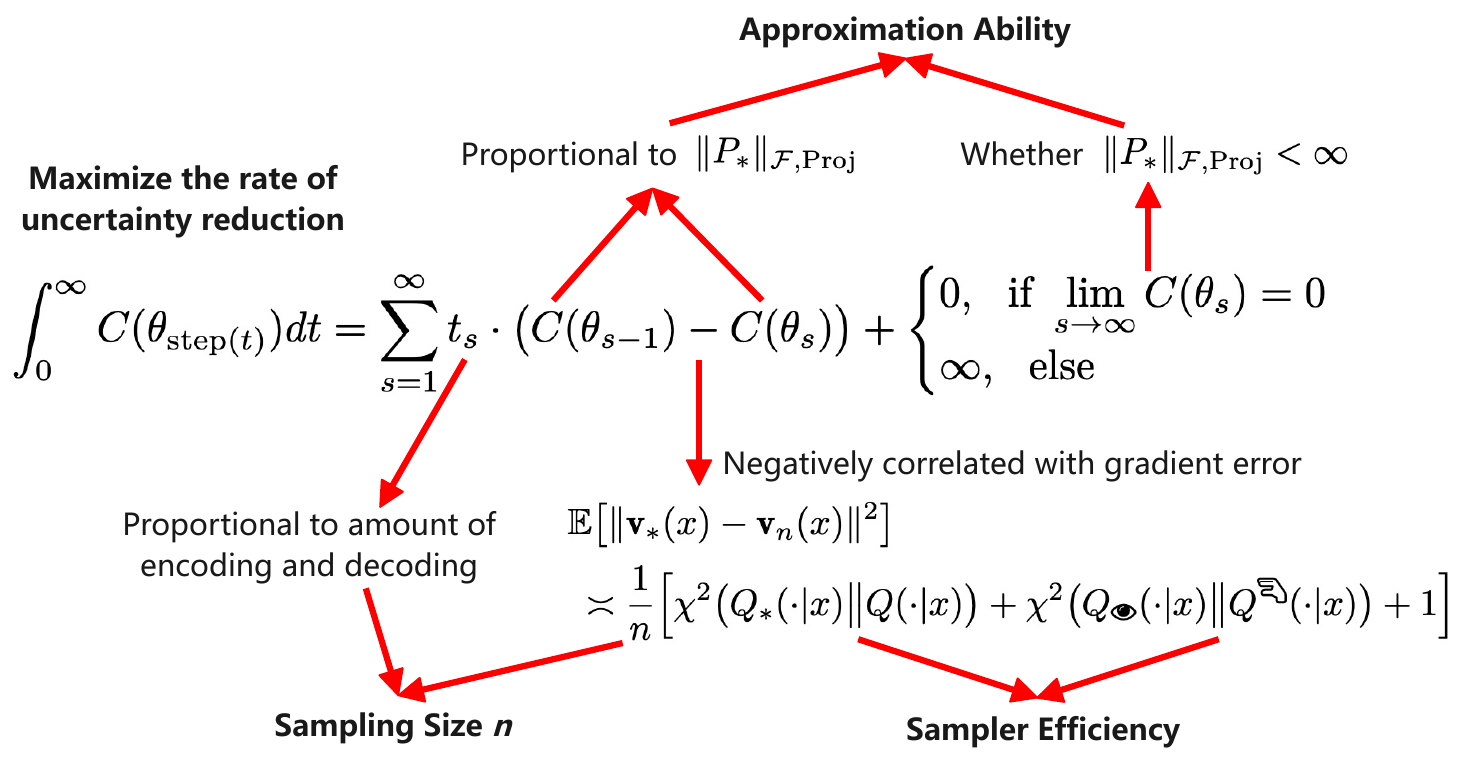}
\caption{Illustration of how the objective (\ref{eq. unified objective}) leads to the need of approximation ability and sampler efficiency, as well as the choice of sampling size.}
\label{fig: unified objective}
\end{figure}

The two objectives of the static theory can be derived from this abstract objective.
A sketch of argument is provided by Figure \ref{fig: unified objective}.
First of all, the abstract model $P_{\theta}$ should prefer parametrizations with greater approximation ability, e.g.\@ the model $\proj\#P_f$ from $\PS_{\text{simple}}$ instead of $P_f$ from $\PS_{\text{plain}}$
(these subspaces are defined in Section \ref{sec. approximation summary}).
Since the former is a strictly larger subspace according to the representation hierarchy, its terminal loss is more ``likely" to be zero
\begin{align*}
& \prob\big[ \min_{(\theta_s)_{s=0}^{\infty} \in \Theta_{\text{simple}}} \lim_{s\to\infty} C(\theta_s) = 0 \big] > \prob\big[ \min_{(\theta_s)_{s=0}^{\infty} \in \Theta_{\text{plain}}} \lim_{s\to\infty} C(\theta_s) = 0 \big] \\
& \Theta_{\text{simple}} = \big\{ (\theta_s)_{s=0}^{\infty} \in \Theta \bigm| \forall s, ~ P_{\theta_s} \in \PS_{\text{simple}} \big\}, \quad \Theta_{\text{plain}} = \big\{ (\theta_s)_{s=0}^{\infty} \in \Theta \bigm| \forall s, ~ P_{\theta_s} \in \PS_{\text{plain}} \big\}
\end{align*}
where informally the target $P_*$ is some random variable in $\PS(\Sigma^*)$.
Thus, the objective (\ref{eq. unified objective}) is less ``likely" to be $\infty$ when using the projection parametrization.
Even if the terminal losses are zero for both $\PS_{\text{simple}}$ and $\PS_{\text{plain}}$,
the estimate (\ref{eq. train loss upper bound}) indicates that the training error is generally proportional to the norm of $P_*$ with respect to the parametrization of the model.
As our separation theorems have justified the norm comparison (\ref{eq. norm separation}), it implies that $\PS_{\text{simple}}$ may enjoy much smaller finite-time losses than $\PS_{\text{plain}}$, for large steps $s$ asymptotically.

Secondly, whenever possible, the training process should adopt approximate computations that save much time with little deterioration in performance.
In our setting, this means reducing the elapsed time $t_s$ while maintaining the uncertainty reduction $C(\theta_{s-1}) - C(\theta_s)$.
Latent sampling helps to reduce each $t_{s+1}-t_s$ from something superpolynomial in the input length $|x|$ to the practical $O(n|x|^2)$ with sampling size $n$
(The computation complexity of Transformers is $O(|x|^2)$.
Assume that the projection is chosen such that $\sup_{x\in\Sigma^{\geq 1}}\max_{z\in\fip(x)} |z|/|x| < \infty$, e.g.\@ for Example \ref{ex. think} this means that the constants $c_t$ are bounded).
Meanwhile, approximating the optimal samplers ($Q\to Q_*$ and $Q^{\see}\to Q_{\eye}$) helps to reduce the estimation error of the training gradient, and thus alleviate the loss in the gain $C(\theta_{s-1}) - C(\theta_s)$.


Thirdly, the loss (\ref{eq. unified objective}) may help to determine many details of the implementation of $\proj\#P_f$.
Choices such as the following are difficult to decide because they simultaneously increase (or decrease) both accuracy and time cost:
\begin{itemize}
\item The amount of sampling $n$ over the latent sequences $\fip(x)$, for the training of both the latent (\ref{eq. minibatch loss inquisitive}) and the sampler(s) (\ref{eq. minibatch sampler}).
\item For projections similar to Example \ref{ex. think}, how frequently should the sampler insert non-empty thoughts $y^{(t)}$, and how long should each thought be.
This can be interpreted as the balancing of fast and slow thinking.
\item Whether to use two samplers $Q,Q^{\see}$ (\ref{eq. minibatch loss inquisitive}) or just one (\ref{eq. minibatch loss one sampler}).
\item Whether the two samplers share the same model, and whether the sampler and the latent distribution share the same model, and so on.
\end{itemize}
If we manage to optimize these variables in an end-to-end fashion,
then the optimizers of (\ref{eq. unified objective}) may reproduce many human traits.
For instance, the optimized sampler can automatically and flexibly switch to fast thinking (not inserting thoughts $y^{(t)}$) when reading easy texts, and slow thinking otherwise (producing long thoughts, conditioned on a large portion of the text $x$),
and think for more trials (larger $n$) when dealing with difficult or multifaceted problems.
A more concrete example of using the objective (\ref{eq. unified objective}) to balance fast and slow thinking will be provided in Section \ref{sec. balance fast slow}.



\section{Derivation and Improvement of Existing Models}
\label{sec. to existing models}

To show that our theory encompasses the slow thinking models used in practice, this section derives the design, training and inference of DeepSeek-R1 \cite{guo2025deepseek} in a first-principles manner, up to implementation-level specifics.
We also discuss possible ways to improve the model, including the optimal samplers
and a more expressive form of slow thinking suitable for pretraining.

Basically, a slow thinking model is determined by three factors: Its projection $\proj$ (or equivalently, lifting) and its positions in the representation hierarchy (Figure \ref{fig: sampler hierarchy}) and sampler hierarchy (Figure \ref{fig: representation hierarchy}).
The combination for DeepSeek-R1 consists of a $\proj$ that resembles Example \ref{ex. think}, the ``forgetful latent" that will be defined, and the identity sampler (Definition \ref{def. identity sampler}).
This derivation from our general theory is a bit lengthy compared to a direct derivation in the particular setting of DeepSeek-R1.
However, such detour makes it easier to improve the model by climbing the sampler hierarchy and representation hierarchy.
There will be three stages of improvements, ranked by the amount of modifications:
ascending the sampler hierarchy, then the representation hierarchy, and transcending the static theory.

\subsection{Derivation of Representation}
\label{sec. deepseek representation}

To begin with, we specify the tuple $(\Sigma,\Omega,\proj,P,Q)$ that determines a slow thinking model (cf.\@ Section \ref{sec. sampling summary}).


Let $\Sigma$ be the vocabulary of DeepSeek-V3 \cite{liu2024deepseekv3}.
Assume for convenience that $\Sigma$ contains an ``end-of-sentence" token $\eos$.
Let $\sot, \eot \notin \Sigma$ be two new tokens that will be used to denote the starts and ends of the inserted thoughts.
Set $\Omega = \Sigma \cup \{\sot, \eot\}$.

Similar to Example \ref{ex. think}, define the projection as follows.
\begin{align}
\nonumber
\mathcal{T} &= \{ \sot x \eot \mid x \in\Sigma^{\leq c} \} \\
\nonumber
\dom(\proj) &= \big\{ (y^{(t)} x_t )_{t=1}^{\infty} \bigm| x_t\in\Sigma, ~ y^{(t)} \in \mathcal{T} \cup \{\varnothing\} \big\} \\
\label{eq. deepseek projection}
\proj &\colon  (y^{(t)} x_t)_{t=1}^{\infty} \mapsto (x_t)_{t=1}^{\infty}
\end{align}
where $c$ is the upper bound on thought length and DeepSeek-R1 sets $c=2^{15}=32768$.
By Proposition \ref{prop. segment-wise partial function}, $\proj\colon \Omega^{\omega}\rightharpoonup\Sigma^{\omega}$ is a continuous surjective partial function with closed domain.
For convenience, denote the set of liftings by
\begin{align}
\nonumber
\Omega^*_{\mathcal{T}} &= \bigcup_{x\in\Sigma^*}\fip(x) \\
\nonumber
&= \big\{ (y^{(t)} x_t )_{t=1}^T \bigm| T\in\N, ~ x_t\in\Sigma, ~ y^{(t)} \in \mathcal{T} \cup \{\varnothing\} \big\} \\
\label{eq. union of lifting}
&= \big\{ x^{(0)} (y^{(t)} x^{(t)} )_{t=1}^T \bigm| x^{(0)} \in \Sigma^*, ~ T\in\N, ~ x^{(t)}\in\Sigma^{\geq 1}, ~ y^{(t)} \in \mathcal{T} \big\}
\end{align}
Then, the prefix projection $\proj^{<\omega}$ (Definition \ref{def. prefix projection}) satisfies $\Omega^*_{\mathcal{T}} \subseteq \dom(\proj^{<\omega})$ and
\begin{equation*}
\proj^{<\omega}\colon  (y^{(t)} x_t)_{t=1}^T \mapsto (x_t)_{t=1}^T
\end{equation*}

We enforce that the texts and latent sequences follow a multi-round query-response format.
Define
\begin{align*}
\Sigma^{\omega}_{\text{\faCode}} &:= \big\{ (x_q^{(t)} x_r^{(t)})_{t=1}^{\infty} \bigm| x_q^{(t)}, x_r^{(t)} \in \Sigma^*_{\square} \big\} \subseteq \Sigma^{\omega} \\
\Omega^{\omega}_{\text{\faCode}} &:= \big\{ (x_q^{(t)} y^{(t)} x_r^{(t)})_{t=1}^{\infty} \bigm| x_q^{(t)}, x_r^{(t)} \in \Sigma^*_{\square}, ~ y^{(t)} \in \mathcal{T} \big\} \subseteq \Omega^{\omega} \\
\Sigma^*_{\square} &:= \{ x \eos \mid x \in \Sigma^*, ~ \eos \notin x \}
\end{align*}
Each $x_q^{(t)}$ represents a user query, $x_r^{(t)}$ represents a model response,
and $y^{(t)}$ is the thought generated by the model before responding.
Furthermore, denote their prefixes by
\begin{align*}
\Sigma^*_{\text{\faCode}} &:= \{x \in \Sigma^* \mid \exists x' \in \Sigma^{\omega}_{\text{\faCode}} , ~ x \sqsubseteq x' \},
\quad \Omega^*_{\text{\faCode}} := \{z \in \Omega^*_{\mathcal{T}} \mid \exists z' \in \Omega^{\omega}_{\text{\faCode}} , ~ z \sqsubseteq z' \}
\end{align*}
The observable data will belong to $\Sigma^*_{\text{\faCode}}$, and the latent distribution $P$ will be supported on $\Omega^{\omega}_{\text{\faCode}}$.
(In practice, the queries $x_q^{(t)}$ are provided by users and need not be modeled by $P$.
Still, putting them here helps to make things simple and does not interfere with the derivation.
The training targets for $P$ will only involve the responses $x_r^{(t)}$ and thoughts $y^{(t)}$.)

One can check that $\Omega^{\omega}_{\text{\faCode}} \subseteq \dom(\proj)$, and that $\proj$ when restricted to $\Omega^{\omega}_{\text{\faCode}}$ is surjective over $\Sigma^{\omega}_{\text{\faCode}}$ and has the form
\begin{equation*}
\proj \colon  (x_q^{(t)} y^{(t)} x_r^{(t)})_{t=1}^{\infty} \mapsto (x_q^{(t)} x_r^{(t)})_{t=1}^{\infty}
\end{equation*}
For any $x \in \Sigma^*_{\text{\faCode}}$, express it as $x = (x_q^{(t)} x_r^{(t)})_{t=1}^{T-1} x^T$ where $x^T\in\Sigma^*$ contains at most one $\eos$.
We consider only latent sequences $z\in\fip(x)$ that follow the format of $\Omega^*_{\text{\faCode}}$, which are given by
\begin{align}
\label{eq. deepseek fip}
\begin{split}
\fip(x) \cap \Omega^*_{\text{\faCode}} &=
\begin{cases}
\big\{ (x_q^{(t)} y^{(t)} x_r^{(t)})_{t=1}^T \bigm| y^{(t)} \in \mathcal{T} \big\} ~~ \text{if} ~ x^T = x_q^T x_r^T ~ \text{for some} ~ x_q^T \in \Sigma^*_{\square}, ~ x_r^{T} \neq \varnothing \\
\big\{ (x_q^{(t)} y^{(t)} x_r^{(t)})_{t=1}^{T-1} x^T \bigm| y^{(t)} \in \mathcal{T} \big\} ~~ \text{else}
\end{cases}
\end{split}
\end{align}
Thus, the next-segments are restricted to the following subset of $\nsm_x$
\begin{equation}
\label{eq. deepseek nsm}
\nsm^{\text{\faCode}}_x(z,x') = \begin{cases}
\{ yx' \mid y \in \mathcal{T} \} ~~ \text{if} ~ x^T \in \Sigma^*_{\square} ~\text{and}~ x' \neq \varnothing \\
\{x'\} ~~ \text{else}
\end{cases}
\end{equation}
for any $z \in \fip(x)$ and any $x' \in \Sigma^*$ that is the prefix of some response $x'' \in \Sigma^*_{\square}$.

We need a parametrized latent distribution that is supported on $\Sigma^{\omega}_{\text{\faCode}}$, namely $P_f\in\PS(\Sigma^{\omega}_{\text{\faCode}})$.
Equivalently, for each $z\in\sprt P_f$, its delimiters should always appear in the following order repeatedly
\begin{equation*}
\dots \eos \sot \dots \eot \dots \eos
\end{equation*}
and the length of the subsequence between each pair of $\sot,\eot$ should not exceed $c$.
Similar to Definition \ref{def. softmax with masking}, this can be achieved by masking the output logits of $f$.
For instance, if the most recent delimiter is $\sot$, then the next-token set, a subset of (\ref{eq. admissible next-token}), is $\nsm_{\text{\faCode}}^{\leq 1}(z)=\Omega\backslash\{\eos,\sot\}$ and the output logits of these two tokens are set to $-\infty$, so the next delimiter must be $\eot$.
Also, if this $\sot$ is the $c$-th token behind, then $\nsm_{\text{\faCode}}^{\leq 1}(z)=\{\eot\}$ and all logits except that of $\eot$ are set to $-\infty$, so the next token is forced to be $\eot$.

Using concatenation (\ref{eq. concatenation}), we can speak of sets like $\mathcal{T} \Sigma^*_{\square}$.
Given any latent distribution $P\in\PS(\Sigma^{\omega}_{\text{\faCode}})$, one can simply notations with the next-segment probabilities (Definition \ref{def. next-segment distribution})
\begin{align*}
(\proj\#P)^{\leq |x|+|x_r^{(T)}|}(x_r^{(T)}|x) &= (\proj\#P)(x_r^{(T)} \mid x\to\Sigma^*_{\square}) \\
P^{\leq |z|+|y^{(T)}|+|x_r^{(T+1)}|}(y^{(T)}x_r^{(T)}|z) &= P(y^{(T)}x_r^{(T)} \mid z \to \mathcal{T} \Sigma^*_{\square}) \\
&= P(y^{(T)} \mid z \to \mathcal{T}) ~ P(x_r^{(T)} \mid zy \to \Sigma^*_{\square})
\end{align*}
where $x = (x_r^{(t)}x_q^{(t)})_{t=1}^{T-1} \in \Sigma^*_{\text{\faCode}}$ is any prefix, $z \in \fip(x)$, $y^{(T)} \in \mathcal{T}$ and $x_r^{(T)}\in\Sigma^*_{\square}$.
One can check that the condition for Definition \ref{def. next-segment distribution} is satisfied.

One important simplification implemented by DeepSeek-R1 is to ignore the thoughts from earlier rounds in multi-round conversations \cite{deepseek2025multiround}.
We define this property as follows.
Note that for any $z\in\Omega^*_{\mathcal{T}}$, the intersection $[z\sot] \cap \dom(\proj)$ is nonempty and each sequence in it has the form $zyw$ for some $y\in\mathcal{T}$ and $w\in\Omega^{\omega}$.

\begin{definition}[Forgetful latent]
\label{def. forgetful}
Given the projection from (\ref{eq. deepseek projection}) and any probability measure (or latent distribution) $P\in\PS(\dom(\proj))$,
define the set $S(P) = \{z\in\Omega^*_{\mathcal{T}} \mid P^{\leq|z|+1}(z\sot)>0\}$,
and for any $z \in S(P)$ define the random variable
$$zY(z)W(z) \sim P\big(\cdot\big|[z\sot]\big)$$
where $Y(z)$ ranges in $\mathcal{T}$.
The latent distribution $P$ is called \textit{forgetful} if for any $z,z' \in S(P)$ such that $\proj^{<\omega}(z) = \proj^{<\omega}(z')$, the following holds
\begin{equation*}
\law\big(Y(z)W(z)\big) = \law\big(Y(z')W(z')\big)
\end{equation*}
In that case, we denote $P\big(\cdot\big|[z\sot]\big)$ by $P\big(\cdot\big|[\proj^{<\omega}(z)\sot]\big)$ for any $z\in\Omega^*_{\mathcal{T}}$.
In plain terms, if a latent context $z$ is followed by a nonempty thought, then this thought and the subsequent sequence depend only on the observable context $\proj^{<\omega}(z)$,
ignoring any thoughts in $z$.
So effectively the earlier thoughts are all forgotten.
\end{definition}

It follows that if $P$ is a forgetful latent,
then for any $z=x^{(0)}(y^{(t)}x^{(t)})_{t=1}^T \in \Omega^*_{\mathcal{T}}$, which is written in the format of (\ref{eq. union of lifting}), and any $y^{(T+1)} \in \mathcal{T}$ and $x^{(T+1)}\in \Sigma^*$, we have
\begin{equation}
\label{eq. forgetful latent}
P^{\leq |z|+|y^{(T+1)}|+|x^{(T+1)}|} \big( y^{(T+1)} x^{(T+1)} \bigm| z \big) = P^{\leq |x^{(\leq T+1)}|+|y^{(T+1)}|} \big( y^{(T+1)} x^{(T+1)} \bigm| x^{(\leq T)} \big)
\end{equation}
where $x^{(\leq T)} = x^{(0)}\dots x^{(T)} = \proj^{<\omega}(z)$.
If furthermore $P\in\PS(\Omega^{\omega}_{\text{\faCode}})$,
then for any latent sequence $z \in \Omega^*_{\text{\faCode}}$ in the form $z=(x_q^{(t)}y^{(t)}x_r^{(t)})_{t=1}^T x_q^{(T+1)}$, we have
\begin{equation}
\label{eq. forgetful latent format}
P^{\leq |z|} \big( y^{(T)} x_r^{(T)} x_q^{(T+1)} \bigm| (x_q^{(t)}y^{(t)}x_r^{(t)})_{t=1}^{T-1} x_q^{(T)} \big) = P^{\leq |x|+|y^{(T)}|} \big( y^{(T)} x_r^{(T)} x_q^{(T+1)} \bigm| (x_q^{(t)}x_r^{(t)})_{t=1}^{T-1} x_q^{(T)} \big)
\end{equation}
where $x=(x_q^{(t)}x_r^{(t)})_{t=1}^T x_q^{(T+1)} = \proj^{<\omega}(z)$.
In terms of LLM inference, given the conversation history $(x_q^{(t)}y^{(t)}x_r^{(t)})_{t=1}^{T-1} x_q^{(T)}$,
a forgetful latent model $P_f$ encodes only the observable context $(x_q^{(t)}x_r^{(t)})_{t=1}^{T-1} x_q^{(T)}$ in the prefill stage, and then generates the thought $y^{(T+1)}$ and response $x_r^{(T+1)}$.
Specifically, the implementation of DeepSeek-R1 is to delete all $y^{(t)}$ and discard all key-value cache, and then after receiving the query $x_q^{(T)}$, encode $(x_q^{(t)}x_r^{(t)})_{t=1}^{(T)} x_q^{(T)}$ from scratch \cite{deepseek2025multiround}.
The reason for removing earlier thoughts is not mentioned in \cite{deepseek2025multiround}, but the possible rationale is to avoid the compute cost and potential degradation of performance due to excessively long contexts \cite{liu2024lost,barbero2024glasses}.

For a more concrete illustration of forgetfulness, one can imagine the first query $x_q^{(1)}$ of the user to be a math problem,
the thought $y^{(1)}$ generated by the LLM to be a detailed proof, and its reply $x_r^{(1)}$ to be the final answer.
If in the second query $x_q^{(2)}$, the user asks for clarification of the proof steps, then since $y^{(1)}$ has been discarded, the LLM would have to generate a new proof in order to answer the query.

Recall that the distribution parametrization $P_f$ is defined by the tokenwise conditional distributions (\ref{eq. softmax conditional}).
As an abuse of notation, if we say that $P_f$ is forgetful or satisfies (\ref{eq. forgetful latent}),
we are assuming that, whenever the conditional probability (\ref{eq. softmax conditional}) is computed for any token in a segment of the form $y^{(T)}x_T$, the earlier thoughts are always set to $y^{(t)}=\varnothing$ when the context $(y^{(t)}x_t)_{t=1}^{T-1}$ is input to $f$.
So this is a slight modification of the default distribution parametrization (Definition 
\ref{def. softmax with masking}).


Forgetfulness simplifies the computation of likelihoods.
For any context $x = (x_q^{(t)}x_r^{(t)})_{t=1}^T x_q^{(T+1)}$,
Theorem \ref{thm. conditional latent distribution} and (\ref{eq. deepseek nsm}, \ref{eq. forgetful latent format}) imply that the conditional probability of any response $x_r^{(T+1)} \in \Sigma^*_{\square}$ is given by
\begin{align}
\nonumber
(\proj\#P_f)( x_r^{(T+1)} \mid x \to\Sigma^*_{\square}) &= \int P_f \big( \nsm_x(z,x_r^{(T+1)}) \bigm| z \big) ~ dQ_*(z|x) \\
\nonumber
&= \int \sum_{y^{(T+1)}\in\mathcal{T}} P_f \big( y^{(T+1)} x_r^{(T+1)} \bigm| z \to \mathcal{T}\Sigma^*_{\square} \big) ~ dQ_* ( z | x ) \\
\label{eq. deepseek conditional sum}
&= \sum_{y^{(T+1)}\in\mathcal{T}} P_f \big( y^{(T+1)} x_r^{(T+1)} \bigm| z \to \mathcal{T}\Sigma^*_{\square} \big) \\
\label{eq. deepseek conditional int}
&= \int P_f(x_r^{(T+1)}\mid xy^{(T+1)}\to\Sigma^*_{\square}) ~ dP_f(y^{(T+1)} \mid x\to\mathcal{T})
\end{align}
Thus, a summation over $\mathcal{T}^{T+1}$ is reduced to over $\mathcal{T}$.

\subsection{Derivation of Sampling}
\label{sec. deepseek sampling}

Next, we derive the sampler $Q$ of DeepSeek-R1 as well as the posterior sampler $Q_*$.

Fix any context $x = (x_q^{(t)}x_r^{(t)})_{t=1}^T x^{(T+1)} \in \Sigma^*_{\text{\faCode}}$,
where $x^{(T+1)}$ is a possibly empty prefix of some query $x_q^{(T+1)} \in\Sigma^*_{\square}$.
By (\ref{eq. deepseek fip}), we have the canonical isomorphism
\begin{equation}
\label{eq. deepseek canonical isomorphism}
\fip(x) \cap \Omega^*_{\text{\faCode}} ~ \simeq ~ \{x\} \times \mathcal{T}^T
\end{equation}
Thus, any latent sequence $z \in \fip(x)$ that can be generated by $P_f$ (and thus is in $\Omega^*_{\text{\faCode}}$) can be identified with some $(y^{(t)})_{t=1}^T \in \mathcal{T}^T$,
and $Q_*(\cdot|x)$ can be identified as a distribution over the product space $\mathcal{T}^T$.
Forgetfulness (Definition \ref{def. forgetful}) implies that
\begin{align*}
Q_*(z|x) \propto P_f^{\leq |z|}(z) &\propto \prod_{t=1}^{T-1} P_f \big( y^{(t)} x_r^{(t)} x_q^{(t+1)} \bigm| (x_q^{(s)} y^{(s)} x_r^{(s)})_{s=1}^{t-1} x_q^{(t)} \to \mathcal{T}\Sigma^*_{\square}\Sigma^*_{\square} \big) \\
& \quad \times P_f^{\leq |z|} \big( y^{(T)} x_r^{(T)} x^{(T+1)} \bigm| (x_q^{(t)} y^{(t)} x_r^{(t)})_{t=1}^{T-1} x_q^{(T)} \big) \\
&= \prod_{t=1}^{T-1} P_f \big( y^{(t)} x_r^{(t)} x_q^{(t+1)} \bigm| (x_q^{(s)} x_r^{(s)})_{s=1}^{t-1} x_q^{(t)} \to \mathcal{T}\Sigma^*_{\square}\Sigma^*_{\square} \big) \\
& \quad \times P_f^{\leq |x|+|y^{(T)}|} \big( y^{(T)} x_r^{(T)} x^{(T+1)} \bigm| (x_q^{(t)} x_r^{(t)})_{t=1}^{T-1} x_q^{(T)} \big)
\end{align*}
So $Q_*(\cdot|x) \in \PS(\mathcal{T}^T)$ is simply a product distribution
\begin{align}
\nonumber
Q_*(\cdot|x) &= \Big( \bigotimes_{t=1}^{T-1} Q_*^t\big(\cdot\big| (x_q^{(s)} x_r^{(s)})_{s=1}^t x_q^{(t+1)} \big) \Big) \otimes Q_*^T(\cdot|x) \\
\nonumber
Q_*^t\big( y^{(t)} \big| (x_q^{(s)} x_r^{(s)})_{s=1}^t x_q^{(t+1)} \big) &= \frac{P_f \big( y^{(t)} x_r^{(t)} x_q^{(t+1)} \bigm| (x_q^{(s)} x_r^{(s)})_{s=1}^{t-1} x_q^{(t)} \to \mathcal{T}\Sigma^*_{\square}\Sigma^*_{\square} \big)}{(\proj\#P_f) \big( x_r^{(t)} x_q^{(t+1)} \bigm| (x_q^{(s)} x_r^{(s)})_{s=1}^{t-1} x_q^{(t)} \to \Sigma^*_{\square} \Sigma^*_{\square} \big)} ~~ \text{for} ~~ t<T \\
\label{eq. deepseek posterior}
Q_*^T(y^{(T)}|x) &= \frac{P_f^{\leq |x|+|y^{(T)}|} \big( y^{(T)}x_r^{(T)}x^{(T+1)} \bigm| (x_q^{(t)}x_r^{(t)})_{t=1}^{T-1} x_q^{(T)} \big)}{(\proj\#P_f)^{\leq |x|} \big( x_r^{(T)}x^{(T+1)} \bigm|(x_q^{(t)}x_r^{(t)})_{t=1}^{T-1} x_q^{(T)} \big)}
\end{align}

DeepSeek-R1 uses the identity sampler $Q_{\id}$ (Definition \ref{def. identity sampler}) for both inference and training.
The canonical isomorphism (\ref{eq. deepseek canonical isomorphism}) implies that $\proj$ satisfies the assumption of Definition \ref{def. identity sampler}, so $Q_{\id}$ is well-defined.
Then, $Q_{\id}$ can be derived by combining (\ref{eq. identity sampler}) and (\ref{eq. forgetful latent format}),
\begin{align}
\nonumber
Q_{\id}(\cdot|x) &= \bigotimes_{t=1}^T Q_{\id}^t\big(\cdot\big| (x_q^{(s)} x_r^{(s)})_{s=1}^{t-1} x_q^{(t)} \big) \\
\label{eq. deepseek sampler}
Q_{\id}^t\big( y^{(t)} \big| (x_q^{(s)} x_r^{(s)})_{s=1}^{t-1} x_q^{(t)} \big) &= P_f\big( y^{(t)} \bigm| (x_q^{(s)} x_r^{(s)})_{s=1}^{t-1} x_q^{(t)} \to \mathcal{T} \big)
\end{align}
With the identity sampler, the sampling of DeepSeek-R1 does not involve any other model or prompt,
and the thought $y^{(t)}$ is sampled directly by decoding $P_f$ conditioned on the user's latest query plus the earlier conversations.
In particular, the conditioning does not involve the response $x_r^{(t)}$.
Still, there is much similarity between (\ref{eq. deepseek posterior}) and (\ref{eq. deepseek sampler}):
Suppose the last segment $x^{(T+1)}$ is a valid query $x_q^{(T+1)} \in \Sigma^*_{\square}$ and that a response $x_r^{(T+1)} \in \Sigma^*_{\square}$ is provided.
Then,
\begin{alignat}{2}
\nonumber
Q_*^{T+1}(y^{(T+1)} & \mid xx_r^{(T+1)}) &&\propto P_f(y^{(T+1)} \mid x\to\mathcal{T}) ~ P_f(x_r^{(T+1)} \mid xy^{(T+1)}\to\Sigma^*_{\square}) \\
\label{eq. deepseek sampler last}
Q_{\id}^{T+1}(y^{(T+1)} & \mid x) &&= P_f(y^{(T+1)} \mid x\to\mathcal{T})
\end{alignat}
which differ only by one term.

Now we estimate the conditional probability $p_* := (\proj\#P_f)(x_r^{(T+1)}|x\to\Sigma^*_{\square})$ given a context $x=(x_q^{(t)} x_r^{(t)})_{t=1}^T x_q^{(T+1)} \in \Sigma^*_{\text{\faCode}}$ and a response $x_r^{(T+1)}\in\Sigma^*_{\square}$.
Recall that $p_*$ can be expressed as an integral (\ref{eq. deepseek conditional int}) over $\mathcal{T}$ and the underlying distribution is exactly $Q_{\id}^T(\cdot|x)$ (\ref{eq. deepseek sampler last}).
Therefore, an unbiased Monte-Carlo estimator can be defined simply by
\begin{align}
\label{eq. deepseek estimator}
p_n &:= \frac{1}{n} \sum_{i=1}^n P_f \big( x_r^{(T+1)} \bigm| x Y_i \to \Sigma^*_{\square}\big), \quad \{Y_i\}_{i=1}^n \iidsample Q_{\id}^{T+1}(\cdot|x)
\end{align}
Here we see another advantage of the identity sampler; while the generic Monte-Carlo estimator (\ref{eq. Monte-Carlo estimator}) consists of quotients and thus may require two encoding operations, the identity sampler helps to reduce the quotients into conditional probabilities (of $x_r^{(T+1)}$ on $x Y_i$) and thus each sample only needs to be encoded once.

Similar to the result in Section \ref{sec. sampler}, the variance of this estimator is given by
\begin{equation*}
var(p_n) = \frac{p_*^2}{n} ~ \chi^2\big( Q_*^{T+1}(\cdot|xx_r^{(T+1)}) \bigm\| Q_{\id}^{T+1}(\cdot|x) \big)
\end{equation*}
As usual, the variance-minimizing sampler is the posterior $Q_*^{T+1}$.
The training and inference of DeepSeek-R1 always use $n=1$.

\subsection{Derivation of Training}
\label{sec. deepseek training}

This section derives the training of DeepSeek-R1,
specifically, the policy gradient training of the R1-Zero model \cite{guo2025deepseek}.

For simplicity, suppose only one sample is used in the training step, instead of a data distribution $P_*$ or mini-batch.
Suppose as before that the sample consists of a conversation context $x=(x_q^{(t)} x_r^{(t)})_{t=1}^T x_q^{(T+1)} \in \Sigma^*_{\text{\faCode}}$ and a response $x_r^{(T+1)}\in\Sigma^*_{\square}$.
We want to maximize the log-likelihood $\log p_*$.

Since the training and inference of DeepSeek-R1 use the same sampler, the discussion of Section \ref{sec. training latent} indicates that the mini-batch loss for the latent distribution would be (\ref{eq. minibatch loss one sampler}).
In terms of the estimator $p_n$ (\ref{eq. deepseek estimator}), this loss can be written as
\begin{equation*}
L_n(P) = - \log\Big(\frac{1}{n}\sum_{i=1}^n q_i \Big) = - \log p_n
\end{equation*}
DeepSeek-R1 uses $n=1$, so the objective simplifies to
\begin{equation}
\label{eq. deepseek latent loss}
\min_{\theta} -\log P_f(x_r^{(T+1)} \mid xY\to\Sigma^*_{\square}), \quad Y \sim P_f(\cdot|x\to\mathcal{T})
\end{equation}
Meanwhile, the sampler $Q_{\id}$ is trained by the reverse KL divergence (\ref{eq. reverse KL sampler}) derived in Section \ref{sec. sampler}.
In the current setting, this loss becomes
\begin{align*}
\KL\big(Q_{\id}^{T+1}(\cdot|x) \bigm\| \detach(Q_*^{T+1}(\cdot|xx_r^{(T+1)}))\big)
\end{align*}
The posterior $Q_*^{T+1}$ is detached from gradient computation, since in theory it depends only on the latent distribution $P_f$ and is a static target for the inference sampler to approximate.
Yet, since $P_f$ and the identity sampler $Q_{\id}^{T+1}$ now share the same model and parameter, the detach operation should be explicitly included.
Then, by the derivation of Section \ref{sec. sampler training}, the mini-batch version of the loss would be
\begin{align*}
L_n &= - \frac{1}{n} \sum_{i=1}^n \detach(R_i) \log Q_{\id}^{T+1}(Y_i|x), \quad R_i = \log P_f(x_r^{(T+1)}\mid xY_i \to \Sigma^*_{\square}), \quad \{Y_i\}_{i=1}^n \iidsample Q_{\id}^{T+1}(\cdot|x) \\
&= - \frac{1}{n} \sum_{i=1}^n \detach(R_i) \log P_f(Y_i|x\to\mathcal{T}), \quad \{Y_i\}_{i=1}^n \iidsample P_f(\cdot|x\to\mathcal{T})
\end{align*}
With $n=1$, the objective becomes
\begin{equation}
\label{eq. deepseek sampler loss}
\min_{\theta} ~ \detach\big(-\log P_f(x_r^{(T+1)}\mid xY \to \Sigma^*_{\square}) \big) \cdot \log P_f(Y|x\to\mathcal{T}), \quad Y \sim P_f(\cdot|x\to\mathcal{T})
\end{equation}
Hence, training involves two components, a supervised finetuning objective  (\ref{eq. deepseek sampler loss}) for generating useful thoughts $Y$, and a policy gradient objective (\ref{eq. deepseek latent loss}) for producing the right answer conditioned on a thought.

Next, we consider a relaxation adopted by DeepSeek-R1,
such that the target response is generalized from a single $x_r^{(T+1)}$ to a subset $\Sigma^*_{\text{\faSmileO}} \subseteq \Sigma^*_{\square}$.
This relaxation can be used when $x$ is a closed-ended problem with a deterministic answer, so that the correctness of any response $x_r\in\Sigma^*_{\square}$ can be decided with some simple rule.
For instance, if $x$ is a math question with answer $5$, then $\Sigma^*_{\text{\faSmileO}}$ are responses that ends with ``\texttt{{\textbackslash}box\{5\}}$\eos$".
This relaxation leads to a slightly different objective.

The conditional probability becomes
\begin{align*}
p_*^{\text{\faSmileO}} &:= (\proj\#P_f)(\Sigma^*_{\text{\faSmileO}} \mid x\to\Sigma^*_{\square}) \\
&= \int P_f(\Sigma^*_{\text{\faSmileO}} \mid xy\to\Sigma^*_{\square}) dP_f(y \mid x\to\mathcal{T}) \\
&= \int \mathbbm{1}\{x_r\in\Sigma^*_{\text{\faSmileO}}\} ~ dP_f(yx_r \mid x\to\mathcal{T}\Sigma^*_{\square})
\end{align*}
The second line follows from (\ref{eq. deepseek conditional int}).
As before, the ideal objective is $L^{\fasmile}(\theta) := -\log p_*^{\text{\faSmileO}}$ with gradient
\begin{equation}
\label{eq. deepseek relaxed gradient}
\nabla_{\theta} L^{\fasmile} = - \frac{1}{p_*^{\text{\faSmileO}}} \int \mathbbm{1}\{x_r\in\Sigma^*_{\text{\faSmileO}}\} ~ \nabla_{\theta} \log P_f(yx_r \mid x\to\mathcal{T}\Sigma^*_{\square}) ~ dP_f(yx_r \mid x\to\mathcal{T}\Sigma^*_{\square})
\end{equation}
To fit this gradient, one may try to define a two-sampler mini-batch loss $L^{\fasmile}_n$ by replicating the calculation of (\ref{eq. minibatch loss inquisitive}).
The posterior sampler and inquisitive sampler that minimize the gradient estimation error $\E[ \|\nabla_{\theta}L^{\fasmile}-\nabla_{\theta}L^{\fasmile}_n\|^2 ]$ are given by
\begin{align*}
Q_*^{\mathcal{TR}}(\cdot|x) &= \mathbbm{1} \{x_r \in \Sigma^*_{\text{\faSmileO}} \} P_f(yx_r \mid x\to\mathcal{T}\Sigma^*_{\square}) / p_*^{\text{\faSmileO}} \\
Q_{\eye}^{\mathcal{TR}}(\cdot|x)
&\propto \mathbbm{1} \{x_r \in \Sigma^*_{\text{\faSmileO}} \} ~ \big\| \nabla_{\theta} \log P_f(yx_r \mid x\to\mathcal{T}\Sigma^*_{\square}) \big\| ~ P_f(yx_r \mid x\to\mathcal{T}\Sigma^*_{\square})
\end{align*}
Instead, DeepSeek-R1 fits (\ref{eq. deepseek relaxed gradient}) with a one-sampler loss and continues to employ the identity sampler $Q_{\id}$, which now becomes
\begin{equation*}
Q_{\id}^{\mathcal{TR}}( \cdot | x) := P_f( \cdot | x\to\mathcal{T}\Sigma^*_{\square})
\end{equation*}
Since this is exactly the distribution in (\ref{eq. deepseek relaxed gradient}), the one-sampler loss can be defined as
\begin{align}
\label{eq. deepseek loss relaxed}
L^{\fasmile}_n(\theta) &= - \frac{1}{\detach(p_n) n} \sum_{i=1}^n \mathbbm{1} \{X_r^{(i)} \in \Sigma^*_{\text{\faSmileO}}\} \log P_f(Y^{(i)}X_r^{(i)} \mid x\to\mathcal{T}\Sigma^*_{\square}) \\
\nonumber
\text{where} \quad p_n &= \frac{1}{n} \sum_{i=1}^n \mathbbm{1} \{X_r^{(i)} \in \Sigma^*_{\text{\faSmileO}} \} P_f(Y^{(i)} X_r^{(i)} \mid x\to\mathcal{T}\Sigma^*_{\square}) \\
\nonumber
&\{ Y^{(i)} X_r^{(i)} \}_{i=1}^n \iidsample Q_{\id}^{\mathcal{TR}}( \cdot | x) = P_f( \cdot | x\to \mathcal{T} \Sigma^*_{\square})
\end{align}
Effectively, the reward for each sample is $R_i = \mathbbm{1} \{X_r^{(i)} \in \Sigma^*_{\text{\faSmileO}} \} / p_n$, and in practice it is replaced by the simpler ``group relative" advantage
\begin{equation}
\label{eq. group relative advantage}
A_i = \frac{R_i - \text{mean}(R_1,\dots R_n)}{\text{std}(R_1,\dots R_n)} = \frac{r_i - \text{mean}(r_1,\dots r_n)}{\text{std}(r_1,\dots r_n)}, \quad r_i = \mathbbm{1} \{X_r^{(i)} \in \Sigma^*_{\text{\faSmileO}} \}
\end{equation}
To improve training stability, one can add clipping and regularization to the log-likelihoods $\log P_f$.
Then, this loss becomes the Group Relative Policy Optimization (GRPO) objective \cite{shao2024deepseekmath} of R1-Zero.

Note that
\begin{align*}
&\quad \KL\big( \detach(Q_*^{\mathcal{TR}}( \cdot | x)) \bigm\| Q_{\id}^{\mathcal{TR}}( \cdot | x) \big) + \text{constant} \\
&= - \frac{1}{p_*^{\text{\faSmileO}}} \int_{\Sigma^*_{\text{\faSmileO}}} \log P_f(yx_r \mid x\to\mathcal{T}\Sigma^*_{\square}) ~ dP_f(yx_r \mid x\to\mathcal{T}\Sigma^*_{\square}) \\
&\approx - \frac{1}{p_*^{\text{\faSmileO}} n} \sum_{i=1}^n \mathbbm{1} \{X_r^{(i)} \in \Sigma^*_{\text{\faSmileO}}\} \log P_f(Y^{(i)}X_r^{(i)} \mid x\to\mathcal{T}\Sigma^*_{\square}), \quad \{ Y^{(i)} X_r^{(i)} \}_{i=1}^n \iidsample Q_{\id}^{\mathcal{TR}}( \cdot | x)
\end{align*}
which is identical to (\ref{eq. deepseek loss relaxed}) up to coefficient.
Thus, training a slow thinking model with the loss (\ref{eq. deepseek loss relaxed}) can be interpreted as fitting the posterior sampler $Q_*^{\mathcal{TR}}$.
As analyzed in Remark \ref{remark. policy collapse}, such training may be prone to policy collapse, and it could be helpful to implement a train sampler that fits the inquisitive sampler $Q^{\mathcal{TR}}_{\eye}$ and train the model $P_f$ with its samples instead.



\subsection{Derivation of Inference}
\label{sec. deepseek inference}

The inference of DeepSeek-R1 can be derived from the first decoding algorithm of Section \ref{sec. decoding}.

Suppose as before that the conversation context is $x=(x_q^{(t)} x_r^{(t)})_{t=1}^T x_q^{(T+1)} \in \Sigma^*_{\text{\faCode}}$.
Decoding means generating a response $X_r^{(T+1)} \sim (\proj\#P_f)(\cdot|x\to\Sigma^*_{\square})$.
In this context, the method (\ref{eq. decode stream}) becomes
\begin{equation*}
\text{Return}~ X_r, \quad YX_r \sim P_f(\cdot|Z\to\mathcal{T}\Sigma^*_{\square}), \quad Z \sim Q(\cdot|x)
\end{equation*}
for any sampling distribution $Q$ over $\fip(x)$.
By forgetfulness (Definition \ref{def. forgetful}), the next-segment distribution $P_f(\cdot|Z\to\mathcal{T}\Sigma^*_{\square})$ depends only on $x$.
So the decoding process simplifies to
\begin{equation}
\label{eq. deepseek decoding}
X_r \sim P_f(\cdot|xY\to\Sigma^*), \quad Y \sim P_f(\cdot|x\to\mathcal{T})
\end{equation}
and no sampler is needed.

Hence, we have derived the training and inference of DeepSeek-R1,
with encoding given by (\ref{eq. deepseek estimator}), decoding given by (\ref{eq. deepseek decoding}),
and training by either (\ref{eq. deepseek latent loss}, \ref{eq. deepseek sampler loss}) with a prescribed target or (\ref{eq. deepseek loss relaxed}, \ref{eq. group relative advantage}) with relaxed targets.


\subsection{Stage One: Improving Sampling Efficiency}
\label{sec. quick improvement}

Next, we move on to the interesting topic of improving the model.
This section introduces the first stage that concerns the sampler hierarchy.
These easy improvements maintain the existing representation of forgetful latent, so they apply only to the training process.

Before going into the technicalities, let us consider a motivating example.

\begin{figure}[!ht]
\centering
\includegraphics[width=1\linewidth]{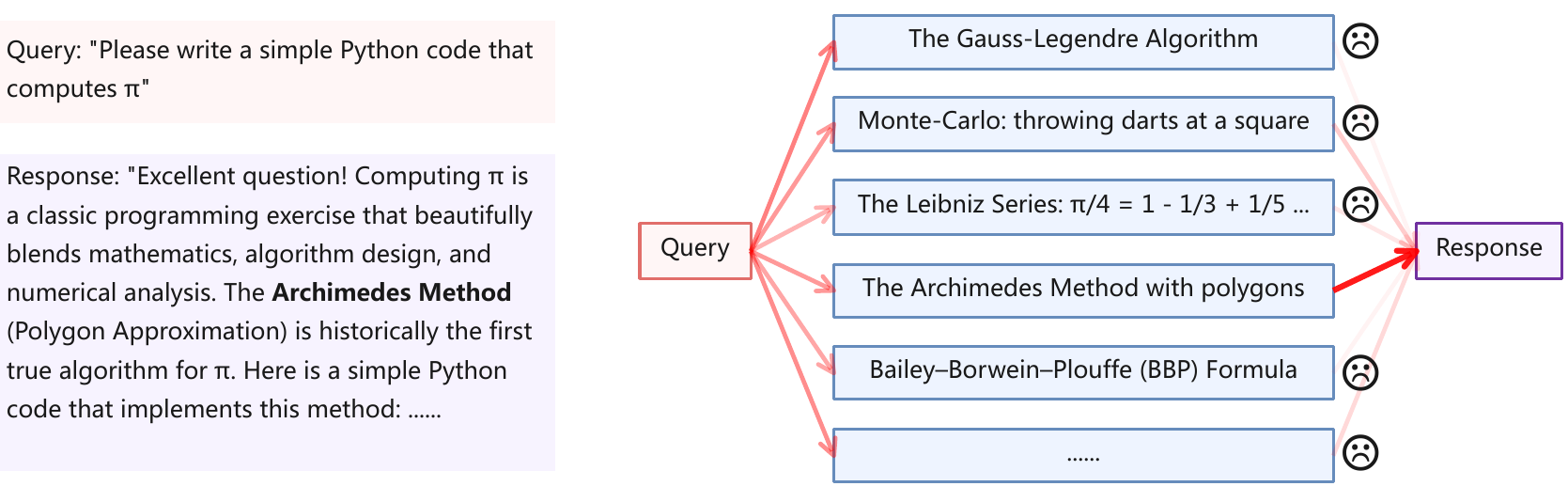}
\caption{Left: A pair of query $x_q$ and response $x_r$ for finetuning slow thinking models.
The response is based on the Archimedes method for computing $\pi$.
Right: The blue boxes represent the possible thoughts that can be generated by the sampler.
Here we assume that a predictive sampler is used, which can only see the query when producing thoughts, and thus the sampled methods for computing $\pi$ are rather uniformly distributed over the popular methods.
Only one thought guessed the Archimedes method, and the notation \faFrownO~indicates that the other thoughts are unhelpful or misleading for producing $x_r$.
The shades of the arrows represent the likelihoods $P_f(y|x_q\to\mathcal{T})$ and $P_f(x_r|x_qy\to\Sigma^*_{\square})$.
The former are positive and approximately uniform, while the latter are all vanishingly small except for the right thought.
If instead a well-trained explanatory sampler is used, then most of the sampled thoughts would focus on the Archimedes method.}
\label{fig: compute pi}
\end{figure}

\begin{example}
\label{ex. compute pi}
Suppose a slow thinking model is finetuned on a pair of user query and high-quality response $(x_q, x_r)$.
As depicted in Figure \ref{fig: compute pi} (left), it is about computing the constant $\pi$ in Python.
Training proceeds by
\begin{enumerate}
    \item Sampling multiple thoughts from the sampler.
    \item Encouraging the latent model to generate the response $x_r$ conditioned on $x_q$ and each thought.
    \item Guiding the sampler to produce thoughts that are helpful for step 2.
\end{enumerate}
(Recall that the formal description of this training process is provided by (\ref{eq. minibatch loss inquisitive}) and (\ref{eq. minibatch sampler}) for the general case, and by (\ref{eq. deepseek latent loss}) and (\ref{eq. deepseek sampler loss}) for the special case with the identity sampler.)
The training efficiency depends heavily on the usefulness of the thoughts for deriving $x_r$,
and thus on the efficiency of the sampler at producing relevant thoughts.
Namely, if the sampler quickly produces a reasonable thought that can effectively lead to $x_r$,
then in step 2 the latent model gets an easy job and in step 3 the sampler has a clear target to fit.
However, things could fall apart if the query is open-ended with multiple possible answers.
As shown in Figure \ref{fig: compute pi} (right), there are many simple methods to compute $\pi$, and it is very unlikely that a randomly chosen method coincides with that of the provided response $x_r$.
Thus, if a thought is generated without the knowledge of $x_r$, it is probably irrelevant to $x_r$ and cannot help with training.
This is the case for all predictive samplers, including the identity sampler $Q_{\id}(\cdot|x_q)$, which can only see $x_q$.
The causal samplers $\Q_{\proj}^{\text{\faPlay}}$ are no better;
by (\ref{eq. causal factorization}), they can see the first token of $x_q$, which in this case is ``Excellent" and is not really informative.
Hence, in this case, it appears that only the explanatory samplers, which can attend to the full response $x_r$, are capable of building a bridge between $x_q$ and $x_r$.
\end{example}


Example \ref{ex. compute pi} (and similarly Examples \ref{ex. cat trajectory} and \ref{ex. night heron}) indicate that explanatory samplers are indispensable for the proper encoding of sequences.
Since Section \ref{sec. deepseek inference} indicates that (the prefill stage of) the inference of DeepSeek-R1 does not involve latent sampling, we only need to consider the application of explanatory samplers to training.

\begin{figure}[!ht]
\centering
\includegraphics[width=0.8\linewidth]{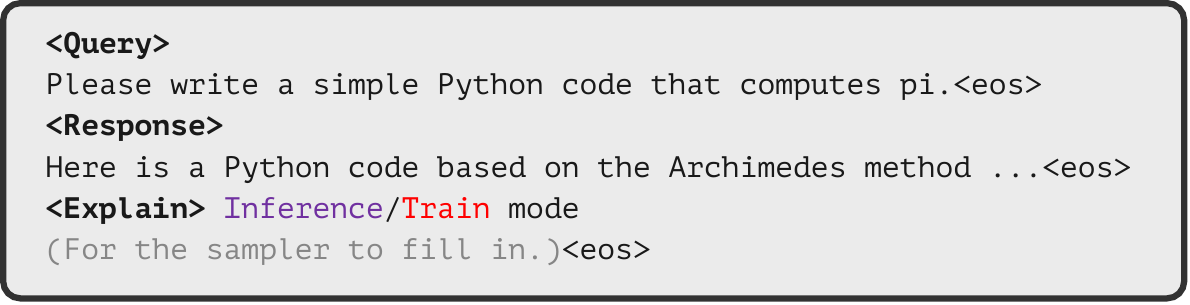}
\caption{One possible prompt format for the explanatory sampler $Q(\cdot|xx_r)$.}
\label{fig: explanatory sampler format}
\end{figure}

Same as the set-up of Section \ref{sec. deepseek training}, suppose the training sample consists of a query $x$ and response $x_r$.
To implement an explanatory sampler, one can use any prompt that contains both $x$ and $x_r$, for instance the one in Figure \ref{fig: explanatory sampler format}.
Optionally, one can instantiate two samplers, the inference sampler $Q$ and train sampler $Q^{\see}$;
one simple way is to add a tag in the prompt that indicates ``inference mode" or ``train mode" as in the figure.
To save storage, the sampler(s) can share the same underlying model (DeepSeek-V3) with the latent distribution $P_f$;
namely, the underlying model plays $P_f$ if the prompt is simply $x$, and plays the (inference or train) sampler if the prompt is the following (with the inference or train tag).

The general losses (\ref{eq. minibatch loss inquisitive}) and (\ref{eq. minibatch sampler}) can be adapted to the current setting.
First, an estimator for the conditional probability $p_*$ can be defined based on the summation (\ref{eq. deepseek conditional sum})
\begin{equation*}
p_n := \frac{1}{n}\sum_{i=1}^n \frac{P_f(Y^{(i)}x_r \mid x\to\mathcal{T}\Sigma^*_{\square})}{Q(Y^{(i)}|xx_r)}, \quad \{ Y^{(i)} \}_{i=1}^n \iidsample Q(\cdot|xx_r)
\end{equation*}
If the train sampler is used, an estimator for the gradient $\nabla_{\theta}(-\log p_*)$ can be defined by
\begin{equation*}
\vb_n := - \frac{1}{p_n n} \sum_{i=1}^n \frac{P_f(Y^{(i)}_{\eye} x_r \mid x\to\mathcal{T}\Sigma^*_{\square}) ~ \nabla_{\theta} \log P_f(Y^{(i)}_{\eye} x_r \mid x\to\mathcal{T}\Sigma^*_{\square})}{Q^{\see}(Y^{(i)}_{\eye}|xx_r)}, \quad \{ Y^{(i)}_{\eye} \}_{i=1}^n \iidsample Q^{\see}(\cdot|xx_r)
\end{equation*}
It follows that, in the one-sampler case, the latent distribution $P_f$ can use the following loss
\begin{equation}
\label{eq. explanatory latent loss inference sampler}
L_n(\theta) = - \log\Big( \frac{1}{n}\sum_{i=1}^n \frac{P_f(Y^{(i)}x_r \mid x\to\mathcal{T}\Sigma^*_{\square})}{\detach(Q)(Y^{(i)}|xx_r)} \Big)
\end{equation}
based on (\ref{eq. minibatch loss one sampler}), and in the two-sampler case, can use the loss
\begin{equation}
\label{eq. explanatory latent loss two sampler}
L_n(\theta) = - \detach(p_n)^{-1} \frac{1}{n} \sum_{i=1}^n \frac{P_f(Y^{(i)}_{\eye} x_r \mid x\to\mathcal{T}\Sigma^*_{\square})}{\detach(Q^{\see})(Y^{(i)}_{\eye}|xx_r)}
\end{equation}
based on (\ref{eq. minibatch loss inquisitive}).
If the optimizer is insensitive to the norm of the gradient $\nabla_{\theta} L_n$ and if the train step does not involve any other train data,
then to save compute the loss (\ref{eq. explanatory latent loss two sampler}) can be modified to
\begin{equation}
\label{eq. explanatory latent loss train sampler}
L_n(\theta) = - \log\Big( \frac{1}{n} \sum_{i=1}^n \frac{P_f(Y^{(i)}_{\eye} x_r \mid x\to\mathcal{T}\Sigma^*_{\square})}{\detach(Q^{\see})(Y^{(i)}_{\eye}|xx_r)} \Big)
\end{equation}
In the case $n=1$, the loss (\ref{eq. explanatory latent loss inference sampler}) is equivalent in gradient to
\begin{equation}
\label{eq. explanatory latent loss inference sampler one sample}
L_1(\theta) = - \log P_f(Yx_r \mid x\to\mathcal{T}\Sigma^*_{\square}), \quad Y \sim Q(\cdot|xx_r)
\end{equation}
Note that this is similar to (\ref{eq. deepseek latent loss}), except that the latent distribution now needs to fit the thought $Y$ as well.
For theoretical interest, one can note that (\ref{eq. deepseek latent loss}) is a special case of (\ref{eq. explanatory latent loss inference sampler one sample}):
If we set the sampler to the identity sampler $Q(\cdot|xx_r) = P_f(\cdot|x\to\mathcal{T})$, then
\begin{align*}
\E_Y\big[ \nabla_{\theta} L_1 \big] &= - \int \nabla_{\theta} \log P_f(x_r \mid xy \to\Sigma^*_{\square}) + \nabla_{\theta} \log P_f(y|x\to\mathcal{T}) ~ dP_f(y|x\to\mathcal{T}) \\
&= \E_Y\big[ \nabla_{\theta} (- \log P_f(x_r \mid xY \to\Sigma^*_{\square})) \big] + 0
\end{align*}
so the two losses has the same expected gradient.

Meanwhile, the inference sampler is trained by (\ref{eq. minibatch sampler}), which becomes
\begin{equation}
\label{eq. deepseek explanatory sampler loss}
L_{\text{sampler}}(\theta) = - \frac{1}{n} \sum_{i=1}^n \detach\Big(\log \frac{P_f(Y^{(i)}x_r \mid x\to\mathcal{T}\Sigma^*_{\square})}{Q(Y^{(i)}|xx_r)}\Big) ~ \log Q(Y^{(i)}|xx_r)
\end{equation}
The train sampler $Q^{\see}$, if used, is trained by the reward (\ref{eq. train sampler reward}), so the loss becomes
{\small
\begin{equation*}
L_{\text{sampler}}^{\see}(\theta) = - \frac{1}{n} \sum_{i=1}^n \detach\Big(\log \frac{P_f(Y^{(i)}_{\eye} x_r \mid x \to \mathcal{T}\Sigma^*_{\square}) ~ \| \nabla_{\theta} \log P_f(Y^{(i)}_{\eye} x_r \mid x \to \mathcal{T}\Sigma^*_{\square}) \|}{Q^{\see}(Y^{(i)}_{\eye}|xx_r)}\Big) \log Q^{\see}(Y^{(i)}_{\eye}|xx_r)
\end{equation*}}
In summary, the underlying model is trained on one of these $L_n$, plus $L_{\text{sampler}}$ and optionally $L_{\text{sampler}}^{\see}$.

There is a subtlety about the sampling size $n$.
To reduce variance, one can use the following mini-batch version of (\ref{eq. explanatory latent loss inference sampler one sample})
\begin{equation}
\label{eq. explanatory latent loss inference sampler 1xn}
L_1^{(n)}(\theta) = - \frac{1}{n}\sum_{i=1}^n \log P_f(Y^{(i)} x_r \mid x\to\mathcal{T}\Sigma^*_{\square}), \quad \{ Y^{(i)} \}_{i=1}^n \iidsample Q(\cdot|xx_r)
\end{equation}
It resembles (\ref{eq. explanatory latent loss inference sampler}) except that the averaging takes place outside of $\log$.
The distinction is that effectively (\ref{eq. explanatory latent loss inference sampler}) trains the $n$-sample estimator $p_n$ for inference, whereas (\ref{eq. explanatory latent loss inference sampler 1xn}) only trains the 1-sample estimator $p_1$.

\begin{remark}
\label{remark. justify explanatory}
One might worry if the explanatory sampler $Q(\cdot|xx_r)$ could help the latent model $P_f$ to cheat during training and become degenerate.
Specifically, it may appear that $Q(\cdot|xx_r)$ could simply copy-and-paste the answer $x_r$ to the sampled thought $Y$, and then $P_f$ can learn to copy $Y$ to produce $x_r$, thus getting a perfect score $P_f(x_r|xY\to\Sigma^*_{\square}) \approx 1$ while learning nothing.
This is not the case.
Note that the reward in the sampler loss $L_{\text{sampler}}$ is proportional to
\begin{equation*}
\log P_f(Y \mid x \to\mathcal{T}) + \log P_f(x_r \mid xY \to\Sigma^*_{\square})
\end{equation*}
So a good thought $Y$ is like a bridge, connecting the prompt $x$ on one side (maximizing the first term) and the response $x_r$ on the other side (maximizing the second term).
Cheating (setting $Y=x_r$) is unhelpful for the first term, and will be avoided by the sampler.
If $x$ is a math problem, for instance, and if $P_f$ is an autoregressive model, then $Y$ would be a unidirectional derivation that proceeds step-by-step to the final answer $x_r$.
Similarly, since the losses $L_n$ are proportional to these two terms, the latent model $P_f$ is trained to adapt to these bridges.
Besides, there is an another way to see that the explanatory samplers are appropriate;
no matter what sampler is used, the Monte-Carlo estimator $p_n$ always approximates the same probability value (\ref{eq. deepseek conditional sum}), so in principle the sampler can use any model and condition on any information.
\end{remark}

A further improvement could be using a larger sampling size for the training of the latent model,
namely setting $n>1$ in the loss $L_n$ (\ref{eq. explanatory latent loss inference sampler}) or (\ref{eq. explanatory latent loss two sampler}).
Despite that (\ref{eq. explanatory latent loss inference sampler}) and (\ref{eq. explanatory latent loss inference sampler 1xn}) use almost the same compute, the latter has a potential disadvantage, that $P_f$ would try to fit each $Y^{(i)} x_r$ regardless of the quality of $Y^{(i)}$ and thus could receive harmful training signal.
This problem could be severe if $x$ is a difficult problem or if $x_r$ covers multiple keypoints, so that the sampled thoughts have a much higher probability to be irrelevant, incorrect or incomplete.
In comparison, inside the summation of (\ref{eq. explanatory latent loss inference sampler}), the low-quality thoughts $Y^{(i)}$ correspond to smaller summands and thus are ignored during the forward pass and backpropagation.
With a larger $n$, the hit rate (the probability that at least one of the thoughts $\{Y^{(i)}\}_{i=1}^n$ is a good thought) approaches $1$, making it more likely for $P_f$ to receive a meaningful training signal.
Still, as discussed in Remark \ref{remark. sequential sampling}, even if $n=1$, it is possible for a sampler to implicitly produce multiple thoughts and thus increase the hit rate.
However, this strategy might not always work, as it could make the thoughts overly lengthy and it has been observed empirically that the performance of DeepSeek-R1 is negatively correlated with its thought length when above some threshold \cite{marjanovic2025thoughtology}.


As a final remark, all the above improvements are only for the training process.
Inference does not need to be modified, since the latent model is forgetful and thus decoding does not involve any sampler.
Nevertheless, suppose one is not interested in the sampling of $(\proj\#P_f)(\cdot|x\to\Sigma^*_{\square})$ but wants the maximum likelihood response
\begin{equation*}
x_r^* = \argmax_{x_r \in \Sigma^*_{\square}} (\proj\#P_f)(x_r|x\to\Sigma^*_{\square})
\end{equation*}
Then, the decoding method (\ref{eq. deepseek decoding}) can be modified to best-of-$K$ sampling for some large $K$
\begin{align*}
X_r^* = X_r^{(k_*)}, \quad k_* = \argmax_k (\proj\#P_f)(X_r^{(k)}|x\to\Sigma^*_{\square}), \quad \{Y^{(k)}X_r^{(k)}\}_{k=1}^K \iidsample P_f(\cdot|x\to\mathcal{T}\Sigma^*_{\square})
\end{align*}
If these likelihoods are estimated by
$$(\proj\#P_f)(X_r^{(k)}|x\to\Sigma^*_{\square}) \approx \frac{1}{n_k} \sum_{i=1}^{n_k} \frac{P_f( Y^{(k,i)}X_r^{(k)} \mid x\to\mathcal{T}\Sigma^*_{\square})}{Q(Y^{(k,i)} \mid xX_r^{(k)})}, \quad \{ Y^{(k,i)} \}_{i=1}^{n_k} \iidsample Q(Y^{(i,k)} \mid xX_r^{(k)})$$
then in order to improve precision and efficiency, one needs an explanatory sampler and large sampling sizes $n_k$.

\subsection{Stage Two: Improving Approximation Ability}
\label{sec. short-term improvement}

This section introduces the second stage of improvements, those that require a modified representation.
Abstractly speaking, the motivation is to climb the representation hierarchy (Figure \ref{fig: representation hierarchy}) and make the model more expressive.
Intuitively speaking, we want the model to learn reasoning more effectively during pretraining, and to perform more sophisticated reasoning during inference.


We advocate for the following two modifications, which need to be applied together to be effective.
\begin{enumerate}
\item Ubiquitous thinking: The thoughts $y^{(t)}$ can be inserted anywhere in the text, not just between each pair of query and response in a conversation setting.
Using the terminology of Section \ref{sec. deepseek representation}, the latent distribution $P_f$ can be supported on the entire $\dom(\proj)$, not just $\Omega^{\omega}_{\text{\faCode}}$.
A natural training objective is needed to encourage the sampler to insert thoughts in an efficient manner, assigning long thoughts only to difficult texts that require deep reasoning.

\item Persistent thinking: Each thought $y^{(t)}$ may have lasting effect on the processing of later texts,
e.g.\@ an epiphany we had when reading chapter 5 of ``Differential Geometry" may become helpful for understanding chapter 9.
In terms of Definition \ref{def. forgetful} and equation (\ref{eq. forgetful latent}), the latent distribution $P_f$ is no longer forgetful, and the likelihood of each segment $y^{(T+1)}x^{(T+1)}$ may depend on the previous thoughts $(y^{(t)})_{t=1}^T$.
\end{enumerate}
We first discuss the representation of persistent and ubiquitous slowing thinking, then formulate its explanatory sampler, and derive a training objective that provides an intrinsic motivation to balance fast and slow thinking.
One may compare the inserted thoughts $(y^{(t)})_{t=1}^T$ to the ``working memory" of a human who is reading the text $x$, since both are persistent, continuously evolving internal states.

\subsubsection{Persistent and Ubiquitous Thinking}

The motivation for ``ubiquitous thinking" is to improve the reading ability of LLMs, specifically, enable them to perform deep and efficient slow thinking when encoding any text worthy of reasoning.
One beneficial outcome is that we can adapt slow thinking to pretraining to improve data efficiency.
Ideally, an LLM should develop most of its reasoning ability in an unsupervised manner during pretraining, while post-training is only for adapting the LLM to specific tasks.
For the state-of-the-art LLMs, a considerable portion of their pretrain datasets consists of texts that are rich in reasoning, such as textbooks, research papers and codes \cite{weber2024redpajama,grattafiori2024llama3,yang2024qwen25math}.
So LLMs should perform slow thinking to squeeze all the reasoning out of these texts, learn it, and master all kinds of reasoning by the end of pretraining.
Another benefit is that LLMs can perform slow thinking in the prefill stage of inference to thoroughly understand the context.

The motivation for ``persistent thinking" is to overcome the limitation of forgetfulness (Definition \ref{def. forgetful}) 
and enhance the approximation ability of the model.
As as illustration of this limitation, consider the ``cups and ball" game of Figure \ref{fig: cups and ball}.
Instead of watching the magician swapping the cups, suppose we are only presented with a written list of the swappings performed.
Conceptually, it is impossible to always answer correctly within a bounded amount of time if the number of swappings is unbounded, following the discussion in Section \ref{sec. separation I}.

The functioning of a forgetful model is just like this modified ``cups and ball" game.
Denote by $x$ the list of swappings and $x_r$ the correct answer.
Given a forgetful latent $P_f$, the likelihood of $x_r$ is given by (\ref{eq. deepseek conditional int}):
\begin{equation*}
(\proj\#P_f)(x_r|x\to\Sigma^*_{\square}) = \int P_f(x_r|xy\to\Sigma^*_{\square}) dP_f(y|x\to\mathcal{T})
\end{equation*}
namely, the model first samples a bounded-length thought $y\in\mathcal{T}$ and then produces the answer.
Thus, the above argument indicates that forgetful models cannot approximate the compositional function (\ref{eq. A5 composition function}), and therefore cannot approximate the hidden Markov model constructed in the proof of Theorem \ref{thm. separation I}.
Hence, we arrive at an informal proof that $\PS_{\text{HMM}}$ is not contained in the following subspace
\begin{equation*}
\PS_{\text{forget}} = \big\{ \proj\#P_f \bigm| f \in \TF_{<\omega}(\Omega), ~ P_f ~\text{is forgetful in the sense of} ~ (\ref{eq. forgetful latent}) \big\}
\end{equation*}
It seems probable that this argument can be made rigorous based on the proof of Theorem \ref{thm. separation III}.
Since $\mathcal{T}$ is finite, even smaller than the polynomial-size $\fip(x)$ of Theorem \ref{thm. separation III}, forgetful models can probably be simulated by uniform $\TC^0$ circuits, and thus have poor approximation ability.
Since it is not central to this paper, we do not elaborate on the formal proof.

\begin{figure}[!ht]
\centering
\includegraphics[width=0.55\linewidth]{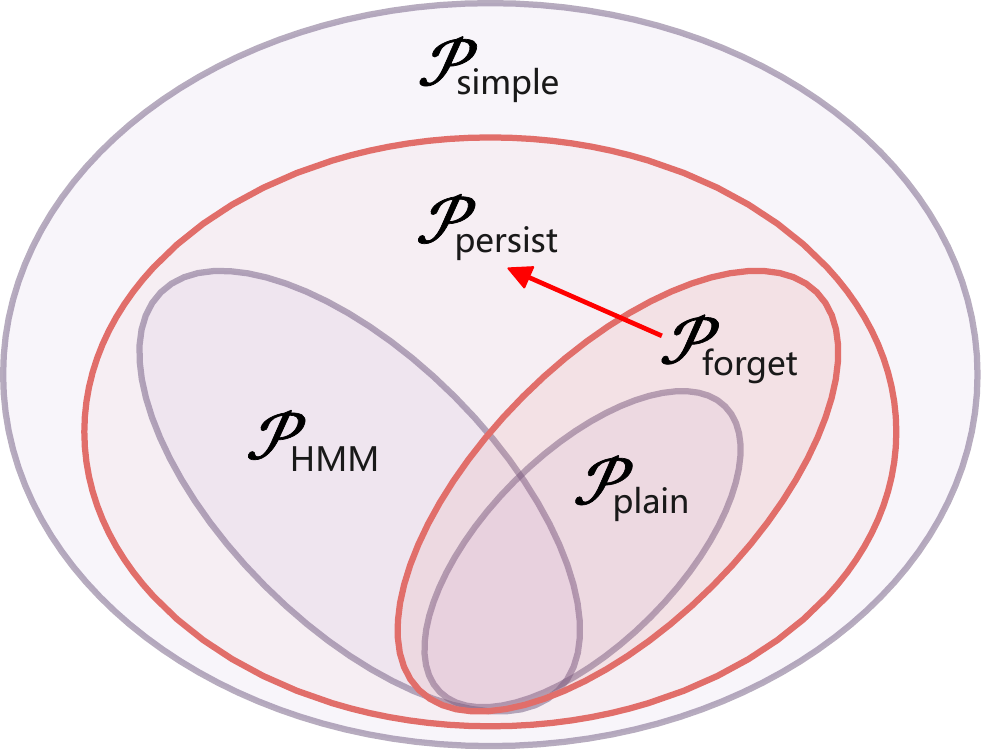}
\caption{The positions of the forgetful models ($\PS_{\text{forget}}$) and persistent thinking models ($\PS_{\text{persist}}$) in the representation hierarchy (Figure \ref{fig: representation hierarchy}), based on the informal arguments in this section.
The arrow indicates that removing the forgetfulness constraint greatly enhances expressivity.}
\label{fig: forgetful in hierarchy}
\end{figure}

Removing the forgetfulness constraint enables these models to approximate all HMMs.
As an informal argument, define the subspace of ``persistent thinking models" by
\begin{equation*}
\PS_{\text{persist}} = \big\{ \proj\#P_f \bigm| f \in \TF_{<\omega}(\Omega) \big\}
\end{equation*}
$\PS_{\text{persist}}$ should be able to approximate $\PS_{\text{forget}}$, since the Transformer $f$ can simply ignore the earlier thoughts $(y^{(t)})_{t=1}^{T-1}$ (which are enclosed by $\sot,\eot$) when encoding each segment $y^{(T)}x^{(T)}$.
Similar to the proof of Theorem \ref{thm. separation II}, $\PS_{\text{persist}}$ can simulate any HMM as follows:
Let $(Z_t,X_t)$ denote the pair of latent and observable variables of the HMM at step $t$.
A persistent thinking model can represent this pair as a segment $Y^{(t)}X_t$ with $Y^{(t)} = \sot Z_t \eot $, and then produce each segment in a Markov manner.
If the latent vocabulary $\Omega'$ of the HMM is larger than $\Sigma$, then $Y^{(t)}$ can use $\lceil \log_{|\Sigma|}|\Omega'|\rceil$ tokens to represent $Z_t$.
In conclusion, we have obtained the relations depicted in Figure \ref{fig: forgetful in hierarchy}.


The relation between $\PS_{\text{forget}}$ and $\PS_{\text{persist}}$ is similar to that of $\PS_{\text{poly}}$ and $\PS_{\text{simple}}$ from Section \ref{sec. approximation summary}, such that the addition or removal of one constraint greatly influences the approximation ability of a distribution family, especially whether it contains the HMMs.
Nevertheless, the relation between $\PS_{\text{poly}}$ and $\PS_{\text{forget}}$ is not obvious, and it seems neither can approximate the other.

\begin{figure}[!ht]
\centering
\includegraphics[width=0.85\linewidth]{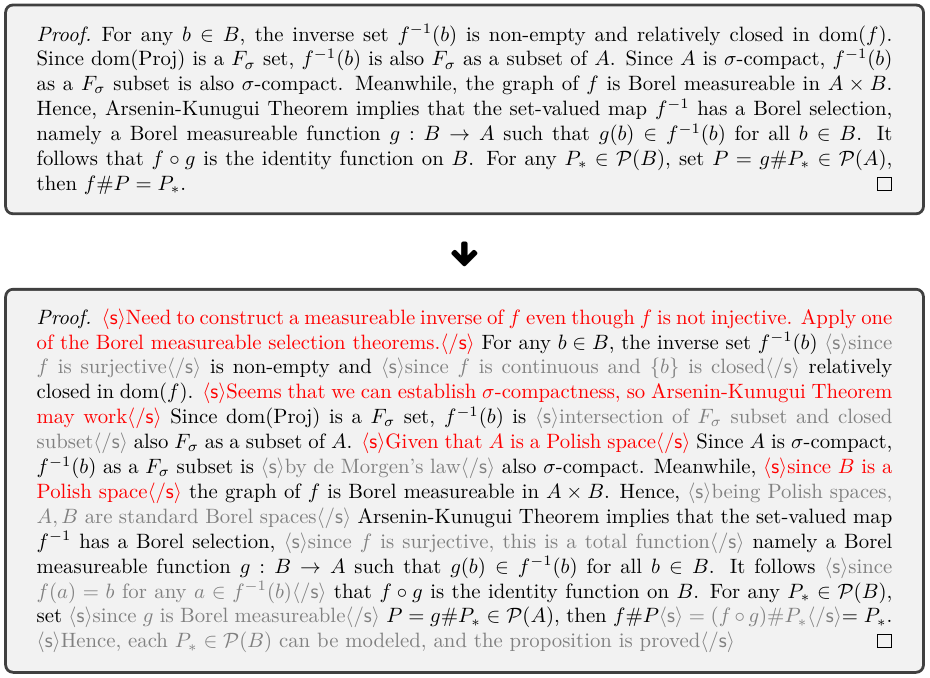}
\caption{Persistent and ubiquitous thinking.
Top: An example of pretraining data (a paragraph taken from Appendix \ref{appendix. measure theory} of this paper) that demands some reasoning to be fully understood.
Bottom: A possible sample $Z$ from the posterior distribution $Q_*(\cdot|x)$.
The colored texts enclosed by $\sot,\eot$ are the latent thoughts, inserted wherever needed.
The {\color{red}highlighted} thoughts are ones that have lasting effect on the encoding and decoding of much later texts, separated by one or more delimiters $\sot$.
This long-range dependency would not be possible if the latent distribution is forgetful.
This is a conceptual illustration, not the output of any model, and the thoughts do not need to follow human languages.}
\label{fig: persistent ubiquitous thinking}
\end{figure}

Hence, the use of persistent and ubiquitous thinking is justified by its enhancement of approximation ability.
Formally, its representation is straightforward, as the latent distribution $P$ is now allowed to range in $\PS(\dom(\proj))$, instead of $\PS(\Omega^{\omega}_{\text{\faCode}})$ or the forgetful distributions.
An illustration of persistent and ubiquitous thinking (specifically a sample of the posterior $Q_*(\cdot|x)$ given an unconstrained latent) is provided by Figure \ref{fig: persistent ubiquitous thinking}.
Since this is designed mainly for pretraining, a minimal format should suffice.

\begin{remark}
Certainly one may strengthen a forgetful model by allowing the length bound $c$ to grow with input length $|x|$, as in Example \ref{ex. think}.
The resulting models might be as expressive as the persistent thinking models, as the former can be seen as a modification of the latter that postpones all thoughts to the end of the sequence $z\in\fip(x)$.
Even if this may work in theory, its implementation would be highly costly.
Given a persistent thinking model, let $Z \in Q_*(\cdot|x)$ be its latent sample, and express it as $Z = (Y^{(t)}x_t)_{t=1}^{|x|}$.
Inside the computation of the encoding of $Z$, the output logits at each token $x_T$ may depend on any of the previous thoughts $(Y^{(t)})_{t=1}^T$.
In order to obtain the same output from a forgetful latent with function parameter $f_{\text{forget}}$, in the worst case $f_{\text{forget}}$ needs to take the permuted sequence $x_{<T}Y^{(1)}\dots Y^{(T)}x_T$ as input, and this encoding needs to be performed separately for each $T=1,\dots |x|$, resulting in an enormous cost.
Therefore, it could be much more efficient to use a persistent thinking model than a forgetful one to achieve a given level of expressivity.
\end{remark}

\subsubsection{Inference of the Explanatory Sampler}

Next, we briefly describe how to construct an explanatory sampler $Q\in\Q_{\proj}$.
Since the latent $P$ can now be supported on the entire $\dom(\proj)$, the latent sequences can range in all of $\fip(x)$.
Since the sampler is explanatory, each inserted thought $Y^{(t)}$ in a sampled latent sequence $Z \sim Q(\cdot|x)$ can depend on all of the input $x$.

\begin{figure}[!ht]
\centering
\includegraphics[width=1\linewidth]{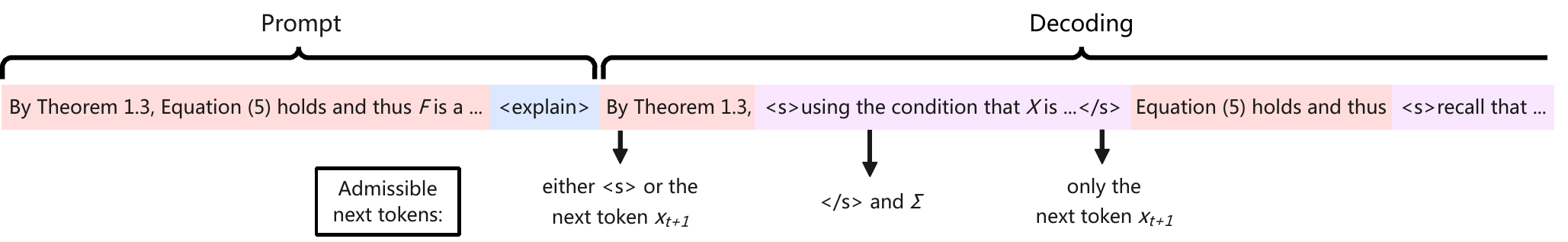}
\caption{The input-output format $x\langle\text{explain}\rangle Z$ for the explanatory sampler that supports persistent and ubiquitous thinking.
The text $x$, including its repetition inside $Z$, is in red, while the inserted thoughts, namely the rest of $Z$, are in purple.
The next-token sets $\nsm^{\leq 1}(z)$ for some prefixes $z$ of $Z$ are annotated below.}
\label{fig: flexible sampler format}
\end{figure}

Here is one straightforward implementation.
Given any text $x\in\Sigma^*$, the format can simply be ``$x\langle\text{explain}\rangle Z$", such that the delimiter $\langle\text{explain}\rangle$ can be either a reserved token or a short instruction, ``$x\langle\text{explain}\rangle$" is the prompt for the sampler $Q$, and $Z$ is the decoded output.
This format allows each token in $Z$ to depend on all of $x$.
If we need to distinguish between the training and inference samplers, then a tag can be included as in Figure \ref{fig: explanatory sampler format}.
To ensure that $Z$ belongs to $\fip(x)$,
the sampler should output in the format of $Z=(Y^{(t)}x_t)_{t=1}^{|x|}$, where each thought $Y^{(t)}$ is either empty or belongs to $\mathcal{T}$.
This can be achieved with the logit masking of Definition \ref{def. softmax with masking} and the next-token set $\nsm^{\leq 1}(z)$ from (\ref{eq. admissible next-token}).
In the current setting, $\nsm^{\leq 1}(z)$ has the following form:
Let $z \in \Omega^*$ be the current decoding and $a = z_{|z|}$ be the current last token,
\begin{itemize}
\item If $a$ is within a thought or $a=\sot$ (i.e.\@ the most recent delimiter in $z$ is $\sot$)
\begin{itemize}
    \item If this thought has length $<c$ (the distance to this $\sot$ is $<c$), then set $\nsm^{\leq 1}(z)=\Sigma \cup \{\eot\}$
    \item Else, set $\nsm^{\leq 1}(z) = \{\eot\}$
\end{itemize}
\item Else if $a=\eot$
\begin{itemize}
    \item If the most recent observable token $x_t$ (the token before the most recent $\sot$) is not the last token (namely $t<|x|$), then set $\nsm^{\leq 1}(z)=\{x_{t+1}\}$
    \item Else, set $\nsm^{\leq 1}(z) = \{\eos\}$
\end{itemize}
\item Else, $a=x_t$ is some observable token
\begin{itemize}
    \item If $a$ is not the last token ($t<|x|$), set $\nsm^{\leq 1}(z) = \{x_{t+1},\sot\}$
    \item Else, set $\nsm^{\leq 1}(z) = \{\sot,\eos\}$
\end{itemize}
\end{itemize}
This programming is reminiscent of the transition rules of a finite automaton.
An illustration of the next-token sets $\nsm^{\leq 1}(z)$ during decoding is given by Figure \ref{fig: flexible sampler format}.

Considering that decoding is generally slower than encoding for LLMs, the above decoding process can be accelerated, especially when the nonempty thoughts are sparse.
If the next token to be encoded by the autoregressive latent model $P_f$ is either some $x_{t+1}$ or $\eot$, then we additionally encode some subsequent observable tokens,
so the input becomes a chunk $x_{t+1}\dots x_{t+C}$ or $\eot x_{t+1}\dots x_{t+C}$,
for some length $C$, which can be fixed or adaptively adjusted.
Then, $C$ softmax distributions are produced (the output distribution at $\eot$ is a Dirac mass and can be ignored), each is binomial over two tokens $\{x_{t+i},\sot\}$ for $i=1,\dots C$, and we sample all of them;
if none produces $\sot$, then we proceed to encode the next chunk;
else, we discard all the encoding after the first $\sot$ and proceed to decoding the next thought.

\subsubsection{\textit{A Priori} Balancing of Fast and Slow Thinking}
\label{sec. balance fast slow}

Finally, we design the training objective.
The goal is to optimize under the speed-accuracy tradeoff, such that the model is able to minimize the cross entropy loss while keeping the density and lengths of the inserted thoughts under control.
Instead of using \textit{ad hoc} penalties, constraints or workflows \cite{kimi2025kimik15,pandian2025snap,hong2026reconsider,yuan2026yuan30flash}, we derive a training loss directly from the unified objective (\ref{eq. unified objective}).





Recall that $\proj$ admits latent sequences $z \in \fip(x)$ with varying lengths.
For any text $x\in\Sigma^*$, the shortest latent sequence is simply $z=x$, while the longest ones are $z=(y^{(t)}x_t)_{t=1}^{|x|}$ with $|y^{(t)}| = c+2 = 32770$,
such that a thought with maximum length is inserted before each token.
In the worst case, even a short input text can be lifted to a long novel.
Thus, from a cost perspective, it is of great interest whether a given latent distribution $P \in \PS(\dom(\proj))$ favors long latent sequences, as measured by the following average length
\begin{equation*}
\bar{l}(x) := \E_{Z\sim Q_*(\cdot|x)}[|Z|] = \frac{\sum_{z\in\fip(x)}|z| \cdot P^{\leq |z|}(z)}{(\proj\#P)^{\leq |x|}(x)}
\end{equation*}
where $Q_*$ is the posterior sampler of this $P$.

Typical training objectives, such as cross entropy (\ref{eq. cross entropy}), encourage the latent sequences $Z\sim Q_*(\cdot|x)$ to grow as lengthy as possible.
As exemplified by the comparison of $\PS_{\text{poly}}$ and $\PS_{\text{simple}}$, approximation ability is proportional to the amount of latent sequences, which in this case is proportional to the average length $\bar{l}(x)$.
Thus, for a model $\proj\#P$ to fit any target $P_*$ well, its latent length $\bar{l}$ would tend to increase to the maximum allowed, which is $32771\cdot|x|$.
This is in accordance with the empirical observations that slow thinking models tend to overthink during inference \cite{chen2025think23o1,wei2026evolution},
and also their thought lengths keep growing monotonically during training \cite{guo2025deepseek,yu2025DAPO}.
To avoid the enormous cost due to overly lengthy thoughts, a cost-aware training objective is needed.
The quick fixes used in practice are rather artificial;
for instance, if one tries to control $\bar{l}(x)$ by adding a length penalty \cite{kimi2025kimik15,ling2025fasteasydeephard}, the choice of its coefficient and schedule cannot be determined \textit{a priori} and are left to trial-and-error.

In principle, the training objective should be (\ref{eq. unified objective}) that maximizes the rate of loss reduction with respect to the real time.
Section \ref{sec. unified objective} has argued that this objective can compel the model to generate long thoughts only when needed,
similar to how humans balance their fast and slow thinking.

To make the abstract objective (\ref{eq. unified objective}) implementable, we derive its simplified form.
Suppose the time cost for processing a length-$l$ latent sequence during a training step is given by some function $\tau(l)$.
Here, the processing includes the decoding, encoding and backpropagation of this sequence,
and this $\tau$ can be obtained either by actual measurement or calculation.
Considering that in practice the time cost is mainly due to decoding, specifically the loading of model weights from VRAM to SRAM (i.e.\@ the memory wall) that is repeated for each decoded token, one may adopt the approximation $\tau(l) \approx C_{\text{decode}} l$ (and this coefficient can be ignored).
Suppose for simplicity that the sampling size $n$ is fixed.
Denote the mini-batch at each step $s$ by $\{X^{(s,b)}\}_{b=1}^B$,
and the sampled latent sequences for training the latent model and sampler model(s) by $\{Z^{(s,b,i)}\}_{i=1}^n$.
Then, the term $t_{s+1}-t_s$ on the right side of (\ref{eq. unified objective}) can be modeled by the random variable
\begin{equation*}
\frac{1}{N_{\text{parallel}}}\sum_{b=1}^B \sum_{i=1}^n ~ \tau(|Z^{(s,b,i)}|)
\end{equation*}
where $N_{\text{parallel}}$ denotes the amount of parallel processors.
Since the constant coefficients do not affect the objective, we can use the following term instead 
\begin{equation*}
\mathbb{T}_s(\theta_s) = \frac{1}{Bn}\sum_{b=1}^B \sum_{i=1}^n ~ \tau(|Z^{(s,b,i)}|)
\end{equation*}
Then, the objective (\ref{eq. unified objective}) becomes
\begin{equation}
\label{eq. loss x lenth total cost}
\min_{(\theta_s)_{s=0}^{\infty} \in \Theta} \sum_{s=0}^{\infty} C(\theta_s) \mathbb{T}_s(\theta_s)
\end{equation}
and it compels both the reducible uncertainty and latent lengths to be small.

Next, we perform a series of approximations to make this new objective tractable.
The first approximation is to simplify the ``total cost" (\ref{eq. loss x lenth total cost}) to the ``running cost"
\begin{equation}
\label{eq. loss x lenth running cost}
\min_{\theta} ~ C(\theta) \mathbb{T}_s(\theta)
\end{equation}
and train by gradient descent.
A justification of this approximation is provided in Appendix \ref{appendix. from global to local}, showing that in certain settings their solutions can be close.
Still, it should be generally helpful to implement some form of future looking beyond the greedy approach to better approximate the total cost,
e.g.\@ using a value function (with respect to step $s$), finite-horizon optimization \cite{ren2024MPC}, or more general optimal control techniques \cite{li2018optimal}.

The second approximation is to estimate the two terms of $C$ (\ref{eq. reducible uncertainty}),  in particular the entropy $H(P_*)$.
One estimator is based on LLM likelihoods.
Suppose we have a series of LLMs of increasing sizes that are trained on data sampled from $P_*$,
and suppose there is a test dataset $X_{\text{test}}$ that is roughly a set of i.i.d.\@ samples of $P_*$ and that has never been trained on by this LLM series.
One may either prepare these models and $X_{\text{test}}$ by oneself,
or use one of the popular LLM families \cite{qwen2025qwen25,yang2025qwen3,grattafiori2024llama3} and let $X_{\text{test}}$ be texts that are written after the publication of these LLMs.
Then, one computes the cross entropy of each LLM on $X_{\text{test}}$, and fits a power-law curve $f(m)=km^{-\alpha}+h$ between the model sizes and these cross entropies.
The value $f(\infty)=h$ is an estimator of $H(P_*)$.
The mechanism is that, by Proposition \ref{prop. cross entropy to KL}, the cross entropy equals $H(P_*)$ plus some KL divergence,
and one may assume that the latter goes to zero as the model size goes to infinity.
Alternatively, one can consider classical estimators such as ones that are based on string matching and compression \cite{feutrill2021review,wyner2002ergodic}.
Denote the estimated value of $H(P_*)$ by $\tilde{H}(P_*)$.
If only one sampler is used, then the cross-entropy in (\ref{eq. reducible uncertainty}) can be estimated by the mini-batch loss $L_n$ from (\ref{eq. minibatch loss one sampler}), now with the samples $\{X^{(s,b)}\}_{b=1}^B$, and $C(\theta)$ can be estimated by
\begin{equation*}
C_s(\theta) = L_n(\theta) - \tilde{H}(P_*)
\end{equation*}
If both the inference and train samplers are used, 
then $C(\theta)$ can be estimated by
\begin{equation*}
C_s(\theta) = L_n^{\see}(\theta) + \detach\big(L_n(\theta) - L_n^{\see}(\theta)\big) - \tilde{H}(P_*)
\end{equation*}
where $L_n^{\see}$ is the mini-batch loss (\ref{eq. minibatch loss inquisitive}) averaged over the samples $\{X^{(s,b)}\}_{b=1}^B$.
The detached term in $\detach$ ensures that $C_s$ not only approximates $C$ in gradient but also in value,
so that the $\mathbb{T}_s$ term in (\ref{eq. loss x lenth running cost}) gets the right coefficient and thus the right gradient.

The third approximation is to restrict $\theta_s$ to include only the parameter of the latent distribution $P$,
whereas the parameter of the sampler(s) is trained separately by its usual objective in Section \ref{sec. sampler training}.
The consideration is that controlling the latent length $\bar{l}$ is only an objective for $P$, while the goal of the sampler(s) is always to fit the posterior sampler (and inquisitive sampler).
Thus, we modify the term $\mathbb{T}_s$ to use the posterior sampler $Q_*$ of $P_{\theta}$
\begin{equation*}
\mathbb{T}_s^*(\theta) = \frac{1}{B} \sum_{b=1}^B \mathbb{T}^*(\theta, X^{(s,b)}), \quad \mathbb{T}^*(\theta, x) = \E_{Z\sim Q_*(\cdot|x)} \big[\tau(|Z|)\big]
\end{equation*}
Its gradient is given by
\begin{align*}
\nabla_{\theta} \mathbb{T}^*(\theta, x) &= \nabla_{\theta} \frac{\sum_{z\in\fip(x)} \tau(|z|) P_{\theta}^{\leq |z|}(z)}{(\proj\#P_{\theta})^{\leq |x|}(x)} \\
&= \frac{\sum_{z\in\fip(x)} \tau(|z|) P_{\theta}^{\leq |z|}(z) \nabla_{\theta} \log P_{\theta}^{\leq |z|}(z)}{(\proj\#P_{\theta})^{\leq |x|}(x)} \\
&\quad - \frac{\sum_{z\in\fip(x)} \tau(|z|) P_{\theta}^{\leq |z|}(z) \cdot \sum_{z\in\fip(x)} P_{\theta}^{\leq |z|}(z) \nabla_{\theta} \log P_{\theta}^{\leq |z|}(z)}{(\proj\#P_{\theta})^{\leq |x|}(x)^2}
\end{align*}
If the variability of the terms $\tau(|z|)$ is small, then the train sampler $Q^{\see}$ is more suitable for estimating the summations with $\nabla_{\theta}\log P_{\theta}$, and the inference sampler $Q$ is more suitable for the rest of the terms:
\begin{align*}
\nabla_{\theta} \mathbb{T}^*(\theta,x) &= \frac{\int \tau(|z|) \nabla_{\theta} \log P_{\theta}^{\leq |z|}(z) \frac{P_{\theta}^{\leq |z|}(z)}{Q^{\see}(z|x)} dQ^{\see}(z|x)}{\int \frac{P_{\theta}^{\leq |z|}(z)}{Q(z|x)} dQ(z|x)}\\
&\quad - \frac{\int \tau(|z|) \frac{P_{\theta}^{\leq |z|}(z)}{Q(z|x)} dQ(z|x) \cdot \int \nabla_{\theta} \log P_{\theta}^{\leq |z|}(z) \frac{P_{\theta}^{\leq |z|}(z)}{Q^{\see}(z|x)} dQ^{\see}(z|x)}{\big(\int \frac{P_{\theta}^{\leq |z|}(z)}{Q(z|x)} dQ(z|x) \big)^2}
\end{align*}
Thus, similar to Section \ref{sec. inquisitive sampler}, a mini-batch loss can be defined whose gradient approximates $\nabla_{\theta} \mathbb{T}$
\begin{align*}
\tilde{\mathbb{T}}^*_s(\theta) &= \frac{1}{B} \sum_{b=1}^B \frac{\frac{1}{n} \sum_{i=1}^n \tau(|Z^{(s,b,i)}_{\eye}|)\log P_{\theta}^{\leq |Z^{(s,b,i)}_{\eye}|}(Z^{(s,b,i)}_{\eye}) ~\detach(q^{(s,b,i)}_{\eye})}{\frac{1}{n} \sum_{i=1}^n \detach(q^{(s,b,i)})} \\
&\quad - \frac{\frac{1}{n} \sum_{i=1}^n \tau(|Z^{(s,b,i)}|) \detach(q^{(s,b,i)}) \cdot \frac{1}{n} \sum_{i=1}^n \log P_{\theta}^{\leq |Z^{(s,b,i)}_{\eye}|}(Z^{(s,b,i)}_{\eye}) ~\detach(q^{(s,b,i)}_{\eye})}{\big( \frac{1}{n} \sum_{i=1}^n \detach(q^{(s,b,i)}) \big)^2} \\
&q^{(s,b,i)} = \frac{P_{\theta}^{\leq |Z^{(s,b,i)}|}(Z^{(s,b,i)})}{Q(Z^{(s,b,i)}|X^{(s,b)})}, \quad \{Z^{(s,b,i)}\}_{i=1}^n \iidsample Q(\cdot|X^{(s,b)}) \\
&q^{(s,b,i)}_{\eye} = \frac{P_f^{\leq |Z^{(s,b,i)}_{\eye}|}(Z^{(s,b,i)}_{\eye})}{Q^{\see}(Z^{(s,b,i)}_{\eye}|X^{(s,b)})}, \quad \{Z^{(s,b,i)}_{\eye}\}_{i=1}^n \iidsample Q^{\see}(\cdot|X^{(s,b)})
\end{align*}
If only the inference sampler is instantiated, then one simply replace all occurrences of $Q^{\see}$ by $Q$ and all $Z^{(s,b,i)}_{\eye}$ by $Z^{(s,b,i)}$.
Note that there is much overlap between the terms of $\tilde{\mathbb{T}}^*_s(\theta)$ and those of the cross entropy loss, (\ref{eq. minibatch loss inquisitive}) or (\ref{eq. minibatch loss one sampler}), so including the term $\tilde{\mathbb{T}}^*_s(\theta)$ does not bring much additional cost.
Finally, the term becomes $\tilde{\mathbb{T}}^*_s(\theta) + \detach(\mathbb{T}_s(\theta) - \tilde{\mathbb{T}}^*_s(\theta))$ when put into the loss (\ref{eq. loss x lenth running cost}), so that $C(\theta)$ gets the right coefficient.

In summary, the latent distribution $P_{\theta}$ can be trained by gradient descent on the following loss
\begin{equation*}
C_s(\theta) \big(\tilde{\mathbb{T}}^*_s(\theta) + \detach(\mathbb{T}_s(\theta) - \tilde{\mathbb{T}}^*_s(\theta)) \big)
\end{equation*}
for each step $s$.
It is a product of (the estimators of) cross entropy and elapsed time, and is derived from the unified objective (\ref{eq. unified objective}) with little artificial design.
Meanwhile, the sampler(s) is trained by (\ref{eq. minibatch sampler}) and (\ref{eq. train sampler reward}) as usual.

\subsection{Stage Three: Active Lifting}
\label{sec. long-term improvement}

Most of the road map (Figure \ref{fig: climb hierarchy}) has been derived in Sections \ref{sec. quick improvement} and \ref{sec. short-term improvement},
except the entry ``active lifting", which we discuss now.

So far our modeling uses a static format.
The routine is always to start from a manually designed projection $\proj$,
derive its lifting $\fip$ and sampler space $\Q_{\proj}$,
and then train the latent model $P$ and sampler $Q$ (and $Q^{\see}$).
That is why we use the term ``static lifting" in Figure \ref{fig: climb hierarchy}.
While this approach to slow thinking is quite versatile (considering the families $\PS_{\text{simple}}$ and $\PS_{\text{persist}}$ in the representation hierarchy), humans are more active thinkers.
We were not taught what format we should perform slow thinking with (at least our ancestors did not have this luxury), but invented the various ways of thinking on our own.
So it is natural to consider a generalized model with more degrees of freedom, which can develop into the projected distribution $\proj\#P$ and sampling-based inference, as well as many other possible forms.
In a narrow sense, the purpose is to enhance the expressivity of the model and go further on the representation hierarchy.
In a broad sense, a first-principles modeling of the agency of slow thinking may enable the model to invent concepts, schema and languages in an unsupervised manner.

While a detailed plan is provided in Section \ref{sec. perceptual} (in particular Section \ref{sec. adapt to text}), the main idea is simple.
We have seen in Section \ref{sec. unified objective} that the unified objective (\ref{eq. unified objective}) is sufficiently general such that its optimization leads to the maximization of both approximation ability and compute efficiency.
Thus, this objective should also be capable of inducing a projection from scratch.
To design a model space that includes the projections as a subset and supports end-to-end training, we let the latent samplers be the free variable, such that they can map to distributions over any latent sequences, unconstrained by $\fip$.
Then, it can be shown that if a sampler is an optimizer for the unified objective, then it must be restrict to the inverse sets $\proj^{-1}(x)$ of some function $\proj$ (cf.\@ Remark \ref{remark. induce projection}).
Thus, a projection implicitly emerges, and the latent samplers can be seen as an active version of lifting.

This approach seems very general, so Section \ref{sec. perceptual} considers the modeling of data from all modalities.
Both training and inference are discussed for this long-term plan.



\section{The Agency of Perceptual Representation}
\label{sec. perceptual}

Finally, we introduce the general theory of active lifting.
The technical goal is to generalize the distribution representation $\proj\#P$ (and Problem \ref{problem. static}) to a more active model without a prescribed projection.
The broader goal is a universal approach to constructing encoders and generative models for all data modalities,
such that they are able to induce concepts and schema in their latent sequences in a completely unsupervised fashion.

This section tries to capture mathematically the intuition of ``agency and urgency" from Example \ref{ex. capybara}.
Regarding agency, our modeling endows encoders and generative models with an internal time axis, and in particular an encoder needs to search for reasonable ``understandings".
Regarding urgency, a training loss, which resembles minimum-length coding compression, is derived from the unified objective of Section \ref{sec. unified objective}.
We also discuss the possibility of unifying the three existing approaches to image modeling,
which are patch autoregression, diffusion, and the classical objects-and-parts representation.
Still, the present discussion is purely theoretical, and the model design and experimental verification are left for future work.

The basic idea is that a latent sequence $z\in\Omega^*$ can serve as a description of some data sample $x$.
As $z$ grows longer, more information is specified, and eventually an infinite sequence $z \in \Omega^{\omega}$ uniquely pins down this $x$.
The act of perceiving can be modeled as the decoding of latent sequences,
and the act of understanding as the search for latent sequences with high prior probability $P^{\leq |z|}(z)$, where $P$ is the marginal distribution of latent sequences (or the prior knowledge).
The decoding steps give rise to the temporal dimension of perception that is independent of the ``spatial" dimension of the data.
Suppose one is impatient and wants to eliminate as quickly as possible the uncertainty of the scene that one is looking at (e.g.\@ the capybara in Example \ref{ex. capybara}).
Then, each decoded token should carry as much information as possible.
At the same time, these sequences should be understandable for one's mind, i.e.\@ can be approximated by a latent distribution model $P$.
So one needs to invent one's own ``mental language" to optimize for both the efficiency and regularity of latent descriptions.




\subsection{General Setup of Latent Sampling}
\label{sec. incremental modeling}

To begin with, this section defines latent sampling for data distributions of any modality.

Let $\X$ be any topological space,
and we call each element $x \in \X$ a data sample.
Let $\PS(\X)$ be the space of Borel probability measures and denote the target distribution by $P_* \in \PS(\X)$.
This is a very general setting and includes all modalities that we know of.
Later our examples will focus on texts and images:
\begin{itemize}
\item Text:
$\X_{\text{text}} = \Sigma^*$ given a finite set (or vocabulary) $\Sigma$.
This is the setting of our previous sections.

\item Image:
$\X_{\text{image}}$ can be defined as the set of measureable functions from $[0,1]^2$ to $[0,1]^3$, where $[0,1]^2$ and $[0,1]^3$ represent the two-dimensional image domain and the RGB color space.
It may use the weak-$*$ topology of $L^{\infty}$ or the topology induced by some perceptually realistic metric based on \cite{johnson2016perceptualloss,liu2021generic}.
Note that we are not restricted to the coarse grid of square patches that is commonly used in present models \cite{dosovitskiy2020image,caron2021emerging}.
\end{itemize}
One can similarity define the spaces for videos, graphs, molecules, and so on.


Let $\Omega$ be an arbitrary finite set (with size at least 2), and we refer to $\Omega^{\omega}$ with the product topology as the latent space.
We define latent samplers as follows.
\begin{definition}
\label{def. general sampler}
A sampler is a measureable function $Q:\X\times\Omega^*\to\PS(\Omega)$.
Namely, for each $x\in \X$ and $z \in \Omega^*$, we have a conditional distribution $Q(\cdot|x,z)$ over $\Omega$.
The space of samplers is denoted by $\Q_{+1}$.
\end{definition}
Given any sampler $Q\in\Q_{+1}$ and length $T \in \N$, a sampler over $\Omega^T$ can be constructed by the autoregressive decoding of $Q$:
\begin{equation}
\label{eq. image sampler autoregressive}
\forall z \in \Omega^T, \quad Q^{\leq T}(z|x) := \prod_{t=1}^T Q(z_t|x,z_{<t})
\end{equation}
Moreover, by Proposition \ref{prop. from next-token to infinity}, the conditional distributions $\{Q(\cdot|x,z)\}_{z\in\Omega^*}$ given any $x\in\X$ corresponds to a distribution over $\Omega^{\omega}$,
and we refer to this correspondence as the sampler $Q^{\leq \omega}\colon \X\to\PS(\Omega^{\omega})$.
In general, we call $Q^{\leq T}$ and $Q^{\leq \omega}$ the samplers induced by $Q$.

\begin{definition}[Lifted distribution]
\label{def. lifted distribution}
Given any $P_*\in\PS(\X)$ and $Q\in\Q_{+1}$, the lifted distribution is defined as
\begin{equation*}
P_{\text{lift}} = P_* \otimes Q^{\leq\omega} \in \PS(\X\times\Omega^{\omega})
\end{equation*}
which means that, for any bounded continuous test function $f:\X\times\Omega^{\omega}\to\R$,
\begin{equation*}
\int_{\X\times\Omega^{\omega}} f(x,z) dP_{\text{lift}}(x,z) = \int_{\X}\int_{\Omega^{\omega}} f(x,z) dQ^{\leq\omega}(z|x) dP_*(x)
\end{equation*}
\end{definition}


Then, consider the factorization of the lifted distribution.
Let $Q_*^{\leq\omega}$ be the marginal distribution of $P_{\text{lift}}$ over $\Omega^{\omega}$.
By the disintegration theorem \cite[Theorem 3.4]{kallenberg2021foundations}, there exists a unique conditional distribution $P_*^{\leq\omega}$ (a Borel measureable function $\Omega^{\omega}\to\PS(\X)$) that satisfies
\begin{equation}
\label{eq. joint distribution identity}
P_*^{\leq \omega} \otimes Q^{\leq \omega}_* = P_* \otimes Q^{\leq \omega} = P_{\text{lift}}
\end{equation}
The construction becomes explicit if we consider finite sequences.
For any $t\in\N$, define the marginal distribution $Q_*^{\leq t}$ and conditional distribution $P_*^{\leq t}$ by
\begin{align*}
Q^{\leq t}_* &= \int Q^{\leq t}(\cdot|x) dP_*(x) \\
\forall z \in \sprt Q^{\leq t}_*, \quad P_*^{\leq t}(\cdot|z) &= \frac{Q^{\leq t}(z|\cdot)}{Q_*^{\leq t}(z)} P_*
\end{align*}
and $P_*^{\leq t}(\cdot|z)$ with other $z$ can be set arbitrarily.
This $P_*^{\leq t}(\cdot|z)$ is the posterior distribution of data whose latent sequences may possibly start with $z$.
One can check that each $Q^{\leq t}_*$ is the marginal distribution of $Q^{\leq \omega}_*$ over the first $t$ entries, and that
\begin{equation}
\label{eq. joint distribution identity finite}
P_*^{\leq t} \otimes Q^{\leq t}_* = P_* \otimes Q^{\leq t}
\end{equation}
For convenience, we often omit the superscript and denote $P_*(\cdot|z) = P_*^{\leq|z|}(z)$ for any $z\in\Omega^{\leq \omega}$.

This formulation will be used for representation learning.
The sampler $Q^{\leq t}$ can be interpreted as a probabilistic encoder that maps the observed data $x$ to a length-$t$ description $Z \sim Q^{\leq t}(\cdot|x)$.
The next section will derive a training objective, which might capture our intuition of a good description.
For instance, if $x$ is an image, the sample $Z$ from a well-trained sampler would be an accurate and concise description of its contents;
if $x$ is a difficult text, $Z$ would be an understanding that fully explains its logic.
The marginal distribution $Q^{\leq t}_*$ can be interpreted as the prior knowledge, which records the possible descriptions $Z$ of any data from $P_*$.
For instance, if $z \in \Omega^t$ is a description of a cat flying in the sky, then the likelihood $Q^{\leq t}_*(z)$ would be near zero, as it is very unlikely that an image $X \sim P_*$ can induce this $z$ (except in ``Tom \& Jerry").

The challenge is that we cannot assume \textit{a priori} the existence of any language, and the prior $Q^{\leq \omega}_*$ needs to evolve into a language from scratch.
By language, we mean an assignment of token sequences to data $X\sim P_*$ that is both efficient (a short description can roughly pin down $X$) and regular (learnable by parametrized functions such as Transformers).
Perhaps this process can be compared to how our human ancestors developed perceptual concepts and ultimately invented natural languages.
Even if the data are $\X_{\text{text}}$, the latent sequences $z \in \Omega^{\leq\omega}$ are not necessarily in the same language;
instead, they are more likely to resemble the mental languages that form our thoughts \cite{rescorla2024LoTH,fodor1975LoTH}.
While efficiency follows directly from the unified objective (Section \ref{sec. unified objective}), regularity is the result of fitting $Q^{\leq \omega}_*$ with a latent distribution model $P$.
After all, unlike $Q^{\leq t}$, the prior $Q^{\leq t}_*$ is not directly accessible and needs to be approximated.
We will see that training can proceed as long as we have the sampler and latent distribution,
but one can optionally model the posterior data distribution $P_*(\cdot|z)$.
By (\ref{eq. joint distribution identity}), this provides a generative model for $P_*$.

Hence, the static Problem \ref{problem. static} can be generalized as follows, in a slightly informal fashion.
\begin{problem}[active]
\label{problem. active}
Given a target distribution $P_* \in \PS(\X)$,
find a lifted distribution $P_{\text{lift}}\in\PS(\X\times\Omega^{\omega})$ such that the sampler $Q$ and marginal distribution $Q^{\leq\omega}_*$ (and optionally, the posterior $P^{\leq\omega}_*$) belong to some parametrized families that resemble $\PS_{\text{plain}}$.
\end{problem}

To see the connection to the projection parametrization $P_* = \proj\#P$, one can set $\X = \X_{\text{text}}$ and $t=\omega$,
assume that $P_*$ is supported on a prefix-free subset of $\Sigma^*$,
and assume that $P_*(\cdot|z)$ is deterministic, namely $P_*(\cdot|z) = \delta_{f(z)}$ for some function $f\colon \Omega^{\omega}\to\Sigma^*$.
Then, there exists at least one function $\proj\colon \Omega^{\omega}\to\Sigma^{\omega}$ such that $f = \Pi_{\sprt P_*} \circ \proj$,
and thus (\ref{eq. joint distribution identity}) implies that
\begin{equation*}
\Pi_{\sprt P_*} \# ( \proj \# Q^{\leq \omega}_* ) = P_*
\end{equation*}
By Proposition \ref{prop. cross entropy to KL}, $Q^{\leq \omega}_*$ is a minimizer of the cross entropy loss
\begin{equation*}
\min_{P \in \PS(\Omega^{\omega})} -\int \log (\proj\#P)^{\leq |x|}(x) dP_*(x)
\end{equation*}
The sampler $Q^{\leq \omega}$ equals the posterior distribution
\begin{equation*}
\forall x \in \sprt P_*, \quad Q^{\leq \omega}(\cdot|x) = Q^{\leq \omega}_*\big(\cdot\big| \proj^{-1}([x]) \big)
\end{equation*}
Hence, we have the correspondence:
the latent distribution $P$ becomes the prior knowledge $Q^{\leq \omega}_*$ (or the model $P$ that fits $Q^{\leq \omega}_*$),
the projection $\proj$ becomes the conditional distribution $P_*(\cdot|z)$,
and the old sampler $Q \in \Q_{\proj}$ becomes the new sampler $Q \in \Q_{+1}$.
One major difference is that in the previous sections we are first given $\proj$ and then derive $P$ and $Q$, while in the current setting $Q$ is a free variable and $P,P_*(\cdot|z)$ are dependent variables.

As a remark on implementation, the latent distribution $P$ can share the same model with the sampler $Q$.
Specifically, Definition \ref{def. general sampler} can be modified such that $Q$ becomes a function on $\X \cup \{\varnothing\}$,
and then we can set $P = Q^{\leq \omega}(\cdot|\varnothing)$.
So if the sampler does not receive any input data, then $Q^{\leq t}$ plays the latent distribution and can be used to approximate $Q^{\leq t}_*$ for all $t\in\N \cup \{\omega\}$.

\subsection{Active Lifting}
\label{sec. active lifting}

This section derives from the unified objective (Section \ref{sec. unified objective}) the training loss for the latent sampler $Q \in \Q_{+1}$ and latent distribution $P\in\PS(\Omega^{\omega})$,
as a characterization of the notion of urgency from Example \ref{ex. capybara}.

Recall that the unified objective (\ref{eq. unified objective}) has the form $\int_0^{\infty} C_t dt$, where $t$ is the elapsed time and $C$ is a measure of ``reducible uncertainty",
so the objective maximizes the rate of uncertainty reduction over the real time.
For language models, this reduction happens across the training steps (as in Sections \ref{sec. unified objective} and \ref{sec. balance fast slow}),
but now we consider the observation steps instead.
By observation, we mean the autoregressive decoding process $Z_t \sim Q(\cdot|x,Z_{<t})$ given any data $x$.
As a coarse approximation, we assume that the tokens $Z_t$ are decoded at a constant rate, so the elapsed time can be equated with the observation step.
To define the ``reducible uncertainty" of the observation process,
let us consider for any data $x$ and any observation step $t\in \N \cup\{\omega\}$, the posterior data distribution $P_*(\cdot|Z)$ from (\ref{eq. joint distribution identity}) with $Z$ sampled from $Q^{\leq t}(\cdot|x)$.
Intuitively, it represents the uncertainty of the observer after looking at $x$ for $t$ moments.
A description $Z$ has been obtained, but it is insufficient to uniquely determine $x$, leaving a posterior distribution of possible data whose descriptions may start with $Z$.
The tree of possibilities in Figure \ref{fig: capybara} can be interpreted in this way,
e.g.\@ after the first second, the decoded token $Z_1$ carries the meaning of ``a fluffy creature", and $P_*(\cdot|Z_1)$ is the distribution of a lot of mammals, as depicted by the level 1 sub-tree in Figure \ref{fig: capybara}.

A natural measure of uncertainty would be the negative log-likelihood $-\log P_*(x|Z)$.
The technical issue is that, since the space $\X$ could be uncountable, we should use the density function of $P_*(\cdot|Z)$ instead of the probability values.
Since $\X$ in general does not have a ``natural" base measure (such as the Lebesgue measure for $\R^d$), we need to choose a reference measure to define the density function.
With the reference measure set to $P_*$, we define the density of $P_*(\cdot|Z)$ as the Radon-Nikodym derivative $dP_*(\cdot|Z)/dP_*$.
Then, $C_t$ is defined by
\begin{align}
\nonumber
C_t &= \E_{X\sim P_*}\E_{Z \sim Q^{\leq t}(\cdot|X)} \Big[ -\log \frac{d P_*(\cdot|Z)}{d P_*} (X) \Big] \\
\label{eq. reducible uncertainty with prior distribution}
&= \iint_{\Omega^t} \log \frac{Q^{\leq t}_*(z)}{Q^{\leq t}(z|x)} dQ^{\leq t}(z|x)dP_*(x) \\
\nonumber
&= H(Z|X)-H(Z), \quad (X,Z) \sim P_* \otimes Q^{\leq t}
\end{align}
The second line follows from (\ref{eq. joint distribution identity finite}), and the Radon-Nikodym derivative is replaced by log-likelihoods since the space $\Omega^t$ is finite.
The third line formally resembles the mutual information
\begin{equation}
\label{eq. reducible uncertainty mutual information}
- C_t = I(X;Z) = H(X)-H(X|Z) = H(Z)-H(Z|X)
\end{equation}
However, this only holds when $H(X),H(X|Z)$ are well-defined (e.g.\@ when $\X$ is countable), so we do not use the term $I(X;Z)$ to avoid confusion.


Next, we include the latent distribution $P$ in the training objective.
Given a parametrized latent distribution $P \in \PS(\Omega^{\omega})$,
its objective is to fit the prior distribution $Q^{\leq \omega}_*$.
Since the marginals of $Q^{\leq \omega}_*$ are $Q^{\leq t}_*$, Proposition \ref{prop. from next-token to infinity} indicates that it suffices for $P^{\leq t}$ to fit $Q^{\leq t}_*$ for each $t\in\N$,
and an appropriate loss would be $\KL(Q^{\leq t}_* \| P^{\leq t})$.
We rewrite (\ref{eq. reducible uncertainty with prior distribution}) as
\begin{align}
\nonumber
C_t &= \max_{P\in\PS(\Omega^{\omega})} \iint_{\Omega^t} \log \frac{Q^{\leq t}_*(z)}{Q^{\leq t}(z|x)} dQ^{\leq t}(z|x)dP_*(x) - \KL(Q^{\leq t}_*\|P^{\leq t})\\
\label{eq. reducible uncertainty with latent distribution}
&= \max_{P\in\PS(\Omega^{\omega})} \iint_{\Omega^t} \log \frac{P^{\leq t}(z)}{Q^{\leq t}(z|x)} dQ^{\leq t}(z|x)dP_*(x)
\end{align}
It follows that $P = Q^{\leq \omega}_*$ is the unique maximizer of $\sum_{t=0}^{\infty} C_t$.
Technically, the sum $\sum_{t=0}^{\infty}C_t$ could be infinite, but it suffices to assume that for $T\to\infty$ the sets $M_T$ of maximizers/minimizers of $\sum_{t=0}^T C_t$ form a nested sequence with nonempty intersection $\bigcap_{T=1}^{\infty} M_T$,
so that we can speak of the maximizers/minimizers of $\sum_{t=0}^{\infty} C_t$.

Now every term in $C_t$ is implementable, so we plug it in the unified objective (\ref{eq. unified objective}).
A min-max problem is obtained
\begin{align}
\nonumber
\min_{Q \in \Q_{+1}} \int_0^{\infty} C_{\lfloor t \rfloor}(\theta) dt &= \min_{Q\in \Q_{+1}} \sum_{t=0}^{\infty} C_t \\
\nonumber
&= \min_{Q\in \Q_{+1}} \max_{P\in\PS(\Omega^{\omega})} \sum_{t=0}^{\infty} \iint_{\Omega^t} \log \frac{P^{\leq t}(z)}{Q^{\leq t}(z|x)} dQ^{\leq t}(z|x)dP_*(x) \\
\label{eq. QP min-max objective}
&= \min_{Q\in \Q_{+1}} \max_{P\in\PS(\Omega^{\omega})} \iint_{\Omega^t} \sum_{t=0}^{\infty} \log \frac{P^{\leq t}(z_{\leq t})}{Q^{\leq t}(z_{\leq t}|x)} dQ^{\leq \omega}(z|x)dP_*(x)
\end{align}
For the derivation, we assume that each observation step costs exactly one second, so the time integral becomes a discrete sum.
(Since $C_0 = 0$, it can be omitted in the summation).
Whenever we say that $(Q,P)$ is a global optimizer for the objective (\ref{eq. QP min-max objective}), we mean that $Q$ is a global minimizer of the first line and that $P$ equals the sampler $Q^{\leq \omega}_*$ induced by $Q$.
Meanwhile, for gradient-based training, to make sure that the term $dQ^{\leq t}$ is differentiated, one probably needs to apply the trick of policy gradient to define a mini-batch loss, similar to the ones in Section \ref{sec. sampler training}.


\begin{remark}
\label{remark. induce projection}
An informal argument shows that this objective implicitly produces a projection $\proj$ and an autoencoder.
Specifically, if $(Q,P)$ is a global optimizer, then the posterior data distribution $P_*(\cdot|z)$ should be deterministic for $P$-almost all $z \in \Omega^{\omega}$,
namely
\begin{equation}
\label{eq. deterministic posterior}
P_*(\cdot|z) = \delta_{\proj(z)}
\end{equation}
for some measureable function $\proj\colon \Omega^{\omega}\to\X$,
and also the pair $(Q^{\leq\omega},P_*^{\leq\omega})$ forms an autoencoder
\begin{equation}
\label{eq. active lifting autoencoder}
\int P_*(\cdot|z) dQ^{\leq\omega}(z|x) = \delta_x
\end{equation}
for $P_*$-almost all $x$.
The argument has two steps.
First, assume that there exists some measureable surjection $f:\Omega^{\omega}\to\X$.
By Proposition \ref{prop. Cantor measureable surjection}, this holds for $\X_{\text{text}}$, $\X_{\text{image}}$ and much more general spaces.
(In fact, $f$ could be continuous, which might be helpful in future works;
for instance, the Banach–Alaoglu theorem implies that $\X_{\text{image}}$ is compact metrizable, and then the Cantor surjection theorem \cite{alexandroff1927stetige,schoenfeld1974continuous} can be applied.)
Let $\tilde{P}$ be the uniform probability measure over $\Omega^{\omega}$ (which exists by Kolmogorov extension theorem),
and $\tilde{Q}^{\leq\omega}(\cdot|x)$ be the conditional distribution $P(\cdot|f^{-1}(x))$.
Let $\tilde{Q} \in \Q_{+1}$ be the sampler that induces $\tilde{Q}^{\leq \omega}$, which exists by Proposition \ref{prop. from next-token to infinity}.
Let $\tilde{P}_*(\cdot|z)$ be the conditional distribution induced by $\tilde{Q}$ and (\ref{eq. joint distribution identity}),
and one can check that it is deterministic, $\tilde{P}_*(\cdot|z)=\delta_{f(z)}$.
Thus, the pair $(\tilde{Q},\tilde{P})$ is a solution to
\begin{equation*}
\E_{Z\sim \tilde{P}}\E_{X\sim \tilde{P}_*(\cdot|Z)} \big[ - \log \tilde{P}_*(X|Z) \big] = 0
\end{equation*}
%
Second, assume that, with the global optimizer $(Q,P)$, the sequence $\{C_t\}_{t=1}^{\infty}$ is non-increasing and $\lim_{t\to\infty}C_t \geq C_{\omega}$.
This is a plausible assumption, since for each $t$ we can achieve the baseline $C_{t+s}=C_t$ (for all $s\geq 1$) by fixing some token $a \in \Omega$ and setting $Q(\cdot|x,z) \equiv \delta_a$ for all $x\in\X$ and $z \in \Omega^{\geq t}$.
Then, (\ref{eq. reducible uncertainty mutual information}) implies that
\begin{align*}
\inf_{t\in\N} C_t + H(P_*) &\geq C_{\omega} + H(P_*) \\
&= H(X|Z), \quad (X,Z)\sim P_{\text{lift}} \\
&= \E_{Z\sim P}\E_{X\sim P_*(\cdot|Z)} \big[ - \log P_*(X|Z) \big] \\
&\geq 0
\end{align*}
Suppose $P_*(\cdot|z)$ is not deterministic for $P$-almost all $z$.
Then, the lower bound is positive, making $(P,Q)$ suboptimal, a contradiction.
Hence, a projection $\proj$ is induced from the training objective (\ref{eq. QP min-max objective}).
The identity $\proj\#P=Q^{\leq\omega}\otimes P_*$ implies that $\sprt Q^{\leq\omega}(\cdot|x) \subseteq \proj^{-1}(x)$ for $P_*$-almost all $x$ and thus (\ref{eq. active lifting autoencoder}).
The emergence of determinism is reminiscent of the passage from Kantorovich solutions (probabilistic transport) to Monge solutions (deterministic transport) in optimal transport theory \cite{brenier1991polar,villani2003topics}.
\end{remark}


Since the present framework can derive the projection function, it is a more general theory that encompasses the static theory from the preceding sections.
The samplers $Q^{\leq t}$ can be regarded as an ``active lifting":
On one hand, if determinism (\ref{eq. deterministic posterior}) holds, then each $Q^{\leq t}$ maps into the set
\begin{equation*}
\proj^{-1}_t(x) = \{ z_{\leq t} \mid z \in \proj^{-1}(x)\}
\end{equation*}
which resembles the lifting $\fip$ (Definition \ref{def. lifting});
on the other hand, $Q$ is a free variable that determines $\proj$, unlike $\fip$ which is a dependent variable determined by $\proj$.
A detailed comparison can be made:
\begin{alignat*}{3}
\proj &\longrightarrow& \fip &\longrightarrow& P &\longrightarrow Q_*/Q_{\eye} \\
Q &\longrightarrow& Q^{\leq \omega} &\longrightarrow& P &\longrightarrow \proj
\end{alignat*}
The two derivations roughly follow the same routine.
Both start with a conversion between finite-sequence operators and infinite-sequence operators, then obtain the latent distribution $P$ from training, and finally determine the remaining variable.

\subsection{Minimum-Length Regular Coding}
\label{sec. minimum-length coding}

This section analyzes the training objective (\ref{eq. QP min-max objective}),
showing that it resembles compression,
specifically minimum-length coding with some regularity constraint.
If we characterize ``language" (at least descriptive language) as an assignment of descriptions to data samples that is both efficient and learnable,
then this objective compels the model to invent a language on its own.

First, we establish the connection to minimum-length coding.
Let us consider the simple setting when $Q$ and $P$ can range in all of $\Q_{+1}$ and $\PS(\Omega^{\omega})$, instead of being constrained to any parametrized families.
Suppose $(Q,P)$ is a global optimizer,
then $P$ always equals the prior $Q^{\leq \omega}_*$.
Assume for simplicity that $\X$ is a countable set.
Then, we express $C_t$ in terms of (\ref{eq. reducible uncertainty mutual information}) and omit the constant term $H(P_*)$.
The objective (\ref{eq. QP min-max objective}) is equivalent to
\begin{align}
\nonumber
\min_{Q^{\leq\omega}\in\Q^{\omega}}& \sum_{t=0}^{\infty} H(X|Z_{\leq t}), \quad (X,Z) \sim P_* \otimes Q^{\leq \omega} \\
\label{eq. max rate reduce NLL}
= \min_{Q^{\leq\omega}\in\Q^{\omega}}& \iint \sum_{t=0}^{\infty} -\log P(x|z_{\leq t}) dQ^{\leq\omega}(z|x) dP_*(x)
\end{align}
where $\Q^{\omega}$ denotes the space of measureable functions from $\X$ to $\PS(\Omega^{\omega})$.
In terms of the interpretations in Section \ref{sec. unified objective}, we are maximizing the reduction rate of conditional entropy.

\begin{figure}[!ht]
\centering
\subfloat{\includegraphics[width=0.46\textwidth]{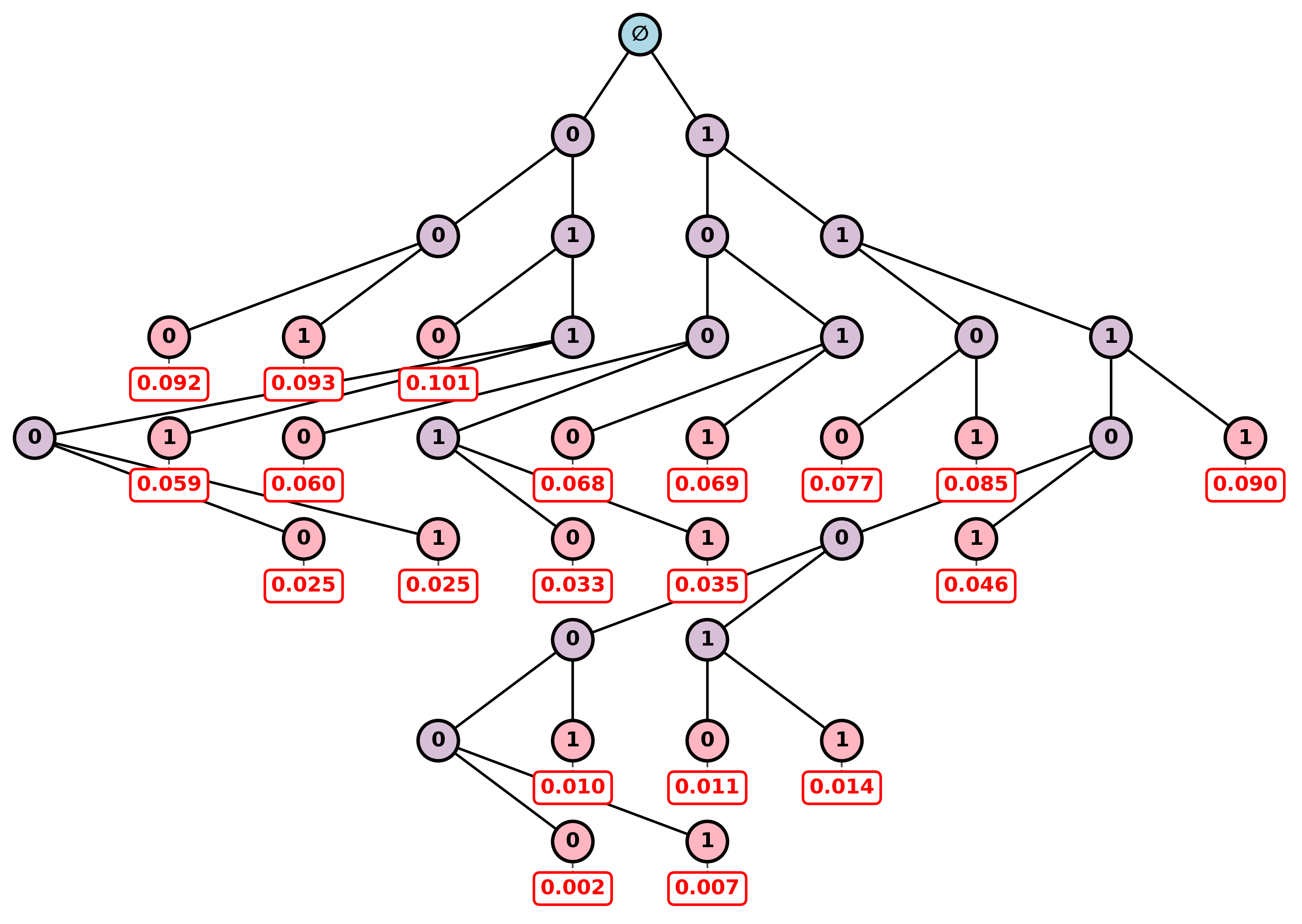}}
\subfloat{\includegraphics[width=0.26\textwidth]{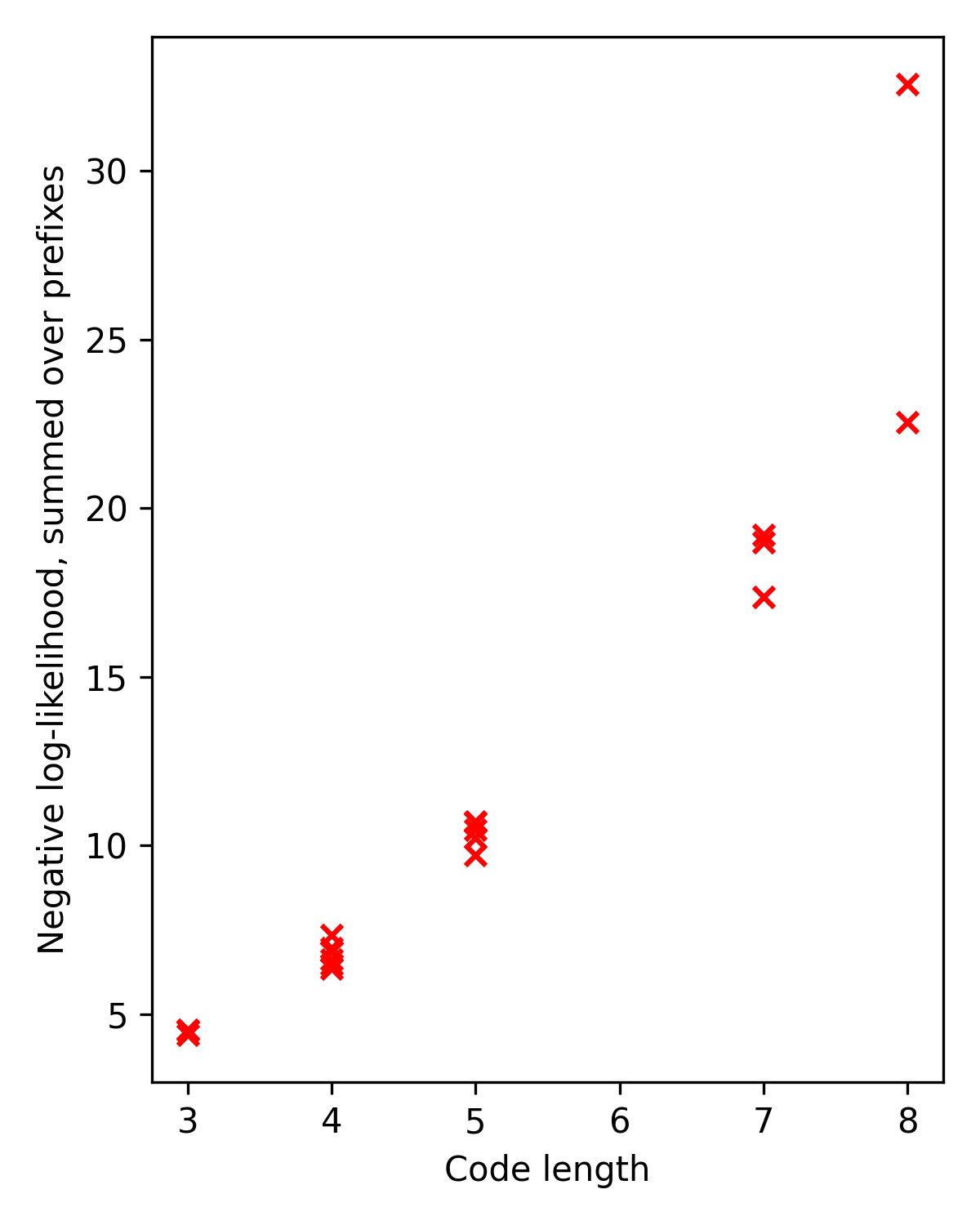}}
\subfloat{\includegraphics[width=0.26\textwidth]{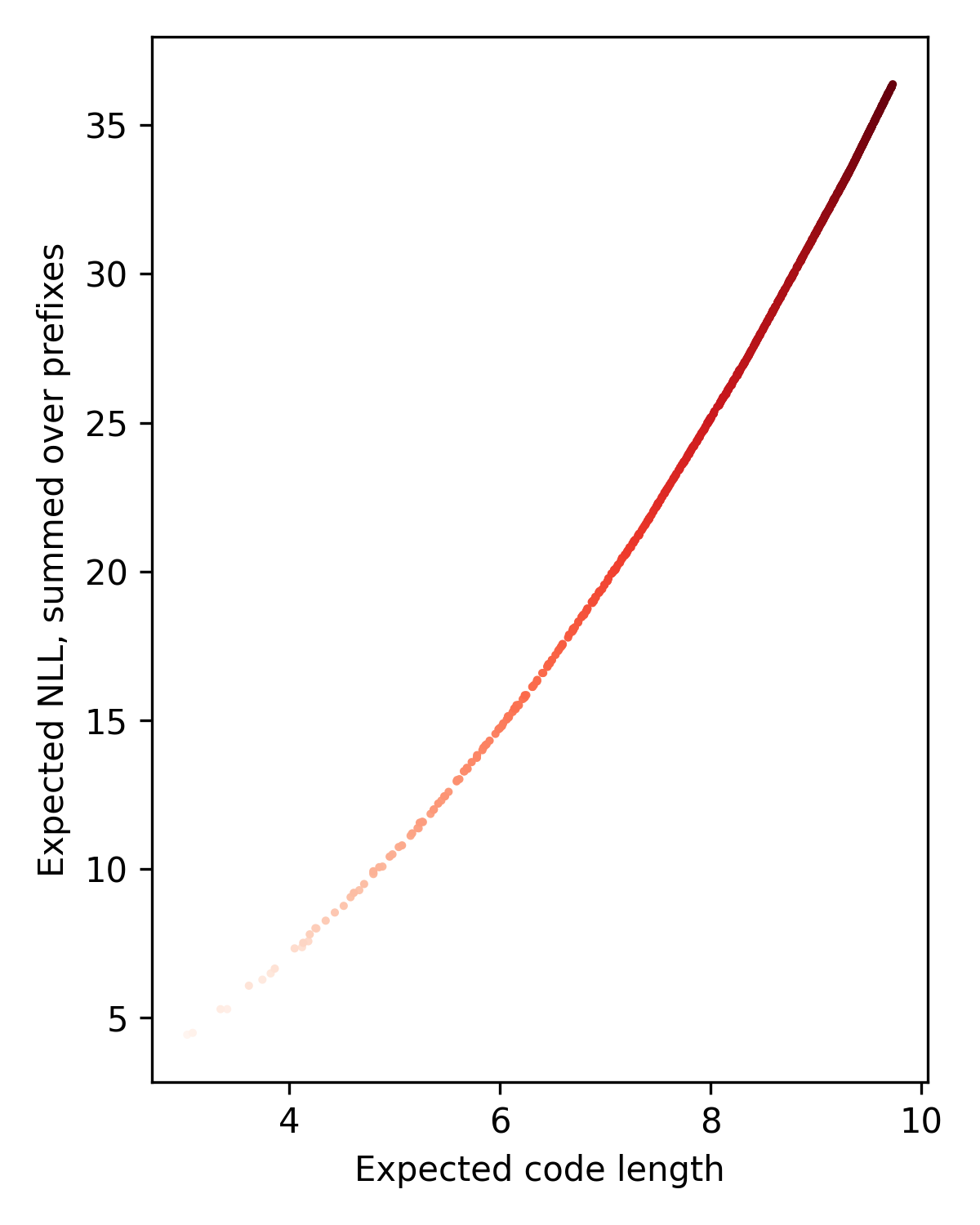}}
\caption{Illustration of the resemblance between minimum-length coding and our training objective.
Suppose $\Omega=\{0,1\}$.
Left: The Huffman tree for some distribution $P_*$ over a set $\X$ with size $|\X|=20$.
It is approximately the minimum-length coding of $P_*$.
Each leaf (red node) represents a code $c(x)$, with its probability $P_*(x)$ annotated underneath.
Middle: A scatter plot depicting the positive correlation (\ref{eq. approximate code length by NLL}).
The two axes are the two sides of (\ref{eq. approximate code length by NLL}), with each $x \in \X$ shown as a point.
These quantities are computed based on the distribution and coding of the left panel.
Right: A scatter plot depicting the positive correlation (\ref{eq. approximate expected code length by expected NLL}).
For each $N\in\{10,\dots 1000\}$, a distribution $P_*$ over $N$ elements is randomly generated by uniformly sampling in $[0,1]^N$ and then normalizing by the sum.
Then, the Huffman coding $c$ for $P_*$ is obtained, and the two sides of (\ref{eq. approximate expected code length by expected NLL}) are computed.
The color gradient indicates the increasing of $N$.}
\label{fig: binary tree}
\end{figure}

A classic problem in information theory is entropy coding, or the minimization of expected code length \cite{shannon1948mathematical,huffman2007method}.
A particular version of this problem is that, given any countable set $\X$, distribution $P_* \in \PS(\X)$ and finite vocabulary $\Omega$ ($|\Omega|>1$),
construct a coding function $c$ to minimize the expected length $\E_{X\sim P_*} [|c(X)|]$.
A coding is defined as an injective function $c\colon \X\to\Omega^*$ whose image set $c(\X)$ is prefix-free.
Shannon's source coding theorem implies that the expected length is lower bounded by entropy $\E_{X\sim P_*}[-\log_{|\Omega|}P_*(X)]$.
An example of an (approximately) optimal coding based on Huffman tree is given by Figure \ref{fig: binary tree} (left).
Note that with any coding $c$ we can make the following approximation
\begin{align}
\label{eq. approximate code length by NLL}
|c(X)| = \sum_{t=0}^{\infty} \mathbbm{1}_{t < |c(X)|} &\approx \sum_{t=0}^{\infty} -\log P_*(X\mid z(X)_{\leq t}) \\
\nonumber
\text{where} \quad z(x) &= c(x) \blank^{\omega} \\
\nonumber
\forall z' \in \Omega^*, \quad P_*(\cdot|z') &= \law(X \mid z' \sqsubseteq z(X))
\end{align}
The coding $z(X)$ is $c(X)$ padded with some blank symbol $\blank \notin \Omega$ so that the prefix $z(X)_{\leq t}$ can be defined for all $t$.
The intuition of this approximation is that for $t < |c(X)|$, there are possibly other $x'$ with the same prefix $c(X)_{\leq t} = c(x')_{\leq t}$, which make $P_*(X|c(X)_{\leq t}) < 1$,
while for $t \geq |c(x)|$, we always have $P_*(X|c(X)_{\leq t})=1$.
Thus, $\mathbbm{1}_{t< |c(X)|}$ can be seen as the binarized version of $-\log P_*(X\mid z(X)_{\leq t})$.
Figure \ref{fig: binary tree} (middle) illustrates the positive correlation between the two sides of (\ref{eq. approximate code length by NLL}).
It follows that
\begin{equation}
\label{eq. approximate expected code length by expected NLL}
\E_{X\sim P_*}\big[|c(X)|\big] \approx \int \sum_{t=0}^{\infty} -\log P_*(x|z(x)_{\leq t}) dP_*(x)
\end{equation}
and this positive correlation is depicted in Figure \ref{fig: binary tree} (right).

Note that our training objective (\ref{eq. max rate reduce NLL}) highly resembles (\ref{eq. approximate expected code length by expected NLL}),
as the sampling distribution $Q^{\leq\omega}(\cdot|x)$ can be seen as a stochastic generalization of the coding $z(x)$.
Hence, one may claim that the unified objective (\ref{eq. unified objective}) induces minimum-length coding when perception is modeled by latent sampling.
For concreteness, consider a distribution $P_*$ of photographs.
Since the objects are the recurring patterns in photographs, they are the most distinctive features from a minimum-length coding perspective.
Thus, if a sampler $Q$ is trained on $P_*$, one may expected that its latent tokens $Z_t$ will encode these objects and their parts, attributes, relations, etc.
A slightly more rigorous discussion will be provided in Section \ref{sec. visual field}.
Creating concepts from unsupervised perceptual data is known as concept induction \cite{higgins2017beta,ke2025infant},
and this training objective may be sufficient for the sampler to induce concepts.

Next, we consider regularity.
Without loss of generality, suppose the latent distribution $P$ is a Transformer LLM, i.e.\@ $P \in \PS_{\text{plain}}$ from Section \ref{sec. approximation summary}.
Denote $P=P_f$ with function parameter $f$.
By definition of a global optimizer, an optimal sampler $Q$ must ensure that its marginal distribution $Q^{\leq \omega}_*$ is achievable (or at least approximable) by $P_f$.
Thus, Theorem \ref{thm. separation I} (and Figure \ref{fig: representation hierarchy}) implies that $Q^{\leq \omega}_*$ cannot involve a too complicated dependency structure.
For instance, if $P_*$ is a distribution of images of book pages, and each latent token $z_t$ encodes a printed word,
then a sampled sequence $Z \sim Q^{\leq \omega}_*$ corresponds to some arrangement of the words in a book page.
In order for $Z$ to be easily readable by $P_f$, the sampler $Q$ needs to arrange these words in the usual linear order, instead of some arbitrary order.
Similarly, if $P_*$ consists of photographs of daily-life scenes, and each $z_t$ encodes a visual concept, then it might be more tractable for $P_f$ to have these concepts presented in the SVO format (subject-verb-object, such as ``the face contains two eyes") instead of some random VSSOOVOSO\dots.
Hence, the training objective not only induces perceptual concepts in $Q^{\leq \omega}_*$, but also forces $Q^{\leq \omega}_*$ to organize these concepts in some learnable format, grammar, or schema.

\subsection{New Approach to Generative Modeling}
\label{sec. generative}

This section discusses how to turn the posterior data distribution $P_*(\cdot|z)$ into a generative model.
Informally, since the equation (\ref{eq. joint distribution identity}) has only one ``degree of freedom" and the sampler $Q$ is a free variable, it should be straightforward to obtain $P_*(\cdot|z)$, just like fitting $Q^{\leq \omega}_*$ with a latent model $P$.
This task is optional, as the objective (\ref{eq. QP min-max objective}) only requires $Q$ and $P$.

Generally speaking, given a target distribution $P_*$,
a generative model is a solution to
$$\int G(\cdot|z)dP(z) = P_*$$
consisting of a latent distribution $P$ and a conditional distribution $G$ that are
both easy to sample from.
This $G$ is commonly called the generator.
Then, one can sample from $P_*$ by
\begin{equation}
\label{eq. generative model sampling}
X\sim G(\cdot|Z), \quad Z\sim P
\end{equation}
Often, $P$ is set to be unit Gaussian, and $G$ is simplified to a measureable function $\proj$ to solve for $\proj\#P = P_*$.
One of the greatest difficulties for designing a generative model is the non-uniqueness of the solution \cite{yang2023thesis}.
The classical approach, adopted by models such as the normalizing flows \cite{tabak2010density}, variational autoencoder \cite{kingma2013auto} and generative adversarial networks \cite{goodfellow2014generative}, is to allow for any solution,
so their loss landscapes are highly nonconvex and they are prone to training failure \cite[Sections 6.3-6.4]{yang2023thesis}.
An alternative approach, adopted by models such as diffusion \cite{song2019generative} and flow matching \cite{albergo2022building}, is to adopt the trivial solution $G(\cdot|z) \equiv P_*$ and try to extend it to some stochastic path whose velocity field can be expressed as an expectation over $P_*$, thus converting the problem to regression.
Although these two approaches enjoy empirical success, it remains a question of theoretical interest whether one can construct, \textit{a priori}, a non-trivial yet tractable target $P_*(\cdot|z)$ for the generator to fit.

Such a target $P_*(\cdot|z)$ can be obtained from our modeling.
Recall from Remark \ref{remark. induce projection} that the posterior data distribution $P_*^{\leq t}$ for each $t \in \N_+ \cup\{\omega\}$ tends to be deterministic,
as a result of minimizing the general objective $\sum_{t=0}^{\infty} C_t$.
So $P_*^{\leq t}$ is less trivial than $G(\cdot|z) \equiv P_*$.
Besides, Section \ref{sec. minimum-length coding} suggests that the latent distribution $Q^{\leq \omega}_*$ might resemble a descriptive language with concepts and grammar,
so $P_*^{\leq t}$ can be seen as an artist/writer that creates an image/text based on its description.
Thus, $P_*^{\leq t}$ is a non-trivial conditional distribution.

Meanwhile, $P_*^{\leq t}$ is implementable, as we can train a generator to fit it with regression.
Suppose the generator $G$ is a measureable function $\Omega^{\leq\omega}\to\PS(\X)$ and its likelihood $G(x|z)$ is well-defined.
Denote by $G^{\leq t}$ the restriction of $G$ to $\Omega^t$.
Given any sampler $Q$, for each $t$ we have
\begin{align}
\label{eq. train generator and target}
\E_{Z\sim Q^{\leq t}_*} \Big[ \KL \big( P^{\leq t}_*(\cdot|Z) \big\| G^{\leq t}(\cdot|Z) \big) \Big] &= \E_{X\sim P_*} \E_{Z \sim Q^{\leq t}(\cdot|X)}\big[ - \log G^{\leq t}(X|Z) + \log P^{\leq t}_*(X|Z) \big]\\
\label{eq. train generator passive}
&= \E_{X\sim P_*} \E_{Z \sim Q^{\leq t}(\cdot|X)}\big[ - \log G^{\leq t}(X|Z) \big] + \text{constant}
\end{align}
The last line is straightforward to carry out as we only train $G$.

Hence, the training objective (\ref{eq. QP min-max objective}) may lead to a new approach to generative modeling.
The aforementioned two approaches (unprescribed solution and trivial solution) are sometimes called the ``free coupling" and ``product coupling" \cite{yang2023thesis}, where coupling means a joint distribution whose two marginal distributions are $P$ and $P_*$.
Thus, one may refer to this new approach as the ``linguistic coupling", given that Section \ref{sec. minimum-length coding} has compared active lifting to descriptive languages.



\begin{remark}
The training loss (\ref{eq. train generator passive}) is passive in the sense that the sampler $Q$ is fixed and only the generator $G$ is optimized.
Alternatively, one may consider a more active objective that combines (\ref{eq. QP min-max objective}) and (\ref{eq. train generator and target}):
\begin{align}
\nonumber
\min_Q 2\sum_{t=0}^{\infty} C_t &= \min_{Q,G} \max_P \sum_{t=0}^{\infty} 2C_t + \E_{Z\sim Q_*^{\leq t}}\big[\KL\big(P^{\leq t}_*(\cdot|Z) \| G^{\leq t}(\cdot|Z) \big) \big] - \KL(Q_*^{\leq t}\|P^{\leq t}) \\
\label{eq. train generator active}
&= \min_{Q,G} \max_P \iint \sum_{t=0}^{\infty} \log \frac{P^{\leq t}(z_{\leq t})}{Q^{\leq t}(z_{\leq t}|x) G^{\leq t}(x|z_{\leq t})} dQ^{\leq\omega}(z)dP_*(x) + \text{constant}
\end{align}
where the constant $2$ is included to cancel out all occurrences of $Q_*^{\leq t}$.
This objective is active in the sense that while $G$ tries to fit each $P_*^{\leq t}$, the sampler $Q$ is compelled to make $P_*^{\leq t}$ more learnable.
For instance, $Q_*^{\leq\omega}$ may invent more visual concepts (as discussed in Section \ref{sec. minimum-length coding}) to describe details to help with image reconstruction.
In analogy, one may consider the distinction between the visual representations of visual artists and non-artists,
e.g.\@ artists are more capable of recognizing and copying local details, not subject to top-down affects \cite{chamberlain2013local,drake2024artists}.
\end{remark}



\subsection{The Time Axis of Inference}
\label{sec. time axis}

Having discussed training, we move on to inference.
This section characterizes the agency of perception, equipping encoders with a time axis.

As studied in Section \ref{sec. latent sampling}, inference involves the four tasks of $\{$unconditional, conditional$\}\times\{$encoding, generation$\}$, where conditional means $P_*$ conditioned on some event (or measureable subset) $A$.
This section studies unconditional inference, while conditional inference is discussed in the next section, with a focus on text data.
Since unconditional generation is straightforward by (\ref{eq. generative model sampling}),
we consider unconditional encoding, namely estimating the probabilities $P_*(x)$ given any input $x\in\X$.
Suppose $\X$ is a countable set (e.g.\@ $\X_{\text{text}}$ or $\X_{\text{image}}$ with finite resolution), so that the probabilities are meaningful.

Note that for any $x\in\X$ and $t\in\N\cup\{\omega\}$,
\begin{equation*}
P_*(x) = \int P_*^{\leq t}(x|z) dQ^{\leq t}_*(z) = \E_{Z\sim Q^{\leq t}(\cdot|x)} \Big[ \frac{Q^{\leq t}_*(Z) P_*(x|Z)}{Q^{\leq t}(Z|x)} \Big]
\end{equation*}
The identity (\ref{eq. joint distribution identity finite}, \ref{eq. joint distribution identity}) implies that $Q^{\leq t}$ is exactly the variance-minimizing sampler (or posterior sampler) if we use the right-hand side as a Monte-Carlo estimator.
Suppose we have a well-trained latent model $P \approx Q^{\leq\omega}_*$ (trained on the objective (\ref{eq. QP min-max objective}))
and generator $G^{\leq t} \approx P_*^{\leq t}$ (trained on either (\ref{eq. train generator passive}) or (\ref{eq. train generator active})).
Then,
\begin{equation*}
P_*(x) \approx \E_{Z\sim Q^{\leq t}(\cdot|x)} \Big[ \frac{P^{\leq t}(Z) G^{
\leq t}(x|Z)}{Q^{\leq t}(Z|x)} \Big]
\end{equation*}
Thus, a Monte-Carlo estimator can be constructed in practice by
\begin{equation}
\label{eq. active encode fixed t}
p_{t,n} = \frac{1}{n} \sum_{i=1}^n \frac{P^{\leq t}(Z^{(i)}) G^{\leq t}(x|Z^{(i)})}{Q^{\leq t}(Z^{(i)}|x)}, \quad \{Z^{(i)}\}_{i=1}^n \iidsample Q^{\leq t}(\cdot|x)
\end{equation}
This formulation has an intuitive interpretation:
each sampled $Z$ is an understanding of the observation $x$;
$P^{\leq t}(Z)$ decides whether $Z$ is an \textit{a priori} reasonable description;
$G^{\leq t}(x|Z)$ decides whether $Z$ accurately describes the observation;
the sampler $Q$ needs to search for a good understanding $Z$ that satisfies both critics.
Thus, encoding becomes an active process.

To align with the interpretation of $t$ as observation steps,
note that we can incrementally calculate the estimators $p_{t,n}$ for all $t \in \N$.
This may resemble how humans look at a scene and continuously update their understandings.
For convenience, suppose $G$ as a likelihood function satisfies
\begin{equation*}
\forall x\in\X, ~ \forall z\in\Omega^{\omega}, ~ \forall T \in\N, \quad \log G^{\leq T}(x|z_{\leq T}) = \sum_{t=0}^T g(x,z_{\leq t})
\end{equation*}
for some function $g\colon \X\times\Omega^*\to\R$.
Consider the following calculations, for each $i=1,\dots n$ in parallel,
\begin{align*}
Z^{(i)}_{\leq t} &= Z^{(i)}_{<t}Z_t^{(i)}, \quad Z_t^{(i)} \sim Q(\cdot|x,Z^{(i)}_{<t}), \quad Z^{(i)}_0 = \varnothing \\
\log q^{(i)}_t &= \log q^{(i)}_{t-1} + \log P^{\leq t}(Z^{(i)}_t|Z^{(i)}_{<t}) + g(x,Z^{(i)}_{\leq t}) - \log Q(Z^{(i)}_t|x,Z^{(i)}_{<t}) , \quad \log q^{(i)}_0 = 0
\end{align*}
Then, at each step $t$, one obtains
\begin{equation}
\label{eq. active encode incremental t}
p_{t,n} = \frac{1}{n} ~ \text{logsumexp}\big( (\log q^{(i)}_t)_{i=1}^n \big)
\end{equation}

The estimation error $\E[|P_*(x) - p_{t,n}|^2]$ may depend on $t$ in an interesting way.
Given that $P_* = \int P_*^{\leq t}(\cdot|z)dQ^{\leq t}_*(z)$,
we can consider two extremes:
for $t=0$, the posterior is trivial $P_*^{\leq 0}(\cdot|\varnothing) = P_*$,
while for $t=\omega$, Remark \ref{remark. induce projection} suggests that $P_*^{\leq \omega}$ is deterministic.
So $t$ is a switch that controls the variability of the conditional distributions $P_*^{\leq t}(\cdot|z)$.
Intuitively, for larger $t$, the less variable $P_*^{\leq t}$ are easier to fit by the generator $G^{\leq t}$,
so the likelihoods $G^{\leq t}(x|z)$ and thus the estimators $p_{t,n}$ become more accurate.
Once again, we encounter a speed-accuracy tradeoff, as more observation steps simultaneously increase the compute and accuracy of inference.

\subsection{Adaptation to Text Data}
\label{sec. adapt to text}

To show that the Stage Three plan for improving slow thinking models (Section \ref{sec. long-term improvement}) is feasible, we apply the framework of active lifting to text data.
While unconditional inference can be implemented as in Section \ref{sec. time axis} for general modalities, conditional inference (e.g.\@ decoding given a prompt) is the main theme of the text modality and requires a bit more design.
The method introduced in this section is probably applicable to other modalities as well, such as video generation given a prompt and image inpainting.

While it should be doable to allow our modeled distribution $P_{\text{model}}$ to range in $\PS(\Sigma^{\omega})$ as in our static theory,
for active lifting the sampler $Q^{\leq \omega}$ does not \textit{a priori} obey causality in the sense of Remark \ref{remark. causal},
which would complicate our analysis.
Thus, for convenience, let both $P_{\text{model}}$ and the target distribution $P_*$ be supported over finite-length texts $\X_{\text{text}}=\Sigma^*$,
and moreover, suppose that their supports are prefix-free (e.g.\@ each sequence contains exactly one $\eos$ that is placed at the end).
Suppose we have performed the training processes of Sections \ref{sec. active lifting} and \ref{sec. generative} and obtained a well-trained sampler $Q$, latent model $P$ and generator $G$,
such that $P=Q^{\leq\omega}_*$ and $G(\cdot|z) = P_*(\cdot|z)$ for all $z\in\Omega^{\omega}$.
It follows that $P_* = P_{\text{model}} = \int G(\cdot|z)dP(z)$.
Furthermore, as discussed in Remark \ref{remark. induce projection}, it is reasonable to assume that $(Q^{\leq\omega},G)$ forms an autoencoder in the sense of (\ref{eq. active lifting autoencoder}), namely
\begin{equation*}
\int G(\cdot|z) dQ^{\leq\omega}(z|x) = \delta_x
\end{equation*}
for $P_*$-almost all $x$.
As an abuse of notation, denote $[x] = \{x' \in \Sigma^* \mid x\subseteq x'\}$.

Given any prompt $x\in\Sigma^*$ such that $P_*([x])>0$, conditional decoding means sampling from the conditional distribution $P_*(\cdot|[x])$.
To solve this, note that
\begin{align*}
P_*(\cdot|[x]) &= \int G(\cdot|z) dQ^{\leq\omega}_{<\omega}(z|x)
\end{align*}
with the sampler $Q^{\leq\omega}_{<\omega}$ defined by
\begin{align*}
Q^{\leq\omega}_{<\omega}(\cdot|x) &:= \int Q^{\leq\omega}(\cdot|x') dP_*(x'|[x]) = \frac{G([x]|\cdot)}{P_*([x])} P
\end{align*}
So it suffices to fit $Q^{\leq\omega}_{<\omega}$.
Since $\sprt P_*$ is assumed prefix-free, the value $Q^{\leq\omega}(\cdot|x)$ is unused if $x$ is a nonempty proper prefix of any $x' \in \sprt P_*$,
while for the boundary case $x=x'$, we always have $Q^{\leq\omega}(\cdot|x) = Q^{\leq\omega}_{<\omega}(\cdot|x)$.
Thus, it is natural to use $Q^{\leq\omega}(\cdot|x)$ to approximate $Q^{\leq\omega}_{<\omega}(\cdot|x)$.
The training loss can simply be KL-divergence
\begin{align}
\nonumber
\int &\E_{l\sim U[1,|x|-1]} \Big[ \KL\big( \detach(Q^{\leq\omega}_{<\omega})(\cdot|x_{\leq l}) \bigm\| Q^{\leq\omega}(\cdot|x_{\leq t}) \big) \Big] dP_*(x) \\
\label{eq. train prefix latent sampler}
= \iint &\E_{l\sim U[1,|x|-1]} \big[ -\log Q^{\leq\omega}(z|x_{\leq l}) \big] ~ d\detach(Q^{\leq\omega})(z|x) dP_*(x) + \text{constant}
\end{align}
where $U[1,|x|-1]$ denotes the uniform distribution over $\{1,\dots |x|-1\}$, so $x_{\leq l}$ is always a nonempty proper prefix of $x$.
For implementation, one can either train a separate copy of the underlying model of $Q$ in order not to alter the sampler $Q^{\leq\omega}$,
or use the same model and include this loss in the training process of active lifting.

Then, suppose that $Q$ has been trained on (\ref{eq. train prefix latent sampler}) such that $Q^{\leq\omega} \approx Q^{\leq\omega}_{<\omega}$.
Conditional decoding can be performed based on the algorithms in Section \ref{sec. decoding}.
Similar to the straightforward method (\ref{eq. decode once}), one can sample $P_*(\cdot|[x])$ approximately by
\begin{equation}
\label{eq. active conditional decode}
X \sim G(\cdot|Z), \quad Z \sim Q^{\leq\omega}(\cdot|x)
\end{equation}
Also, similar to the resampling method (\ref{eq. resampling posterior}), one can use
\begin{align}
\label{eq. resampling active conditional decode}
\begin{split}
X &\sim G(\cdot|Z^{(I)}), \quad I \sim \text{Categorical}(p_1, \dots p_n), \quad \{ Z^{(i)} \}_{i=1}^n \iidsample Q^{\leq\omega}(\cdot|x)\\
&\text{where} \quad p_i = \frac{G([x]|Z^{(i)})P(Z^{(i)}) \big/ Q^{\leq\omega}(Z^{(i)}|x)}{\sum_{j=1}^n G([x]|Z^{(j)})P(Z^{(j)}) \big/ Q^{\leq\omega}(Z^{(j)}|x)}
\end{split}
\end{align}
The resampling weights are defined based on the property that $Q^{\leq\omega}_{<\omega}(\cdot|x) \propto G([x]|\cdot)P$.
Note that if the decoder $G$ is implemented as an autoregressive model, the term $\log G([x]|z)$ can be easily obtained by summing the token-wise log-likelihoods;
moreover, with $x$ appended to the input sequence of $G$, its decoding would always be a continuation of $x$, namely $X$ always starts with $x$.
Analogous to Proposition \ref{prop. TV decoding error}, one can show that the decoding algorithm (\ref{eq. resampling active conditional decode}) enjoys a small error that goes to $0$ as $n\to\infty$,
\begin{equation*}
\E\big[\TV(P_*(\cdot|[x]), \law(X))\big] \leq \E\big[\TV(Q^{\leq\omega}_{<\omega}(\cdot|x), \law(Z^{(I)}))\big] \leq \sqrt{\frac{6\chi^2\big(Q^{\leq\omega}_{<\omega}(\cdot|x)\big\|Q^{\leq\omega}(\cdot|x)\big) + 2}{n}}
\end{equation*}

Conditional encoding can be similarly performed.
Given any nonempty texts $x_q, x_r \in \Sigma^*$ (e.g.\@ a query and a response) such that $x_q x_r \in \sprt P_*$, we can evaluate the conditional probability
\begin{equation*}
P_*\big(x_q x_r \bigm| [x_q] \big) = \frac{P_*(x_qx_r)}{P_*([x_q])} = \frac{\int G(x_q x_r |z) dP(z)}{\int G([x_q]|z) dP(z)}
\end{equation*}
by the estimator
\begin{align}
\label{eq. active encode conditional}
p_n &= \frac{\frac{1}{n} \sum_{i=1}^n G(x_q x_r |Z^{(i)}) P(Z^{(i)}) / Q^{\leq\omega}(Z^{(i)})}{\frac{1}{n} \sum_{i=1}^n G([x_q] |\tilde{Z}^{(i)}) P(\tilde{Z}^{(i)}) / Q^{\leq\omega}(\tilde{Z}^{(i)})} \\
\nonumber
& \{Z^{(i)}\}_{i=1}^n \iidsample Q^{\leq\omega}(\cdot|x_q x_r), \quad \{\tilde{Z}^{(i)}\}_{i=1}^n \iidsample Q^{\leq\omega}(\cdot|x_q)
\end{align}
One can check that $Q^{\leq\omega}_{<\omega}$ is the variance-minimizing sampler (or posterior sampler) for the denominator.

For the practical implementation of training and inference, one would need to apply a finite-length cutoff to all sampled latent sequences $Z \sim Q^{\leq\omega}(\cdot|x)$.
In theory, the minimum-length coding analogy in Section \ref{sec. minimum-length coding} suggests that the entries $Z_t$ ceases to provide more information when $t$ exceeds the minimum code length of $x$, so it suffices to keep a finite prefix of $Z$.
In practice, one can either choose some large and uniform cutoff length $T$ or implement an adaptive cutoff in the spirit of Section \ref{sec. balance fast slow}.
With all sequences being finite, one may implement all three components $Q,P,G$ by one underlying LLM:
For instance, when playing $Q$, the LLM takes in the prompt $x \sot$ and decodes some $z$;
when playing $P$, the LLM simply encodes $z$;
when playing $G$, the LLM takes in $z \eot x$ and decodes some $X'$, which is a continuation of $x$.
Namely, $X=xX'$ is a sample of $P_*(\cdot|[x])$.

In summary, an active lifting model is specified by the tuple $(\X,\Omega,Q,P,G)$,
with data space $\X$, latent vocabulary $\Omega$, sampler $Q\in\Q_{+1}$, latent model $P\in\PS(\Omega^{\omega})$, and optionally, generator $G\colon\Omega^{\leq\omega}\to\PS(\X)$.
Regarding inference, unconditional encoding can be performed by either (\ref{eq. active encode fixed t}) or (\ref{eq. active encode incremental t}),
conditional encoding by (\ref{eq. active encode conditional}),
unconditional generation by (\ref{eq. generative model sampling}),
and conditional generation by (\ref{eq. active conditional decode}) or (\ref{eq. resampling active conditional decode}).
Regarding training,
one can either train $Q,P$ by (\ref{eq. QP min-max objective}) to obtain an encoder and then train $G$ separately by (\ref{eq. train generator passive}) to obtain a generative model,
or train all three together by (\ref{eq. train generator active}), which resembles an autoencoder.
Optionally, the sampler loss (\ref{eq. train prefix latent sampler}) can be included to enable conditional inference.

\begin{remark}
Apparently the modelings introduced in this paper rely heavily on sequences;
the latent space $\Z=\Omega^{\omega}$ always consists of sequences, and for the majority of our sections, the data space $\X=\Sigma^{\omega}$ is also sequences.
From a measure theoretic point of view, this is a very general setup, since any uncountable Polish space is Borel isomorphic to $\{0,1\}^{\omega}$, so whether the observable or latent data are sequences is mostly a matter of perspective.
Yet, perspective does matter here, as we need a ``perspective" that allows the existence of a lifted distribution that can be realized by neural networks (cf.\@ Problem \ref{problem. active}), preferably with small complexity (cf.\@ Section \ref{sec. learnability}).
Though sufficiently abstract for a first-principles theory, the choice $\Z=\Omega^{\omega}$ is more like an ``axiom" instead of a derived result,
and reflects the dominance of sequence models (specifically Transformers) in machine learning today.
Meanwhile, regarding $\X$, in particular those of non-sequence modalities, our modeling can be described informally as equipping any data with a sequence structure, so that we can perceive/generate they like reading/writing texts.
Still, for now, only Sections \ref{sec. unified objective} and Section \ref{sec. perceptual} (except Section \ref{sec. adapt to text}) are directly applicable to non-sequence data.
\end{remark}

\subsection{Unification of Existing Representations}
\label{sec. visual field}

This section is an informal study of the feature representation of the sampler $Q$, with a focus on image data.
One purpose is to make the general discussion in Section \ref{sec. minimum-length coding} on concept induction and language invention more concrete.
A visualization method called ``visual field trajectory" is introduced.
(However, we emphasize that this diagnostic tool is not part of our model's architecture or training, nor is it based on biological modeling.)
We discuss the relation of our model to the existing computer vision models, and the possibility of unifying their representations.


Let $(Q,P)$ be a global optimizer of the training objective (\ref{eq. QP min-max objective}) with the latent distribution $P=Q^{\leq\omega}_*$ belonging to some parametrized family such as $\PS_{\text{plain}}$.
Based on Remark \ref{remark. induce projection}, we may assume that $P^{\leq\omega}_*$ is deterministic (concentrated on some measureable function $\proj$).
Then, it is reasonable to assume that for $P$-almost all $z \in \Omega^{\omega}$, the posterior distribution $P^{\leq t}_*(\cdot|z_{\leq t})$ converges weakly to $P^{\leq \omega}_*(\cdot|z) = \delta_{\proj(z)}$.
Consider $\X = \X_{\text{image}}$.
Given a random image $X \sim P_*$, let $Z\sim Q^{\leq\omega}(\cdot|X)$ be a latent sample.
Define the pointwise variances over the image domain $u\in [0,1]^2$,
\begin{equation*}
\forall t \in \N, \quad \sigma^2_t(u) = var(X'(u)), \quad X' \sim P^{\leq t}_*(\cdot|Z)
\end{equation*}
Then, it holds almost surely that $\lim_{t\to\infty} \sigma^2_t(u) = 0$ almost everywhere in $[0,1]^2$.
Define the ``visual field" of each token $Z_t$ as the distribution of the reduction in standard deviation
\begin{equation*}
\forall t \in \N_+, \quad F_t(u) = \big(\sigma_{t-1}(u)-\sigma_t(u)\big)_+
\end{equation*}
and define the ``visual field trajectory" of $Z$ as the sequece $(F_t)_{t=1}^{\infty}$.
Conceptually, the introduction of $Z_t$ makes the support of $F_t$ less uncertain, so this is probably the region that the model extracts information from (or whose state the model can infer) at the $t$-th observation step.
Since $P^{\leq 0}_* \equiv  P_*$, the initial variance $\sigma^2_0(u)$ does not depend on $Z$ and is likely a large value everywhere in the domain $[0,1]^2$.
Given that
\begin{equation*}
\sum_{t=1}^{\infty} F_t(u) \geq \sigma_0(u) - \lim_{t\to\infty} \sigma_t(u) = \sigma_0(u)
\end{equation*}
the visual field trajectory can be seen as a ``covering" of the image domain, such that each place $u \in [0,1]^2$ will eventually be attended to by sufficiently many $Z_t$'s to become fully determined.
Besides the spatial domain, one can also consider the frequency domain and define the visual fields with the Fourier basis
\begin{align*}
\forall \xi \in \mathbb{Z}^2, \quad \sigma^{\text{freq}}_t(\xi) &= var\Big( \int_{[0,1]^2} X(u) e^{-2\pi i u^T\xi} du \Big)^{1/2} \\
F_t^{\text{freq}}(\xi) &= \big(\sigma^{\text{freq}}_{t-1}(\xi) - \sigma^{\text{freq}}_t(\xi)\big)_+
\end{align*}


Based on the analogy of minimum-length coding in Section \ref{sec. minimum-length coding}, the tokens $Z_t$ may correspond to the most distinctive features that can rapidly pin down the image $X$, distinguishing it from the rest of $P_*$.
So their visual fields $F_t$ tend to overlap with the visual patterns that are recurring throughout $P_*$.
For instance, if $P_*$ consists of images of daily-life scenes, then each $F_t$ could cover an object or one of its constituent parts,
and if $P_*$ consists of images of book pages, then each $F_t$ could be a printed word, phrase or sentence.
Moreover, based on the notion of regularity discussed in Section \ref{sec. minimum-length coding}, the visual field trajectory $(F_t)_{t=1}^{\infty}$ tends to be an understandable ordering of these regions,
so that the marginal distribution of these trajectories has a simple dependency structure.
For instance, the model would read the printed words in the the raster order (reading from left to right, from top to bottom) instead of some arbitrary order,
and would look first at the entirety of a person, then the face and clothes, then the detailed facial features and so on, instead of jumping abruptly from this person's ears to the other person's shoes.
This ordering of concepts gives rise to visual schema.

%
%

Thus, the training objective (\ref{eq. QP min-max objective}) compels the visual representation of our model to adapt to the characteristics of the image distribution.
Below are three more detailed examples that relate this representation to those of several existing models in computer vision.
%
\begin{example}[Text reading and patch autoregression]
\label{ex. read text}
Following the above discussion, if $P_*$ involves book pages and papers, then the sampler $Q$ would parse $X$ in the raster order, so that the latent model $P$ can read these texts in the linear order.
If we compare the visual fields $F_t$ to the square patches commonly used in computer vision, then their trajectory $(F_t)_t$ correspondes to the popular representation of patch autoregression \cite{van2016conditional,van2017VQVAE,sun2024autoregressive}, so the latter can be seen as a special case of our representation.
Moreover, $Q$ possesses the degree of freedom to read like humans, such as skimming through sentences with low information content (i.e.\@ $F_t$ becomes a large region) and reexamining an earlier paragraph for new insights.
\end{example}
\begin{example}[Diffusion]
$P_*$ could be spatially homogeneous images that exhibit more salient characteristics in the frequency domain, such as the images of textures, forests, and starry skies.
It could be simpler to model them autoregressively in the Fourier space, with each visual field $\sigma^{\text{freq}}_t$ being concentrated at some frequency $\xi_t \in \mathbb{Z}^2$.
The visual field trajectory imposes an ordering on the frequencies $\{\xi_t\}_{t=1}^{\infty}$, such as by increasing magnitude $\|\xi_t\|$.
Since the well-known diffusion models \cite{sohl2015deep,song2021maximum,rombach2022high} can be interpreted as autoregressive models in the frequency domain (in either the default low-to-high order or some manually designed order) \cite{dieleman2024spectral,falck2025fourier},
one may conjecture that these models are also a special case of our modeling.
\end{example}
\begin{example}[Objects and parts]
\label{ex. objects and parts}
Images of daily-life scenes have multiscale compositional contents,
which means that each image may consist of multiple objects, each object may consist of multiple parts, and so on, and that they may possess multiple attributes and relations.
In pursuit of this human-like representation, part-based modeling \cite{felzenszwalb2009object,felzenszwalb2005pictorial,fischler1973representation} was a popular research topic in computer vision before deep learning became dominant.
Even now, people often assess the quality of deep vision models by how closely their learned features resemble the real objects and parts \cite{caron2021emerging,zeiler2014visualizing,han2022vision}.
Our preceding discussions indicate that with a well-trained latent sampler $Q$, the latent tokens $Z_t$ may acquire the role of visual concepts,
and the subsequences of $Z \sim Q^{\leq \omega}$ may serve as visual schema.
These visual schema enjoy both the efficiency of part-based representations and the flexibility of deep vision models.
It is conceptually possible for $Q$ to go beyond manually labeled concepts and schema,
discovering all possible attributes (``having a rough surface attractive to cats"),
relations (``caught goofing off by" could be a relation between an employee and an employer),
compositions (``an umbrella hat"), and so on.
\end{example}


These examples suggest that perhaps a vision model based on latent sequences can help to unify the apparently distinct vision models that people currently know of and establish a larger design space.
Also, in accordance with \cite{wei2025OCR,cheng2025glyph}, Example \ref{ex. read text} advocates for encoding texts with vision models, removing the need of tokenizers and allowing for more flexible reading orders.
Meanwhile, Example \ref{ex. objects and parts} suggests that, since the latent sequences $Z \sim Q^{\leq \omega}(\cdot|x)$ resemble the visual concepts of humans,
they might align well with text tokens, such that it is easier to integrate $Q^{\leq \omega}$, as an image encoder, into vision-language models \cite{alayrac2022flamingo,bai2023qwenvl,lin2025adaptvision} compared to the existing image encoders.


\begin{remark}[Active reading and world model]
\label{remark. world model}
So far we have discussed how the objective of minimum-length regular coding induces the visual field trajectories to visit the contents of an image in an efficient yet understandable order.
What if there are two images?
For instance, $x$ can be a photograph of two paintings, placed side by side.
Minimum-length coding urges the visual field trajectory to visit the most distinctive parts of both paintings, probably switching frequently between them.
Meanwhile, regularity prevents the trajectory from switching too rapidly, in order to obtain reasonable visual schema.
This tradeoff compels the sampler $Q$ to actively search for the best ``itinerary" that moves at a moderate and adaptive pace.
We can go further to consider an infinitely large ``image" that contains infinitely many scenes (such as the internet or the physical world).
An optimal itinerary for this ``image" might be:
First, visit the websites for learning languages;
then, read all the interesting books in the world;
finally, design and launch a spaceship to see the universe, etc.
Hence, the combination of the unified objective (\ref{eq. unified objective}) and latent sampling might become a technical solution to active reading and world models.
\end{remark}







\section{Preliminary Experiment}
\label{sec. experiment}

The implementation of the numerous by-products of this paper are left for future work.
Nevertheless, we perform a preliminary experiment to test the Stage One improvement of slow thinking from Section \ref{sec. quick improvement}.
The theory of active lifting holds as long as its derivations are valid, but experiments are helpful for demonstrating that the theory is implementable with the existing technology.

Recall that the Stage One improvement substitutes the identity sampler with explanatory sampler(s) to improve the sampling efficiency of training.
One consequence is that the model could make greater progress when trained with the same data and sampling size $n$, since the sampling errors in the training gradients are reduced.
We verify this prediction, with progress measured by the reduction in test loss.

We consider the simple setting when only the inference sampler $Q$ is instantiated, which fits the posterior sampler $Q_*$.
By Section \ref{sec. quick improvement}, a model with an explanatory sampler would train its latent distribution on (\ref{eq. explanatory latent loss inference sampler}) and its sampler on (\ref{eq. deepseek explanatory sampler loss}).
Putting them together, the latent model $P_{\theta}$ and sampler $Q_{\psi}$ are trained simultaneously on the losses
\begin{align}
\label{eq. experiment latent loss}
L_{\text{latent}}(\theta) &= - \frac{1}{B} \sum_{b=1}^B \log\Big(\frac{1}{n} \sum_{i=1}^n \frac{P_{\theta}^{\leq|X^{(b)} Y_i^{(b)}X_r^{(b)}|}(Y_i^{(b)}X_r^{(b)}|X^{(b)})}{\detach \big( Q_{\psi}(Y_i^{(b)}|X^{(b)}{\color{purple}X_r^{(b)}}) \big)}\Big) \\
\nonumber
L_{\text{sampler}}(\psi) &= -\frac{1}{Bn}\sum_{b=1}^B\sum_{i=1}^n \detach\Big(\log\frac{P_{\theta}^{\leq|X^{(b)} Y_i^{(b)}X_r^{(b)}|}(Y_i^{(b)}X_r^{(b)}|X^{(b)})}{Q_{\psi}(Y_i^{(b)}|X^{(b)}{\color{purple}X_r^{(b)}})}\Big) \log Q_{\psi}(Y_i^{(b)}|X^{(b)}{\color{purple}X_r^{(b)}}) \\
\label{eq. experiment sampling}
\{Y_i^{(b)}\}_{i=1}^n &\iidsample Q_{\psi}(\cdot|X^{(b)}{\color{purple}X_r^{(b)}})
\end{align}
where $\{(X^{(b)},X_r^{(b)})\}_{b=1}^B$ denotes the mini-batch of a training step.
Formally speaking, if all the colored terms ${\color{purple}X_r^{(b)}}$ are removed, then we get the control group that uses the predictive sampler.
Note that the test loss is computed by (\ref{eq. experiment latent loss}) if the mini-batch is replaced by the test dataset.
Note that since $L_{\text{sampler}}$ is a policy gradient loss, one may consider more advanced implementations such as GRPO \cite{shao2024deepseekmath} and DAPO \cite{yu2025DAPO}.
Still, we stick with plain policy gradient, partly for simplicity, and partly because its success would imply that explanatory samplers do not require sophisticated losses to work.
Meanwhile, the classical supervised finetuning (SFT) uses the following loss
\begin{equation*}
L(\theta) = - \frac{1}{B} \sum_{b=1}^B \log P_{\theta}^{\leq|X^{(b)} X_r^{(b)}|}(X_r^{(b)}|X^{(b)})
\end{equation*}
Thus, we run three training methods:
SFT that represents fast thinking,
RL finetuning of slow thinking with a predictive sampler,
and RL finetuning with an explanatory sampler.
Denote their test losses at the end of training by $L_{\text{fast}}$, $L_{\text{pred}}$ and $L_{\text{expl}}$.
We show that $L_{\text{fast}} - L_{\text{expl}} > L_{\text{fast}} - L_{\text{pred}}$, so that slowing thinking with explanatory sampler achieves greater gain.
It is fair for the control group to use the predictive sampler instead of the identity sampler, since the former is more general and makes the control group stronger (cf.\@ Section \ref{sec. prediction vs explanation}).

The Qwen2.5-7B base model \cite{qwen2025qwen25} is adopted for all components, including the latent distribution $P_{\theta}$ and sampler $Q_{\psi}$.
Since a clean experiment is needed, such that the components should not know how to produce or encode the latent thoughts prior to finetuning, we prefer the base model, produced directly from pretraining, over its post-trained version \cite{qwen2025qwen25}.
For convenience, $P_{\theta}$ and $Q_{\psi}$ use separate copies of this LLM, instead of sharing one copy.
One detail to be aware of is that usually the output logits of Transformer LLMs are modified by default during decoding \cite{huggingface2026generationconfig}, e.g.\@ with lower temperature and top-$k$ cutoff.
We explicitly turn off all modifications
in the sampling step (\ref{eq. experiment sampling}), so that the thoughts $Y_i^{(b)}$ are sampled exactly from $Q_{\psi}(\cdot|X^{(b)}{X_r^{(b)}})$ and thus the above losses are valid.

The training data are sampled from the OpenWebMath dataset \cite{paster2023openwebmath}, which consists of math-themed web texts and is commonly used for pretraining.
We use pretraining data instead of SFT data, because as advocated by Section \ref{sec. short-term improvement}, it is reasonable to expect that slow thinking will eventually be applied to pretraining, instead of being limited to post-training.
For convenience, we set $|X^{(b)}| = |X_r^{(b)}|=128$.
Specifically, we keep only the first 256 tokens of each sample, treating the first half as the query $X^{(b)}$ and the second half as the response $X_r^{(b)}$.
We also fix the thought length to $|Y_i^{(b)}|=128$.
Since $L_{\text{latent}}$ is an estimator of the cross entropy loss (\ref{eq. cross entropy}), we normalize it by the response length $|X_r^{(b)}|=128$ during training, as is common in practice.
Similarly, we normalize the two log terms in $L_{\text{sampler}}$ by 128.
The sampling size is set to $n=8$ as in \cite{guo2025deepseek}.

\begin{figure}[!ht]
\centering
\subfloat{\includegraphics[width=0.48\textwidth]{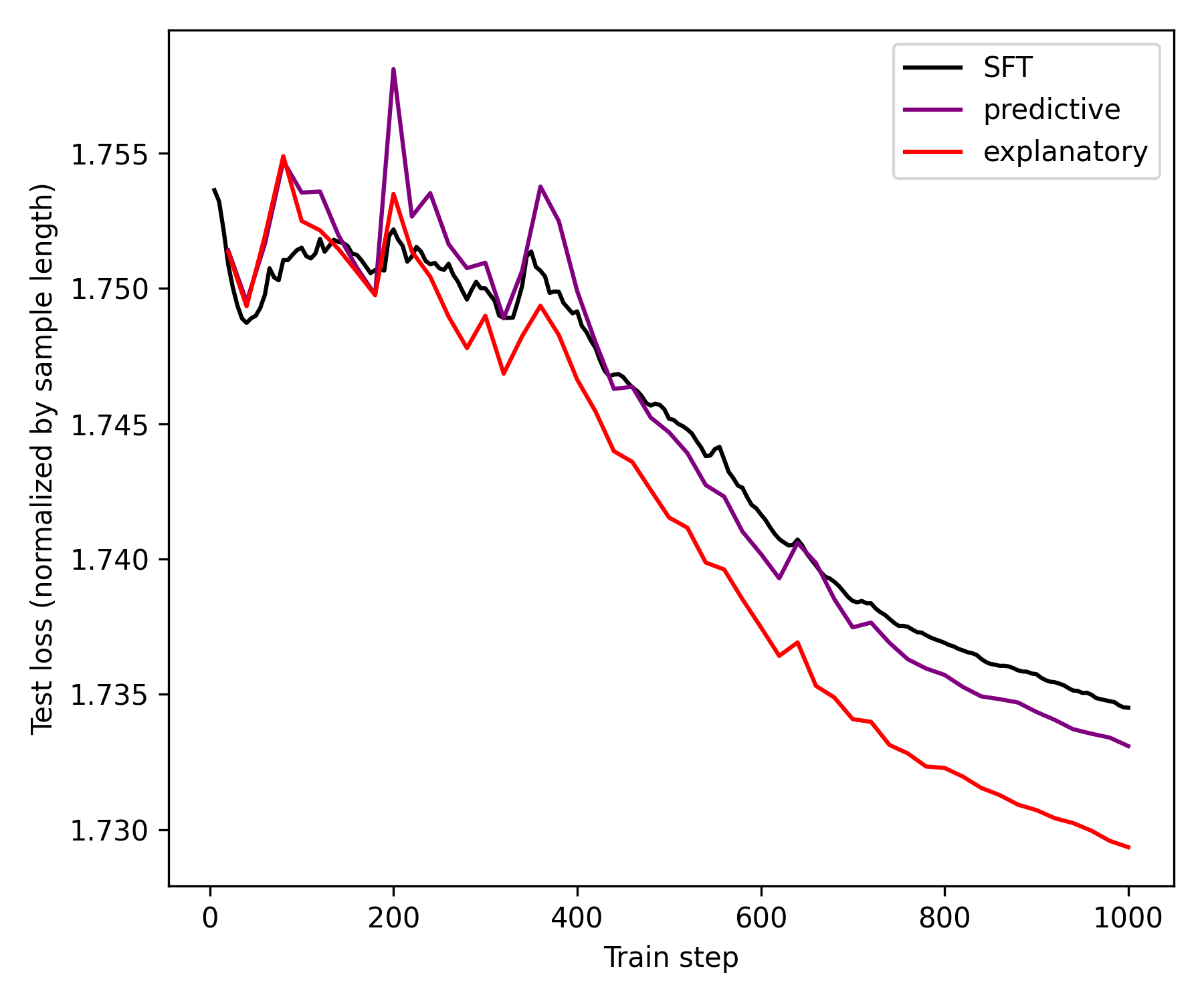}}
\quad
\subfloat{\includegraphics[width=0.48\textwidth]{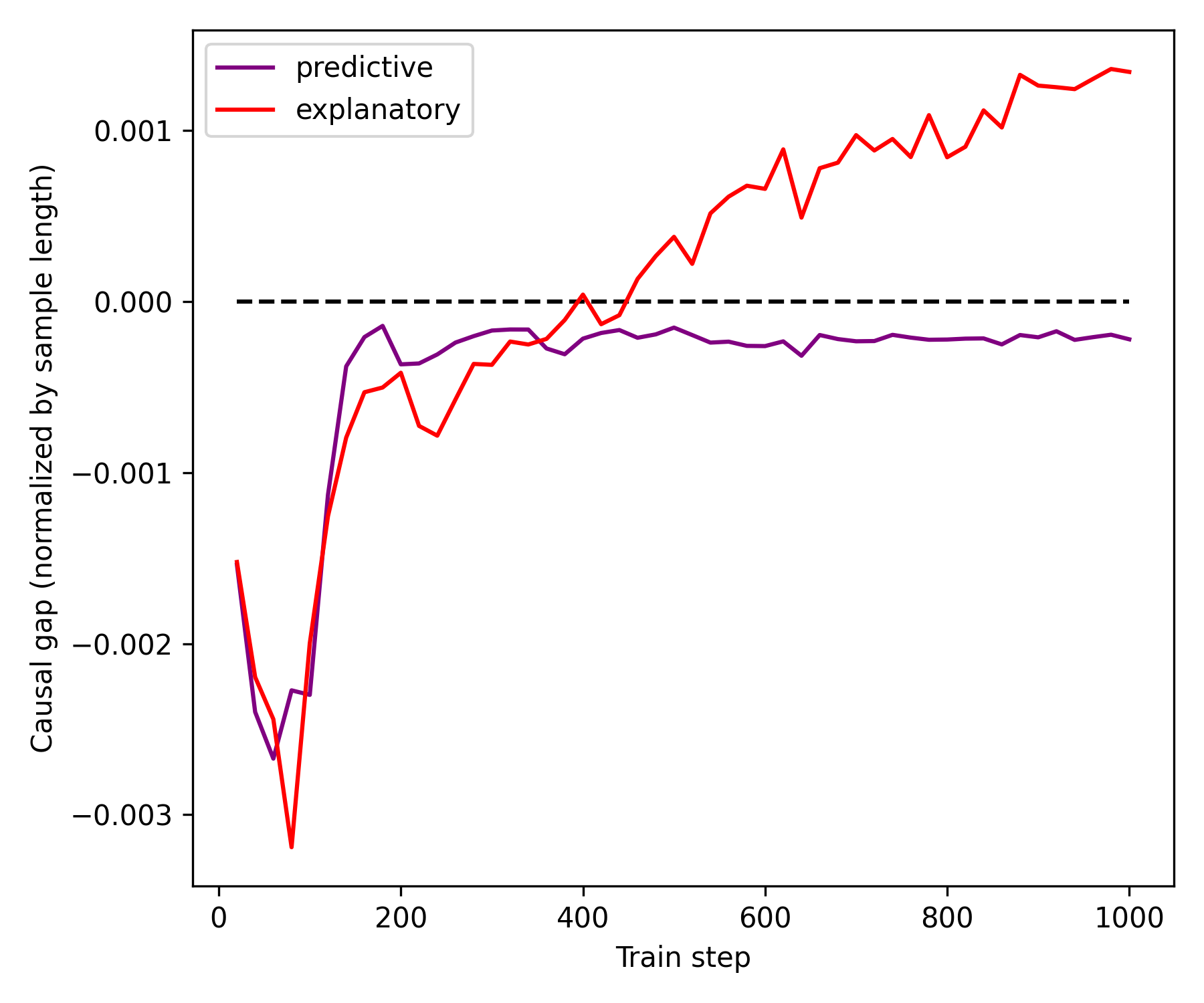}}
\caption{Comparison of the effectiveness of the predictive and explanatory samplers for the training of slow thinking models.
Left: The test losses during training.
Right: The average causal gap (\ref{eq. average causal gap}).
Both are normalized by the response length $|X_r^{(b)}|=128$.}
\label{fig: prelim experiment}
\end{figure}

The test loss curves are plotted in Figure \ref{fig: prelim experiment} (Left).
The implementation details are listed in Appendix \ref{appendix. experiment}.
The relative improvement of using explanatory sampler can be estimated by
\begin{equation*}
\frac{L_{\text{fast}} - L_{\text{expl}}}{L_{\text{fast}} - L_{\text{pred}}} - 1 = \frac{1.7345-1.7294}{1.7345-1.7331} - 1 \approx 264\%
\end{equation*}
Given the limited reasoning content in the pretraining data, one may expect a higher relative improvement when finetuning is performed on difficult texts.
We reemphasize that the use of explanatory samplers for training and prefill is justifiable by Remark \ref{remark. justify explanatory}.

For a closer inspection, note that (the value of) the loss (\ref{eq. experiment latent loss}) can be rewritten as follows
\begin{align*}
L_{\text{latent}}(\theta) &= \log n - \frac{1}{B} \sum_{b=1}^B \text{logsumexp}\Big( \big( \log P_{\theta}^{\leq|X^{(b)} Y_i^{(b)}X_r^{(b)}|}(X_r^{(b)}|X^{(b)}Y_i^{(b)}) - r_i^{(b)} \big)_{i=1}^n \Big) \\
r_i^{(b)} &= \begin{cases}
\log Q_{\psi}(Y_i^{(b)}|X^{(b)}X_r^{(b)}) - \log P_{\theta}^{\leq|X^{(b)} Y_i^{(b)}|}(Y_i^{(b)}|X^{(b)}) \quad \text{with explanatory sampler} \\
\log Q_{\psi}(Y_i^{(b)}|X^{(b)}) - \log P_{\theta}^{\leq|X^{(b)} Y_i^{(b)}|}(Y_i^{(b)}|X^{(b)}) \quad \text{with predictive sampler}
\end{cases}
\end{align*}
For explanatory samplers, this $r_i^{(b)}$ is the gap between the \textit{a priori} likelihood $P_{\theta}^{\leq|X^{(b)} Y_i^{(b)}|}(Y_i^{(b)}|X^{(b)})$ and the \textit{a posteriori} likelihood $Q_{\psi}(Y_i^{(b)}|X^{(b)}X_r^{(b)})$ given the knowledge of $X_r^{(b)}$.
A positive $r_i^{(b)}$ suggests that the thought $Y_i^{(b)}$ contains some information of $X_r^{(b)}$, so we refer to $r_i^{(b)}$ as the causal gap.
Consider the weighted average of casual gaps as a training statistic
\begin{equation}
\label{eq. average causal gap}
\frac{1}{B} \sum_{b=1}^B r_i^{(b)} p_i^{(b)}, \quad p_i^{(b)} = \frac{\tilde{p}_i^{(b)}}{\sum_{j=1}^n \tilde{p}_j^{(b)}}, \quad \tilde{p}_i^{(b)} = \frac{P_{\theta}^{\leq|X^{(b)} Y_i^{(b)}X_r^{(b)}|}(Y_i^{(b)}X_r^{(b)}|X^{(b)})}{Q_{\psi}(Y_i^{(b)}|X^{(b)}{\color{purple}X_r^{(b)}})}
\end{equation}
where the colored terms ${\color{purple}X_r^{(b)}}$ are removed for the control group.
This statistic is plotted in Figure \ref{fig: prelim experiment} (Right).
As expected, the averaged causal gap with the explanatory sampler eventually becomes positive.



\section{Related Work}
\label{sec. related work}

Many related works have already been mentioned in this paper, and this section lists the remaining ones.

\vs
\noindent
\textbf{Slow thinking Models:}
General discussions of the current status of slow thinking models are provided by \cite{zhang2026system2,pan2025survey}.
Notable phenomenological studies of the mechanisms of slow thinking include \cite{hu2025complexnetwork,marjanovic2025thoughtology,wang2025multistep}.
One major problem of slow thinking is policy collapse \cite{yu2025DAPO,cui2025entropy}, and current solutions focus on token-wise operations that encourage exploration \cite{huang2025spark,wang2025beyond,cheng2025exploration}.
We identify a possible cause (Remark \ref{remark. policy collapse}) and take the different approach of using the inquisitive sampler to reduce the sampling variance of the training gradient.
Another problem of slow thinking is to avoid overthinking and reduce inference cost \cite{sui2025stopover} and the engineering solutions include \cite{zhang2025light,liao2025reward,ding2025dynamic}.
Section \ref{sec. balance fast slow} takes the different approach of deriving an intrinsic loss function from the unified objective.
Meanwhile, similar to our latent sampling, several works have studied slow thinking as a sampling problem \cite{yue2025does,karan2026reasoning}, notably with \cite{zhou2025variational} that also discovered the posterior sampling distribution.
Designs that resemble the explanatory sampler have been proposed in earlier works \cite{ling2017program,lin2025RAVR}, in the specific setting with question-answer data, though they are introduced as engineering heuristics instead of from a unified derivation.
Some models perform slow thinking in the feature space, a method known as continuous chain-of-thought \cite{chen2025continuous,hao2025coconut}, and it would be interesting to see if our framework with discrete latent spaces can be extended to the continuous setting.


\vs
\noindent
\textbf{First-principles modeling:}
Theories that try to model intelligence at the fundamental level were proposed even in the early days of AI \cite{wiener2019cybernetics,von2017general,von2012computer,turing2007computing}.
More recent works include the pattern theory that models real-world data \cite{grenander1996elements,mumford1994pattern},
the free-energy principle that models how the brain minimizes surprise \cite{friston2010free,friston2023free},
a derivation of animal behaviors from a progressive refinement of feedback control loops \cite{cisek2019resynthesizing,cisek2022evolution},
and a derivation of the Transformer architecture from compression \cite{ma2024white,ma2025textbook}.

\vs
\noindent
\textbf{Latent variable models:}
Broadly speaking, our active lifting theory concerns the unsupervised discovery of meaningful latent variables that can explain a data distribution, which is a general topic in machine learning.
For Euclidean data, the classical methods include clustering and principle components \cite{omran2007overview,alpaydin2020introduction,tabak2018explanation}.
For sequence data, besides HMMs, we have conditional random fields \cite{lafferty2001CRF} and probabilistic context-free grammar \cite{collins2003head}.
Prior to deep learning, latent structure models were popular in natural language processing, whose structures include parse trees \cite{blei2003latent}, document topics \cite{blei2003latent}, bilingual word correspondences \cite{brown1993translation}, etc.
Similarly, information-theoretic training objectives were devised to guide computer vision models to develop disentangled, semantically meaningful latent variables \cite{chen2016infogan,higgins2017beta}.

\vs
\noindent
\textbf{Bounded-depth circuits:}
Our representation hierarchy largely depends on the characterization of Transformers as bounded-depth circuits.
This boundedness has been studied by multiple works \cite{merrill2022saturated,strobl2023averagehard,chen2025circuit}.
Besides slow thinking, other ways for LLMs to tackle this constraint include using recurrent layers \cite{geiping2025scaling,hao2025coconut,zeng2025pretraining}, adopting the classical recurrent neural networks \cite{beck2024xlstm}, and test-time updates \cite{sun2025TTT,peng2025rwkv7goose}.



\section{Conclusion}
\label{sec. conclusion}

This paper presents a mathematical theory called ``active lifting", as a first-principles modeling of the cognitive function of slow thinking, or more generally, active perception.
The general theory starts from the sampling of latent sequences and the maximization of uncertainty reduction rate, and transforms data distributions to lifted distributions on a sequence space $\Omega^{\omega}$ to make them easier to model (Problems \ref{problem. intro}, \ref{problem. static} and \ref{problem. active}).
Thus, inference (or more generally, encoding) is endowed with an internal time axis, along which latent samplers produce sequences to ``describe" the observations.
Meanwhile, training is to construct a lifted distribution that is both regular (learnable by neural networks, or more generally, bounded-depth Boolean circuits) and efficient (analogous to the compression method of minimum-length coding), thus resembling the invention of descriptive languages.
While active lifting implicitly develops a deterministic projection function, one can consider the sub-theory for text data with a prescribed projection $\proj\colon \Omega^{\omega}\rightharpoonup\Sigma^{\omega}$.
This static theory leads to the representation hierarchy, a cascade of subspaces of the space of data distributions $\PS(\Sigma^{\omega})$, and the sampler hierarchy, a cascade of subspaces of the sampler space $\Q_{\proj}$.
The existing slow thinking models can be derived from the lower levels of the two hierarchies, while ascending these hierarchies makes a model more expressive (capable of modeling complex data distributions) and efficient (improving the performance of training and inference with fixed data and compute).

For future work, one direction could be the first-principles modeling of the data distributions.
It would be interesting to try to derive the familiar characteristics of natural language texts and natural images, without using any empirical knowledge of them.
Such modeling could help to make the arguments of Section \ref{sec. perceptual} more rigorous, e.g.\@ defining the visual concepts and visual schema as well as their emergence during the training process of active lifting.
In practice, one might try to recreate the process by which our human ancestors invented natural languages.
Meanwhile, it could be useful to study the dynamics of the latent sampler along the perceptual time axis.
While an indirect characterization is provided by our visual field trajectory, an in-depth analysis would be beneficial.
One possible outcome could be a more formal formulation of the agency of perception, which might extend to the agency of behavior.
The latter can help to characterize the design space of information-seeking behaviors (Figure \ref{fig: cognitive modeling}), and thus implement the world model discussed in Remark \ref{remark. world model}.

\section*{Acknowledgement}

This work is supported by the NSFC Major Research Plan - Interpretable and General Purpose Next-generation Artificial Intelligence of China (No.\@ 92270001).
We thank Linpeng Tang, Liangkai Hang, Qihang Wang, Boyu Chen, Zhiyu Li, Jihao Long, Jiequn Han and Yang Li for helpful discussions.
We thank the editor and reviewers for their constructive feedback.

\bibliography{main}

@phdthesis{yang2023thesis,
    author    = {Yang, Hongkang},
    title     = {{Mathematical Theory of Machine Learning Models for Estimating Probability Distributions}},
    school    = {Princeton University},
    year      = {2023},
    address   = {Princeton, New Jersey, USA},
    month     = {May},
    url       = {http://arks.princeton.edu/ark:/88435/dsp01ks65hg462}
}

@article{e2022barron,
  title={The {Barron} space and the flow-induced function spaces for neural network models},
  author={E, Weinan and Ma, Chao and Wu, Lei},
  journal={Constructive Approximation},
  volume={55},
  number={1},
  pages={369--406},
  year={2022},
  publisher={Springer}
}

@misc{e2019residual,
  title={A Priori Estimates of the Population Risk for Residual Networks}, 
  author={Weinan E and Chao Ma and Qingcan Wang},
  year={2019},
  eprint={1903.02154},
  archivePrefix={arXiv},
  primaryClass={cs.LG},
  url={https://arxiv.org/abs/1903.02154}, 
}

@article{e2020multilayer,
  title={On the {Banach} Spaces Associated with Multi-Layer {ReLU} Networks: Function Representation, Approximation Theory and Gradient Descent Dynamics},
  author={E, Weinan and Wojtowytsch, Stephan},
  volume={1},
  DOI={10.4208/csiam-am.20-211},
  number={3},
  journal={CSIAM Transactions on Applied Mathematics},
  year={2020},
  month={Sep},
  pages={387–440},
  url={https://arxiv.org/abs/2007.15623}
}

@article{e2020comparative,
  title={A comparative analysis of optimization and generalization properties of two-layer neural network and random feature models under gradient descent dynamics},
  author={E, Weinan and Ma, Chao and Wu, Lei},
  journal={Science China Mathematics},
  volume={63},
  number={7},
  pages={1235--1258},
  year={2020},
  publisher={Science in China Press}
}

@inproceedings{rahimi2008uniform,
  title={Uniform approximation of functions with random bases},
  author={Rahimi, Ali and Recht, Benjamin},
  booktitle={2008 46th annual allerton conference on communication, control, and computing},
  pages={555--561},
  year={2008},
  organization={IEEE}
}

@article{dieuleveut2017harder,
  title={Harder, better, faster, stronger convergence rates for least-squares regression},
  author={Dieuleveut, Aymeric and Flammarion, Nicolas and Bach, Francis},
  journal={Journal of Machine Learning Research},
  volume={18},
  number={101},
  pages={1--51},
  year={2017}
}

@misc{chiang2025TC0,
  title={{Transformers in Uniform TC$^0$}}, 
  author={David Chiang},
  year={2025},
  eprint={2409.13629},
  archivePrefix={arXiv},
  primaryClass={cs.CC},
  url={https://arxiv.org/abs/2409.13629}, 
}

@article{wu2022spectral,
  author  = {Lei Wu and Jihao Long},
  title   = {A spectral-based analysis of the separation between two-layer neural networks and linear methods},
  journal = {Journal of Machine Learning Research},
  year    = {2022},
  volume  = {23},
  number  = {119},
  pages   = {1--34},
  url     = {http://jmlr.org/papers/v23/21-092.html}
}

@InProceedings{safran2019depth,
  title = 	 {Depth Separations in Neural Networks: What is Actually Being Separated?},
  author =       {Safran, Itay and Eldan, Ronen and Shamir, Ohad},
  booktitle = 	 {Proceedings of the Thirty-Second Conference on Learning Theory},
  pages = 	 {2664--2666},
  year = 	 {2019},
  editor = 	 {Beygelzimer, Alina and Hsu, Daniel},
  volume = 	 {99},
  series = 	 {Proceedings of Machine Learning Research},
  month = 	 {25--28 Jun},
  publisher =    {PMLR},
}

@InProceedings{daniely2017depth,
  title = 	 {Depth Separation for Neural Networks},
  author = 	 {Daniely, Amit},
  booktitle = 	 {Proceedings of the 2017 Conference on Learning Theory},
  pages = 	 {690--696},
  year = 	 {2017},
  editor = 	 {Kale, Satyen and Shamir, Ohad},
  volume = 	 {65},
  series = 	 {Proceedings of Machine Learning Research},
  month = 	 {07--10 Jul},
  publisher =    {PMLR},
}

@article{bach2017breaking,
  author  = {Francis Bach},
  title   = {Breaking the Curse of Dimensionality with Convex Neural Networks},
  journal = {Journal of Machine Learning Research},
  year    = {2017},
  volume  = {18},
  number  = {19},
  pages   = {1--53},
  url     = {http://jmlr.org/papers/v18/14-546.html}
}

@book{berlinet2011reproducing,
  title={Reproducing kernel Hilbert spaces in probability and statistics},
  author={Berlinet, Alain and Thomas-Agnan, Christine},
  year={2011},
  publisher={Springer Science \& Business Media}
}

@article{aronszajn1950theory,
  title={Theory of reproducing kernels},
  author={Aronszajn, Nachman},
  journal={Transactions of the American mathematical society},
  volume={68},
  number={3},
  pages={337--404},
  year={1950}
}

@article{merrill2023parallelism,
  title={The parallelism tradeoff: Limitations of log-precision transformers},
  author={Merrill, William and Sabharwal, Ashish},
  journal={Transactions of the Association for Computational Linguistics},
  volume={11},
  pages={531--545},
  year={2023},
  publisher={MIT Press One Broadway, 12th Floor, Cambridge, Massachusetts 02142, USA~…}
}

@misc{strobl2023averagehard,
      title={Average-Hard Attention {Transformers} are Constant-Depth Uniform Threshold Circuits}, 
      author={Lena Strobl},
      year={2023},
      eprint={2308.03212},
      archivePrefix={arXiv},
      primaryClass={cs.CL},
      url={https://arxiv.org/abs/2308.03212}, 
}

@article{merrill2022saturated,
  title={Saturated transformers are constant-depth threshold circuits},
  author={Merrill, William and Sabharwal, Ashish and Smith, Noah A},
  journal={Transactions of the Association for Computational Linguistics},
  volume={10},
  pages={843--856},
  year={2022},
  publisher={MIT Press One Broadway, 12th Floor, Cambridge, Massachusetts 02142, USA}
}

@article{vaswani2017attention,
  title={Attention is all you need},
  author={Vaswani, Ashish and Shazeer, Noam and Parmar, Niki and Uszkoreit, Jakob and Jones, Llion and Gomez, Aidan N and Kaiser, {\L}ukasz and Polosukhin, Illia},
  journal={Advances in neural information processing systems},
  volume={30},
  year={2017}
}

@article{shazeer2020glu,
  title={{GLU variants improve Transformer}},
  author={Shazeer, Noam},
  journal={arXiv preprint arXiv:2002.05202},
  year={2020}
}

@article{raffel2020exploring,
  title={Exploring the limits of transfer learning with a unified text-to-text transformer},
  author={Raffel, Colin and Shazeer, Noam and Roberts, Adam and Lee, Katherine and Narang, Sharan and Matena, Michael and Zhou, Yanqi and Li, Wei and Liu, Peter J},
  journal={Journal of machine learning research},
  volume={21},
  number={140},
  pages={1--67},
  year={2020}
}

@article{su2024roformer,
  title={Roformer: Enhanced transformer with rotary position embedding},
  author={Su, Jianlin and Ahmed, Murtadha and Lu, Yu and Pan, Shengfeng and Bo, Wen and Liu, Yunfeng},
  journal={Neurocomputing},
  volume={568},
  pages={127063},
  year={2024},
  publisher={Elsevier}
}

@article{press2021train,
  title={Train short, test long: Attention with linear biases enables input length extrapolation},
  author={Press, Ofir and Smith, Noah A and Lewis, Mike},
  journal={arXiv preprint arXiv:2108.12409},
  year={2021}
}

@inproceedings{nguyen2019prenorm,
    title = "Transformers without Tears: Improving the Normalization of Self-Attention",
    author = "Nguyen, Toan Q.  and
      Salazar, Julian",
    editor = {Niehues, Jan  and
      Cattoni, Rolando  and
      St{\"u}ker, Sebastian  and
      Negri, Matteo  and
      Turchi, Marco  and
      others},
    booktitle = "Proceedings of the 16th International Conference on Spoken Language Translation",
    month = nov # " 2-3",
    year = "2019",
    address = "Hong Kong",
    publisher = "Association for Computational Linguistics",
    url = "https://aclanthology.org/2019.iwslt-1.17/",
}

@misc{nie2025llada,
  title={Large Language Diffusion Models}, 
  author={Shen Nie and Fengqi Zhu and Zebin You and Xiaolu Zhang and Jingyang Ou and Jun Hu and Jun Zhou and Yankai Lin and Ji-Rong Wen and Chongxuan Li},
  year={2025},
  eprint={2502.09992},
  archivePrefix={arXiv},
  primaryClass={cs.CL},
  url={https://arxiv.org/abs/2502.09992}, 
}

@article{dziri2023faith,
  title={Faith and fate: Limits of transformers on compositionality},
  author={Dziri, Nouha and Lu, Ximing and Sclar, Melanie and Li, Xiang Lorraine and Jiang, Liwei and Lin, Bill Yuchen and Welleck, Sean and West, Peter and Bhagavatula, Chandra and Le Bras, Ronan and others},
  journal={Advances in Neural Information Processing Systems},
  volume={36},
  pages={70293--70332},
  year={2023}
}

@article{peng2024limitations,
  title={On limitations of the transformer architecture},
  author={Peng, Binghui and Narayanan, Srini and Papadimitriou, Christos},
  journal={Collegium Beatus Rhenanus},
  year={2024},
  publisher={Springer Nature Switzerland}
}

@article{donatelli2023compositionality,
  title={Compositionality in computational linguistics},
  author={Donatelli, Lucia and Koller, Alexander},
  journal={Annual Review of Linguistics},
  volume={9},
  number={1},
  pages={463--481},
  year={2023},
  publisher={Annual Reviews}
}

@article{karchmer1995super,
  title={Super-logarithmic depth lower bounds via the direct sum in communication complexity},
  author={Karchmer, Mauricio and Raz, Ran and Wigderson, Avi},
  journal={Computational Complexity},
  volume={5},
  number={3},
  pages={191--204},
  year={1995},
  publisher={Springer}
}

@article{gavinsky2017toward,
  title={Toward better formula lower bounds: The composition of a function and a universal relation},
  author={Gavinsky, Dmitry and Meir, Or and Weinstein, Omri and Wigderson, Avi},
  journal={SIAM Journal on Computing},
  volume={46},
  number={1},
  pages={114--131},
  year={2017},
  publisher={SIAM}
}

@book{arora2009computational,
  title={Computational complexity: a modern approach},
  author={Arora, Sanjeev and Barak, Boaz},
  year={2009},
  publisher={Cambridge University Press}
}

@book{vollmer1999introduction,
  title={Introduction to circuit complexity: a uniform approach},
  author={Vollmer, Heribert},
  year={1999},
  publisher={Springer Science \& Business Media}
}

@unpublished{barrington2000lecture,
  author  = {Barrington, David Mix and Maciel, Alexis},
  title   = {Lecture 6: Arithmetic and Threshold Circuits},
  year    = {2000},
  url     = {{https://people.clarkson.edu/~alexis/PCMI/}},
  urldate = {2025-10-09},
  note    = {Advanced course on computational complexity. IAS/Park City Mathematics Institute}
}

@article{jerabek2012root,
    author = {Emil Jeřábek},
    title = {Root finding with threshold circuits},
    journal = {Theoretical Computer Science},
    volume = {462},
    pages = {59-69},
    year = {2012},
    issn = {0304-3975},
    doi = {https://doi.org/10.1016/j.tcs.2012.09.001},
}

@article{goldmann1998simulating,
author = {Goldmann, Mikael and Karpinski, Marek},
title = {Simulating Threshold Circuits by Majority Circuits},
journal = {SIAM Journal on Computing},
volume = {27},
number = {1},
pages = {230-246},
year = {1998},
doi = {10.1137/S0097539794274519},
}

@article{hesse2002uniform,
    author = {William Hesse and Eric Allender and David A. Mix Barrington},
    title = {Uniform constant-depth threshold circuits for division and iterated multiplication},
    journal = {Journal of Computer and System Sciences},
    volume = {65},
    number = {4},
    pages = {695-716},
    year = {2002},
    note = {Special Issue on Complexity 2001},
    issn = {0022-0000},
    doi = {https://doi.org/10.1016/S0022-0000(02)00025-9},
}

@article{guo2025deepseek,
  title={{DeepSeek-R1} incentivizes reasoning in {LLMs} through reinforcement learning},
  author={Guo, Daya and Yang, Dejian and Zhang, Haowei and Song, Junxiao and Wang, Peiyi and Zhu, Qihao and Xu, Runxin and Zhang, Ruoyu and Ma, Shirong and Bi, Xiao and others},
  journal={Nature},
  volume={645},
  number={8081},
  pages={633--638},
  year={2025},
  publisher={Nature Publishing Group UK London}
}

@article{chandra1984constant,
  title={Constant depth reducibility},
  author={Chandra, Ashok K and Stockmeyer, Larry and Vishkin, Uzi},
  journal={SIAM Journal on Computing},
  volume={13},
  number={2},
  pages={423--439},
  year={1984},
  publisher={SIAM}
}

@article{gabrie2022adaptive,
  title={Adaptive {Monte Carlo} augmented with normalizing flows},
  author={Gabri{\'e}, Marylou and Rotskoff, Grant M and Vanden-Eijnden, Eric},
  journal={Proceedings of the National Academy of Sciences},
  volume={119},
  number={10},
  pages={e2109420119},
  year={2022},
  publisher={National Academy of Sciences}
}

@misc{zelikman2024quietstar,
  title={{Quiet-STaR}: Language Models Can Teach Themselves to Think Before Speaking}, 
  author={Eric Zelikman and Georges Harik and Yijia Shao and Varuna Jayasiri and Nick Haber and Noah D. Goodman},
  year={2024},
  eprint={2403.09629},
  archivePrefix={arXiv},
  primaryClass={cs.CL},
  url={https://arxiv.org/abs/2403.09629}, 
}

@misc{murdza2024hausdorff,
  title={Hausdorff measure of zeros of polynomials}, 
  author={Andrew Murdza and Khai T. Nguyen and Etienne Phillips},
  year={2024},
  eprint={2312.17462},
  archivePrefix={arXiv},
  primaryClass={math.CA},
  url={https://arxiv.org/abs/2312.17462}, 
}

@misc{fischer2016polynomial,
  TITLE = {Is the set $P^{-1}(\{0\})$ a set of measure zero for any multivariate polynomial?},
  AUTHOR = {Daniel Fischer},
  HOWPUBLISHED = {Mathematics Stack Exchange},
  NOTE = {(version: 2016-12-17)},
  URL = {https://math.stackexchange.com/q/2062328}
}

@book{lipton2004inference,
    title={Inference to the Best Explanation},
    author={Lipton, Peter},
    year={2004},
    publisher={Routledge Taylor \& Francis Group},
    address={London},
    edition={2nd}
}

@article{davey2024inferring,
  title={On Inferring Explanations and Inference to the Best Explanation},
  author={Davey, Kevin},
  journal={Episteme},
  volume={21},
  number={4},
  pages={1120--1137},
  year={2024},
  publisher={Cambridge University Press}
}

@article{van2001effects,
  title={The effects of readers’ goals on inference generation and memory for texts},
  author={Van den Broek, Paul and Lorch, Robert F and Linderholm, Tracy and Gustafson, Mary},
  journal={Memory \& cognition},
  volume={29},
  number={8},
  pages={1081--1087},
  year={2001},
  publisher={Springer}
}

@article{yeari2017online,
  title={Online and offline inferential and textual processing of poor comprehenders: Evidence from a probing method},
  author={Yeari, Menahem and Elentok, Shiri and Schiff, Rachel},
  journal={Journal of Experimental Child Psychology},
  volume={155},
  pages={12--31},
  year={2017},
  publisher={Elsevier}
}

@book{josephson1996abductive,
  title={Abductive inference: Computation, philosophy, technology},
  author={Josephson, John R and Josephson, Susan G},
  year={1996},
  publisher={Cambridge University Press}
}

@article{campos2011distinction,
  title={On the distinction between {Peirce}’s abduction and {Lipton}’s inference to the best explanation},
  author={Campos, Daniel G},
  journal={Synthese},
  volume={180},
  number={3},
  pages={419--442},
  year={2011},
  publisher={Springer}
}

@article{liu2024deepseekv3,
  title={{DeepSeek-V3} technical report},
  author={Liu, Aixin and Feng, Bei and Xue, Bing and Wang, Bingxuan and Wu, Bochao and Lu, Chengda and Zhao, Chenggang and Deng, Chengqi and Zhang, Chenyu and Ruan, Chong and others},
  journal={arXiv preprint arXiv:2412.19437},
  year={2024}
}

@misc{deepseek2025multiround,
  author={{DeepSeek Inc.}},
  title={{DeepSeek API Docs. Reasoning model (deepseek-reasoner)}},
  url={https://api-docs.deepseek.com/guides/reasoning_model},
  year = {2025},
  note = {[Accessed 12-12-2025]},
}

@misc{torch2025clipgrad,
  author = {PyTorch},
  title = {torch/nn/utils/clip\_grad.py},
  year = {2025},
  howpublished = {\url{https://github.com/pytorch/pytorch/blob/5c61c25545eb8be7e1245b1acdd3ac901961fc17/torch/nn/utils/clip_grad.py#L233}},
  note = {Accessed: 2025-12-22.},
}

@article{yang2024memory,
    title={Memory$^3$: {Language} modeling with explicit memory},
    volume={3},
    DOI={10.4208/jml.240708},
    number={3},
    journal={Journal of Machine Learning},
    author={Yang, Hongkang and Lin, Zehao and Wang, Wenjin and Wu, Hao and Li, Zhiyu and Tang, Bo and Wei, Wenqiang and Wang, Jinbo and Tang, Zeyun and Song, Shichao and Xi, Chenyang and Yu, Yu and Chen, Kai and Xiong, Feiyu and Tang, Linpeng and E, Weinan},
    year={2024},
    month={Sep},
    pages={300–346}
}

@misc{shao2024deepseekmath,
      title={{DeepSeekMath}: Pushing the Limits of Mathematical Reasoning in Open Language Models}, 
      author={Zhihong Shao and Peiyi Wang and Qihao Zhu and Runxin Xu and Junxiao Song and Xiao Bi and Haowei Zhang and Mingchuan Zhang and Y. K. Li and Y. Wu and Daya Guo},
      year={2024},
      eprint={2402.03300},
      archivePrefix={arXiv},
      primaryClass={cs.CL},
      url={https://arxiv.org/abs/2402.03300}, 
}

@article{morokoff1995quasi,
  title={Quasi-{Monte Carlo} integration},
  author={Morokoff, William J and Caflisch, Russel E},
  journal={Journal of computational physics},
  volume={122},
  number={2},
  pages={218--230},
  year={1995},
  publisher={Elsevier}
}

@article{marjanovic2025thoughtology,
  title={{DeepSeek-R1 Thoughtology: Let's think about LLM Reasoning}},
  author={Marjanovi{\'c}, Sara Vera and Patel, Arkil and Adlakha, Vaibhav and Aghajohari, Milad and BehnamGhader, Parishad and Bhatia, Mehar and Khandelwal, Aditi and Kraft, Austin and Krojer, Benno and L{\`u}, Xing Han and others},
  journal={arXiv preprint arXiv:2504.07128},
  year={2025}
}

@misc{weber2024redpajama,
      title={{RedPajama: an Open Dataset for Training Large Language Models}},
      author={Maurice Weber and Daniel Fu and Quentin Anthony and Yonatan Oren and Shane Adams and Anton Alexandrov and Xiaozhong Lyu and Huu Nguyen and Xiaozhe Yao and Virginia Adams and others},
      year={2024},
      eprint={2411.12372},
      archivePrefix={arXiv},
      primaryClass={cs.CL},
      url={https://arxiv.org/abs/2411.12372}, 
}

@misc{yang2024qwen25math,
      title={{Qwen2.5-Math Technical Report: Toward Mathematical Expert Model via Self-Improvement}}, 
      author={An Yang and Beichen Zhang and Binyuan Hui and Bofei Gao and Bowen Yu and Chengpeng Li and Dayiheng Liu and Jianhong Tu and Jingren Zhou and Junyang Lin and others},
      year={2024},
      eprint={2409.12122},
      archivePrefix={arXiv},
      primaryClass={cs.CL},
      url={https://arxiv.org/abs/2409.12122}, 
}

@misc{qwen2025qwen25,
      title={Qwen2.5 Technical Report}, 
      author={An Yang and Baosong Yang and Beichen Zhang and Binyuan Hui and Bo Zheng and Bowen Yu and Chengyuan Li and Dayiheng Liu and Fei Huang and Haoran Wei and others},
      year={2025},
      eprint={2412.15115},
      archivePrefix={arXiv},
      primaryClass={cs.CL},
      url={https://arxiv.org/abs/2412.15115}, 
}

@misc{yang2025qwen3,
      title={Qwen3 Technical Report}, 
      author={An Yang and Anfeng Li and Baosong Yang and Beichen Zhang and Binyuan Hui and Bo Zheng and Bowen Yu and Chang Gao and Chengen Huang and Chenxu Lv and others},
      year={2025},
      eprint={2505.09388},
      archivePrefix={arXiv},
      primaryClass={cs.CL},
      url={https://arxiv.org/abs/2505.09388}, 
}

@misc{grattafiori2024llama3,
      title={{The Llama 3 Herd of Models}},
      author={Aaron Grattafiori and Abhimanyu Dubey and Abhinav Jauhri and Abhinav Pandey and Abhishek Kadian and Ahmad Al-Dahle and Aiesha Letman and Akhil Mathur and Alan Schelten and Alex Vaughan and others},
      year={2024},
      eprint={2407.21783},
      archivePrefix={arXiv},
      primaryClass={cs.AI},
      url={https://arxiv.org/abs/2407.21783},
}

@book{kahneman2011fastslow,
  title={Thinking, fast and slow},
  author={Kahneman, Daniel},
  year={2011},
  publisher={Farrar, Straus and Giroux}
}

@article{kahneman2003maps,
  title={Maps of bounded rationality: Psychology for behavioral economics},
  author={Kahneman, Daniel},
  journal={American economic review},
  volume={93},
  number={5},
  pages={1449--1475},
  year={2003},
  publisher={American Economic Association}
}

@article{stanovich2000individual,
  title={Individual differences in reasoning: Implications for the rationality debate?},
  volume={23},
  DOI={10.1017/S0140525X00003435},
  number={5},
  journal={Behavioral and Brain Sciences},
  author={Stanovich, Keith E. and West, Richard F.},
  year={2000},
  pages={645–665}
}

@book{kechris2012classical,
  title={Classical descriptive set theory},
  author={Kechris, Alexander},
  volume={156},
  year={2012},
  publisher={Springer Science \& Business Media}
}

@article{shi2024thorough,
  title={A thorough examination of decoding methods in the era of {LLMs}},
  author={Shi, Chufan and Yang, Haoran and Cai, Deng and Zhang, Zhisong and Wang, Yifan and Yang, Yujiu and Lam, Wai},
  journal={arXiv preprint arXiv:2402.06925},
  year={2024}
}

@article{meister2023locally,
    title = "Locally Typical Sampling",
    author = "Meister, Clara  and
      Pimentel, Tiago  and
      Wiher, Gian  and
      Cotterell, Ryan",
    journal = "Transactions of the Association for Computational Linguistics",
    volume = "11",
    year = "2023",
    address = "Cambridge, MA",
    publisher = "MIT Press",
    url = "https://aclanthology.org/2023.tacl-1.7/",
    doi = "10.1162/tacl_a_00536",
    pages = "102--121",
}

@article{hewitt2022truncation,
  title={Truncation sampling as language model desmoothing},
  author={Hewitt, John and Manning, Christopher D and Liang, Percy},
  journal={arXiv preprint arXiv:2210.15191},
  year={2022}
}

@article{keskar2019ctrl,
  title={{CTRL}: A conditional transformer language model for controllable generation},
  author={Keskar, Nitish Shirish and McCann, Bryan and Varshney, Lav R and Xiong, Caiming and Socher, Richard},
  journal={arXiv preprint arXiv:1909.05858},
  year={2019}
}

@misc{huggingface2026generationconfig,
    author = {{HuggingFace}},
	title = {Parameters for manipulation of the model output logits},
	howpublished = {\url{https://huggingface.co/docs/transformers/en/main_classes/text_generation#transformers.GenerationConfig.top_k}},
	year = {2026},
	note = {[Accessed 12-01-2026] Logit masking is applied as a default in the code.},
}

@article{feutrill2021review,
  title={A review of {Shannon} and differential entropy rate estimation},
  author={Feutrill, Andrew and Roughan, Matthew},
  journal={Entropy},
  volume={23},
  number={8},
  pages={1046},
  year={2021},
  publisher={MDPI}
}

@article{wyner2002ergodic,
  title={Some asymptotic properties of the entropy of a stationary ergodic data source with applications to data compression},
  author={Wyner, Aaron D and Ziv, Jacob},
  journal={IEEE Transactions on Information Theory},
  volume={35},
  number={6},
  pages={1250--1258},
  year={2002},
  publisher={IEEE}
}

@inproceedings{johnson2016perceptualloss,
  title={Perceptual losses for real-time style transfer and super-resolution},
  author={Johnson, Justin and Alahi, Alexandre and Fei-Fei, Li},
  booktitle={European conference on computer vision},
  pages={694--711},
  year={2016},
  organization={Springer}
}

@InProceedings{liu2021generic,
    author    = {Liu, Yifan and Chen, Hao and Chen, Yu and Yin, Wei and Shen, Chunhua},
    title     = {Generic Perceptual Loss for Modeling Structured Output Dependencies},
    booktitle = {Proceedings of the IEEE/CVF Conference on Computer Vision and Pattern Recognition (CVPR)},
    month     = {June},
    year      = {2021},
    pages     = {5424-5432}
}

@article{van2016conditional,
  title={Conditional image generation with pixelcnn decoders},
  author={Van den Oord, Aaron and Kalchbrenner, Nal and Espeholt, Lasse and Vinyals, Oriol and Graves, Alex and others},
  journal={Advances in neural information processing systems},
  volume={29},
  year={2016}
}

@article{van2017VQVAE,
  title={Neural discrete representation learning},
  author={Van Den Oord, Aaron and Vinyals, Oriol and others},
  journal={Advances in neural information processing systems},
  volume={30},
  year={2017}
}

@article{sun2024autoregressive,
  title={Autoregressive model beats diffusion: {Llama} for scalable image generation},
  author={Sun, Peize and Jiang, Yi and Chen, Shoufa and Zhang, Shilong and Peng, Bingyue and Luo, Ping and Yuan, Zehuan},
  journal={arXiv preprint arXiv:2406.06525},
  year={2024}
}

@article{song2021maximum,
  title={Maximum likelihood training of score-based diffusion models},
  author={Song, Yang and Durkan, Conor and Murray, Iain and Ermon, Stefano},
  journal={Advances in neural information processing systems},
  volume={34},
  pages={1415--1428},
  year={2021}
}

@article{song2019generative,
  title={Generative modeling by estimating gradients of the data distribution},
  author={Song, Yang and Ermon, Stefano},
  journal={Advances in neural information processing systems},
  volume={32},
  year={2019}
}

@inproceedings{sohl2015deep,
  title={Deep unsupervised learning using nonequilibrium thermodynamics},
  author={Sohl-Dickstein, Jascha and Weiss, Eric and Maheswaranathan, Niru and Ganguli, Surya},
  booktitle={International conference on machine learning},
  pages={2256--2265},
  year={2015},
  organization={pmlr}
}

@inproceedings{rombach2022high,
  title={High-resolution image synthesis with latent diffusion models},
  author={Rombach, Robin and Blattmann, Andreas and Lorenz, Dominik and Esser, Patrick and Ommer, Bj{\"o}rn},
  booktitle={Proceedings of the IEEE/CVF conference on computer vision and pattern recognition},
  pages={10684--10695},
  year={2022}
}

@misc{dieleman2024spectral,
  author = {Dieleman, Sander},
  title = {Diffusion is spectral autoregression},
  url = {https://sander.ai/2024/09/02/spectral-autoregression.html},
  year = {2024}
}

@article{falck2025fourier,
  title={A {Fourier} Space Perspective on Diffusion Models},
  author={Falck, Fabian and Pandeva, Teodora and Zahirnia, Kiarash and Lawrence, Rachel and Turner, Richard and Meeds, Edward and Zazo, Javier and Karmalkar, Sushrut},
  journal={arXiv preprint arXiv:2505.11278},
  year={2025}
}

@article{fischler1973representation,
  title={The representation and matching of pictorial structures},
  author={Fischler, Martin A and Elschlager, Robert A},
  journal={IEEE Transactions on computers},
  volume={100},
  number={1},
  pages={67--92},
  year={1973},
  publisher={IEEE}
}

@article{felzenszwalb2005pictorial,
  title={Pictorial structures for object recognition},
  author={Felzenszwalb, Pedro F and Huttenlocher, Daniel P},
  journal={International journal of computer vision},
  volume={61},
  number={1},
  pages={55--79},
  year={2005},
  publisher={Springer}
}

@article{felzenszwalb2009object,
  title={Object detection with discriminatively trained part-based models},
  author={Felzenszwalb, Pedro F and Girshick, Ross B and McAllester, David and Ramanan, Deva},
  journal={IEEE transactions on pattern analysis and machine intelligence},
  volume={32},
  number={9},
  pages={1627--1645},
  year={2009},
  publisher={IEEE}
}

@inproceedings{caron2021emerging,
  title={Emerging properties in self-supervised vision transformers},
  author={Caron, Mathilde and Touvron, Hugo and Misra, Ishan and J{\'e}gou, Herv{\'e} and Mairal, Julien and Bojanowski, Piotr and Joulin, Armand},
  booktitle={Proceedings of the IEEE/CVF international conference on computer vision},
  pages={9650--9660},
  year={2021}
}

@inproceedings{zeiler2014visualizing,
  title={Visualizing and understanding convolutional networks},
  author={Zeiler, Matthew D and Fergus, Rob},
  booktitle={European conference on computer vision},
  pages={818--833},
  year={2014},
  organization={Springer}
}

@article{han2022vision,
  title={Vision gnn: An image is worth graph of nodes},
  author={Han, Kai and Wang, Yunhe and Guo, Jianyuan and Tang, Yehui and Wu, Enhua},
  journal={Advances in neural information processing systems},
  volume={35},
  pages={8291--8303},
  year={2022}
}

@article{wei2025OCR,
  title={{DeepSeek-OCR}: Contexts optical compression},
  author={Wei, Haoran and Sun, Yaofeng and Li, Yukun},
  journal={arXiv preprint arXiv:2510.18234},
  year={2025}
}

@article{cheng2025glyph,
  title={Glyph: Scaling context windows via visual-text compression},
  author={Cheng, Jiale and Liu, Yusen and Zhang, Xinyu and Fei, Yulin and Hong, Wenyi and Lyu, Ruiliang and Wang, Weihan and Su, Zhe and Gu, Xiaotao and Liu, Xiao and others},
  journal={arXiv preprint arXiv:2510.17800},
  year={2025}
}

@misc{bai2023qwenvl,
      title={{Qwen-VL}: A Versatile Vision-Language Model for Understanding, Localization, Text Reading, and Beyond}, 
      author={Jinze Bai and Shuai Bai and Shusheng Yang and Shijie Wang and Sinan Tan and Peng Wang and Junyang Lin and Chang Zhou and Jingren Zhou},
      year={2023},
      eprint={2308.12966},
      archivePrefix={arXiv},
      primaryClass={cs.CV},
      url={https://arxiv.org/abs/2308.12966}, 
}

@article{alayrac2022flamingo,
  title={Flamingo: a visual language model for few-shot learning},
  author={Alayrac, Jean-Baptiste and Donahue, Jeff and Luc, Pauline and Miech, Antoine and Barr, Iain and Hasson, Yana and Lenc, Karel and Mensch, Arthur and Millican, Katherine and Reynolds, Malcolm and others},
  journal={Advances in neural information processing systems},
  volume={35},
  pages={23716--23736},
  year={2022}
}

@article{lin2025adaptvision,
  title={{AdaptVision}: Efficient Vision-Language Models via Adaptive Visual Acquisition},
  author={Lin, Zichuan and Liu, Yicheng and Yang, Yang and Tao, Lvfang and Ye, Deheng},
  journal={arXiv preprint arXiv:2512.03794},
  year={2025}
}

@inproceedings{ke2025infant,
  title={Discovering hidden visual concepts beyond linguistic input in Infant learning},
  author={Ke, Xueyi and Tsutsui, Satoshi and Zhang, Yayun and Wen, Bihan},
  booktitle={Proceedings of the Computer Vision and Pattern Recognition Conference},
  pages={4343--4352},
  year={2025}
}

@inproceedings{higgins2017beta,
  title={$\beta$-{VAE}: Learning basic visual concepts with a constrained variational framework},
  author={Higgins, Irina and Matthey, Loic and Pal, Arka and Burgess, Christopher and Glorot, Xavier and Botvinick, Matthew and Mohamed, Shakir and Lerchner, Alexander},
  booktitle={International conference on learning representations},
  year={2017}
}

@InCollection{rescorla2024LoTH,
	author       =	{Rescorla, Michael},
	title        =	{{The Language of Thought Hypothesis}},
	booktitle    =	{{The Stanford Encyclopedia of Philosophy}},
	editor       =	{Edward N. Zalta and Uri Nodelman},
	howpublished =	{\url{https://plato.stanford.edu/archives/sum2024/entries/language-thought/}},
	year         =	{2024},
	edition      =	{{S}ummer 2024},
	publisher    =	{Metaphysics Research Lab, Stanford University}
}

@book{fodor1975LoTH,
    place={New York},
    title={The Language of Thought},
    publisher={Thomas Y. Crowell Company},
    author={Fodor, Jerry},
    year={1975}
}

@book{kallenberg2021foundations,
  title={Foundations of modern probability},
  author={Kallenberg, Olav},
  year={2021},
  publisher={Springer},
  edition={3rd}
}

@article{chen2016infogan,
  title={{InfoGAN}: Interpretable representation learning by information maximizing generative adversarial nets},
  author={Chen, Xi and Duan, Yan and Houthooft, Rein and Schulman, John and Sutskever, Ilya and Abbeel, Pieter},
  journal={Advances in neural information processing systems},
  volume={29},
  year={2016}
}

@article{huffman2007method,
  title={A method for the construction of minimum-redundancy codes},
  author={Huffman, David A},
  journal={Proceedings of the IRE},
  volume={40},
  number={9},
  pages={1098--1101},
  year={2007},
  publisher={IEEE}
}

@article{shannon1948mathematical,
  title={A mathematical theory of communication},
  author={Shannon, Claude E},
  journal={The Bell system technical journal},
  volume={27},
  number={3},
  pages={379--423},
  year={1948},
  publisher={Nokia Bell Labs}
}

@inproceedings{goodfellow2014generative,
  title={Generative adversarial nets},
  author={Goodfellow, Ian and Pouget-Abadie, Jean and Mirza, Mehdi and Xu, Bing and Warde-Farley, David and Ozair, Sherjil and Courville, Aaron and Bengio, Yoshua},
  booktitle={Advances in neural information processing systems},
  pages={2672--2680},
  year={2014}
}

@article{tabak2010density,
  title={Density estimation by dual ascent of the log-likelihood},
  author={Tabak, Esteban G and Vanden-Eijnden, Eric},
  journal={Communications in Mathematical Sciences},
  volume={8},
  number={1},
  pages={217--233},
  year={2010},
  publisher={International Press of Boston}
}

@article{kingma2013auto,
  title={Auto-encoding variational Bayes},
  author={Kingma, Diederik P and Welling, Max},
  journal={arXiv preprint arXiv:1312.6114},
  year={2013}
}

@article{albergo2022building,
  title={Building Normalizing Flows with Stochastic Interpolants},
  author={Albergo, Michael S and Vanden-Eijnden, Eric},
  journal={arXiv preprint arXiv:2209.15571},
  year={2022}
}

@article{chamberlain2013local,
  title={Local processing enhancements associated with superior observational drawing are due to enhanced perceptual functioning, not weak central coherence},
  author={Chamberlain, Rebecca and McManus, IC and Riley, Howard and Rankin, Qona and Brunswick, Nicola},
  journal={Quarterly Journal of Experimental Psychology},
  volume={66},
  number={7},
  pages={1448--1466},
  year={2013},
  publisher={SAGE Publications Sage UK: London, England}
}

@article{drake2024artists,
  title={Artists have superior local and global processing abilities but show a preference for initially drawing globally.},
  author={Drake, Jennifer E and Riccio, Ariana and Chamberlain, Rebecca and Kozbelt, Aaron},
  journal={Psychology of aesthetics, creativity, and the arts},
  volume={18},
  number={3},
  pages={357},
  year={2024},
  publisher={Educational Publishing Foundation}
}

@article{dosovitskiy2020image,
  title={An image is worth 16x16 words: Transformers for image recognition at scale},
  author={Dosovitskiy, Alexey},
  journal={arXiv preprint arXiv:2010.11929},
  year={2020}
}

@misc{openai2024o1card,
      title={{OpenAI} o1 System Card}, 
      author={Aaron Jaech and Adam Kalai and Adam Lerer and Adam Richardson and Ahmed El-Kishky and Aiden Low and Alec Helyar and Aleksander Madry and Alex Beutel and Alex Carney and others},
      year={2024},
      eprint={2412.16720},
      archivePrefix={arXiv},
      primaryClass={cs.AI},
      url={https://arxiv.org/abs/2412.16720}, 
}

@inproceedings{ling2017program,
    title = "Program Induction by Rationale Generation: Learning to Solve and Explain Algebraic Word Problems",
    author = "Ling, Wang  and
      Yogatama, Dani  and
      Dyer, Chris  and
      Blunsom, Phil",
    editor = "Barzilay, Regina  and
      Kan, Min-Yen",
    booktitle = "Proceedings of the 55th Annual Meeting of the Association for Computational Linguistics (Volume 1: Long Papers)",
    month = jul,
    year = "2017",
    address = "Vancouver, Canada",
    publisher = "Association for Computational Linguistics",
    doi = "10.18653/v1/P17-1015",
    pages = "158--167"
}

@article{wei2022chain,
  title={Chain-of-thought prompting elicits reasoning in large language models},
  author={Wei, Jason and Wang, Xuezhi and Schuurmans, Dale and Bosma, Maarten and Xia, Fei and Chi, Ed and Le, Quoc V and Zhou, Denny and others},
  journal={Advances in neural information processing systems},
  volume={35},
  pages={24824--24837},
  year={2022}
}

@article{zelikman2022star,
  title={{STaR}: Bootstrapping reasoning with reasoning},
  author={Zelikman, Eric and Wu, Yuhuai and Mu, Jesse and Goodman, Noah},
  journal={Advances in Neural Information Processing Systems},
  volume={35},
  pages={15476--15488},
  year={2022}
}

@misc{paster2023openwebmath,
      title={{OpenWebMath}: An Open Dataset of High-Quality Mathematical Web Text}, 
      author={Keiran Paster and Marco Dos Santos and Zhangir Azerbayev and Jimmy Ba},
      year={2023},
      eprint={2310.06786},
      archivePrefix={arXiv},
      primaryClass={cs.AI},
      url={https://arxiv.org/abs/2310.06786}, 
}

@misc{rajbhandari2020zero,
      title={{ZeRO}: Memory Optimizations Toward Training Trillion Parameter Models}, 
      author={Samyam Rajbhandari and Jeff Rasley and Olatunji Ruwase and Yuxiong He},
      year={2020},
      eprint={1910.02054},
      archivePrefix={arXiv},
      primaryClass={cs.LG},
      url={https://arxiv.org/abs/1910.02054}, 
}

@misc{loshchilov2019AdamW,
      title={Decoupled Weight Decay Regularization}, 
      author={Ilya Loshchilov and Frank Hutter},
      year={2019},
      eprint={1711.05101},
      archivePrefix={arXiv},
      primaryClass={cs.LG},
      url={https://arxiv.org/abs/1711.05101}, 
}

@misc{yuan2026yuan30flash,
      title={{Yuan3.0 Flash}: An Open Multimodal Large Language Model for Enterprise Applications}, 
      author={Shawn Wu and Sean Wang and Louie Li and Darcy Chen and Allen Wang and Jiangang Luo and Xudong Zhao and Joseph Shen and Gawain Ma and Jasper Jia and others},
      year={2026},
      eprint={2601.01718},
      archivePrefix={arXiv},
      primaryClass={cs.AI},
      url={https://arxiv.org/abs/2601.01718}, 
}

@article{schoenfeld1974continuous,
  title={Continuous surjections from {Cantor} sets to compact metric spaces},
  author={Schoenfeld, Alan H},
  journal={AMERICAN MATHEMATICAL SOCIETY},
  volume={46},
  number={1},
  year={1974}
}

@article{alexandroff1927stetige,
  title={{\"U}ber stetige Abbildungen kompakter R{\"a}ume},
  author={Alexandroff, Paul},
  journal={Mathematische Annalen},
  volume={96},
  number={1},
  pages={555--571},
  year={1927},
  publisher={Springer}
}

@book{villani2003topics,
  title={Topics in optimal transportation},
  author={Villani, C{\'e}dric},
  number={58},
  year={2003},
  publisher={American Mathematical Soc.}
}

@article{kantorovich1960mathematical,
  title={Mathematical methods of organizing and planning production},
  author={Kantorovich, Leonid V},
  journal={Management science},
  volume={6},
  number={4},
  pages={366--422},
  year={1960},
  publisher={INFORMS}
}

@article{brenier1991polar,
  author = {Brenier, Yann},
  title = {Polar factorization and monotone rearrangement of vector-valued functions},
  journal = {Communications on Pure and Applied Mathematics},
  volume = {44},
  number = {4},
  pages = {375-417},
  year = {1991}
}

@inproceedings{devlin2019BERT,
    title = "{BERT}: Pre-training of Deep Bidirectional Transformers for Language Understanding",
    author = "Devlin, Jacob  and
      Chang, Ming-Wei  and
      Lee, Kenton  and
      Toutanova, Kristina",
    editor = "Burstein, Jill  and
      Doran, Christy  and
      Solorio, Thamar",
    booktitle = "Proceedings of the 2019 Conference of the North {A}merican Chapter of the Association for Computational Linguistics: Human Language Technologies, Volume 1 (Long and Short Papers)",
    month = jun,
    year = "2019",
    address = "Minneapolis, Minnesota",
    publisher = "Association for Computational Linguistics",
    doi = "10.18653/v1/N19-1423",
    pages = "4171--4186",
}

@article{raffel2020T5,
  author  = {Colin Raffel and Noam Shazeer and Adam Roberts and Katherine Lee and Sharan Narang and Michael Matena and Yanqi Zhou and Wei Li and Peter J. Liu},
  title   = {Exploring the Limits of Transfer Learning with a Unified Text-to-Text Transformer},
  journal = {Journal of Machine Learning Research},
  year    = {2020},
  volume  = {21},
  number  = {140},
  pages   = {1--67},
  url     = {http://jmlr.org/papers/v21/20-074.html}
}

@misc{hu2025complexnetwork,
      title={Emergent Slow Thinking in {LLMs} as Inverse Tree Freezing},
      author={Sihan Hu and Xiansheng Cai and Yuan Huang and Zhiyuan Yao and Linfeng Zhang and Pan Zhang and Youjin Deng and Kun Chen},
      year={2026},
      eprint={2509.23629},
      archivePrefix={arXiv},
      primaryClass={cs.AI},
      url={https://arxiv.org/abs/2509.23629}, 
}

@inproceedings{pandian2025snap,
    title = "Snap Out of It: A Dual-Process Approach to Mitigating Overthinking in Language Model Reasoning",
    author = "Pandian, Ashish  and
      Lojo, Nelson  and
      Lai, Wei Xun  and
      Lukas, Jackson",
    editor = "Kamalloo, Ehsan  and
      Gontier, Nicolas  and
      Lu, Xing Han  and
      Dziri, Nouha  and
      Murty, Shikhar  and
      Lacoste, Alexandre",
    booktitle = "Proceedings of the 1st Workshop for Research on Agent Language Models (REALM 2025)",
    month = jul,
    year = "2025",
    address = "Vienna, Austria",
    publisher = "Association for Computational Linguistics",
    doi = "10.18653/v1/2025.realm-1.16",
    pages = "228--249",
    ISBN = "979-8-89176-264-0",
}

@misc{hong2026reconsider,
      title={Reconsidering Overthinking: Penalizing Internal and External Redundancy in CoT Reasoning}, 
      author={Jialiang Hong and Taihang Zhen and Kai Chen and Jiaheng Liu and Junlan Feng and Wenpeng Zhu and Jing Huo and Yang Gao and Depeng Wang and Haitao Wan and others},
      year={2026},
      eprint={2508.02178},
      archivePrefix={arXiv},
      primaryClass={cs.AI},
      url={https://arxiv.org/abs/2508.02178}, 
}

@misc{ling2025fasteasydeephard,
      title={Fast on the Easy, Deep on the Hard: Efficient Reasoning via Powered Length Penalty}, 
      author={Zehui Ling and Deshu Chen and Hongwei Zhang and Yifeng Jiao and Xin Guo and Yuan Cheng},
      year={2025},
      eprint={2506.10446},
      archivePrefix={arXiv},
      primaryClass={cs.CL},
      url={https://arxiv.org/abs/2506.10446}, 
}

@misc{kimi2025kimik15,
      title={Kimi k1.5: Scaling Reinforcement Learning with LLMs}, 
      author={Angang Du and Bofei Gao and Bowei Xing and Changjiu Jiang and Cheng Chen and Cheng Li and Chenjun Xiao and Chenzhuang Du and Chonghua Liao and Chuning Tang and others},
      year={2025},
      eprint={2501.12599},
      archivePrefix={arXiv},
      primaryClass={cs.AI},
      url={https://arxiv.org/abs/2501.12599}, 
}

@misc{chen2025think23o1,
      title={Do NOT Think That Much for 2+3=? On the Overthinking of o1-Like {LLMs}}, 
      author={Xingyu Chen and Jiahao Xu and Tian Liang and Zhiwei He and Jianhui Pang and Dian Yu and Linfeng Song and Qiuzhi Liu and Mengfei Zhou and Zhuosheng Zhang and others},
      year={2025},
      eprint={2412.21187},
      archivePrefix={arXiv},
      primaryClass={cs.CL},
      url={https://arxiv.org/abs/2412.21187}, 
}

@misc{wei2026evolution,
      title={The Evolution of Thought: Tracking {LLM} Overthinking via Reasoning Dynamics Analysis}, 
      author={Zihao Wei and Liang Pang and Jiahao Liu and Wenjie Shi and Jingcheng Deng and Shicheng Xu and Zenghao Duan and Fei Sun and Huawei Shen and Xueqi Cheng},
      year={2026},
      eprint={2508.17627},
      archivePrefix={arXiv},
      primaryClass={cs.CL},
      url={https://arxiv.org/abs/2508.17627}, 
}

@misc{yu2025DAPO,
      title={{DAPO}: An Open-Source {LLM} Reinforcement Learning System at Scale}, 
      author={Qiying Yu and Zheng Zhang and Ruofei Zhu and Yufeng Yuan and Xiaochen Zuo and Yu Yue and Weinan Dai and Tiantian Fan and Gaohong Liu and Lingjun Liu and others},
      year={2025},
      eprint={2503.14476},
      archivePrefix={arXiv},
      primaryClass={cs.LG},
      url={https://arxiv.org/abs/2503.14476}, 
}

@misc{cui2025entropy,
      title={The Entropy Mechanism of Reinforcement Learning for Reasoning Language Models}, 
      author={Ganqu Cui and Yuchen Zhang and Jiacheng Chen and Lifan Yuan and Zhi Wang and Yuxin Zuo and Haozhan Li and Yuchen Fan and Huayu Chen and Weize Chen and Zhiyuan Liu and Hao Peng and Lei Bai and Wanli Ouyang and Yu Cheng and Bowen Zhou and Ning Ding},
      year={2025},
      eprint={2505.22617},
      archivePrefix={arXiv},
      primaryClass={cs.LG},
      url={https://arxiv.org/abs/2505.22617}, 
}

@misc{huang2025spark,
      title={Low-probability Tokens Sustain Exploration in Reinforcement Learning with Verifiable Reward}, 
      author={Guanhua Huang and Tingqiang Xu and Mingze Wang and Qi Yi and Xue Gong and Siheng Li and Ruibin Xiong and Kejiao Li and Yuhao Jiang and Bo Zhou},
      year={2025},
      eprint={2510.03222},
      archivePrefix={arXiv},
      primaryClass={cs.LG},
      url={https://arxiv.org/abs/2510.03222}, 
}

@ARTICLE{zhang2026system2,
    author={Zhang, Duzhen and Li, Zhong-Zhi and Zhang, Ming-Liang and Zhang, Jiaxin and Liu, Zengyan and Yao, Yuxuan and Xu, Haotian and Zheng, Junhao and Chen, Xiuyi and Zhang, Yingying and others},
    journal={ IEEE Transactions on Pattern Analysis \& Machine Intelligence },
    title={{ From System 1 to System 2: A Survey of Reasoning Large Language Models }},
    year={2026},
    volume={48},
    number={03},
    ISSN={1939-3539},
    pages={3335-3354},
    doi={10.1109/TPAMI.2025.3637037},
    publisher={IEEE Computer Society},
    address={Los Alamitos, CA, USA},
    month=mar
}

@misc{pan2025survey,
      title={A Survey of Slow Thinking-based Reasoning LLMs using Reinforced Learning and Inference-time Scaling Law}, 
      author={Qianjun Pan and Wenkai Ji and Yuyang Ding and Junsong Li and Shilian Chen and Junyi Wang and Jie Zhou and Qin Chen and Min Zhang and Yulan Wu and Liang He},
      year={2025},
      eprint={2505.02665},
      archivePrefix={arXiv},
      primaryClass={cs.AI},
      url={https://arxiv.org/abs/2505.02665}, 
}

@inproceedings{wang2025multistep,
    title = "Understanding the Language Model to Solve the Symbolic Multi-Step Reasoning Problem from the Perspective of Buffer Mechanism",
    author = "Wang, Zhiwei  and
      Wang, Yunji  and
      Zhang, Zhongwang  and
      Zhou, Zhangchen  and
      Jin, Hui  and
      Hu, Tianyang  and
      Sun, Jiacheng  and
      Li, Zhenguo  and
      Zhang, Yaoyu  and
      Xu, Zhi-Qin John",
    editor = "Christodoulopoulos, Christos  and
      Chakraborty, Tanmoy  and
      Rose, Carolyn  and
      Peng, Violet",
    booktitle = "Findings of the Association for Computational Linguistics: EMNLP 2025",
    month = nov,
    year = "2025",
    address = "Suzhou, China",
    publisher = "Association for Computational Linguistics",
    doi = "10.18653/v1/2025.findings-emnlp.893",
    pages = "16446--16474",
    ISBN = "979-8-89176-335-7",
}

@inproceedings{
wang2025beyond,
    title={Beyond the 80/20 Rule: High-Entropy Minority Tokens Drive Effective Reinforcement Learning for {LLM} Reasoning},
    author={Shenzhi Wang and Le Yu and Chang Gao and Chujie Zheng and Shixuan Liu and Rui Lu and Kai Dang and Xiong-Hui Chen and Jianxin Yang and Zhenru Zhang and others},
    booktitle={The Thirty-ninth Annual Conference on Neural Information Processing Systems},
    year={2025},
    url={https://openreview.net/forum?id=yfcpdY4gMP}
}

@misc{cheng2025exploration,
      title={Reasoning with Exploration: An Entropy Perspective}, 
      author={Daixuan Cheng and Shaohan Huang and Xuekai Zhu and Bo Dai and Wayne Xin Zhao and Zhenliang Zhang and Furu Wei},
      year={2025},
      eprint={2506.14758},
      archivePrefix={arXiv},
      primaryClass={cs.CL},
      url={https://arxiv.org/abs/2506.14758}, 
}

@misc{sui2025stopover,
      title={Stop Overthinking: A Survey on Efficient Reasoning for Large Language Models}, 
      author={Yang Sui and Yu-Neng Chuang and Guanchu Wang and Jiamu Zhang and Tianyi Zhang and Jiayi Yuan and Hongyi Liu and Andrew Wen and Shaochen Zhong and Na Zou and Hanjie Chen and Xia Hu},
      year={2025},
      eprint={2503.16419},
      archivePrefix={arXiv},
      primaryClass={cs.CL},
      url={https://arxiv.org/abs/2503.16419}, 
}

@inproceedings{zhang2025light,
    title = "{L}ight{T}hinker: Thinking Step-by-Step Compression",
    author = "Zhang, Jintian  and
      Zhu, Yuqi  and
      Sun, Mengshu  and
      Luo, Yujie  and
      Qiao, Shuofei  and
      Du, Lun  and
      Zheng, Da  and
      Chen, Huajun  and
      Zhang, Ningyu",
    editor = "Christodoulopoulos, Christos  and
      Chakraborty, Tanmoy  and
      Rose, Carolyn  and
      Peng, Violet",
    booktitle = "Proceedings of the 2025 Conference on Empirical Methods in Natural Language Processing",
    month = nov,
    year = "2025",
    address = "Suzhou, China",
    publisher = "Association for Computational Linguistics",
    doi = "10.18653/v1/2025.emnlp-main.673",
    pages = "13307--13328",
    ISBN = "979-8-89176-332-6",
}

@InProceedings{liao2025reward,
  title = 	 {Reward-Guided Speculative Decoding for Efficient {LLM} Reasoning},
  author =       {Liao, Baohao and Xu, Yuhui and Dong, Hanze and Li, Junnan and Monz, Christof and Savarese, Silvio and Sahoo, Doyen and Xiong, Caiming},
  booktitle = 	 {Proceedings of the 42nd International Conference on Machine Learning},
  pages = 	 {37555--37572},
  year = 	 {2025},
  editor = 	 {Singh, Aarti and Fazel, Maryam and Hsu, Daniel and Lacoste-Julien, Simon and Berkenkamp, Felix and Maharaj, Tegan and Wagstaff, Kiri and Zhu, Jerry},
  volume = 	 {267},
  series = 	 {Proceedings of Machine Learning Research},
  month = 	 {13--19 Jul},
  publisher =    {PMLR},
  url = 	 {https://proceedings.mlr.press/v267/liao25f.html},
}

@inproceedings{ding2025dynamic,
    title = "Dynamic Parallel Tree Search for Efficient {LLM} Reasoning",
    author = "Ding, Yifu  and
      Jiang, Wentao  and
      Liu, Shunyu  and
      Jing, Yongcheng  and
      Guo, Jinyang  and
      Wang, Yingjie  and
      Zhang, Jing  and
      Wang, Zengmao  and
      Liu, Ziwei  and
      Du, Bo  and others",
    editor = "Che, Wanxiang  and
      Nabende, Joyce  and
      Shutova, Ekaterina  and
      Pilehvar, Mohammad Taher",
    booktitle = "Proceedings of the 63rd Annual Meeting of the Association for Computational Linguistics (Volume 1: Long Papers)",
    month = jul,
    year = "2025",
    address = "Vienna, Austria",
    publisher = "Association for Computational Linguistics",
    doi = "10.18653/v1/2025.acl-long.550",
    pages = "11233--11252",
    ISBN = "979-8-89176-251-0",
}

@inproceedings{yue2025does,
    title={Does Reinforcement Learning Really Incentivize Reasoning Capacity in {LLM}s Beyond the Base Model?},
    author={Yang Yue and Zhiqi Chen and Rui Lu and Andrew Zhao and Zhaokai Wang and Yang Yue and Shiji Song and Gao Huang},
    booktitle={The Thirty-ninth Annual Conference on Neural Information Processing Systems},
    year={2025},
    url={https://openreview.net/forum?id=4OsgYD7em5}
}

@inproceedings{karan2026reasoning,
    title={Reasoning without Training: Your Base Model is Smarter Than You Think},
    author={Aayush Karan and Yilun Du},
    booktitle={The Fourteenth International Conference on Learning Representations},
    year={2026},
    url={https://openreview.net/forum?id=Vsgq2ldr4K}
}

@misc{chen2025continuous,
      title={Reasoning Beyond Language: A Comprehensive Survey on Latent Chain-of-Thought Reasoning}, 
      author={Xinghao Chen and Anhao Zhao and Heming Xia and Xuan Lu and Hanlin Wang and Yanjun Chen and Wei Zhang and Jian Wang and Wenjie Li and Xiaoyu Shen},
      year={2025},
      eprint={2505.16782},
      archivePrefix={arXiv},
      primaryClass={cs.CL},
      url={https://arxiv.org/abs/2505.16782}, 
}

@misc{hao2025coconut,
      title={Training Large Language Models to Reason in a Continuous Latent Space}, 
      author={Shibo Hao and Sainbayar Sukhbaatar and DiJia Su and Xian Li and Zhiting Hu and Jason Weston and Yuandong Tian},
      year={2025},
      eprint={2412.06769},
      archivePrefix={arXiv},
      primaryClass={cs.CL},
      url={https://arxiv.org/abs/2412.06769}, 
}

@book{von2012computer,
  title={The computer and the brain},
  author={Von Neumann, John and Kurzweil, Ray},
  year={2012},
  publisher={Yale university press}
}

@incollection{von2017general,
  title={The general and logical theory of automata},
  author={Von Neumann, John},
  booktitle={Systems research for behavioral science},
  pages={97--107},
  year={2017},
  publisher={Routledge}
}

@book{wiener2019cybernetics,
  title={Cybernetics or Control and Communication in the Animal and the Machine},
  author={Wiener, Norbert},
  year={2019},
  publisher={MIT press}
}

@incollection{turing2007computing,
  title={Computing machinery and intelligence},
  author={Turing, Alan M},
  booktitle={Parsing the Turing test: Philosophical and methodological issues in the quest for the thinking computer},
  pages={23--65},
  year={2007},
  publisher={Springer}
}

@book{grenander1996elements,
  title={Elements of pattern theory},
  author={Grenander, Ulf},
  year={1996},
  publisher={JHU Press}
}

@inproceedings{mumford1994pattern,
  title={Pattern theory: a unifying perspective},
  author={Mumford, David},
  booktitle={First European Congress of Mathematics: Paris, July 6-10, 1992 Volume I Invited Lectures (Part 1)},
  pages={187--224},
  year={1994},
  organization={Springer}
}

@book{ma2025textbook,
  title={Learning Deep Representations of Data Distributions},
  author={Buchanan, Sam and Pai, Druv and Wang, Peng and Ma, Yi},
  month=aug,
  year={2025},
  publisher={Online},
  note={\url{https://ma-lab-berkeley.github.io/deep-representation-learning-book/}.}
}

@article{ma2024white,
  author  = {Yaodong Yu and Sam Buchanan and Druv Pai and Tianzhe Chu and Ziyang Wu and Shengbang Tong and Hao Bai and Yuexiang Zhai and Benjamin D. Haeffele and Yi Ma},
  title   = {White-Box Transformers via Sparse Rate Reduction: Compression Is All There Is?},
  journal = {Journal of Machine Learning Research},
  year    = {2024},
  volume  = {25},
  number  = {300},
  pages   = {1--128},
  url     = {http://jmlr.org/papers/v25/23-1547.html}
}

@article{friston2010free,
  title={The free-energy principle: a unified brain theory?},
  author={Friston, Karl},
  journal={Nature reviews neuroscience},
  volume={11},
  number={2},
  pages={127--138},
  year={2010},
  publisher={Nature publishing group}
}

@article{friston2023free,
  title={The free energy principle made simpler but not too simple},
  author={Friston, Karl and Da Costa, Lancelot and Sajid, Noor and Heins, Conor and Ueltzh{\"o}ffer, Kai and Pavliotis, Grigorios A and Parr, Thomas},
  journal={Physics Reports},
  volume={1024},
  pages={1--29},
  year={2023},
  publisher={Elsevier}
}

@article{omran2007overview,
  title={An overview of clustering methods},
  author={Omran, Mahamed GH and Engelbrecht, Andries P and Salman, Ayed},
  journal={Intelligent Data Analysis},
  volume={11},
  number={6},
  pages={583--605},
  year={2007},
  publisher={SAGE Publications Sage UK: London, England}
}

@book{alpaydin2020introduction,
  title={Introduction to machine learning},
  author={Alpaydin, Ethem},
  year={2020},
  publisher={MIT press}
}

@inproceedings{lafferty2001CRF,
    author = {Lafferty, John D. and McCallum, Andrew and Pereira, Fernando C. N.},
    title = {Conditional Random Fields: Probabilistic Models for Segmenting and Labeling Sequence Data},
    year = {2001},
    isbn = {1558607781},
    publisher = {Morgan Kaufmann Publishers Inc.},
    address = {San Francisco, CA, USA},
    booktitle = {Proceedings of the Eighteenth International Conference on Machine Learning},
    pages = {282–289},
    numpages = {8},
    series = {ICML '01}
}

@article{collins2003head,
  title={Head-driven statistical models for natural language parsing},
  author={Collins, Michael},
  journal={Computational linguistics},
  volume={29},
  number={4},
  pages={589--637},
  year={2003},
  publisher={MIT Press One Rogers Street, Cambridge, MA 02142-1209, USA journals-info~…}
}

@article{brown1993translation,
    title = "The Mathematics of Statistical Machine Translation: Parameter Estimation",
    author = "Brown, Peter F.  and
      Della Pietra, Stephen A.  and
      Della Pietra, Vincent J.  and
      Mercer, Robert L.",
    editor = "Hirschberg, Julia",
    journal = "Computational Linguistics",
    volume = "19",
    number = "2",
    year = "1993",
    address = "Cambridge, MA",
    publisher = "MIT Press",
    url = "https://aclanthology.org/J93-2003/",
    pages = "263--311"
}

@article{blei2003latent,
  title={Latent dirichlet allocation},
  author={Blei, David M and Ng, Andrew Y and Jordan, Michael I},
  journal={Journal of machine Learning research},
  volume={3},
  number={Jan},
  pages={993--1022},
  year={2003}
}

@article{tabak2018explanation,
  title={Explanation of variability and removal of confounding factors from data through optimal transport},
  author={Tabak, Esteban G and Trigila, Giulio},
  journal={Communications on Pure and Applied Mathematics},
  volume={71},
  number={1},
  pages={163--199},
  year={2018},
  publisher={Wiley Online Library}
}

@inproceedings{chen2025circuit,
    title = "Circuit Complexity Bounds for {R}o{PE}-based Transformer Architecture",
    author = "Chen, Bo  and
      Li, Xiaoyu  and
      Liang, Yingyu  and
      Long, Jiangxuan  and
      Shi, Zhenmei  and
      Song, Zhao  and
      Zhang, Jiahao",
    editor = "Christodoulopoulos, Christos  and
      Chakraborty, Tanmoy  and
      Rose, Carolyn  and
      Peng, Violet",
    booktitle = "Proceedings of the 2025 Conference on Empirical Methods in Natural Language Processing",
    month = nov,
    year = "2025",
    address = "Suzhou, China",
    publisher = "Association for Computational Linguistics",
    doi = "10.18653/v1/2025.emnlp-main.561",
    pages = "11080--11097",
    ISBN = "979-8-89176-332-6",
}

@inproceedings{geiping2025scaling,
    title={Scaling up Test-Time Compute with Latent Reasoning: A Recurrent Depth Approach},
    author={Jonas Geiping and Sean Michael McLeish and Neel Jain and John Kirchenbauer and Siddharth Singh and Brian R. Bartoldson and Bhavya Kailkhura and Abhinav Bhatele and Tom Goldstein},
    booktitle={The Thirty-ninth Annual Conference on Neural Information Processing Systems},
    year={2025},
    url={https://openreview.net/forum?id=S3GhJooWIC}
}

@article{zeng2025pretraining,
  title={Pretraining Language Models to Ponder in Continuous Space},
  author={Zeng, Boyi and Song, Shixiang and Huang, Siyuan and Wang, Yixuan and Li, He and He, Ziwei and Wang, Xinbing and Li, Zhiyu and Lin, Zhouhan},
  journal={arXiv preprint arXiv:2505.20674},
  year={2025}
}

@article{beck2024xlstm,
  title={{xLSTM}: Extended long short-term memory},
  author={Beck, Maximilian and P{\"o}ppel, Korbinian and Spanring, Markus and Auer, Andreas and Prudnikova, Oleksandra and Kopp, Michael and Klambauer, G{\"u}nter and Brandstetter, Johannes and Hochreiter, Sepp},
  journal={Advances in Neural Information Processing Systems},
  volume={37},
  pages={107547--107603},
  year={2024}
}

@misc{peng2025rwkv7goose,
      title={{RWKV-7 "Goose"} with Expressive Dynamic State Evolution}, 
      author={Bo Peng and Ruichong Zhang and Daniel Goldstein and Eric Alcaide and Xingjian Du and Haowen Hou and Jiaju Lin and Jiaxing Liu and Janna Lu and William Merrill and others},
      year={2025},
      eprint={2503.14456},
      archivePrefix={arXiv},
      primaryClass={cs.CL},
      url={https://arxiv.org/abs/2503.14456}, 
}

@inproceedings{sun2025TTT,
  title={Learning to (Learn at Test Time): {RNN}s with Expressive Hidden States},
  author={Yu Sun and Xinhao Li and Karan Dalal and Jiarui Xu and Arjun Vikram and Genghan Zhang and Yann Dubois and Xinlei Chen and Xiaolong Wang and Sanmi Koyejo and others},
  booktitle={Forty-second International Conference on Machine Learning},
  year={2025},
  url={https://openreview.net/forum?id=wXfuOj9C7L}
}

@article{hendler2008avoiding,
  title={Avoiding another {AI} winter},
  author={Hendler, James},
  journal={IEEE Intelligent Systems},
  volume={23},
  number={02},
  pages={2--4},
  year={2008},
  publisher={IEEE Computer Society}
}

@book{crevier1993ai,
  title={{AI}: the tumultuous history of the search for artificial intelligence},
  author={Crevier, Daniel},
  year={1993},
  publisher={Basic Books, Inc.}
}

@article{agar2020science,
  title={What is science for? The {Lighthill} report on artificial intelligence reinterpreted},
  author={Agar, Jon},
  journal={The British Journal for the History of Science},
  volume={53},
  number={3},
  pages={289--310},
  year={2020},
  publisher={Cambridge University Press}
}

@article{mitchell2021ai,
  title={Why {AI} is harder than we think},
  author={Mitchell, Melanie},
  journal={arXiv preprint arXiv:2104.12871},
  year={2021}
}

@article{gholami2024wall,
  title={{AI} and memory wall},
  author={Gholami, Amir and Yao, Zhewei and Kim, Sehoon and Hooper, Coleman and Mahoney, Michael W and Keutzer, Kurt},
  journal={IEEE Micro},
  volume={44},
  number={3},
  pages={33--39},
  year={2024},
  publisher={IEEE}
}

@misc{wu2025combating,
  title={Combating the Memory Walls: Optimization Pathways for Long-Context Agentic {LLM} Inference}, 
  author={Haoran Wu and Can Xiao and Jiayi Nie and Xuan Guo and Binglei Lou and Jeffrey T. H. Wong and Zhiwen Mo and Cheng Zhang and Przemyslaw Forys and Wayne Luk and others},
  year={2025},
  eprint={2509.09505},
  archivePrefix={arXiv},
  primaryClass={cs.AR},
  url={https://arxiv.org/abs/2509.09505},
}

@misc{wolters2024CIM,
  title={Memory Is All You Need: An Overview of Compute-in-Memory Architectures for Accelerating Large Language Model Inference}, 
  author={Christopher Wolters and Xiaoxuan Yang and Ulf Schlichtmann and Toyotaro Suzumura},
  year={2024},
  eprint={2406.08413},
  archivePrefix={arXiv},
  primaryClass={cs.AR},
  url={https://arxiv.org/abs/2406.08413},
}

@inproceedings{zeng2024flightllm,
  title={{FlightLLM}: Efficient large language model inference with a complete mapping flow on {FPGAs}},
  author={Zeng, Shulin and Liu, Jun and Dai, Guohao and Yang, Xinhao and Fu, Tianyu and Wang, Hongyi and Ma, Wenheng and Sun, Hanbo and Li, Shiyao and Huang, Zixiao and others},
  booktitle={Proceedings of the 2024 ACM/SIGDA International Symposium on Field Programmable Gate Arrays},
  pages={223--234},
  year={2024}
}

@inproceedings{narayanan2021efficient,
  title={Efficient large-scale language model training on {GPU} clusters using {Megatron-LM}},
  author={Narayanan, Deepak and Shoeybi, Mohammad and Casper, Jared and LeGresley, Patrick and Patwary, Mostofa and Korthikanti, Vijay and Vainbrand, Dmitri and Kashinkunti, Prethvi and Bernauer, Julie and Catanzaro, Bryan and others},
  booktitle={Proceedings of the international conference for high performance computing, networking, storage and analysis},
  pages={1--15},
  year={2021}
}

@inproceedings{li2018optimal,
  title={An optimal control approach to deep learning and applications to discrete-weight neural networks},
  author={Li, Qianxiao and Hao, Shuji},
  booktitle={International Conference on Machine Learning},
  pages={2985--2994},
  year={2018},
  organization={PMLR}
}

@misc{ren2024MPC,
  title={Unifying back-propagation and forward-forward algorithms through model predictive control}, 
  author={Lianhai Ren and Qianxiao Li},
  year={2024},
  eprint={2409.19561},
  archivePrefix={arXiv},
  primaryClass={cs.LG},
  url={https://arxiv.org/abs/2409.19561}, 
}

@article{barbero2024glasses,
  title={Transformers need glasses! Information over-squashing in language tasks},
  author={Barbero, Federico and Banino, Andrea and Kapturowski, Steven and Kumaran, Dharshan and Ara{\'u}jo, Jo{\~a}o G and Vitvitskyi, Alex and Pascanu, Razvan and Veli{\v{c}}kovi{\'c}, Petar},
  journal={Advances in Neural Information Processing Systems},
  volume={37},
  pages={98111--98142},
  year={2024}
}

@article{liu2024lost,
  title={Lost in the middle: How language models use long contexts},
  author={Liu, Nelson F and Lin, Kevin and Hewitt, John and Paranjape, Ashwin and Bevilacqua, Michele and Petroni, Fabio and Liang, Percy},
  journal={Transactions of the association for computational linguistics},
  volume={12},
  pages={157--173},
  year={2024}
}

@inproceedings{barrington1986bounded,
  title={Bounded-width polynomial-size branching programs recognize exactly those languages in ${NC}^1$},
  author={Barrington, David A},
  booktitle={Proceedings of the eighteenth annual ACM symposium on Theory of computing},
  pages={1--5},
  year={1986}
}

@book{straubing2012finite,
  title={Finite automata, formal logic, and circuit complexity},
  author={Straubing, Howard},
  year={2012},
  publisher={Springer Science \& Business Media}
}

@article{allender2010amplifying,
  title={Amplifying lower bounds by means of self-reducibility},
  author={Allender, Eric and Kouck{\`y}, Michal},
  journal={Journal of the ACM (JACM)},
  volume={57},
  number={3},
  pages={1--36},
  year={2010},
  publisher={ACM New York, NY, USA}
}

@inproceedings{yao1989circuits,
  title={Circuits and local computation},
  author={Yao, Andrew C},
  booktitle={Proceedings of the twenty-first annual ACM symposium on Theory of computing},
  pages={186--196},
  year={1989}
}

@inproceedings{li2024chain,
  title={Chain of thought empowers transformers to solve inherently serial problems},
  author={Li, Zhiyuan and Liu, Hong and Zhou, Denny and Ma, Tengyu},
  booktitle={The Twelfth International Conference on Learning Representations},
  year={2024}
}

@article{liu2022transformers,
  title={Transformers learn shortcuts to automata},
  author={Liu, Bingbin and Ash, Jordan T and Goel, Surbhi and Krishnamurthy, Akshay and Zhang, Cyril},
  journal={arXiv preprint arXiv:2210.10749},
  year={2022}
}

@article{strobl2024formal,
  title={What formal languages can transformers express? a survey},
  author={Strobl, Lena and Merrill, William and Weiss, Gail and Chiang, David and Angluin, Dana},
  journal={Transactions of the Association for Computational Linguistics},
  volume={12},
  pages={543--561},
  year={2024},
  publisher={MIT Press 255 Main Street, 9th Floor, Cambridge, Massachusetts 02142, USA~…}
}

@article{cisek2019resynthesizing,
  title={Resynthesizing behavior through phylogenetic refinement},
  author={Cisek, Paul},
  journal={Attention, perception, \& psychophysics},
  volume={81},
  number={7},
  pages={2265--2287},
  year={2019},
  publisher={Springer}
}

@article{cisek2022evolution,
  title={Evolution of behavioural control from chordates to primates},
  author={Cisek, Paul},
  journal={Philosophical Transactions of the Royal Society B: Biological Sciences},
  volume={377},
  number={1844},
  year={2022},
  publisher={The Royal Society}
}

@misc{lin2025RAVR,
  title={{RAVR}: Reference-Answer-guided Variational Reasoning for Large Language Models}, 
  author={Tianqianjin Lin and Xi Zhao and Xingyao Zhang and Rujiao Long and Yi Xu and Zhuoren Jiang and Wenbo Su and Bo Zheng},
  year={2025},
  eprint={2510.25206},
  archivePrefix={arXiv},
  primaryClass={cs.AI},
  url={https://arxiv.org/abs/2510.25206},
}

@article{zhou2025variational,
  title={Variational Reasoning for Language Models},
  author={Zhou, Xiangxin and Liu, Zichen and Wang, Haonan and Du, Chao and Lin, Min and Li, Chongxuan and Wang, Liang and Pang, Tianyu},
  journal={arXiv preprint arXiv:2509.22637},
  year={2025}
}
\bibliographystyle{plainurl}

\appendix

\section{Background on Circuit Complexity}
\label{appendix. circuit complexity}

This section provides a brief introduction to the key concepts from circuit complexity theory that are used in this paper.
Standard materials on Boolean circuits can be found in \cite{vollmer1999introduction,arora2009computational,straubing2012finite},
while materials on \textsf{DLOGTIME}-uniform $\TC^0$ circuits can be found in \cite{barrington2000lecture,jerabek2012root}.

A Boolean circuit is a mapping $\{0,1\}^n\to\{0,1\}^m$, where the output length $m$ defaults to 1 unless stated otherwise.
It can be defined as a directed acyclic graph (DAG) with $n$ sources (vertices without incoming edges) and $m$ sinks (vertices without outgoing edges).
Each non-source vertex is a gate, either AND, OR or NOT.
Each AND or OR gate can have unbounded fan-in (incoming edges), and each non-sink gate can have unbounded fan-out (outgoing edges).
The circuit size is the number of gates, and circuit depth the longest path in this DAG.
A set $\{C_n\}_{n=0}^{\infty}$ is called a circuit family if each $C_n$ has $n$ sources.

A language is a subset of binary strings $\{0,1\}^* = \bigcup_{n=0}^{\infty}\{0,1\}^n$.
For instance, PARITY consists of strings with an even number of 1's.
A circuit family $\{C_n\}_{n=0}^{\infty}$ decides or recognizes a language $L$ if for any string $x$, $x$ belongs to $L$ if and only if $C_{|x|}(x)=1$.
So different circuits are responsible for inputs of different lengths.

\begin{definition}
An $\AC^0$ circuit family $\{C_n\}$ consists of bounded-depth polynomial-size Boolean circuits, i.e.\@ whose circuit depths are $O(1)$ in $n$ and circuit sizes are $O(n^k)$ for some $k\in\N$.
$\AC^0$ denotes the set of languages decidable by $\AC^0$ circuit families.
\end{definition}

\begin{definition}
An $\NC^1$ circuit family $\{C_n\}$ consists of polynomial-size Boolean circuits whose circuit depths are $O(\log n)$ and whose AND and OR gates all have fan-in $=2$.
$\NC^1$ denotes the set of languages decidable by $\NC^1$ circuit families.
\end{definition}

To avoid degenerate examples, some uniformity condition is usually imposed on circuit families.
For instance, a circuit family $\{C_n\}$ is $\mathbf{P}$-uniform if there exists a polynomial-time Turing machine that maps each string $1^n$ to a description of $C_n$.
It is known that the set of languages decidable by $\mathbf{P}$-uniform circuit families is exactly the complexity class $\mathbf{P}$ \cite{arora2009computational}.

\begin{definition}
A circuit family $\{C_n\}$ is \textsf{DLOGTIME}-uniform if there exists a deterministic Turing machine such that, given the input $(n,i,j)$, it determines in $O(\log n)$ time the type of the $i$-th gate of $C_n$ and whether $(i,j)$ is a directed edge in $C_n$.
\end{definition}

In particular, it follows that a \textsf{DLOGTIME}-uniform circuit family must have a polynomial circuit size, since otherwise the gate numbers $i,j$ cannot be read in $O(\log n)$ time.

A threshold gate is a Boolean function $\{0,1\}^n\to\{0,1\}$ that outputs $1$ if and only if $\geq k$ (or $\leq k$) of its input bits are $1$, for some $n$ and $k$.
Since the AND, OR and NOT gates can each be implemented by a threshold gate, a threshold circuit can be defined as a Boolean circuit all of whose gates are threshold gates.
There are several other ways to define threshold circuits and they are equivalent up to some constant change in depth and polynomial change in size \cite{goldmann1998simulating}.

\begin{definition}
A $\TC^0$ circuit family consists of bounded-depth polynomial-size threshold circuits.
$\TC^0$ denotes the set of languages decidable by $\TC^0$ circuit families.
\end{definition}

It is known that $\AC^0 \subsetneq \TC^0 \subseteq \NC^1$ \cite{vollmer1999introduction}.
The conjecture $\TC^0 \subsetneq \NC^1$ is widely believed to be true \cite{yao1989circuits,allender2010amplifying,strobl2024formal}.

\textsf{DLOGTIME}-uniform $\TC^0$ circuits are of particular interest, as they can implement many arithmetical operations \cite{hesse2002uniform,jerabek2012root}.
In particular, each Transformer network can be modeled as a \textsf{DLOGTIME}-uniform $\TC^0$ circuit family \cite{chiang2025TC0}.

\vs
In order to model a function $f\colon  \Omega^* \to \Sigma^*$, where $\Omega,\Sigma$ are arbitrary finite sets, some adaptations need to be made.
First, since the inputs and outputs of Boolean circuits are binary, we represent each element of $\Omega$ (or $\Sigma$) as a binary string of length $\lceil \log_2|\Omega|\rceil$ (or $\lceil \log_2|\Sigma|\rceil$), and thus speak of the input segment $\big[i\lceil \log_2|\Omega|\rceil, (i+1)\lceil \log_2|\Omega|\rceil \big)$ as the $i$-th input token (and the $i$-th output token is similarly defined).

Second, the circuits need to handle variable-length outputs.
As each $C_t$ models $f$ restricted to $\Omega^t$, its output length has a finite upper bound $b_t = \max_{z\in\Sigma^t} |f(z)|$, so we let $C_t$ output $b_t$ tokens.
Since some outputs may have length $<b_t$, we include a $\blank$ symbol and allow $C_t$ to output $\big(\Sigma \cup \{\blank\} \big)^{b_t}$.
Define the function $\text{strip}\colon  \big(\Sigma \cup \{\blank\} \big)^* \to \Sigma^*$ that strips all occurrences of $\blank$ from a string.
Then, we say that $\{C_t\}$ models $f$ if $\text{strip}\circ C_t = f|_{\Omega^t}$ for each $t$.

With non-determinism, these circuits can be extended to model set-valued functions $f\colon  \Sigma^* \to 2^{\Omega^*}$.

\begin{definition}
\label{def. non-deterministic circuit}
A non-deterministic circuit $C\colon \{0,1\}^n \to 2^{\{0,1\}^m}$ is a nonempty-set-valued function.
It can be modeled by a circuit $\tilde{C}\colon  \{0,1\}^{n+k}\to \{0,1\}^m$ with $k$ auxiliary input bits for some $k$, and
\begin{equation*}
\forall x \in \{0,1\}^n, \quad C(x) = \big\{ \tilde{C}(x,y) \bigm| y \in \{0,1\}^k \big\}
\end{equation*}
\end{definition}

Hence, we say that a non-deterministic circuit family $\{C_t\}$ models $f$ if $$\forall t, \forall x \in \Sigma^t, \quad \{ \text{strip}(z) \mid z \in C_t(x) \} = f(x)$$

Note that if $f\colon  \Omega^* \to \Sigma^*$ can be modeled by a \textsf{DLOGTIME}-uniform circuit family, then its output length is at most polynomial, $|f(x)| = O(\poly(|x|))$.
If $f\colon  \Sigma^* \to 2^{\Omega^*}$ can be modeled by a \textsf{DLOGTIME}-uniform nondeterministic circuit family, then the circuits can have at most $O(\poly(|x|))$ auxiliary input bits, and thus the output set has size $2^{O(\poly(|x|))}$.

\section{Measure-Theoretic Proofs}
\label{appendix. measure theory}

\begin{lemma}
\label{lemma. cylinder criteria}
For any $P_1,P_2 \in \PS(\Sigma^{\omega})$, if $P_1([x]) = P_2([x])$ for all $x\in\Sigma^*$, then $P_1=P_2$.
For any sequence $\{P_n\}_{n=1}^{\infty} \subseteq \PS(\Sigma^{\omega})$ and $P_* \in \PS(\Sigma^{\omega})$,
we have $\{P_n\}_{n=1}^{\infty}$ converges weakly to $P_*$ if and only if $P_n([x])\to P_*([x])$ for all $x\in\Sigma^*$.
\end{lemma}

\begin{proof}
Denote by $\mathcal{I}=\{\mathbbm{1}_{[x]} \mid x\in\Sigma^* \}$ the set of all indicator functions of cylinders.
Denote by $\text{span}(\mathcal{I})$ the linear span of $\mathcal{I}$ (i.e.\@ linear combinations of all finite subsets).
Since $\Sigma^{\omega}$ is compact and Hausdorff, the Stone–Weierstrass theorem implies that $\text{span}(\mathcal{I})$ is dense in the space of continuous functions $C(\Sigma^{\omega},\R)$ with respect to the uniform metric.
Hence, $P_1([x])=P_2([x])$ (or $P_n([x])\to P_*([x])$) for all $x \in \Sigma^*$ is equivalent to $\int f dP_1 = \int f dP_2$ (or $\int f dP_n \to \int f dP_*$) for all $f \in C(\Sigma^{\omega},\R)$.
\end{proof}

\begin{proof}[Proof of Proposition \ref{prop. weak topology}]
Note that $D_0(P_*,P_n)\to 0$ is equivalent to $\KL(P_*^{\leq t} \| P_n^{\leq t}) \to 0$ for all $t$.
Since these marginal distributions are over finite sets $\Sigma^t$, the latter condition is equivalent to $P_n([x]) \to P_*([x])$ for all $x\in\Sigma^*$.
Then, Lemma \ref{lemma. cylinder criteria} concludes the proof.
\end{proof}

\begin{proof}[Proof of Proposition \ref{prop. from next-token to infinity}]
First, we construct $\mathfrak{I}$.
Since $\Sigma^{\omega}$ is a compact metric space, Prokhorov's theorem indicates that the weak topology of $\PS(\Sigma^{\omega})$ is compact and metrizable.
For any $t\in\N$ and $p\in\PS(\Sigma^t)$, denote
\begin{equation*}
[p] = \{P \in \PS(\Sigma^{\omega}) \mid P^{\leq t} = p \}
\end{equation*}
This set is always nonempty.
Since the projection $\Sigma^{\omega}\to\Sigma^t$ is continuous, $[p]$ is a compact subset of $\PS(\Sigma^{\omega})$.
Given any $P_{+1}\in\PS_{+1}$, for any $T\in\N$ we define $P_{+1}^{\leq T} \in \PS(\Sigma^T)$ as follows
\begin{equation*}
\forall x \in \Sigma^T, \quad P_{+1}^{\leq T}(x) := \prod_{t=1}^T P_{+1}(x_t | x_{<t})
\end{equation*}
Then, $[P_{+1}^{\leq T+1}] \subseteq [P_{+1}^{\leq T}]$, and thus $\{[P_{+1}^{\leq T}] \}_{T=1}^{\infty}$ is a nested sequence of compact sets.
It follows that the limiting set must be nonempty
\begin{equation*}
S_{\infty}(P_{+1}) := \bigcap_{T=1}^{\infty} [P_{+1}^{\leq T}] \neq \varnothing
\end{equation*}
(Otherwise, $\PS(\Sigma^{\omega})\backslash [P_{+1}^{\leq T}]$ forms a countable cover of a compact space that does not have a finite subcover).
For any $P_1,P_2 \in S_{\infty}(P_{+1})$, we have $P_1^{\leq T} = P_2^{\leq T} = P_{+1}^{\leq T}$ for all $T\in\N$, and thus $P_1([x])=P_2([x])$ for all $x\in\Sigma^*$.
Then, Lemma \ref{lemma. cylinder criteria} implies that $P_1=P_2$, and thus $S_{\infty}(P_{+1})$ is a one point set.
Denote this point by $\mathfrak{I}(P_{+1})$.
Since $\mathfrak{I}(P_{+1})^{\leq t+1} = P_{+1}^{\leq t+1}$, we always have $\mathfrak{I}(P_{+1})^{\leq t+1}(\cdot|x) = P_{+1}^{\leq t+1}(\cdot|x)$ for any $x\in\Sigma^t$ with $\mathfrak{I}(P_{+1})^{\leq t}(x)>0$.

Then, we show that $\mathfrak{I}$ is surjective.
For any $P\in\PS(\Sigma^{\omega})$, construct a $P_{+1}\in\PS_{+1}$ as follows.
For any $T\in\N$ and $x\in\Sigma^T$, if $P([T])>0$, then set $P_{+1}(\cdot|x) = P^{\leq T+1}(\cdot|T)$; otherwise, set it arbitrarily.
One can check that $P \in [P_{+1}^{\leq T}]$ for each $T$ and thus is in $S_{\infty}(P_{+1})$.
It follows that $\mathfrak{I}(P_{+1}) = P$.
\end{proof}

\begin{proposition}
\label{prop. surjective projection}
Given any measureable spaces $A,B$ and a measureable partial function $f\colon A\rightharpoonup B$,
\begin{enumerate}
\item If $\PS(B) = \{ f\#P \mid P \in \PS(A) \}$ where $\PS$ denotes the space of probability measures, then $f$ is surjective.
\item Suppose $A,B$ are Polish spaces (separable and completely metrizable), $f$ is continuous, and its domain $\dom(f)$ is an $F_{\sigma}$ set (a countable union of closed sets).
If $f$ is surjective, then $\PS(B) = \{ f\#P \mid P \in \PS(A) \}$, where $\PS$ denotes the space of Borel probability measures.
\end{enumerate}
\end{proposition}

This proposition is relevant to our setting, since $\Sigma^{\omega},\Omega^{\omega}$ (with product topology) are Polish spaces, and for all examples considered in this paper the projections $\proj$ are continuous and the domains $\dom(\proj)$ are closed (and thus $F_{\sigma}$).
In fact, the condition of $f$ being continuous might be unnecessary, if one applies \cite[Exercise 13.5]{kechris2012classical}.

\begin{proof}[Proof of Proposition \ref{prop. surjective projection}]
For the first statement, suppose $f$ is not surjective.
Choose some $b \in B$ that does not belong to the image of $f$, and let $\delta_b \in \PS(B)$ be the Dirac mass.
Suppose there exists some $P\in\PS(A)$ such that $\delta_b = f\#P$.
Then, $1 = \delta_b(\{b\}) = P(f^{-1}(\{b\})) = P(\varnothing)$.
A contradiction.
Thus, such $P$ cannot exist and $\PS(B)$ is not contained in $\{ f\#P \mid P \in \PS(A) \}$.

Next, we prove the second statement.
For any $b\in B$, the inverse set $f^{-1}(b)$ is non-empty by surjectivity, and relatively closed in $\dom(f)$ by continuity.
Since $\dom(\proj)$ is an $F_{\sigma}$ set, $f^{-1}(b)$ is also $F_{\sigma}$ as a subset of $A$.
By Theorem 5.3 of \cite{kechris2012classical}, $A$ is $\sigma$-compact (being a countable union of compact sets), and thus $f^{-1}(b)$ as an $F_{\sigma}$ subset is also $\sigma$-compact.
Meanwhile, Proposition 12.4 of \cite{kechris2012classical} implies that the graph of $f$ is Borel measureable in $A \times B$.
Hence, the conditions for Theorem 35.46 of \cite{kechris2012classical} are met, and it implies that the set-valued map $f^{-1}$ has a Borel selection, namely a Borel measureable function $g\colon B\to A$ such that $g(b) \in f^{-1}(b)$ for all $b \in B$.
It follows that $f \circ g$ is the identity function on $B$.
For any $P_* \in \PS(B)$, set $P = g\# P_* \in \PS(A)$, then $f\#P = P_*$.
\end{proof}

\begin{proposition}
\label{prop. Cantor measureable surjection}
Let $\Omega$ be a finite set with $|\Omega|>1$, and $\X$ be a Polish space (i.e.\@ separable and completely metrizable).
There exists a Borel measureable surjection $f$ from $\Omega^{\omega}$ to $\X$.
\end{proposition}

\begin{proof}
Denote by $\C=2^{\omega}$ the Cantor space and by $[0,1]^{\omega}$ the Hilbert cube.
By \cite[Theorem 4.14]{kechris2012classical}, there exists a homeomorphism $\psi$ between $\X$ and some $G_{\delta}$ subset (countable intersection of open subsets) of $[0,1]^{\omega}$.
By Borel isomorphism theorem \cite[Theorem 15.6]{kechris2012classical}, there exists some Borel isomorphism $\phi$ between $[0,1]$ and $\C$.
Denote by $\phi^{\omega}$ the Borel isomorphism between $[0,1]^{\omega}$ and $\C^{\omega}$, with $\phi$ applied entry-wise.
Let $o\colon\N\to\N^2$ be any bijection, and define a homeomorphism $\tilde{o}:\C^{\omega}\to\C$ by
\begin{equation*}
\tilde{o}\big((x^{(i)})_{i=1}^{\infty}\big) = \big( x^{(o(i)_1)}_{o(i)_2} \big)_{i=1}^{\infty}
\end{equation*}
Then, define $g\colon \X\to\C$ by $g = \tilde{o} \circ \phi^{\omega} \circ \psi$.
This is a Borel injection from $\X$ to $\C$.
Note that $\X,\C$ are standard Borel spaces, $\psi(\X)$ is a Borel subset, and $\tilde{o} \circ \phi^{\omega}$ is a Borel injection.
So \cite[Corollary 15.2]{kechris2012classical} implies that $g(\X)$ is a Borel subset of $\C$, and that $\tilde{o} \circ \phi^{\omega}$ is a Borel isomorphism between $\psi(\X)$ and $g(\X)$.
It follows that $g$ is a Borel isomorphism between $\X$ and $g(\X)$.

Fix any point $c_0 \in g(\X)$.
View $\C$ as a subspace of $\Omega^{\omega}$ and define the Borel surjection $h\colon\Omega^{\omega}\to g(\X)$
\begin{equation*}
h(c) = \begin{cases}
c, \quad c \in g(\X) \\
c_0, \quad c \notin g(\X)
\end{cases}
\end{equation*}
Then, $f = g^{-1} \circ h$ is a Borel surjection from $\Omega^{\omega}$ to $\X$.

\end{proof}

\section{Properties of Divergences}

\begin{proposition}
\label{prop. sampler loss minimizer}
Let $\X$ be a countable set and $P$ be a finite measure over $\X$ with $P(\X)>0$.
For any integer $n \geq 1$,
the following function over $\PS(\X)$
\begin{equation*}
L^{(n)}(Q) = \E [ - \log p_n ], \quad p_n = \frac{1}{n} \sum_{i=1}^n \frac{P(Z^{(i)})}{Q(Z^{(i)})}, \quad \{ Z^{(i)} \}_{i=1}^n \iidsample Q 
\end{equation*}
has the unique minimizer $Q_* := P/P(\X)$.
\end{proposition}

\begin{proof}
Replace $P$ by $Q_*$ in $p_n$.
This only affects $L^{(n)}$ by an additive constant, so the minimizers are unaffected.
Since $-\log$ is strictly convex, by Jensen's inequality
\begin{align*}
L^{(n)}(Q) &\geq -\log \E[p_n] = -\log Q_*(\sprt Q) \geq 0
\end{align*}
Since $L^{(n)}(Q_*)=0$, one minimizer is $Q_*$.
Let $Q$ be any minimizer.
To make the first inequality an equality, Jensen's inequality implies that the random variable $p_n$ is almost surely constant, and thus $Q = c Q_*$ over $\sprt Q$ for some constant $c>0$.
To make the second inequality an equality, $Q$ must satisfy $\sprt Q_* \subseteq \sprt Q$.
Hence, $c=1$ and $Q=Q_*$.
\end{proof}

\begin{lemma}
\label{lemma. KL event}
For any measurable space $\Omega$, let $\PS(\Omega)$ be the space of probability distributions over $\Omega$.
For any measurable subset $E\subseteq \Omega$ and any $P_*,P \in \PS(\Omega)$, if $P_*(E)>0$, then
\begin{align*}
\KL(P_*\|P) &\geq P_*(E) ~ \KL\big( P_*|_E \big\| P|_E \big) \\
\KL_{0.01}(P_*\|P) &\geq P_*(E) ~ \KL_{0.01}\big( P_*|_E \big\| P|_E \big)
\end{align*}
where $P|_E$ denotes the conditional distribution of $P$ conditioned on $E$, and $\KL_{0.01}$ is defined by (\ref{eq. KL 0.01}).
\end{lemma}

\begin{proof}
First, consider the plain $\KL$ divergence.
If $P(E)=0$, then $\KL(P_*\|P)=\infty$, so the inequality trivially holds.
So we can assume that $P(E)>0$.
For any $p\in[0,1]$, denote by bin$(p)$ the Bernoulli distribution with probability $p$.
Then,
\begin{align*}
\KL(P_*\|P) &= \int \log \frac{P_*(x)}{P(x)} dP_*(x) = \int_E \log \frac{P_*|_E(x)P_*(E)}{P|_E(x)P(E)} dP_*(x) + \int_{E^C} \log \frac{P_*|_{E^C}(x)P_*(E^C)}{P|_{E^C}(x)P(E^C)} dP_*(x)\\
&= \KL\big(\text{bin}(P_*(E)) \big\| \text{bin}(P(E)) \big) + P_*(E) ~ \KL\big(P_*|_E \big\| P|_E \big) + P_*(E^C) ~ \KL\big( P_*|_{E^C} \big\| P|_{E^C} \big) \\
&\geq P_*(E) ~ \KL\big( P_*|_E \big\| P|_E \big)
\end{align*}
Next, consider $\KL_{0.01}$.
\begin{align*}
\KL_{0.01}(P_*\|P) &= \inf_Q \Big\{ \KL(P_*\|Q) \Bigm| Q\in\PS(\Omega) , ~ \|\log(Q/P)\|_{\infty} \leq 0.01 \Big\}\\
&\geq \inf_Q \Big\{ P_*(E) ~ \KL\big( P_*|_E \big\| Q|_E \big) \Bigm| Q\in\PS(\Omega) , ~ \|\log(Q/P)\|_{\infty} \leq 0.01 \Big\} \\
&\geq \inf_Q \Big\{ P_*(E) ~ \KL\big( P_*|_E \big\| Q|_E \big) \Bigm| Q\in\PS(\Omega) , ~ \|\log\big( Q|_E / P|_E \big)\|_{\infty} \leq 0.01 \Big\} \\
&= P_*(E) ~ \KL_{0.01} \big( P_*|_E \big\| P|_E \big)
\end{align*}
\end{proof}

\begin{lemma}
\label{lemma. KL mapping}
For any measureable spaces $\Omega$ and $S$, let $B$ be a conditional distribution such that for any $\omega \in \Omega$, $B_{\omega}$ is a distribution over $S$ and the mapping $\omega\mapsto B_{\omega}$ is measureable.
For any distributions $P$ over $\Omega$, denote by $B\# P$ the distribution $\int B_{\omega} dP(\omega)$.
Then, for any distributions $P_*,P$ over $\Omega$,
\begin{align*}
\KL(P_*\|P) &\geq \KL(B\#P_*\|B\#P)\\
\KL_{0.01}(P_*\|P) &\geq \KL_{0.01}(B\#P_*\|B\#P)
\end{align*}
\end{lemma}

\begin{proof}
Given any non-negative finite measures $a,b$ over $\Omega$, the inequality
\begin{equation*}
\KL\Big( \frac{a}{\int a} \Big\| \frac{b}{\int b} \Big) \geq 0
\end{equation*}
implies that
\begin{equation*}
\int a \log \frac{a}{b} \geq \int a ~\log \frac{\int a}{\int b}
\end{equation*}
For any $s \in S$, setting $a(\omega) = B_{\omega}(s)P_*(\omega)$ and $b(\omega) = B_{\omega}(s)P(\omega)$, we obtain
\begin{align*}
\KL(B\#P_*\|B\#P) &= \int_S \int_{\Omega} B_{\omega} dP_*(\omega) ~\log \frac{\int_{\Omega} B_{\omega} dP_*(\omega)}{\int_{\Omega} B_{\omega} dP(\omega)} \\
&\leq \int_S \int_{\Omega} B_{\omega} dP_*(\omega) \log \frac{B_{\omega} P_*(\omega)}{B_{\omega} P(\omega)} \\
&= \int_{\Omega} \Big(\int_S B_{\omega}\Big) dP_*(\omega) \log \frac{P_*(\omega)}{P(\omega)}\\
&= \KL(P_*\|P)
\end{align*}
Next, consider $\KL_{0.01}$.
For any $P,Q \in \PS(\Omega)$, if $Q\leq C\cdot P$ (or $Q\geq C\cdot P$) everywhere in $\Omega$ for some constant $C>0$, then $B\#Q \leq C \cdot B\#P$ (or $\geq$) everywhere in $S$.
It follows that
$$\Big[\inf \frac{B\#Q}{B\#P}, ~\sup \frac{B\#Q}{B\#P}\Big] \subseteq \Big[\inf \frac{Q}{P}, ~\sup \frac{Q}{P}\Big]$$
so $\|\log B\#Q/ B\#P \|_{\infty} \leq \|\log Q/P\|_{\infty}$.
Then,
\begin{align*}
\KL_{0.01}(P_*\|P) &= \inf_Q \Big\{ \KL(P_*\|Q) \Bigm| Q\in\PS(\Omega) , ~ \|\log(Q/P)\|_{\infty} \leq 0.01 \Big\}\\
&\geq \inf_Q \Big\{ \KL\big( B\#P_* \big\| B\#Q \big) \Bigm| Q\in\PS(\Omega) , ~ \|\log(Q/P)\|_{\infty} \leq 0.01 \Big\} \\
&\geq \inf_Q \Big\{ \KL\big( B\#P_* \big\| B\#Q \big) \Bigm| Q\in\PS(\Omega) , ~ \|\log\big( B\#Q / B\#P \big)\|_{\infty} \leq 0.01 \Big\} \\
&= \KL_{0.01} \big( B\#P_* \big\| B\#P \big)
\end{align*}
\end{proof}

\begin{lemma}
\label{lemma. chi-square}
Let $\chi^2(p\|q) = \int (d p/d q)^2 dq-1$ be the chi-square divergence.
For any measureable spaces $A,B$ and distributions $p,q \in \PS(A\times B)$,
\begin{equation*}
\chi^2(p\|q) \geq \chi^2\Big(\int_B p \Bigm\| \int_B q \Big)
\end{equation*}
\end{lemma}
\begin{proof}
By Jensen's inequality, the chi-square divergence is always nonnegative.
Then,
\begin{align*}
\chi^2(p\|q) &= \int \frac{d p}{d q}(a)^2 \int\frac{d p(\cdot|a)}{d q(\cdot|a)}(b)^2 dq(b|a) dq(a) - 1 \\
&= \int \frac{d p}{d q}(a)^2 \Big( \chi^2 \big( p(\cdot|a) \big\| q(\cdot|a) \big) + 1 \Big) dq(a) - 1 \\
&\geq \int \frac{d p}{d q}(a)^2 \cdot 1 ~ dq(a) - 1 \\
&= \chi^2\Big(\int_B p \Bigm\| \int_B q \Big)
\end{align*}
\end{proof}

\section{Properties of Projections}
\label{appendix. projection}

\begin{proof}[Proof of Proposition \ref{prop. inverse projection}]
Recall that the product topologies over $\Sigma^{\omega}$ and $\Omega^{\omega}$ are induced by the metric $d_{\omega}(x,y)=2^{-\inf\{i\mid x_i\neq y_i\}}$.
Fix any $x\in\Sigma^*$, we have
$$\inf \big\{ d_{\omega}(x_1, x_2) \bigm| x_1\in [x], x_2 \notin [x] \big\} \geq 2^{-|x|}$$
By Tychonoff's theorem, $\Omega^{\omega}$ is compact, and thus the closed subset $\dom(\proj)$ is also compact.
It follows that $\proj$ is uniformly continuous, and there exists some $T$ such that
\begin{equation*}
\forall z,z' \in \dom(\proj), \quad d_{\omega}(z,z') \leq 2^{-T} ~ \to ~ d_{\omega}\big(\proj(z),\proj(z')\big) < 2^{-|x|}
\end{equation*}
Thus, for any $z \in \Omega^T$, the set $[z] \cap \dom(\proj)$ is either contained in $\proj^{-1}([x])$ or disjoint from it.

Consider the tree with depth $T$ (with the root level being depth $0$) and with $|\Omega|$ children for each non-leave node.
So it can be identified with the set $\Omega^{\leq T}$.
For each $z \in \Omega^T$, label the corresponding leave by
\begin{itemize}
\item $1$, if $\varnothing \subsetneq [z] \cap \dom(\proj) \subseteq \proj^{-1}([x])$.
\item $0$, if $[z] \cap \dom(\proj)=\varnothing$.
\item $-1$, else.
\end{itemize}
We call this the step $T$.
Then, recursively for $t=T-1, \dots 0$, label each node at depth $t$ by
\begin{itemize}
\item $1$, if all of its children are $\geq 0$ and at least one child is $1$.
Then, relabel all of its children by $-1$.
\item $0$, if all children are $0$.
\item $-1$, else.
\end{itemize}
For each step $t=T,\dots 0$, denote by $I(t) \subseteq \Omega^T$ the subset of nodes that are currently labeled by $1$.
Denote the resulting set by $I_x^* = I(0)$.
One can show by induction on $t=T, \dots 0$ that $I(t)$ is non-empty, finite, prefix-free and satisfies (\ref{eq. disjoint union of cylinders}), and $[z]\cap \dom(\proj) \neq \varnothing$ for each $z \in I(t)$.
Thus, the three properties in Proposition \ref{prop. inverse projection} are satisfied by $I_x^*$.

Suppose $I_x \subseteq \Omega^*$ is any subset that also satisfies these three properties.
For any $z \in I_x$, the node $z_{\leq \min(T,|z|)}$ in the tree has been labeled by $1$ in the above process, so either this node or one of its ancestors belongs to $I_x^*$.
It follows that $I_x^* \preceq I_x$ and thus $I_x^*$ is the unique minimal set.

Next, consider the function $\fip$.
For any $x,x'\in \Sigma^*$ with $x\sqsubseteq x'$, and for any $z' \in \fip(x')$, since $\proj([z']) \subseteq [x'] \subseteq [x]$, we have $[z'] \subseteq \bigsqcup_{z\in \fip(x)}[z]$.
Since cylinders either contain one another or are disjoint, $[z']$ must be contained in $[z]$ for some $z\in \fip(x)$, and thus $z' \sqsubseteq z$.
This proves monotonicity.

For any $x\in \Sigma^{\omega}$,
denote $S_t = \bigsqcup_{z\in \fip(x_{\leq t})} [z] \cap \dom(\proj)$.
Then, $S_0 \supseteq S_1 \supseteq S_2 \supseteq \dots$ and $\proj^{-1}(x) \subseteq S_t$ for all $t$.
Thus, $\proj^{-1}(x) \subseteq \bigcap_{t=0}^{\infty} S_t$.
Meanwhile, for any $z \in \bigcap_{t=0}^{\infty} S_t$, since $\proj(z) \in [x_{\leq t}]$ for all $t$, we have $\proj(z)_t = x_t$ for all $t$, and thus $\proj(z) = x$.
Hence, $\bigcap_{t=0}^{\infty} S_t \subseteq \proj^{-1}(x)$.
This proves convergence.
\end{proof}

\begin{proof}[Proof of Lemma \ref{lemma. segments prefix-free}]

Regarding property 1, the set $\nsm_x(z,y)$ is finite and prefix-free since
\begin{equation*}
\big\{ zs \bigm| s \in \nsm_x(z,y) \big\} \subseteq \fip(xy)
\end{equation*}
and $\fip(xy)$ is finite and prefix-free by Proposition \ref{prop. inverse projection}.

Regarding property 2,
recall that by the construction of Proposition \ref{prop. inverse projection},
\begin{equation*}
\forall x \in \Sigma^*, ~\forall z \in \fip(x), \quad [z] \cap \dom(\proj) \neq \varnothing
\end{equation*}
Fix any $x$ and $z\in\fip(x)$, this condition implies that
\begin{equation}
\label{eq. segment intersect dom proj}
\forall y \in \Sigma^*, ~\forall s \in \nsm_x(z,y), \quad [zs] \cap \dom(\proj) \neq \varnothing
\end{equation}

Fix any prefix-free subset $S\subseteq \Sigma^*$.
For any $y,y'\in\Sigma^*$ ($y\neq y'$),
\begin{equation*}
\forall s \in \nsm_x(z,y) \cap \nsm_x(z,y'), \quad \proj\big([zs]\cap\dom(\proj)) \subseteq [xy] \cap [xy'] = \varnothing
\end{equation*}
so $[zs] \cap \dom(\proj) = \varnothing$, which by (\ref{eq. segment intersect dom proj}) implies that such $s$ cannot exist.
Hence, the sets $\nsm_x(z,y),y\in S$ are disjoint.

Consider any segments $s \in \nsm_x(z,y)$ and $s'\in\nsm_x(z,y')$.
Suppose $s \sqsubseteq s'$, then $\proj([zs'] \cap \dom(\proj)) \subseteq [xy] \cap [xy'] = \varnothing$, and thus $[zs'] \cap \dom(\proj) = \varnothing$.
Again (\ref{eq. segment intersect dom proj}) implies that such $s'$ cannot exist and $\nsm(z,y')=\varnothing$.
Hence, the set $\bigsqcup_{y\in\Sigma} \nsm_x(z,y)$ is prefix-free.

Regarding property 3, it suffices to consider the case of two segments ($k=2$), $x=x^{(1)}x^{(2)}$.
For $k>2$, the argument can be applied recursively to the pairs $(x^{(<k)},x^{(k)})$, $(x^{(<k-1)},x^{(<k)})$, $(x^{(<k-2)},x^{(<k-1)})$, $\dots$.
By monotonicity (\ref{eq. monotone}), there exists some $z^{(1)} \in \fip(x^{(1)})$ such that $z^{(1)} \sqsubseteq z$.
Since $\fip(x^{(1)})$ is prefix-free, this $z^{(1)}$ is unique.
Set $z^{(2)} = \big(z^{(1)}\big)^{-1} z$.
Then, by definition (\ref{eq. next-segment set}), $z^{(2)} \in \nsm_{x^{(1)}}(z^{(1)},x^{(2)})$.
Hence, $z=z^{(1)}z^{(2)}$ is a unique decomposition.

Regarding property 4, given that $\proj(w) \in \proj([z]) \subseteq [x]$, we have $\proj(w)_{\leq |x|+1} = xa$ for some $a\in\Sigma$.
Thus, $w \in \bigsqcup_{s\in\nsm_x(z,a)} [zs]$.
So $zs \sqsubseteq w$ for some $s\in\nsm_x(z,a)$.
By property 2, the set $\bigsqcup_{a\in\nsm(z,a)}$ is prefix-free, so $s$ is unique.
\end{proof}

\begin{proof}[Proof of Lemma \ref{lemma. prefix projection}]
$\proj^{<\omega}$ is monotone in the sense that for any $z',z''\in\Omega^*$, if $z' \sqsubseteq z''$, then $\proj^{<\omega}(z') \sqsubseteq \proj^{<\omega}(z'')$.
Fix any $z \in \dom(\proj)$.
Thus, $\proj^{<\omega}(z_{\leq 1}) \sqsubseteq \proj^{<\omega}(z_{\leq 2}) \sqsubseteq \dots$.
So the limiting sequence $x := \bigcup_{t=1}^{\infty} \proj^{<\omega}(z_{\leq t}) \in \Sigma^{\leq\omega}$ is well-defined.
Since $\proj$ is continuous around $z$,
there exists some $T(t)<\infty$ for any $t$,
such that $\proj([z_{\leq {T(t)}}]) \subseteq [\proj(z)_{\leq t}]$.
Thus,
$$|x| \geq \lim_{t\to\infty} \big| \proj^{<\omega}(z_{\leq T_t}) \big| = \lim_{t\to\infty} \big| \text{gcd}\big( \proj( [z_{\leq T_t}] ) \big) \big| \geq \lim_{t\to\infty} t = \infty$$
So $x \in \Sigma^{\omega}$.
Meanwhile, $\proj([z_{\leq {T_t}}]) \subseteq [\proj(z)_{\leq t}]$ implies that $x_{\leq t} = \proj(z)_{\leq t}$ for each $t$.
So $x = \proj(z)$.
\end{proof}

\begin{proof}[Proof of Theorem \ref{thm. conditional latent distribution}]
For any $x \in \Sigma^*$, equation (\ref{eq. disjoint union of cylinders}) implies that
\begin{align*}
(\proj\#P)^{\leq |x|}(x) &= P\big(\proj^{-1}([x])\big) = P\Big(\bigsqcup_{z\in\fip(x)}[z] \cap \dom(\proj) \Big)\\
&= \sum_{z\in\fip(x)} P([z]) = \sum_{z\in\fip(x)} P^{\leq |z|}(z)
\end{align*}
Similarly, for any $y\in\Sigma^*$,
\begin{align*}
(\proj\#P)^{\leq |x|+|y|}(y|x) &= \frac{(\proj\#P)^{\leq |x|+|y|}(xy)}{(\proj\#P)^{\leq |x|}(x)} = \frac{\sum_{z\in\fip(xy)} P^{\leq |z|}(z)}{\sum_{z\in\fip(x)} P^{\leq |z|}(z)}
\end{align*}
where
\begin{equation*}
\sum_{z'\in\fip(xy)} P^{\leq |z'|}(z') = \sum_{z\in\fip(x)} \sum_{s\in\nsm_x(z,y)} P^{\leq |z|+|s|}(zs) = \sum_{z\in\fip(x)} P^{|z|}(z) \sum_{s\in\nsm_x(z,y)} P^{\leq |z|+|s|}(s|z)
\end{equation*}
Hence,
\begin{equation*}
(\proj\#P)^{\leq |x|+|y|}(y|x) = \sum_{z\in\fip(x)} \frac{P^{|z|}(z)}{(\proj\#P)^{\leq |x|}(x)} \Big( \sum_{s\in\nsm_x(z,y)} P^{\leq |z|+|s|}(s|z) \Big)
\end{equation*}
\end{proof}

\begin{proof}[Proof of Proposition \ref{prop. segment-wise partial function}]
Since each $S_t \subseteq \Omega^{l_t}$, they are finite sets.
For each $t$, define
\begin{equation*}
S_{\leq t} = \bigcup_{s_1\in S_1} \dots \bigcup_{s_t\in S_t} [s_1 \dots s_t]
\end{equation*}
Since this is a finite union of cylinders (which are closed sets), $S_{\leq t}$ is closed.
Since $S = \bigcap_{t=1}^{\infty} S_{\leq t}$, it is also closed.

Since $f_t(s_t)$ is never the empty string, $\proj$ maps into $\Sigma^{\omega}$.
Let $d_{\omega}$ be the metric (\ref{eq. metric product topology}).
For any sequence $\{(s_t^{(i)})_{t=1}^{\infty}\}_{i=1}^n \subseteq S$ that converges in $d_{\omega}$ to some limit $(s_t)_{t=1}^{\infty} \in S$.
By definition of $d_{\omega}$, these sequences converge pointwise, so each segment $s_t^{(i)}$ eventually becomes fixed at $s_t$ as $i\to\infty$.
Thus, each $f_t(s_t^{(i)})$ eventually becomes $f_t(s_t)$,
and $\proj((s_t^{(i)})_{t=1}^{\infty})$ converges to $\proj((s_t)_{t=1}^{\infty})$ in $d_{\omega}$.
So $\proj$ is continuous over $S$.
\end{proof}

\section{Proof of the Separation Theorems}

\subsection{Part I}
\label{append. separation I}

To construct the HMM of Theorem \ref{thm. separation I}, we need a target function that is ``inherently sequential".
Let $\sigma_0=(1,2)(3,4)$ and $\sigma_1=(1,3,5)$ be permutations from the alternating group $A_5$ (using the cycle notation).
Define a function $F\colon\{0,1\}^*\to\{1,\dots 5\}$ by
\begin{equation}
\label{eq. A5 composition function}
F(i_1,\dots,i_t) = \sigma_{i_t} \circ \dots \circ \sigma_{i_1}(1)
\end{equation}
The group $A_5$ is chosen since it is non-solvable,
and it is a well-known result in circuit complexity theory that the wording problem of any non-solvable group is $\NC^1$-complete \cite{barrington1986bounded,straubing2012finite}.
Specifically, let $G$ be any non-solvable group and define the following set of sequences of group elements
\begin{equation*}
L_G = \big\{ (g_1, \dots g_n) \in G^* \bigm| g_n \circ \dots \circ g_1  = \id \big\}
\end{equation*}
then any $\NC^1$ language $L$ is strongly reducible to $L_G$, meaning that $L$ can be recognized by a constant-depth polynomial-size circuit family with NOT-gates, AND/OR-gates with fan-in 2, and gates that recognize $L_G$ \cite{straubing2012finite}.
As a straightforward corollary, computing $F$ also requires $\NC^1$ circuits.

\begin{lemma}
\label{lemma. A5 is NC1}
Assume that $\TC^0 \subsetneq \NC^1$.
The function $F$ cannot be computed by $\TC^0$ circuits.
\end{lemma}

\begin{proof}
Using the terminology of \cite{straubing2012finite}, given any subset (or language) $L \subseteq \{0,1\}^*$, denote by $\equiv_L$ the syntactic congruence over $\{0,1\}^*$ induced by $L$, namely for any $x,y\in\{0,1\}^*$, $x \equiv_L y$ if and only if
\begin{equation*}
\forall u, v \in \{0,1\}^*, \quad uxv \in L \leftrightarrow uyv \in L
\end{equation*}
and denote by $M(L)$ the syntactic monoid, namely the quotient $\{0,1\}^* / \equiv_L$.

Define $\mathfrak{F} \colon \{0,1\}^*\to A_5$ by $\mathfrak{F}(i_1,\dots,i_t) = \sigma_{i_t} \circ \dots \circ \sigma_{i_1}$ so that $F \equiv \mathfrak{F}(\cdot)(1)$.
Since $\sigma_0,\sigma_1$ generate $A_5$, the subset $\mathfrak{F}^{-1}(g)$ is nonempty for each $g\in A_5$.
Consider the language $L_F = F^{-1}(1)$.
Then, one can check that $x \equiv_{L_F} y$ if and only if $\mathfrak{F}(x) = \mathfrak{F}(y)$.
Specifically, for the ``only if" part, if $\mathfrak{F}(x) \neq \mathfrak{F}(y)$,
we can choose $k \in \{1,\dots 5\}$ and $z \in \{0,1\}^*$ such that $\mathfrak{F}(x)(k) \neq \mathfrak{F}(y)(k)$ and
$$1 = \mathfrak{F}(z)\big( \mathfrak{F}(x)(k) \big) = \mathfrak{F}(zx)(k)$$
so $zx \in L_F$.
Meanwhile, $\mathfrak{F}(zy)(k) = \mathfrak{F}(z)( \mathfrak{F}(y)(k) ) \neq 1$ since $\mathfrak{F}(z)$ is a bijection, so $zy \notin L$ and thus $x \not\equiv_{L_F} y$.
Hence, the syntactic monoid $M(L_F)$ is isomorphic to $A_5$.
By \cite[Theorem IX.1.5]{straubing2012finite}, $L_F$ is $\NC^1$-complete.

Suppose for contradiction that $F$ can be computed by $\TC^0$ circuits.
Then, the subset $F^{-1}(1) = L_F$ can be recognized by some $\TC^0$ circuit family, e.g.\@ append an output gate to each circuit of $F$ that outputs $1$ or $0$ if $F(x)=1$ or $\neq 1$.
It follows that all $\NC^1$ languages, being strongly reducible to the $\TC^0$ language $L_F$, can be recognized by $\TC^0$ circuits, a contradiction to $\TC^0 \subsetneq \NC^1$.
\end{proof}

\begin{proof}[Proof of Theorem \ref{thm. separation I}]
In terms of Definition \ref{def. HMM}, we define an HMM as follows.
Denote by $\sigma_2$ the group identity $\id$ of $A_5$, following the notations $\sigma_0,\sigma_1$ of (\ref{eq. A5 composition function}).
The latent vocabulary is $\Omega=\{1,2,3,4,5\}\times\{0,1,2\}$, with each state denoted by $z=(v,w)$.
The observable vocabulary is $\Sigma=\{1,2,3,4,5\} \cup \{0,-1\}$, with the last two elements representing $\sigma_0,\sigma_1$.
Regarding the transition matrix $A\in\R^{15\times 7}$, each row $A_{(v,w)}$ has three non-zero entries, $(\sigma_w(v),w')$ with $w'=0,1,2$, and their probabilities are $\frac{1}{4},\frac{1}{4},\frac{1}{2}$.
The observation matrix $B\in\R^{15\times 7}$ is deterministic, such that each row $B_{(v,w)}$ has only one nonzero entry.
If $w\in\{0,1\}$, then this entry is $-w$, else this entry is $v$.
The initial distribution $\lambda\in\R^{15}$ has three non-zero entries, $(1,w)$ with $w=0,1,2$, again with probabilities $\frac{1}{4},\frac{1}{4},\frac{1}{2}$.

In plain terms, this HMM iteratively applies random permutations to a hidden element.
At each step, it either reveals this element ($v$), or applies a new permutation and reveals this permutation ($w$), each case with $1/2$ probability.

Suppose for contradiction that inequality (\ref{eq. separation 1}) does not hold.
Then, there exists some configuration $(L,H,d_h,p)$ that achieves equality.
By Theorem 13 of \cite{chiang2025TC0}, any $f\in\TF(L,H,d_h,p)$ is equivalent to a \textsf{DLOGTIME}-uniform $\TC^0$ circuit family.
Thus, there exists some constant $d_f$ such that the minimum size of a threshold circuit with depth $\leq d_f$ that simulates $f$ over length-$t$ inputs is polynomial in $t$.
Note that this bound is uniform over $\TF(L,H,d_h,p)$:
Denote by $f=f_{\theta}$ the parametrization of transformers in $\TF(L,H,d_h,p)$, where $\theta$ is a $O(LH^2d_h^2)$-dimensional vector of $p$-bit numbers.
Since each entry of $\theta$ is only involved in at most $t$ floating-point multiplications or additions for inputs of length $t$, the mapping $(\theta,x) \mapsto f_{\theta}(x)$ can also be implemented by a \textsf{DLOGTIME}-uniform $\TC^0$ circuit family.
Denote its circuit depth upper bound by $d$ and its circuit sizes by a polynomial $s(t)$ of input length $t$.
Then, all $f \in \TF(L,H,d_h,p)$ are uniformly bounded in circuit depth by $d$ and in size by $s(t)$.

Denote by $\bar{s}(t)$ the minimum circuit size for any family of threshold circuits with depth upper bound $d$ that computes the function $F$ (\ref{eq. A5 composition function}).
Denote by $s_p$ the minimum size of a threshold circuit that computes the argmax of three floating point numbers with precision $p$ bits (this argmax is always unique in later construction, so we do not need to consider ties).
Lemma \ref{lemma. A5 is NC1} indicates that $\bar{s}$ must be superpolynomial, namely $\lim_{t\to\infty} \bar{s}(t) / t^k$ for all $k \in \N$.
Thus, there exists some $T$ such that $\bar{s}(T) > s(T) + s_p$.

By construction, $P_{A,B,\lambda}^{\leq T}(x)=4^{-T}$ for any $x\in\{0,-1\}^T$, and $P_{A,B,\lambda}^{\leq T+1}(\{1,\dots 5\} \mid x) = \frac{1}{2}$ for any $x\in\Sigma^T$.
Denote by $P|_S$ the conditional distribution of a distribution $P$ conditioned on an event $S$ with $P(S)>0$.
Define the event $E=\{x_{T+1}\geq 1\}$.
Consider the terms $D_t$ in (\ref{eq. semi-divergence}).
For any distribution $P\in\PS(\Sigma^{\omega})$,
\begin{align*}
D_{T+1}(P_{A,B,\lambda},P) &\geq \max_{x \in \{0,-1\}^T} 4^{-T} ~ \KL_{0.01}\big(P_{A,B,\lambda}^{\leq T+1}(\cdot|x) \big\| P^{\leq T+1}(\cdot|x) \big) \\
&\geq \max_{x \in \{0,-1\}^T} \frac{4^{-T}}{2} ~ \KL_{0.01}\big( P_{A,B,\lambda}^{\leq T+1}|_E (\cdot|x) \big\| P^{\leq T+1}|_E (\cdot|x) \big)
\end{align*}
where the second inequality follows from Lemma \ref{lemma. KL event}.
Note that for any $x \in \{0,-1\}^T$,
\begin{equation*}
P_{A,B,\lambda}^{\leq T+1}|_E (\cdot|x) = \delta_{F(-x_1,\dots -x_T)}
\end{equation*}
So for any probability vector $q \in \R^3$,
\begin{equation*}
\KL\big( P_{A,B,\lambda}^{\leq T+1}|_E (\cdot|x) \bigm\| q \big) = -\log q_{F(-x_1,\dots -x_T)}
\end{equation*}
Then,
\begin{align*}
\KL_{0.01}\big( P_{A,B,\lambda}^{\leq T+1}|_E (\cdot|x) \bigm\| P^{\leq T+1}|_E (\cdot|x) \big) \geq - \log P^{\leq T+1}|_E \big( F(-x_1,\dots -x_T) \bigm| x \big) - 0.01
\end{align*}
It follows that
\begin{align*}
\min_{i \in \{0,1\}^T} \log P^{\leq T+1}|_E \big( F(i_1,\dots i_T) \bigm| -i_1,\dots -i_T \big) \geq -2^{2T+1} D_{T+1}(P_{A,B,\lambda},P) - 0.01
\end{align*}

For any $0<\ep<2^{-(T+1)}$, let $f_{\ep} \in \TF(L,H,d_h,p)$ be a solution to $D(P_{A,B,\lambda},P_f)<\ep$.
By definition (\ref{eq. infinite divergence}), $D_{T+1} < 2^{T+1} \ep$.
Then,
\begin{align*}
\min_{i \in \{0,1\}^T} P_{f_{\ep}}^{\leq T+1}|_E \big( F(i_1,\dots i_T) \bigm| -i_1,\dots -i_T \big) > \exp \big( -2^{3T+2} \ep - 0.01 \big)
\end{align*}
Denote by $g_{\ep}$ the modification of $f_{\ep}$ that removes the rows corresponding to $\{1,\dots 5\}$ of the input embedding matrix and the columns corresponding to $\{0,-1\}$ of the LM head matrix.
Setting $\ep < 0.5 / 2^{3T+2}$, we obtain
\begin{equation}
\label{eq. prob > 0.6}
\min_{i\in\{0,1\}^T} P_{g_{\ep}}^{\leq T+1} \big(F_T(i_1,\dots,i_T) \bigm| i_1,\dots i_T\big) > 0.6
\end{equation}
Let $C_T$ be the threshold circuit corresponding to $\text{argmax} \circ g_{\ep}$ over length $T$ binary inputs.
Then, the size of $C_T$ is bounded by $s(T)+s_p < \bar{s}(T)$.
However, (\ref{eq. prob > 0.6}) indicates that $\text{argmax} \circ g_{\ep}$, and thus $C_T$, computes $F_T$, violating the minimality of $\bar{s}(T)$.
A contradiction.
Hence, (\ref{eq. separation 1}) must hold.
\end{proof}

\subsection{Part II}
\label{appendix. separation II}

\begin{proof}[Proof of Theorem \ref{thm. separation II}]
Following Definition \ref{def. HMM}, let $\Omega$ and $\Sigma$ be the hidden space and observation space.
Denote the hidden-observable joint distribution of HMM by
$$\prob_{A,B,\lambda} = \law\big( ((Z_t,X_t))_{t=1}^{\infty} \big) \in \PS\big((\Omega\times\Sigma)^{\omega}\big)$$
Let the vocabulary of Transformers be $(\Omega\times\Sigma) \cup \{\bos\}$.
As in Example \ref{ex. product}, define the projection by a total function $\proj\colon  (\Omega\times\Sigma)^{\omega}\to \Sigma^{\omega}$
\begin{equation*}
\proj\colon  \big( (z_t,x_t) \big)_{t=1}^{\infty} \mapsto (x_t)_{t=1}^{\infty}
\end{equation*}
Thus, $\proj$ is a simple projection and $\proj\#\prob_{A,B,\lambda}=P_{A,B,\lambda}$.

Set $d_h^*=|\Omega|+1$, while $p^*\geq 1$ is to be specified.
For any $d_h\geq d_h^*, p\geq p^*$, we construct a Transformer $f \in \TF(1,1,d_h,p)$ as follows.
Since there is only one head ($H=1$), the hidden dimension equals $d_h$.
Denote by $\R_p$ the set of floating-point numbers with $p$ bits, and the nearest rounding by $\text{cast}_p\colon \R\cup\{\pm\infty\}\to\R_p$.
We can assume that as $p\to\infty$, $\text{cast}_p$ converges pointwise to the identity function on $\R$, and also $\exp(\text{cast}_p(-\infty))$ converges to $0$.
Since $p\geq 1$, we require for simplicity that $\{0,1\}\subseteq \R_p$.

Since there is no layer ($L=0$), the function $f$ reduces to $W^I_{(z_{-1},x_{-1})} W^O$, where
$W^I\in\R_p^{(|\Omega|\cdot|\Sigma|+1)\times d_h}$ is the embedding matrix,
$W^O\in\R_p^{d_h \times |\Omega|\cdot|\Sigma|}$ is the LM head matrix, and $(z_{-1},x_{-1})$ is the last token of the input sequence.
(Even if $L>0$, we can set all parameters of the attention and MLP layers to zero.
Since Transformers use residual connection, these layers become identity functions.)

$W^I$ is constructed as follows.
Each row is a one-hot vector (i.e.\@ one entry is 1, all others are 0).
For each token $(z,x)\in\Omega\times\Sigma$, the row of $(z,x)$ is one-hot at the $z$-th entry, while the row of $\bos$ is one-hot at the $(|\Omega|+1)$-th entry.

For any vectors $v\in\R^{|\Omega|},w\in\R^{|\Sigma|}$, define the outer product $v\otimes w$ by a $\R^{|\Omega|\cdot |\Sigma|}$ vector whose $(|\Sigma|\cdot i+j)$-th entry is $v_i w_j$.
Define the matrices $P^O = \R^{(|\Omega|+1)\times |\Omega|\cdot|\Sigma|}$ and $W^O$ as follows:
For each $z \in \Omega$,
\begin{align*}
P^O_z &= A_z \otimes (A_z B), \quad P^O_{|\Omega|+1} = \lambda \otimes (\lambda B)\\
W^O_{\leq |\Omega|+1} &= \text{cast}_p\circ\log(P^O), \quad W^O_{> |\Omega|+1} \equiv 0
\end{align*}
where $\text{cast}_p$ and $\log$ are applied entrywise, and the initial distribution $\lambda$ is used as a row vector.
It follows that
$$\lim_{p\to\infty} \big\| P^O - \text{softmax}(W^O) \|_{\infty} = 0$$
where softmax is applied row-wise.
Thus, we can choose $p_*$ such that
$$\sup_{p\geq p_*} \max_{i=1,\dots |\Omega|+1} \KL_{0.01}\big( P^O_i \bigm\| \text{softmax}(W^O_i) \big) = 0$$

Lemma \ref{lemma. KL mapping} implies that
\begin{align*}
\sup_{t\geq 2} &~ D_t(P_{A,B,\lambda},\proj\#P_f) \leq \sup_{t\geq 2} D_t(\prob_{A,B,\lambda},P_f) \\
&\leq \sup_{t\geq 2} \max_{y \in (\Omega\times\Sigma)^{t-1}} \KL_{0.01} \big( \prob_{A,B,\lambda}^{\leq t} (\cdot|y ) \bigm\| P_f^{\leq t} (\cdot | y ) \big)\\
&= \max_{(z,x) \in \Omega\times\Sigma} \KL_{0.01} \big( \prob^{\leq 1}_{A,B,\lambda}((z,x)) \bigm\| P_f^{\leq 1}((z,x)) \big)\\
&= \max_{z\in\Omega} ~\KL_{0.01}(P^O_z\|W^O_z)\\
&= 0
\end{align*}
Similarly, for $t=1$,
\begin{equation*}
D_1(P_{A,B,\lambda},\proj\#P_f) \leq D_1(\prob_{A,B,\lambda},P_f) = \KL_{0.01}(P^O_{|\Omega|+1}\|W^O_{|\Omega|+1}) = 0
\end{equation*}
Hence, $D(P_{A,B,\lambda},\proj\#P_f) = 0$.



\end{proof}

\subsection{Part III}
\label{appendix. separation III}

\begin{proof}[Proof of Theorem \ref{thm. separation III}]
The proof is similar to the proof of Theorem \ref{thm. separation I} (Appendix \ref{append. separation I}), except that a slightly different circuit is constructed to simulate $F$.
For convenience, \textsf{DLOGTIME}-uniform $\TC^0$ is abbreviated into uniform $\TC^0$.

Again, suppose for contradiction that equality can be achieved in (\ref{eq. separation III}) for some polynomially simple projection $\proj$ and some configuration $(L,H,d_h,p)$.
Let $s_t = O(\poly(t))$ be the uniform upper bound on circuit sizes for all $f\in\TF(L,H,d_h,p)$ from the proof of Theorem \ref{thm. separation I}.
Let $\{C^{\proj}_t\}_{t=0}^{\infty}$ be a uniform $\TC^0$ nondeterministic circuit family with variable-length output that simulates $\fip$.
Denote by $l_t$ and $a_t$ the number of output tokens and auxiliary bits of $C^{\proj}_t$.
Since $\proj$ is a simple projection, $l_t = O(\poly(t))$, and since $\proj$ is polynomially simple, $a_t = O(\log t)$.

For any $f\in\TF(L,H,d_h,p)$, let $\{C^f_t\}_{t=0}^{\infty}$ be a uniform $\TC^0$ circuit family that simulates $f$.
Let $P_f$ be the parametrized distribution from Definition \ref{def. softmax with masking}.
A new circuit family $\{C^{f,\text{sum}}_t\}$ that is uniform $\TC^0$ can be constructed from $\{C^f_t\}$ that simulate the following function
\begin{equation*}
\forall z\in \Omega^*, \quad z \mapsto \log P_f^{\leq |z|}(z) = \sum_{t=1}^{|z|} \log \text{softmax}\big(f(z_{\leq t})\big)_{z_t} - \infty \cdot \mathbbm{1}_{z\notin\dom(\proj^{<\omega})}
\end{equation*}
since logsoftmax and iterated addition can be implemented by uniform $\TC^0$ circuits \cite{chiang2025TC0},
and since the domain $\dom(\proj^{<\omega})$ is uniform $\TC^0$ by Definition \ref{def. simple projection}.

Then, construct a circuit family $\{C^{f,\proj}_t\}$ that computes $\log (\proj\#P_f)^{\leq t}(\cdot|x_{<t})$ as follows.
Although much computation (and thus circuit size) can be saved by using the identity (\ref{eq. latent conditional}) of Theorem \ref{thm. conditional latent distribution}, we use brute-force computation to simplify our arguments.
Fix any $t$, denote the variable input by $x \in \Sigma^t$,
\begin{enumerate}
\item The circuit first splits into $|\Sigma|$ branches.
For $i=1,\dots |\Sigma|$, the $i$-th branch gets the input $y=x_{<t}\sigma_i$, where $\sigma_i$ is the $i$-th element of $\Sigma$.

\item Within each branch, we enumerate $2^{a_t}$ copies of $C^{\proj}_t$ and provide each with a unique configuration of auxiliary bits $\in \{0,1\}^{a_t}$.
Each copy of $C^{\proj}_t$ is followed by a uniform $\TC^0$ circuit $C^{\blank}_{l_t}$ that accepts $l_t$ input tokens and moves all the $\blank$ tokens to the end.
Then, the outputs of these $2^{a_t}$ branches are sent to a uniform $\TC^0$ circuit that detects duplicate outputs, marking the $i$-th output with $1$ if it differs from the first $i-1$ outputs and otherwise with $0$.
A weighted sum will be performed later over these $2^{a_t}$ branches with those $0,1$ weights, so effectively a deduplication is performed.
Now, the set $\fip(y)$ is obtained.

\item Next, each of the $2^{a_t}$ branches is followed by $l_t+1$ circuits, $C^{f,\text{sum}}_0,\dots C^{f,\text{sum}}_{l_t}$, such that if the input has $l$ non-blank tokens, only the circuit $C^{f,\text{sum}}_l$ is used to process this input.
Such routing can be similarly performed using weighted summation.
Then, a uniform $\TC^0$ circuit performs logsumexp on the outputs of these $2^{a_t}$ branches, ignoring the branches marked as duplicates.
By identity (\ref{eq. latent marginal}), we have obtained
\begin{equation*}
\text{logsumexp} \big\{ \log P^{\leq |z|}(z) \bigm| z\in\proj^{-1}_{\omega}(y) \big\} = \log \sum_{z\in\proj^{-1}_{\omega}(y)} P^{\leq |z|}(z) = \log (\proj\#P_f)^{\leq |y|}(y)
\end{equation*}

\item Performing logsumexp over these $|\Sigma|$ branches produces $\log (\proj\#P_f)^{<t}(x_{<t})$.
Subtracting it obtains the $|\Sigma|$-dimensional vector $\log (\proj\#P_f)^{\leq t}(\cdot|x_{<t})$
\end{enumerate}
By construction, $\{C^{f,\proj}_t\}$ is uniform $\TC^0$, as there are only $|\Sigma| 2^{a_t} (l_t+1)=O(\poly(t))$ branches.
Since the polynomial size bound $s_t$ is uniform for all $f \in \TF(L,H,d_h,p)$, the circuit size of $\{C^{f,\proj}_t\}$ is also uniformly polynomial over $\TF(L,H,d_h,p)$.

It follows that the rest of the proof of Theorem \ref{thm. separation I} can be applied to conclude this proof.
\end{proof}

\section{Properties of Decoding}

\begin{proof}[Proof of Proposition \ref{prop. decode error bound}]
Since $\sprt Q(\cdot|x) \subseteq \sprt Q_*(\cdot|x)$, it happens almost surely that $Q_*(Z^{(0)}|x)>0$ and thus
$$P([Z^{(0)}]) = Q_*(Z^{(0)}|x) \cdot (\proj\#P)^{\leq |x|}(x) >0$$
So the conditional distribution $P(\cdot|[Z^{(0)}]) = P|_{[Z^{(0)}]} / P([Z^{(0)}])$ is well-defined.
It follows that the random variables $Z^{(\omega)}$ and $X^{(\omega)}$ are well-defined.

Define a conditional distribution $B\colon \fip(x)\to\PS(\Sigma^{\omega})$ as follows:
\begin{equation*}
\forall z \in \fip(x), \quad B_z := \proj\# P(\cdot|[z])
\end{equation*}
Since $\fip(x)$ is a finite set, $B$ is trivially a measureable function.
Note that
\begin{equation*}
(\proj\#P)(\cdot|x) = \law(Z), \quad Z\sim P(\cdot|[Z']), \quad Z' \sim Q_*(\cdot|x)
\end{equation*}
So
\begin{equation*}
B\# Q_*(\cdot|x) = \int B_z dQ_*(z|x) = (\proj\#P)(\cdot|x)
\end{equation*}
Similarly, $B\# Q(\cdot|x) = \law(X^{(\omega)})$.
Then, Lemma \ref{lemma. KL mapping} concludes the proof.
\end{proof}

\begin{proof}[Proof of Proposition \ref{prop. stream decode equivalence}]
Denote $Z^{\omega} = \bigcup_{t=1}^{\infty} Z^{(t)}$.
It happens almost surely that $Z^{(\omega)} \in \sprt P \subseteq \dom(\proj)$.
Thus, $[Z^{(t)}] \cap \dom(\proj) \neq \varnothing$ and $Z^{(t)} \in \dom(\proj^{<\omega})$ for all $t$.
So $X^{(t)}$ are well-defined.


By Lemma \ref{lemma. prefix projection},
\begin{equation*}
\bigcup_{t=1}^{\infty} X^{(t)} = \bigcup_{t=1}^{\infty} \proj^{<\omega}(Z^{(t)}) = \proj(Z^{(\omega)}) = X^{(\omega)}
\end{equation*}
and thus this limit $\in\Sigma^{\omega}$.
\end{proof}

\begin{lemma}
\label{lemma. TV inequality}
For any measureable spaces $\X,\mathcal{Y}$ and probability measures $P,Q\in\PS(\X\times \mathcal{Y})$,
the total variation distance satisfies
\begin{equation*}
\text{TV}(P,Q) \geq TV\Big(\int_{\mathcal{Y}} P, \int_{\mathcal{Y}} Q  \Big)
\end{equation*}
Also, for any measureable function $G\colon \X\to\PS(\mathcal{Y})$ (i.e.\@ conditional distribution) and $P,Q \in \PS(\X)$,
\begin{equation*}
\text{TV}(P,Q) \geq \text{TV}\Big(\int G(\cdot|x)dP(x), \int G(\cdot|x)dQ(x)\Big)
\end{equation*}
In particular, for any measureable function $f\colon \X\to\mathcal{Y}$ and $P,Q \in \PS(\X)$,
\begin{equation*}
\text{TV}(P,Q) \geq \text{TV}(f\#P,f\#Q)
\end{equation*}
Furthermore, if $\X$ is a countable set, then for any $P,Q\in\PS(\X)$
\begin{equation*}
\text{TV}(P,Q) \leq \frac{1}{2} \|P-Q\|_{L^2}
\end{equation*}
\end{lemma}

\begin{proof}
Denote the $\sigma$-field of $\X$ by $\sigma(\X)$.
For the first inequality,
\begin{align*}
\text{TV}(P,Q) &= \sup_{A \in \sigma(\X\times\mathcal{Y})} \big| P(A)-Q(A) \big| \geq \sup_{A \in \sigma(\X)} \big| P(A\times Y)-Q(A\times\mathcal{Y}) \big| = TV\Big(\int_{\mathcal{Y}} P, \int_{\mathcal{Y}} Q  \Big)
\end{align*}
For the second inequality,
\begin{align*}
\text{TV}(P,Q) &= \sup_{A \in \sigma(\X)} \big| P(A)-Q(A) \big| \geq \sup_{A \in \sigma(\mathcal{Y})} \Big| \int G(A|x)dP(x) - \int G(A|x) dQ(x) \Big| = TV(f\#P, f\#Q)
\end{align*}
For the last inequality, using Jensen's inequality we have
\begin{equation*}
\text{TV}(P,Q) = \frac{1}{2} \| P-Q \|_{L^1} \leq \frac{1}{2} \| P-Q \|_{L^2}
\end{equation*}
\end{proof}

\begin{lemma}
\label{lemma. one-hot estimator}
Let $\X$ be a countable set and $p,q \in \PS(\X)$ be probability distributions such that $p \ll q$
(i.e.\@ $q_x >0$ if $p_x>0$ for all $x\in\X$).
Define the random distribution
\begin{equation*}
p_n = \frac{\sum_{i=1}^n w_i \delta_{X_i}}{\sum_{i=1}^n w_i}, \quad w_i = \frac{p_{X_i}}{q_{X_i}}, \quad X_1, \dots X_n \iidsample q
\end{equation*}
Then,
\begin{equation*}
\E\big[ \| p - p_n \|^2_{L^2} \big] \leq \frac{24\chi^2(p\|q) + 8}{n}
\end{equation*}
If $p=q$, then
\begin{equation*}
\E\big[ \| p - p_n \|^2_{L^2} \big] = \frac{1-\|p\|^2_{L^2}}{n}
\end{equation*}
\end{lemma}

\begin{proof}
Denote $a_n = \frac{1}{n}\sum_{i=1}^n w_i$, $\Delta^i =  w_i (p -\delta_{X_i})$ and $\Delta_n = \frac{1}{n} \sum_{i=1}^n \Delta^i$.
Then, $p - p_n = \Delta_n / a_n$ and
\begin{equation*}
\E[\Delta_n] = \E[\Delta^1] \equiv 0, \quad \E[a_n]=\E[w_1]=1, \quad var(a_n) = \frac{var(w_1)}{n} = \frac{\chi^2(p\|q)}{n}
\end{equation*}
For any $\lambda\in(0,1)$, by Chebyshev's inequality,
\begin{equation*}
\prob[a_n \leq \lambda] \leq \frac{var(a_n)}{(\E[a_n] - \lambda)^2} = \frac{ \chi^2(p\|q) }{ (1-\lambda)^2 n }
\end{equation*}
Meanwhile,
\begin{align*}
\E\big[\| \Delta_n \|^2 \big] &= \frac{1}{n} \E\big[\| \Delta^1 \|^2 \big] = \frac{1}{n} \E_{X\sim q} \Big[ \frac{p_X^2}{q_X^2} \cdot \sum_x p_x^2 -2 p_x \delta_X(x) + \delta_X(x) \Big] \\
&= \frac{1}{n} \E_{X\sim q} \Big[ \frac{p_X^2}{q_X^2} \cdot \big(\|p\|^2 - 2 p_X + 1 \big) \Big] \leq \frac{2}{n} \E_{X\sim q} \Big[ \frac{p_X^2}{q_X^2} \Big] \\
&\leq \frac{2\big( \chi^2(p\|q) + 1 \big)}{n}
\end{align*}
Hence,
\begin{align*}
\E[\|p-p_n\|^2] &= \E\Big[ \mathbbm{1}_{a_n \geq \lambda} \Big\| \frac{\Delta_n}{a_n} \Big\|^2 \Big] + \E\big[ \mathbbm{1}_{a_n < \lambda} \|p-p_n\|^2 \big] \\
&\leq \lambda^{-2} \E[\|p - \tilde{p}_n\|^2] + 4\prob[a_n<\lambda] \\
&\leq \frac{2 \chi^2(p\|q) + 2}{\lambda^2 n} + \frac{4\chi^2(p\|q)}{(1-\lambda)^2 n}
\end{align*}
Setting $\lambda=1/2$ gives the first bound.
The second result is straightforward since $p_n$ simplifies to $\frac{1}{n}\sum_{i=1}^n \delta_{X_i}$
\begin{equation*}
\E\big[\|p-p_n\|^2\big] = \frac{1}{n}\E_{X\sim p}[\|p-\delta_X\|^2] = \frac{1}{n}\E_{X\sim p} \big[ \|p\|^2 - 2 p_X + 1 \big] = \frac{1-\|p\|^2}{n}
\end{equation*}
\end{proof}

\begin{proof}[Proof of Proposition \ref{prop. TV decoding error}]
Denote
\begin{align*}
Q_n &= \frac{\sum_{i=1}^n w_i \delta_{Z^{(i,0)}}}{\sum_{i=1}^n w_i} \in \PS(\fip(x)), \quad w_i = \frac{P^{\leq |Z^{(i,0)}|}(Z^{(i,0)}) }{ Q(Z^{(i,0)}|x) }
\end{align*}
Then, $\law(Z^{(0)}|Z^{(1,0)},\dots Z^{(n,0)}) = Q_n$ and
\begin{align*}
\law(X^{(\omega)}) = \proj\#\int P(\cdot|[z]) dQ_n(z), \quad (\proj\#P)(\cdot|[x]) = \proj\# \int P(\cdot|[z]) dQ_*(z|x)
\end{align*}
It follows from Lemma \ref{lemma. TV inequality} that
\begin{align*}
\text{TV}\big( (\proj\#P)(\cdot|[x]), \law(X^{(\omega)}) \big) &\leq \text{TV}\Big( \int P(\cdot|[z]) dQ_*(z|x), \int P(\cdot|[z]) dQ_n(z)\Big) \\
&\leq \text{TV}\big( Q_*(\cdot|x), Q_n(\cdot|x) \big) \\
&\leq \frac{1}{2} \big\| Q_*(\cdot|x) - Q_n(\cdot|x) \big\|_{L^2}
\end{align*}
Then, Lemma \ref{lemma. one-hot estimator} implies that
\begin{equation*}
\E\Big[ \big\| Q_*(\cdot|x) - Q_n(\cdot|x) \big\|_{L^2} \Big] \leq \E\Big[ \big\| Q_*(\cdot|x) - Q_n(\cdot|x) \big\|_{L^2}^2 \Big]^{1/2} \leq \sqrt{\frac{24\chi^2\big( Q_*(\cdot|x) \big\| Q(\cdot|x)\big) + 8}{n}}
\end{equation*}
\end{proof}

\section{Unbiased Decoding Algorithm}
\label{appendix. unbiased decode}


Supplementary to the simple but biased decoding algorithms of Section \ref{sec. decoding},
this section describes an algorithm based on an unbiased estimation of the next-token probabilities $(\proj\#P)^{\leq |x|+1}(\cdot|x)$.

In preparation, we first consider the ideal setting with unlimited compute and calculate each likelihood exactly.
An implementation is provided by Algorithm \ref{alg: decoding ideal}.

\begin{algorithm}[ht]
\caption{Decoding algorithm, ideal setting}
\label{alg: decoding ideal}
\begin{algorithmic}[1]
\State \textbf{Input:} Prompt $x \in \Sigma^*$, projection $\proj$, distribution $P$

\State \textbf{Initialization:} Set $X^{(0)}=x$ and compute the cached values $\{P^{\leq |z|}(z) \mid z \in \fip(x)\}$

\For{$t=0,1,2\dots$}

\State For each $z \in \fip(X^{(t)})$ and $a \in \Sigma$, compute $$P(\nsm(z,a)|z) = \sum \big\{ P^{\leq |zs|} (s|z) \bigm| s\in\nsm_{X^{(t)}}(z,a) \big\}$$
\quad \quad and save the terms $P^{\leq |zs|} (s|z)$ for line \ref{enum. cache update}
\label{enum. segment conditional}

\State Set $Q_*(z|X^{(t)}) \propto P^{\leq |z|}(z)$ for each $z \in \fip(X^{(t)})$, with $P^{\leq |z|}(z)$ taken from the cache
\label{enum. latent posterior}

\State For each $a \in \Sigma$, compute $$(\proj\#P)^{\leq |x|+t+1}(a|X^{(t)}) = \int P(\nsm(z,a)|z) \text{d}Q_*(z|X^{(t)})$$
\label{enum. integrate segment probability}

\State Sample $X_{t+1} \sim (\proj\#P)^{\leq |x|+t+1}(\boldsymbol{\cdot}|X^{(t)})$, set $X^{(t+1)} = X^{(t)}X_{t+1}$, display $\proj^{<\omega}(X^{(t+1)})$


\State For each $z \in \fip(X^{(t)})$ and $s\in\nsm_{X^{(t)}}(z, X_{t+1})$, compute $$P^{\leq |zs|}(zs) = P^{\leq |z|}(z) P^{\leq |zs|}(s|z)$$
\quad \quad and replace the cached term $P^{\leq |z|}$ by the terms $P^{\leq |zs|}(zs)$
\label{enum. cache update}

\EndFor
\State \textbf{return} $X^{(\omega)}$
\end{algorithmic}
\end{algorithm}

Note that, except for the initialization, line \ref{enum. segment conditional} is the only place that calls the distribution $P$ (or the model that parametrizes $P$).
The cache updated at line \ref{enum. cache update} works for line \ref{enum. latent posterior} of the next step since
\begin{equation*}
\fip(X^{(t+1)}) = \big\{ zs \bigm| z \in \fip(X^{(t)}), ~s\in\nsm_{X^{(t)}}(z, X_{t+1}) \big\}
\end{equation*}
The resulting $X^{(\omega)}$ is a sample of $(\proj\#P)(\cdot|[x])$.
Alternatively, one can terminate whenever the decoded token $X_t=\eos$, some ``end of sentence" symbol, and return $X^{(t)}$.

Sampling-based decoding replaces lines \ref{enum. segment conditional} and \ref{enum. integrate segment probability} of Algorithm \ref{alg: decoding ideal} with Monte-Carlo estimators.
Replacing line \ref{enum. integrate segment probability} is necessary, since $\fip(x)$ must be intractably large as aforementioned.
However, the cost of line \ref{enum. segment conditional} depends on the size of $\nsm(z,a)$ and thus on the choice of $\proj$,
e.g.\@ its size is constant for Example \ref{ex. product}, while unbounded for Example \ref{ex. think}.
For simplicity, we replace line \ref{enum. integrate segment probability} anyway.

Ideally, this algorithm should proceed incrementally to save compute, similar to the decoding algorithms of Section \ref{sec. decoding}.
Specifically, when estimating line \ref{enum. integrate segment probability} by sampling, with some sampler $Q$, the algorithm preferably should obtain each sample $Z^{(t,i)}$ by updating the earlier sample $Z^{(t-1,i)}$, instead of sampling $Q(\cdot|X^{(t)})$ from scratch for each $t$.
So, an ideal algorithm is to first sample $\{ Z^{(0,i)} \}_{i=1}^n$ from $Q(\cdot|X^{(0)})$ with $X^{(0)}=x$, and then incrementally extend $Z^{(0,i)}$ to $Z^{(t,i)}$ and $X^{(0)}$ to $X^{(t)}$.
These $Z^{(t,i)}$ should remain i.i.d.\@ samples of $Q(\cdot|X^{(t)})$ for all $t$ to enable Monte-Carlo estimation.

Technically, such incremental procedure requires a conditional distribution $B(\cdot|z)$ such that, given a sample $Z \sim Q(\cdot|X^{(t-1)})$, the distribution of $ZS$ with $S \sim B(\cdot|Z)$ is exactly $Q(\cdot|X^{(t)})$.
However, the following proposition indicates that this constraint makes the sampler $Q$ suboptimal.

\begin{proposition}
\label{prop. impossible incremental}
Given the setting of Theorem \ref{thm. conditional latent distribution} and any sampler $Q \in \Q_{\proj}$,
if for any $x\in\Sigma^*$ and $a\in\Sigma$ such that $xa\in\dom(Q)$, there exists a conditional distribution $B\colon \fip(x)\to\PS(\Omega^*)$
such that
\begin{equation*}
Q(\cdot|xa) = \law(ZS), \quad S \sim B(\cdot|Z), \quad Z \sim Q(\cdot|x)
\end{equation*}
then $Q$ is a causal sampler and in general must differ from the posterior sampler $Q_*$.
\end{proposition}

\begin{proof}
The existence of $B$ implies that
\begin{equation*}
Q(\cdot|x) = \Pi_{\fip(x)}\#Q(\cdot|xa)
\end{equation*}
Applying this equality recursively over the next token $a$, we obtain the identity (\ref{eq. causal sampler identity}) for any $x,x' \in \dom(Q)$ such that $x \sqsubseteq x'$.
Thus, $Q$ is a causal sampler.
By Theorem \ref{thm. posterior not causal}, $Q_*$ is in general not a causal sampler.
\end{proof}

Hence, in order to be able to approximate the variance-minimizing sampler $Q_*$, we sample $Q(\cdot|X^{(t)})$ from scratch for each $t$.
The resulting sampling-based decoding algorithm is Algorithm \ref{alg: decoding sampling}.

\begin{algorithm}[ht]
\caption{Decoding with latent sampling}
\label{alg: decoding sampling}
\begin{algorithmic}
\State \textbf{Input:} Prompt $x \in \Sigma^*$, projection $\proj$, distribution $P$, sampler $Q$, sample size $n$
\State \textbf{Initialization:} Set $X^{(0)} = x$

\For{$t=0,1,2\dots$}

\For{$i=1,\dots,n$ in parallel}
\State Sample $\{Z^{(t,i)}\}_{i=1}^n \iidsample Q(\cdot|X^{(t)})$
\State Compute the weights $w_i^{(t)} = P^{\leq |Z^{(t,i)}|}(Z^{(t,i)}) / Q(Z^{(t,i)}|X^{(t)})$
\State Set sample size for next segments $m_i = \big\lceil n (w_i^{(t)})^2 / \sum_j (w^{(t)}_j)^2 \big\rceil$ \Comment{In total, $n\leq \sum_i m_i <2n$}
\State Call Algorithm \ref{alg: segment probability} on $(\proj,P,X^{(t)},Z^{(t,i)},m_i)$ to obtain an estimator $p^{(t,i)}_m$
\EndFor

\State Estimate the next-token distribution by 
\begin{equation}
\label{eq. Monte-Carlo next-token}
p^{(t)}_{n,m} = \sum_{i=1}^n w_i^{(t)} p^{(t,i)}_m  \bigm/ \sum_{i=1}^n w_i^{(t)}
\end{equation}

\State Sample $X_{t+1} \sim p^{(t)}_{n,m}$, set $X^{(t+1)} = X^{(t)}X_{t+1}$, display $\proj^{<\omega}(X^{(t+1)})$





\EndFor
\State \textbf{return} $X^{(\omega)}$
\end{algorithmic}
\end{algorithm}

\begin{algorithm}[ht]
\caption{Estimation of probability vector $P(\nsm_x(z,\boldsymbol{\cdot})|z)$}
\label{alg: segment probability}
\begin{algorithmic}
\State \textbf{Input:} Projection $\proj$, distribution $P$, sequences $x\in \Sigma^*$ and $z \in \fip(x)$, sample size $m$
\State \textbf{Initialization:} Set sequences $S^{(0,1)},\dots,S^{(0,m)}=\varnothing$
\For{$i=1,\dots,m$ in parallel}
\For{$t=0,1,2\dots$}
\If{$S^{(t,i)} \in \nsm_x(z,a)$ for some $a$} \Comment{This always happens within some $c_{|x|}$ steps}
\State Set $A_i = a$ \Comment{and this $a$ is always unique (cf.\@ Lemma \ref{lemma. estimate latent conditional next token})}
\State \textbf{break}
\EndIf
\State Sample $Z_{t+1} \sim P^{\leq |z|+t+1}(\cdot|zS^{(t,i)})$ and set $S^{(t+1,i)} = S^{(t,i)} Z_{t+1}$
\EndFor
\EndFor
\State \textbf{return} the $|\Sigma|$-dimensional probability vector $\frac{1}{m}\sum_{i=1}^m \mathbf{e}_{A_i}$
\end{algorithmic}
\end{algorithm}

The following proposition establishes the correctness and sample efficiency of Algorithm \ref{alg: decoding sampling}.

\begin{proposition}
\label{prop. estimate conditional}
Given any continuous surjective partial function $\proj\colon \Omega^{\omega}\rightharpoonup\Sigma^*$, any distribution $P\in\PS(\dom(\proj))$,
any conditional distribution $Q(\cdot|x)$ whose support equals $\fip(x)$ for all $x\in\Sigma^*$,
and any $t\geq 0$,
the probability vector $p^{(t)}_{n,m}$ (\ref{eq. Monte-Carlo next-token}) is an unbiased estimator for the next-token distribution $(\proj\#P)^{\leq |x|+t+1}(\cdot|X^{(t)})$,
conditioned on $X^{(t)}=xy$ for any $x\in\Sigma^*$ and $y\in\Sigma^t$ with $(\proj\#P)^{|x|+t}(xy)>0$.
Its variance is bounded by
\begin{align*}
var\big( p^{(t)}_{n,m} \bigm| X^{(t)}=xy \big) < \frac{32 \chi^2\big( Q_*(\cdot|x) \big\| Q(\cdot|x) \big) + 12}{n}
\end{align*}
where $\chi^2$ is the chi-square divergence.
\end{proposition}


\begin{proof}

Let $\Delta = \Delta_{|\Sigma|}$ denote the probability simplex with dimension $|\Sigma|$.
Define the function
$$g\colon \fip(X^{(t)})\to\Delta, \quad g(z) = P(\nsm_{X^{(t)}}(z,\boldsymbol{\cdot})|z)$$
Define the following $|\Sigma|$-dimensional vectors
\begin{align*}
p_*^{(t)} &= (\proj\#P)^{\leq |x|+t+1}(\cdot|X^{(t)}) \\
p_n^{(t)} &= \sum_{i=1}^n w_i^{(t)} g(Z^{(t,i)}) \Bigm/ \sum_{i=1}^n w_i^{(t)}, \quad \tilde{p}^{(t)}_{n,m} = \sum_{i=1}^n w_i^{(t)} p^{(t,i)}_m \Bigm/ \sum_{i=1}^n w_i^{(t)}
\end{align*}


By Lemma \ref{lemma. estimate latent conditional next token}, Algorithm \ref{alg: segment probability} is unbiased, so
\begin{align*}
\forall z\in\fip(xy), \quad \E\big[ p^{(t,i)}_m \bigm| X^{(t)}=xy, ~Z^{(t,i)}=z \big] &= g(z) \\
\E\big[ p^{(t)}_{n,m} \bigm| X^{(t)}=xy \big] &= p_*^{(t)}
\end{align*}
Thus,
\begin{align*}
var\big( p^{(t)}_{n,m} \bigm| X^{(t)}=xy \big) &= \E \big[ \| p_*^{(t)} - \tilde{p}_{n,m}^{(t)} \|^2 \big| X^{(t)}=xy \big]\\
&= \E\big[ \| p_*^{(t)} - p_n^{(t)} \|^2 \big| X^{(t)}=xy \big] + \E\big[ \| p_n^{(t)} - \tilde{p}_{n,m}^{(t)} \|^2 \big| X^{(t)}=xy \big] + 0
\end{align*}
By Theorem \ref{thm. conditional latent distribution},
\begin{align*}
p_*^{(t)} &= \int g(z) dQ_*(z|X^{(t)}) = \E_{Z\sim Q(\cdot|X^{(t)})} \Big[ \frac{Q_*(Z|X^{(t)})}{Q(Z|X^{(t)})} g(Z) \Big]\\
p_n^{(t)} &= \sum_{i=1}^n \tilde{w}_i g(Z^{(t,i)}) \Bigm/ \sum_{i=1}^n \tilde{w}_i, \quad \tilde{w}_i = \frac{Q_*(Z^{(t,i)}|X^{(t)})}{Q(Z^{(t,i)}|X^{(t)})}
\end{align*}
Then, Lemma \ref{lemma. one-hot estimator} implies that
\begin{equation*}
\E\big[ \| p_*^{(t)} - p_n^{(t)} \|^2 \bigm| X^{(t)}=xy \big] \leq \frac{24 \chi^2\big( Q_*(\cdot|x) \big\| Q(\cdot|x) \big) + 8}{n}
\end{equation*}


Meanwhile,
since the errors $g(Z^{(t,i)}) - p^{(t,i)}_m$ are independent conditioned on $Z_1^{(t)},\dots,Z_n^{(t)}$ and $X^{(t)}$,
we have
\begin{align*}
\E\big[ \| p_n^{(t)} - \tilde{p}_{n,m}^{(t)} \|^2 \big| X^{(t)}=xy \big] &= \E \Big[ \frac{\sum_{i=1}^n \tilde{w}_i^2 ~ \E_{p^{(t,i)}_m} \big[ \| g(Z^{(t,i)}) - p^{(t,i)}_m \|^2 \big]}{\big( \sum_{i=1}^n \tilde{w}_i \big)^2} \Big] \\
& < \E \Big[ \frac{\sum_{i=1}^n \tilde{w}_i^2 / m_i}{\big( \sum_{i=1}^n \tilde{w}_i \big)^2} \Big] \leq \E \Big[ \frac{\sum_{i=1}^n \tilde{w}_i^2}{\big( \sum_{i=1}^n \tilde{w}_i \big)^2} \Big] \\
&\leq \frac{8 var(\tilde{w}_1)/\E[\tilde{w}_1]^2 + 4}{n} = \frac{8 \chi^2\big( Q_*(\cdot|x) \| Q(\cdot|x) \big) + 4}{n}
\end{align*}
where the first inequality follows from Lemma \ref{lemma. estimate latent conditional next token},
the second inequality from the definition of the sample size $m_i$,
and the third inequality from Lemma \ref{lemma. coefficient of variation}.
Combining the two upper bounds completes the proof.
\end{proof}


\begin{lemma}
\label{lemma. estimate latent conditional next token}
For any continuous surjective partial function $\proj\colon \Omega^{\omega}\rightharpoonup\Sigma^*$ and distribution $P\in\PS(\dom(\proj))$, Algorithm \ref{alg: segment probability} is well-defined such that
it terminates $P$-almost surely with steps $t \leq c_{|x|}$ for some constant $c_{|x|}$ that depends only on input length $|x|$,
and that the assigned token $a$ is always unique.
The output probability vector $p_m = \frac{1}{m}\sum_{i=1}^m \mathbf{e}_{A_i}$ is an unbiased and strongly consistent estimator for $P(\nsm_x(z,\boldsymbol{\cdot})|z)$, with variance
\begin{equation*}
var(p_m) := \E\big[\big\|p_m - P(\nsm_x(z,\boldsymbol{\cdot})|z) \big\|^2\big] < \frac{1}{m}
\end{equation*}
\end{lemma}

\begin{proof}[Proof of Lemma \ref{lemma. estimate latent conditional next token}]
Fix any inputs $x\in\Sigma^*$ and $z\in\fip(x)$.
Consider the sample $Z\sim P(\cdot|[z])$.
Property 4 of Lemma \ref{lemma. segments prefix-free} indicates that there is exactly one $a\in\Sigma$ and $s \in \nsm_x(z,a)$ such that $zs \sqsubseteq Z$.
Property 1 of Lemma \ref{lemma. segments prefix-free} implies that the set $\bigcup_{x'\in\Sigma^{|x|}}\bigcup_{z'\in\fip(x')} \bigsqcup_{a\in\Sigma}\nsm_{x'}(z',a)$ is finite,
and thus $|s|$ is bounded by some constant $c_{|x|}$.
This proves that Algorithm \ref{alg: segment probability} always returns within $c_{|x|}$ steps.

Effectively, Algorithm \ref{alg: segment probability} is sampling the next-segment distribution (Definition \ref{def. next-segment distribution}) such that
$$\{S^{(t_i,i)} \}_{i=1}^m \iidsample P\Big( \cdot \Big| z \to \bigsqcup_{a\in\Sigma} \nsm_x(z,a) \Big)$$
where $t_i \leq c_{|x|}$ denotes the termination time of each $S_i$.
By (\ref{eq. next token condition on latent}),
\begin{equation*}
A_i = \proj_{x,z}(S^{(t_i,i)}) \iidsample P\big(\nsm_x(z,\cdot)\big|z\big)
\end{equation*}
So the following is an unbiased and strongly consistent estimator
\begin{equation*}
p_m = \frac{1}{m} \sum_{i=1}^m \mathbf{e}_{A_i} \approx P\big(\nsm_x(z,\boldsymbol{\cdot})\big|z\big)
\end{equation*}
where $\mathbf{e}_j$ denotes the unit vector of coordinate $j$.
Its variance is given by
\begin{align*}
var(p_m) &= \E\big[\big\|p_m - P(\nsm_x(z,\boldsymbol{\cdot})|z) \big\|^2\big] \leq \sup_{q \in \Delta_{|\Sigma|}} \frac{1}{m} \E_{A\sim q} \big[ \| \mathbf{e}_A - q \|^2 ]\\
&= \frac{1}{m} \sup_{q \in \Delta_{|\Sigma|}} \sum_{i=1}^{|\Sigma|} q_i(1-q_i) = \frac{1}{m} \Big(1 - \frac{1}{|\Sigma|}\Big)
\end{align*}
where $\Delta_k$ denotes the $k$-dimensional probability simplex.
\end{proof}

\begin{lemma}
\label{lemma. coefficient of variation}
Given a distribution over positive numbers $P \in \PS(\R_+)$ with mean $\mu$ and variance $\sigma^2$, we have
\begin{equation*}
\E\Big[\frac{\sum_{i=1}^n X_i^2}{\big(\sum_{i=1}^n X_i \big)^2}\Big] \leq \frac{8\sigma^2/\mu^2+4}{n}, \quad X_1, \dots X_n \iidsample P
\end{equation*}
\end{lemma}

\begin{proof}
Denote $\mu_n = \frac{1}{n}\sum_{i=1}^n X_i$ and $s_n = \frac{1}{n}\sum_{i=1}^n X_i^2$.
We always have $s_n \leq n\mu_n^2$.
For any $\lambda\in(0,1)$, by Chebyshev's inequality,
\begin{equation*}
\prob[\mu_n \leq \lambda\mu] \leq \frac{var(\mu_n)}{(\E[\mu_n]-\lambda\mu)^2} = \frac{\sigma^2/\mu^2}{(1-\lambda)^2 n}
\end{equation*}
Then,
\begin{align*}
n \cdot \E\Big[\frac{\sum_{i=1}^n X_i^2}{\big(\sum_{i=1}^n X_i \big)^2}\Big] &= \E\Big[\mathbbm{1}_{\mu_n < \lambda\mu} \frac{s_n}{\mu_n^2} \Big] + \E\Big[\mathbbm{1}_{\mu_n \geq \lambda\mu} \frac{s_n}{\mu_n^2} \Big] \\
&\leq n\cdot \prob[\mu_n \leq \lambda\mu] + \frac{\E[s_n]}{\lambda^2\mu^2} \\
&\leq \frac{\sigma^2/\mu^2}{(1-\lambda)^2} + \frac{\sigma^2/\mu^2+1}{\lambda^2}
\end{align*}
Setting $\lambda=1/2$ finishes the proof.
\end{proof}

\section{KL Divergence from Finite to Infinite Sequences}
\label{appendix. KL finite to infinite}

This section provides a supplementary result to Section \ref{sec. training}, that the cross entropy loss (\ref{eq. cross entropy}) can be interpreted in terms of the KL divergence with the latent distribution $P$, not just with $\proj\#P$.


We assume that $P_*$ is supported on a prefix-free subset.
This is typical in practice, for instance the samples for pretraining are usually arranged to have the same length, and the samples for finetuning are appended with an $\eos$ token that only appears at the end, so the sample sets in both cases are prefix-free.

\begin{proposition}
\label{prop. cross entropy to KL}
Given any $P_*\in\PS(\Sigma^*)$ with prefix-free support $\sprt P_*$ and finite entropy $H(P_*) = -\int \log P_* dP_* < \infty$,

\begin{enumerate}

\item The cross entropy loss can be expressed as
\begin{equation*}
\forall P \in \PS(\Sigma^*), \quad -\int \log P^{\leq |x|}(x)dP_*(x) = \min_{P'\in\PS_*} \KL(P'\|P) + H(P_*)
\end{equation*}
where $\PS_* \subseteq \PS(\Sigma^{\omega})$ is defined by
\begin{align*}
\PS_* &:= \{P' \mid P'\in\PS(\Sigma^{\omega}), ~\Pi_{\sprt P_*}\#P'=P_*\}\\
&= \Big\{ \int B_x dP_*(x) \Bigm| B\colon \Sigma^*\to\PS(\Sigma^{\omega}), ~\forall x\in\sprt P_*, ~\sprt B_x \subseteq [x] \Big\}
\end{align*}
where $\Pi_{\sprt P_*}$ is defined by Lemma \ref{lemma. prefix-free projection}.
The unique minimizer is given by $B_x = P(\cdot|[x])$.

\item Given any surjective continuous partial function $\proj\colon \Omega^{\omega}\rightharpoonup\Sigma^{\omega}$,
the loss $L$ (\ref{eq. cross entropy}) can be expressed as
\begin{equation*}
\forall P \in \PS(\dom(\proj)), \quad L(P) = \min_{P'\in\proj^{-1}\#\PS_*} \KL(P'\|P) + H(P_*)
\end{equation*}
where $\proj^{-1}\#\PS_* \subseteq \PS(\Sigma^{\omega})$ is defined by
\begin{align*}
\proj^{-1}\#\PS_* &:= \big\{P' \bigm| P'\in\PS(\Sigma^{\omega}), ~(\Pi_{\sprt P_*} \circ \proj) \#P' = P_* \big\}\\
&= \Big\{ \int B_x dP_*(x) \Bigm| B\colon \Sigma^*\to\PS(\Omega^{\omega}), ~\forall x\in\sprt P_*, ~\sprt B_x \subseteq \bigsqcup_{z\in\fip(x)}[z] \Big\}
\end{align*}
The unique minimizer is given by
$$B_x = \int P(\cdot|[z]) dQ_*(z|x) = P\Big(\cdot\Big|\bigsqcup_{z\in\fip(x)}[z]\Big)$$
where $Q_*$ is the posterior distribution (\ref{eq. latent posterior}) of $P$.

\end{enumerate}
\end{proposition}

\begin{proof}
Regarding property 1, note that
\begin{equation*}
-\int \log P^{\leq |x|}(x) dP_*(x) = \KL\big(P_* \big\| \Pi_{\sprt P_*}\#P \big) + H(P_*)
\end{equation*}
Meanwhile, by Lemma \ref{lemma. prefix-free projection},
\begin{align*}
\KL\big(P_* \big\| \Pi_{\sprt P_*}\#P \big) &= \KL\big(P_* \big\| \Pi_{\sprt P_*}\#P \big) + \int \KL\big(P(\cdot|[x]) \big\| P(\cdot|[x]) \big) dP_*(x) \\
&= \KL\big(P_* \big\| \Pi_{\sprt P_*}\#P \big) + \min_{P^{(2)}_*(\cdot|x)}\int \KL\big(P^{(2)}_*(\cdot|x) \big\| P(\cdot|[x]) \big) dP_*(x) \\
&= \min_{P' \in \PS_*} \KL(P'\|P)
\end{align*}

Regarding property 2, we similarly have
\begin{align*}
L(P) - H(P_*) &= \KL\big(P_* \big\| \Pi_{\sprt P_*}\#\proj\#P \big) = \min_{P' \in \PS_*} \KL(P'\|\proj\#P) \\
&\leq \min_{P' \in \proj^{-1}\#\PS_*} \KL(P'\|P)
\end{align*}
The last inequality follows from Lemma \ref{lemma. KL mapping}.
Since $\sprt P_*$ is prefix-free, Lemma \ref{lemma. segments prefix-free} property 2 implies that the sets $\fip(x)$ for $x\in\sprt P_*$ are disjoint and their union is prefix-free.
Therefore, the following sets are all disjoint
\begin{equation*}
\mathcal{S} = \big\{ [z] \bigm| z\in\fip(x), ~x\in\sprt P_* \big\}
\end{equation*}
Consider the distribution
\begin{align*}
P_{\sqcup} &= \iint P(\cdot|[z]) dQ_*(z|x) dP_*(x) = \iint P(\cdot|[z]) \frac{dP([z])}{(\proj\#P)^{\leq |x|}(x)} dP_*(x) \\
&= \int \frac{\sum_{z\in\fip(x)}\mathbf{1}_{[z]}(\cdot) P(\cdot)}{(\proj\#P)^{\leq |x|}(x)} dP_*(x) = \int P\Big(\cdot\Big|\bigsqcup_{z\in\fip(x)}[z]\Big) dP_*(x) \\
&\in \proj^{-1}\#\PS_*
\end{align*}
Thus, $P$ and $P_{\sqcup}$ share the same conditional distributions over $\bigsqcup_{z\in\fip(x)}[z]$ for each $x\in\sprt P_*$.
Then, Lemma \ref{lemma. prefix-free projection} implies that
\begin{align*}
\KL\big(P_* \big\| \Pi_{\sprt P_*}\#\proj\#P \big) &= \KL\big( P_{\sqcup} \big\| P \big) \geq \min_{P'\in\proj^{-1}\#\PS_*} \KL(P'\|P)
\end{align*}
which completes the proof.
\end{proof}

\begin{lemma}
\label{lemma. prefix-free projection}
Given any prefix-free subset $S\subseteq \Sigma^*$,
\begin{enumerate}
\item The following relation $\Pi_S$ is a well-defined measureable function $\Sigma^{\omega}\to \Sigma^*$
\begin{equation*}
\Pi_S := \big\{ (x,x') \bigm| \forall x \in \Sigma^{\omega}, ~\forall x' \in S, ~x' \sqsubseteq x \big\} ~\cup~ \big\{ (x,\varnothing) \bigm| \forall x \in \Sigma^{\omega}, ~\not\exists x' \in S, ~x' \sqsubseteq x \big\}
\end{equation*}

\item Given any distribution $P^{(1)}_* \in \PS(S)$ and any conditional distribution $P^{(2)}_*(\cdot|x)\colon S\to \PS(\Sigma^{\omega})$ such that $\sprt P^{(2)}_*(\cdot|x) \subseteq [x]$, denote
\begin{equation*}
P_* = \int P^{(2)}_*(\cdot|x) dP^{(1)}_*(x)
\end{equation*}
Then, for any $P \in \PS(\Sigma^{\omega})$,
\begin{equation*}
\KL(P_*\|P) = \KL\big(P_*^{(1)} \big\| \Pi_S\# P \big) + \int \KL\big( P_*^{(2)}(\cdot|x) \big\| P(\cdot|[x]) \big) dP_*^{(1)}(x)
\end{equation*}

\end{enumerate}
\end{lemma}

\begin{proof}
Regarding property 1, if $(x,x_1),(x,x_2) \in \Pi_S$ for distinct $x_1,x_2 \in S$, then $x_1,x_2 \sqsubseteq x$ implies that $x_1 \sqsubseteq x_2$ or $x_2 \sqsubseteq x_1$.
A contradiction to prefix-free.
Thus, $(x,x') \in \Pi_S$ holds for at most one $x' \in S$, and $\Pi_S$ is a well-defined function.
Since the sets $\Pi_S^{-1}(x)$ for $x \in S$ are open cylinders, $\Pi_S^{-1}(\varnothing)$ is closed, and thus $\Pi_S$ is measureable.

Regarding property 2, note that $\Pi_S\#P(x) = P([x])$ for all $x\in S$.
Thus,
\begin{align*}
\KL(P_*\|P) &= \iint \log \frac{P^{(2)}(y|x) P^{(1)}(x)}{P(y|[x])P([x])} dP^{(2)}(y|x) dP^{(1)}(x)\\
&= \int \log \frac{P^{(1)}(x)}{P^{\leq |x|}(x)} + \int \log \frac{P^{(2)}(y|x)}{P(y|[x])} dP^{(2)}(y|x) dP^{(1)}(x) \\
&= \KL\big(P_*^{(1)} \big\| \Pi_S\# P \big) + \int \KL\big( P_*^{(2)}(\cdot|x) \big\| P(\cdot|[x]) \big) dP_*^{(1)}(x)
\end{align*}
\end{proof}





\section{Properties of Gradient Estimation}
\label{appendix. gradient estimation}

This section provides the derivation of the optimal samplers and expected error of the gradient estimator in Section \ref{sec. training}.

To avoid degeneracy, we make the following assumption.
It can be satisfied, for instance, if we implement $P_f$ and the samplers by Transformer LLMs.
\begin{assumption}
\label{assume. gradient nondegeneracy}
The latent distribution $P_f$ satisfies the following:
for any $x\in \dom(Q_*)$ and $z\in\fip(x)$, the likelihood $P^{\leq |z|}(z)>0$,
the function $\log P_f^{\leq|z|}(z)$ is differentiable in $\theta$,
and the gradient vector $\vb_*(x)$ is non-zero.
Meanwhile, the sampling distribution $\tilde{Q}'$ in (\ref{eq. min grad mse}) is constrained to those such that $Q_*(\cdot|x) \ll \tilde{Q}'$ (Since $\fip(x)$ is finite, this is equivalent to $\sprt Q_*(\cdot|x) \subseteq \sprt\tilde{Q}'$).
\end{assumption}

\subsection{Optimal Samplers}

\begin{proposition}
\label{prop. min grad mse}
Given Assumption \ref{assume. gradient nondegeneracy}, the unique minimizer of (\ref{eq. min grad mse}) is the samplers $Q=Q_*$ and $\tilde{Q} = Q_{\eye}$, with the latter defined by (\ref{eq. inquisitive sampler}).
\end{proposition}

\begin{proof}[Proof of Proposition \ref{prop. min grad mse}]
Fix any $x\in\dom(Q_*)$.
Denote the numerator and denominator of $\vb_n(x)$ by $\w_n$ and $q_n$,
and denote those of $\vb_*(x)$ by $\w$ and $q$.
Then, $q = (\proj \# P_f)^{\leq |x|}(x) > 0$.
Note that
\begin{align*}
\E[q_n] &= \sum_{z \in \sprt Q'} P_f^{\leq |z|}(z) = (\proj\#P_f)^{\leq |x|}(x) \cdot Q_*(\sprt Q') \in (0,q]
\end{align*}
Similarly, since $\sprt Q_*(\cdot|x) \subseteq \sprt \tilde{Q}'$ by assumption, we always have $\E[\w_n]=\w$.
One can check that $q_n = q$ almost surely if and only if $Q' = Q_*(\cdot|x)$.
Meanwhile,
\begin{align}
\nonumber
\frac{n}{q^2} \cdot \E\big[\|\w_n-\w\|^2] &= \frac{\E\big[\|\w_1\|^2] - \|\w\|^2}{q^2} = \int \frac{Q_*(z|x)^2 ~\|\nabla_{\theta}\log P_f^{\leq|z|}(z)\|^2}{\tilde{Q}'(z)^2} d\tilde{Q}'(z) - \|\vb_*(x)\|^2 \\
\nonumber
&= C_{\uparrow}^2 \int \frac{Q_{\eye}(z|x)^2}{\tilde{Q}'(z)^2} -1 ~d\tilde{Q}'(z) + (C_{\uparrow}^2-\|\vb_*(x)\|^2) \\
\label{eq. grad nonimator var}
&= C_{\uparrow}^2 ~ \chi^2\big(Q_{\eye}(\cdot|x)\big\|\tilde{Q}'\big) + (C_{\uparrow}^2-\|\vb_*(x)\|^2)
\end{align}
where we define
\begin{align*}
C_{\uparrow} &:= \int \big\| \nabla_{\theta} \log P_f^{\leq |z|}(z) \big\| dQ_*(z|x) \geq \Big\| \int \nabla_{\theta} \log P_f^{\leq |z|}(z) dQ_*(z|x) \Big\| = \|\vb_*(x)\| > 0
\end{align*}
So $Q_{\eye}(\cdot|x)$ can be characterized as the unique minimizer
\begin{equation*}
Q_{\eye}(\cdot|x) = \argmin_{\substack{\tilde{Q}'\in\PS(\fip(x))\\\sprt Q_*(\cdot|x) \subseteq \sprt\tilde{Q}'}} \E\big[\|\w_n-\w\|^2]
\end{equation*}
If $\w_n, q_n$ are seen as random variables parametrized by the distributions $\tilde{Q}'$ and $Q'$,
then all conditions of Lemma \ref{lemma. min quotient estimator} are satisfied,
so it implies that the unique minimizer to (\ref{eq. min grad mse}) is the pair $(Q_*,Q_{\eye})$.
\end{proof}

\begin{lemma}
\label{lemma. min quotient estimator}
Let $\w \neq \mathbf{0}$ be any vector in some Hilbert space $(\HS,\|\cdot\|)$, and let $q>0$ be a constant.
Let $\w_{\phi}$ and $q_{\psi}$ be independent random variables that range in $\HS$ and $\R_+$ and are parametrized by parameters $\phi$ and $\psi$.
Assume that $\E[\w_{\phi}]=\w$ for any $\phi$,
that $\E[q_{\psi}] \leq q$ for any $\psi$,
and that there exists a unique $\psi_*$ such that $q_{\psi_*} = q$ almost surely.
If $\min_{\phi} \E[\|\w_{\phi}-\w\|^2]$ has a unique minimizer $\phi_*$,
then $(\phi_*,\psi_*)$ is the unique minimizer to
\begin{equation*}
\min_{\phi,\psi} ~ \E\Big[\Big\| \frac{\w_{\phi}}{q_{\psi}} - \frac{\w}{q} \Big\|^2\Big]
\end{equation*}
\end{lemma}

\begin{proof}
By Jensen's inequality and the assumptions,
\begin{align*}
\E\Big[\Big\| \frac{\w_{\phi}}{q_{\psi}} - \frac{\w}{q} \Big\|^2\Big] &= \E\Big[\Big\| \frac{\w_{\phi}}{q_{\psi}} - \frac{\w}{q_{\psi}} \Big\|^2\Big] + \E\Big[\Big\| \frac{\w}{q_{\psi}} - \frac{\w}{q} \Big\|^2\Big] \\
&\geq \frac{\E[\|\w_{\phi}-\w\|^2]}{\E[q_{\psi}]^2} + \|\w\|^2 \E\big[(q_{\psi}^{-1}-q^{-1})^2\big] \\
&\geq q^{-2} \E[\|\w_{\phi}-\w\|^2] + \|\w\|^2 \E\big[(q_{\psi}^{-1}-q^{-1})^2\big] \\
&\geq q^{-2} \min_{\phi} \E[\|\w_{\phi}-\w\|^2]
\end{align*}
and the first two inequalities become equalities if and only if $q_{\psi} = q$ almost surely.
Thus, $(\phi_*,\psi_*)$ achieves the minimum value and is a minimizer.
Any other choice of $\phi$ or $\psi$ would get a strictly larger value, so $(\phi_*,\psi_*)$ is the unique minimizer.
\end{proof}

\subsection{Expected Error}
\label{appendix. gradient estimation error}

This section provides a derivation of the error estimate (\ref{eq. grad mse one sample}).
All calculations are formal (except for Lemma \ref{lemma. quotient estimator}), such that technical conditions such as integrability and convergence are ignored.

Denote for $i=1,\dots n$,
\begin{align*}
\w^{(i)} = \frac{\nabla_{\theta} P_f^{\leq |Z^{(i)}_{\eye}|}(Z^{(i)}_{\eye})}{Q^{\see}(Z^{(i)}_{\eye}|X^{(i)})}, \quad \w_n &= \frac{1}{n} \sum_{i=1}^n \w^{(i)}, \quad \w_* = \nabla_{\theta} (\proj\#P_f)^{\leq |x|}(x) = \E[\w^{(1)}] \\
q^{(i)} = \frac{P_f^{\leq |Z^{(i)}|}(Z^{(i)})}{Q(Z^{(i)}|X^{(i)})}, \quad q_n &= \frac{1}{n} \sum_{i=1}^n q^{(i)}, \quad q_* = (\proj\#P_f)^{\leq |x|}(x) = \E[q^{(1)}]
\end{align*}
so that $\vb_* = \w_* / q_*$ and $\vb^{\see}_n = \w_n / q_n$.
Then, by (\ref{eq. grad nonimator var}) and (\ref{eq. Monte-Carlo var}),
\begin{align*}
var(\w^{(1)}) &= q_*^2 \Big( C_{\uparrow}^2 ~\chi^2\big( Q_{\eye}(\cdot|x) \big\| Q^{\see}(\cdot|x) \big) + (C_{\uparrow}^2 - C_{\downarrow}^2) \Big) \\
var(q^{(1)}) &= q_*^2 ~ \chi^2\big( Q_*(\cdot|x) \big\| Q(\cdot|x) \big)
\end{align*}
Using Lemma \ref{lemma. quotient estimator} in a formal manner (such that its condition is not verified), we have
\begin{align*}
\E[q_n^{-1}] &= q_*^{-1} + \frac{var(q^{(1)})}{n q_*} + O(n^{-2}) \\
\E[\big( q_n^{-1} - q_*^{-1})^2 \big] &= \frac{var(q^{(1)})}{n q_*^2} + O(n^{-2})
\end{align*}
Thus,
\begin{align*}
\E[q_n^{-2}] &= \E\big[(q_n^{-1} - q_*^{-1})^2\big] - q_*^{-2} + 2q_*^{-1} \E[q_n^{-1}] = q_*^{-2} + \frac{3 var(q^{(1)})}{nq_*^{-2}}+ O(n^{-2})
\end{align*}
The MSE becomes
\begin{align*}
\E\big[\| \vb_* - \vb^{\see}_n \|^2\big] &= \E\Big[ \Big\| \big( \frac{\w_*}{q_*} - \frac{\w_*}{q_n} \big) + \big( \frac{\w_*}{q_n} - \frac{\w_n}{q_n} \big) \Big\|^2 \Big] \\
&= \E\Big[ \big\| \frac{\w_*}{q_*} - \frac{\w_*}{q_n} \big\|^2 \Big] + \E\Big[ \big\| \frac{\w_*}{q_n} - \frac{\w_n}{q_n} \big\|^2 \Big] + 0 \\
&= \|\w_*\|^2 ~ \E[\big( q_n^{-1} - q_*^{-1})^2 \big] + var(\w_n) ~ \E[q_n^{-2}] \\
&= q_*^2 C_{\downarrow}^2 \frac{var(q^{(1)})}{n q_*^2} + \frac{var(\w^{(1)})}{n} q_*^{-2} + O(n^{-2}) \\
&= \frac{q_*^2 C_{\downarrow}^2 ~ \chi^2\big( Q_*(\cdot|x) \big\| Q(\cdot|x) \big)}{n} + \frac{C_{\uparrow}^2 ~\chi^2\big( Q_{\eye}(\cdot|x) \big\| Q^{\see}(\cdot|x) \big) + (C_{\uparrow}^2 - C_{\downarrow}^2)}{n} + O(n^{-2})
\end{align*}

\begin{lemma}
\label{lemma. quotient estimator}
Let $X^{(1)},\dots X^{(n)}$ be positive and i.i.d.\@ random variables with mean $x_*$.
Assume that $X^{(1)}-x_*$ ranges in $[-0.9x_*,0.9x_*]$.
Denote $X_n = \frac{1}{n}\sum_{i=1}^n X^{(i)}$.
Then
\begin{align*}
\E\Big[ \frac{1}{X_n} - \frac{1}{x_*} \Big] &= \frac{var(X^{(1)})}{n x_*} + O(n^{-2}) \\
\E\Big[ \Big( \frac{1}{X_n} - \frac{1}{x_*} \Big)^2 \Big] &= \frac{var(X^{(1)})}{nx_*^2} + O(n^{-2})
\end{align*}
\end{lemma}

\begin{proof}
Replacing each $X^{(i)}$ by $X^{(i)}/x_*$, we can assume for simplicity that $x_*=1$.
Since $|1-X_n|\leq 0.9$, the Dominated Convergence Theorem implies that
\begin{align*}
\E\Big[ \frac{1}{X_n} - 1 \Big] &= \E \Big[ \sum_{k=1}^{\infty} (1-X_n)^{k} \Big] = \sum_{k=1}^{\infty} \E \big[ (1-X_n)^{k} \big]
\end{align*}
Note that for each multi-index $\mathbf{i} \in \{1,\dots n\}^k$,
\begin{equation*}
\E \Big[ \prod_{j=1}^k (1-X^{\mathbf{i}_j}) \Big] \begin{cases}
= 0 ~\text{if some entry of}~ \mathbf{i} ~\text{appears only once} \\
\in [-0.9^k, 0.9^k] ~\text{else}
\end{cases}
\end{equation*}
It follows that
\begin{align*}
\E \big[ (1-X_n)^1 \big] &= 0, \quad \E \big[ (1-X_n)^2 \big] = \frac{var(X^{(1)})}{n} \\
\big| \E \big[ (1-X_n)^3 \big] \big| &= \Big| \frac{\E\big[ (1-X^{(1)})^3 \big]}{n^2} \Big| \leq \frac{0.9^3}{n^2} \\
\big| \E \big[ (1-X_n)^4 \big] \big| &= \Big| 3 \frac{var(X^{(1)})^2}{n^2} + \frac{\E\big[ (1-X^{(1)})^4 \big]}{n^3} \Big| \leq \frac{3 \cdot 0.9^2}{n^2} + \frac{0.9^4}{n^3}
\end{align*}
Let $A(n,k)$ denote the number of $\mathbf{i} \in \{1,\dots n\}^k$ with no unique entry.
Then, for $nk>1$,
\begin{equation*}
A(n,k) \leq \sum_{j=1}^{\lfloor k/2 \rfloor} \binom{n}{j} \frac{k!}{(k-2j)!} (k-2j)^j \leq 2 (nk^3)^{\lfloor k/2 \rfloor}
\end{equation*}
The first inequality follows from the following counting process:
Let $j$ be the number of distinct entries of $\mathbf{i}$.
There are $\binom{n}{j}$ choices of $j$ sorted numbers $(a_1,\dots a_j)$ and $k!/(k-2j)!$ choices of $2j$ indices of $\mathbf{i}$ to receive the duplicated numbers $(a_1, a_1, a_2, a_2, \dots)$, and the rest $k-2j$ indices can choose freely from $(a_1,\dots a_j)$.
Since there could be over-counting, this is an upper bound.

Set $K_n = \lfloor \log(10n^2) / \log(1/0.9) \rfloor$.
It follows that for each $k \in \{2, \dots K_n \}$,
\begin{equation*}
\Big| \E \big[ (1-X_n)^{k} \big] \Big| \leq \frac{1}{n^k} \sum_{\mathbf{i}\in\{1,\dots n\}^k} \E \Big[ \prod_{j=1}^k \big| 1-X^{\mathbf{i}_j} \big| \Big] \leq 0.9^k \frac{A(n,k)}{n^k} = O(n^{-\lceil k/2 \rceil + 0.1})
\end{equation*}
Meanwhile,
\begin{equation*}
\Big| \sum_{k=K_n+1}^{\infty} \E\big[(1-X_n)^k\big] \Big| \leq \sum_{k=K_n+1}^{\infty} 0.9^k \leq n^{-2}
\end{equation*}
Hence,
\begin{align*}
\E\Big[ \frac{1}{X_n} - 1 \Big] &= \sum_{k=1}^{4} \E\big[(1-X_n)^k\big] + \sum_{k=5}^{K_n} \E\big[(1-X_n)^k\big] + \sum_{k=K_n+1}^{\infty} \E\big[(1-X_n)^k\big] \\
&= \frac{var(X^{(1)})}{n} + O(n^{-2}) + \sum_{k=5}^{K_n} O(n^{-\lceil k/2 \rceil + 0.1}) + O(n^{-2}) \\
&= \frac{var(X^{(1)})}{n} + O(n^{-2})
\end{align*}

Similarly, by the Dominated Convergence Theorem
\begin{align*}
\E\Big[ \Big( \frac{1}{X_n} - 1 \Big)^2 \Big] &= \E \Big[ \Big( \sum_{k=1}^{\infty} (1-X_n)^{k} \Big)^2 \Big] = \E \Big[ \sum_{k=2}^{\infty} (k - 1) (1-X_n)^{k} \Big] = \sum_{k=2}^{\infty} (k - 1) \E \big[ (1-X_n)^{k} \big]
\end{align*}
Set $K_n = \lfloor \log(10n) + a\log\log n + b \rfloor$ for some large constants $a,b$.
As usual,
\begin{align*}
\Big| (k-1) \E \big[ (1-X_n)^{k} \big] \Big| &\leq k ~ 0.9^k \frac{A(n,k)}{n^k} = O(n^{-\lceil k/2 \rceil + 0.1}) \quad \text{for}~ 2 \leq k \leq K_n \\
\Big| \sum_{k=K_n+1}^{\infty} (k-1) \E\big[(1-X_n)^k\big] \Big| &< \frac{(K_n-1)c^{K_n}}{1-c} \leq n^{-2}
\end{align*}
Hence,
\begin{align*}
\E\Big[ \Big( \frac{1}{X_n} - 1 \Big)^2 \Big] &= \sum_{k=2}^{4} (k-1) \E\big[(1-X_n)^k\big] + \sum_{k=5}^{K_n} (k-1) \E\big[(1-X_n)^k\big] + \sum_{k=K_n+1}^{\infty} (k-1) \E\big[(1-X_n)^k\big] \\
&= \frac{var(X^{(1)})}{n} + O(n^{-2}) + \sum_{k=5}^{K_n} O(n^{-\lceil k/2 \rceil + 0.1}) + O(n^{-2}) \\
&= \frac{var(X^{(1)})}{n} + O(n^{-2})
\end{align*}

\end{proof}

\section{The Impossibility of Causal Samplers}

\begin{proof}[Proof of Proposition \ref{prop. causal sampler condition}]
By Lemmas \ref{lemma. segments prefix-free} and \ref{lemma. prefix-free projection}, the partial function $\Pi_{\fip(x)}$ is well-defined.
By monotonicity (\ref{eq. monotone}), its domain includes $\fip(x')$ for all $x'$ such that $x \sqsubseteq x'$.
So the right side of (\ref{eq. causal sampler identity}) is well-defined.

Given $Z\sim Q(\cdot|x)$, note that $Z^{(\leq t)}$ does not depend on $x_{>t}$ if and only if $\law(Z^{(\leq t)}) = Q(\cdot|x_{\leq t})$ for any choice of $x_{>t}$.
This condition is equivalent to (\ref{eq. causal sampler identity}).
So condition 1 is equivalent to condition 2.

Condition 3 implies condition 1 by marginalizing the right side of (\ref{eq. causal factorization}) over $x_{>t}$.
Condition 2 implies condition 3 by defining
\begin{equation*}
\forall a\in\Sigma, ~\forall s \in \nsm_x(z,a), \quad \tilde{Q}(s|xa,z) = \frac{Q(zs|xa)}{Q(z|x)}
\end{equation*}
which is a probability distribution since $Q(z|x) = \sum_{s\in\nsm_x(z,a)} Q(zs|xa)$ by (\ref{eq. causal sampler identity}).
\end{proof}

\begin{proof}[Proof of Theorem \ref{thm. posterior not causal}]
For any $P \in \PS(\Sigma^{\omega})$, condition 2 of Proposition \ref{prop. causal sampler condition} is equivalent to the following property
\begin{align*}
\forall x,y &\in \Sigma^*, ~\forall z \in \fip(x), \quad f_{x,y,z}(P) = 0
\end{align*}
where each function $f_{x,y,z}\colon \PS(\Sigma^{\omega}) \to \R$ is defined by
\begin{align*}
f_{x,y,z}(P) &:= (\proj\#P)^{\leq |x|+|y|}(xy) \cdot P^{\leq |z|}(z) - (\proj\#P)^{\leq |x|}(x) \sum_{s\in\nsm_x(z,y)} P^{\leq |z|+|s|}(zs) \\
&= \Big(\sum_{z' \in \fip(xy)}P([z'])\Big) P([z]) - \Big(\sum_{z' \in \fip(x)}P([z'])\Big) \Big(\sum_{s \in \nsm_x(z,y)}P([zs])\Big)
\end{align*}
Unlike in Proposition \ref{prop. causal sampler condition}, we do not restrict $x$ and $xy$ to be in $\dom(Q_*)$,
because by definition of $\dom(Q_*)$ (\ref{eq. domain of posterior}),
if $xy \notin \dom(Q_*)$, then $P^{\leq |z'|}(z')=0$ for any $z'\in\fip(xy)$,
and thus
\begin{equation*}
f_{x,y,z}(P) = 0 \cdot P^{\leq |z|}(z) - (\proj\#P)^{\leq |x|}(x) \sum_{s\in\nsm_x(z,y)} 0 = 0
\end{equation*}
Similarly, if $x \notin \dom(Q_*)$, then $xy \notin \dom(Q_*)$, so the same result applies.
Hence,
\begin{equation*}
\PS_{\proj}^{\text{\faPlay}} = \big\{P\in\PS(\Sigma^{\omega}) \bigm| \forall x,y \in \Sigma^*, ~\forall z \in \fip(x), ~f_{x,y,z}(P)=0 \big\}
\end{equation*}

As an abuse of notation, for any $t \in \N \cup \{\omega\}$ and any $x\in\Sigma^{\leq t}$,
denote by $[x]$ the cylinder $\{x' \in \Sigma^t \mid x \sqsubseteq x'\}$
(Previously this notation only applies to $t=\omega$).
It follows that the function $f_{x,y,z}$ can be extended to $\PS(\Sigma^t)$ for any $t \geq \max_{z' \in \fip(xy)}|z'|$.
Therefore, for any $t \in \N$,
\begin{align}
\nonumber
\PS_{\leq t}^{\text{\faPlay}} &= \big\{ P^{\leq t} \bigm| P\in\PS(\Sigma^{\omega}), ~\forall x,y \in \Sigma^*, ~\forall z \in \fip(x), ~f_{x,y,z}(P)=0 \big\} \\
\nonumber
&\subseteq \big\{ P^{\leq t} \bigm| P\in\PS(\Sigma^{\omega}), ~\forall x,y \in \Sigma^*, ~\forall z \in \fip(x), ~ t \geq \max_{z'\in\fip(xy)} |z'| \to f_{x,y,z}(P)=0 \big\} \\
\label{eq. simplex polynomial zeros}
&= \big\{ P \bigm| P\in\PS(\Sigma^{\leq t}), ~\forall x,y \in \Sigma^*, ~\forall z \in \fip(x), ~ t \geq \max_{z'\in\fip(xy)} |z'| \to f_{x,y,z}(P)=0 \big\}
\end{align}

For any $t \in \N$, since $\Sigma^t$ is a finite set,
the space $\PS(\Sigma^{\leq t})$ is just a $(|\Sigma|^t-1)$-dimensional probability simplex.
For any $P \in \PS(\Sigma^{\leq t})$ and any $x\in\Sigma^{\leq t}$, the probability $P([x])$ is a finite sum of some entries of the vector $P$.
So each $f_{x,y,z}$ in (\ref{eq. simplex polynomial zeros}) is a polynomial of degree $\leq 2$ over $\PS(\Sigma^{\leq t})$.
By identifying the affine plane
\begin{equation*}
A_t = \big\{ P \in \R^{|\Sigma|^t} \bigm| \sum_i P_i = 1 \big\}
\end{equation*}
with $\R^{|\Sigma|^t-1}$, each $f_{x,y,z}$ can be seen as a polynomial on $\R^{|\Sigma|^t-1}$.

Now, we aim to show that some function $f_{x,y,z}$ in (\ref{eq. simplex polynomial zeros}) is not constant zero.
Consider any $t \geq t_0$ where
\begin{equation*}
t_0 := \max \big\{ |z| \bigm| a\in\Sigma, ~ z\in\fip(x_0a) \big\}
\end{equation*}
which is finite since Lemma \ref{lemma. segments prefix-free} indicates that $\fip$ always outputs finite sets.
Given the condition that $|\fip(x_0)|>1$,
we must have $|\Sigma|>1$; otherwise, $\fip(x) \equiv \{\varnothing\}$, a contradiction.
Define a relation $R$ between $\fip(x_0)$ and $\Sigma$, such that $(z,a) \in R$ if and only if $\nsm_{x_0}(z,a) \neq \varnothing$.
Since $\proj$ is surjective (Assumption \ref{assume. sec sampling}), for any $a\in\Sigma$, $\fip(x_0a) \neq \varnothing$, so $\nsm_{x_0}(z,a)\neq\varnothing$ for some $z\in\fip(x_0)$.
Similarly, for each $z\in\fip(x_0)$, Proposition \ref{prop. inverse projection} implies that $[z]\cap\dom(\proj) \neq \varnothing$, and thus Lemma \ref{lemma. segments prefix-free} property 4 implies that $\nsm_{x_0}(z,a) \neq \varnothing$ for some $a\in\Sigma$.
Hence, the relation $R$ satisfies the condition of Lemma \ref{lemma. distinct pairs}, and thus there exists distinct $z_0,z_1 \in \fip(x_0)$ and distinct $a_0,a_1 \in \Sigma$ such that both $\nsm_{x_0}(z_0,a_0),\nsm_{x_0}(z_1,a_1)$ are non-empty.
Choose $s_0 \in \nsm_{x_0}(z_0, a_0)$ and $s_1 \in \nsm_x(z_1, a_1)$.
Construct $P \in \PS(\Sigma^t)$ such that $P([z_0 s_0]),P([z_1 s_1])>0$ and $P([z_0 s_0])+P([z_1 s_1])=1$.
It follows that
\begin{equation*}
f_{x_0,z_0,a_0}(P) = P([z_0 s_0]) \cdot P([z_0]) - 1 \cdot P([z_0s_0]) = - P([z_0 s_0]) P([z_1 s_1]) < 0
\end{equation*}
Hence, there is at least one non-zero polynomial in (\ref{eq. simplex polynomial zeros}).

It is known that for any dimension $d \geq 1$ and any non-zero polynomial over $\R^d$, its zeros form a negligible subset with respect to the Lebesgue measure \cite{murdza2024hausdorff,fischer2016polynomial}.
Since
\begin{equation*}
\PS_{\leq t}^{\text{\faPlay}} \subseteq \PS(\Sigma^t) \cap \{P\in A_t \mid f_{x_0,z_0,a_0}(P) = 0 \}
\end{equation*}
we conclude that $\PS_{\leq t}^{\text{\faPlay}}$, as the subset of a negligible set, also has Lebesgue measure zero.
\end{proof}

\begin{lemma}
\label{lemma. distinct pairs}
Let $A,B$ be two sets such that $|A|,|B|>1$, and let $R$ be a relation between $A,B$ (namely a subset of $A\times B$).
If each $a\in A$ is contained in at least one pair $r\in R$, and each $b \in B$ is contained in at least one $r\in R$,
then there exist two pairs, $(a,b), (a',b') \in R$, such that $a\neq a'$ and $b\neq b'$.
\end{lemma}

\begin{proof}
Let $a,a'$ be two distinct elements of $A$, and let $(a,b),(a',b')$ be two pairs in $R$.
If $b \neq b'$, then we are done.
Else, since $|B|>1$, there exist some $b'' \in B$ such that $b'' \neq b$.
Choose a pair $(a'',b'') \in R$.
If $a'' \neq a$, then we can go with $(a,b),(a'',b'')$.
Else, we can go with $(a,b''),(a',b)$.
\end{proof}

\begin{proof}[Proof of Proposition \ref{prop. approximable by causal}]
We prove the contrapositive.
Let $\{Q_n\}_{n=1}^{\infty} \subseteq \Q_{\proj}^{\text{\faPlay}}$ be a sequence such that $\dom(Q) \subseteq \dom(Q_n)$ for each $n$ and
\begin{equation*}
\lim_{n\to\infty} \sup_{x\in\dom(Q)} \chi^2\big( Q(\cdot|x) \big\| Q_n(\cdot|x) \big) = 0
\end{equation*}
Then, for any $x\in\dom(Q)$, since $\sprt Q(\cdot|x) \subseteq \fip(x)$ is a finite set by Lemma \ref{lemma. segments prefix-free},
the conditional distribution $Q_n(\cdot|x)$ converges pointwise to $Q(\cdot|x)$.
For any $x,x'\in\dom(Q)$ such that $x\sqsubseteq x'$, taking limit at both sides of the identity
\begin{equation*}
Q_n(\cdot|x) = \Pi_{\fip(x)}\# Q_n(\cdot|x')
\end{equation*}
we obtain (\ref{eq. causal sampler identity}).
Hence, Proposition \ref{prop. causal sampler condition} implies that $Q \in \Q_{\proj}^{\text{\faPlay}}$.
\end{proof}

\section{From Global Objective to Instantaneous Loss}
\label{appendix. from global to local}

In Section \ref{sec. balance fast slow}, we approximate the optimization of the total cost (\ref{eq. loss x lenth total cost}) with gradient descent on the running cost (\ref{eq. loss x lenth running cost}).
This appendix provides an informal discussion of when this approximation may hold.

To simplify the analysis, we use continuous time and express problem (\ref{eq. loss x lenth total cost}) abstractly by the value function
\begin{align}
\label{eq. optimal control}
V(\x_0,t_0) &= \min_{p \in \mathfrak{P}(\x_0,t_0)} \int_{t_0}^{\infty} f(p(t)) dt \\
\nonumber
\mathfrak{P}(\x_0,t_0) &= \big\{ p\colon  [t_0,\infty)\to\R^d \bigm| p(t_0)=\x_0, ~ \|p\|_{\text{Lip}} \leq 1 \big\}
\end{align}
Specifically, the train steps $s$ are converted to continuous time $t$ (as an abuse of notation, this is not the elapsed time in Section \ref{sec. balance fast slow}), the parameter $\theta$ becomes a point $\x \in \R^d$,
the update trajectory $(\theta_s)_{s=0}^{\infty}$ becomes a path $p\colon [t_0,\infty)\to\R^d$,
the admissible set of trajectories $\Theta$ becomes the set of paths $\mathfrak{P}$ (with prescribed initialization),
and the running cost $C(\theta_s) \mathbb{T}_s(\theta_s)$ is generalized to a nonnegative smooth function $f$.
The Lipschitz bound $\|p\|_{\text{Lip}} \leq 1$ represents the constraint on $\Theta$;
since typically one avoids having excessively large parameter updates when training LLMs (e.g.\@ with gradient clipping and normalization), it seems acceptable to model the update trajectories as 1-Lipschitz paths.

To ensure that the total cost (\ref{eq. loss x lenth total cost}) is finite, the running cost must converge to zero.
So we assume for convenience that the set of minimizers $S = \{\x \mid f(\x) = 0\}$ is nonempty.
It follows that the value function $V$ is finite and nonnegative, and $V=0$ if and only if on $S$.
Also, the definition of $V$ implies that it does not depend on $t$, so $\partial_t V \equiv 0$ and we write $\nabla V = \nabla_{\x}V$.
By Hamilton-Jacobi-Bellman equation,
\begin{equation*}
0 = - \partial_t V = \min_{\|\vb\| \leq 1} \nabla_{\x} V \cdot \vb + f = - \|\nabla V\| + f
\end{equation*}
So the value function satisfies the eikonal equation $|\nabla V | = f$.

The minimizing path of (\ref{eq. optimal control}) is driven by $\dot{p}_t = - \frac{\nabla V(p_t)}{\| \nabla V(p_t) \|}$.
Meanwhile, gradient descent on the running cost can be modeled by the path $\dot{p}_t = - \nabla f(p_t)$.
We say that two paths are equal if their images overlap, so the problem becomes under what condition there exists a non-negative function $\lambda$ such that $\nabla V = \lambda \nabla f$.

Suppose the gradient norm of $f$ depends only on the value of $f$, namely there exist some differentiable function $\psi$ such that $\|\nabla f\| = \psi(f)$.
Then, the velocity field $\frac{f}{\|\nabla f\|} \nabla f$ is a gradient field: In terms of differential forms
\begin{equation*}
d\Big( \frac{f}{\|\nabla f\|} d f \Big) = 0 + d \psi(f)^{-1} \wedge fdf = 0
\end{equation*}
So there exists some function $\tilde{V}$ such that $\nabla \tilde{V} = \frac{f}{\|\nabla f\|} \nabla f$, and in particular, $|\nabla \tilde{V}| = f$.
By uniqueness of the solution to $|\nabla V| = f$, the functions $V,\tilde{V}$ are equal up to additive constant.
Thus, $\nabla V = \lambda \nabla f$ with $\lambda = f/\|f\|$.


Hence, it is possible for the gradient descent path to coincide with the optimal path for the total cost.
Despite that the condition $\|\nabla f\| = \psi(f)$ probably does not hold in practice, this argument indicates that our approximation is at least not entirely unfounded.

\section{Implementation of the Preliminary Experiment}
\label{appendix. experiment}

This appendix lists the implementation details of Section \ref{sec. experiment}.

For all components of all three models, the maximum learning rate is $10^{-5}$, weight decay is $0.1$ to align with \cite{qwen2025qwen25}, gradient clipping is $1.0$, batch size is 1920, and latent sampling size (not applicable to SFT) is $n=8$.
All models are trained for $1000$ steps, and each evaluation takes $10$ steps.
The training samples and test samples are disjoint.
During training, the SFT model is evaluated every 5 steps, and the two slow thinking models are evaluated every 20 steps, so the curves in Figure \ref{fig: prelim experiment} have different resolutions.
The learning rate schedule is the commonly used cosine schedule, such that learning rate increases linearly from $0$ to the maximum value in the first $5\%$ steps, and then decay to $10\%$ of the maximum value.
The optimizer is AdamW \cite{loshchilov2019AdamW} with the default hyparameters $\beta=(0.9, 0.999)$.
To save GPU memory, we use the gradient parallelism and optimizer parallelism of DeepSpeed ZeRO-2 \cite{rajbhandari2020zero}.

\begin{figure}[!ht]
\centering
\includegraphics[width=0.9\linewidth]{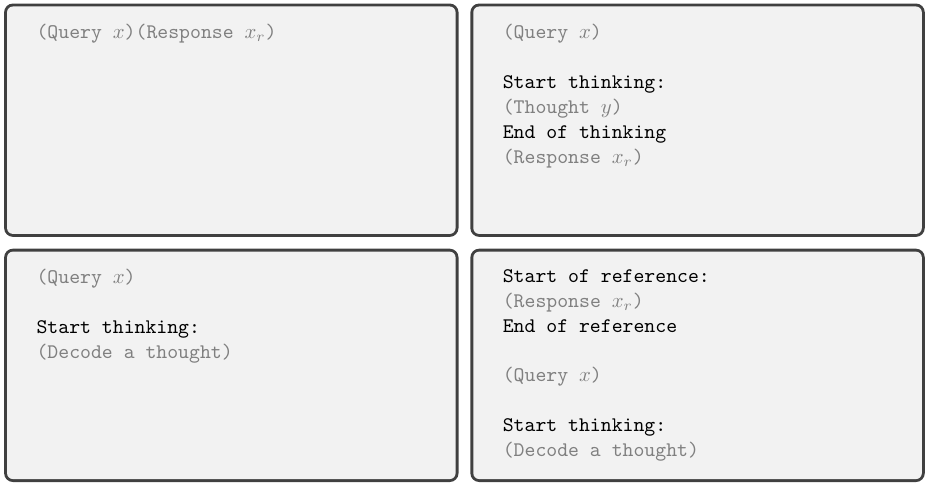}
\caption{The input formats of the SFT model, the latent models, the predictive sampler, and the explanatory sampler, respectively.}
\label{fig: enc dec format}
\end{figure}

Figure \ref{fig: enc dec format} shows the input formats of all trained models.
Denote these four formats by $xx_r$, $x\sot y \eot x_r$, $x\sot y$ and $\sor x_r \eor x\sot y$, with the delimiters denoted by $\sot,\eot,\sor,\eor$.

One issue is that in general, adapting an LLM to a new format could cause a large increase in the losses at the beginning of training, which could last for a few thousand steps.
Thus, a smooth initialization is needed, and we adapt the gating trick of \cite{zelikman2024quietstar} to our setting.
Empirically, this trick largely alleviates the initial loss increase of training slow thinking models, as their loss curves are close to the SFT curve in Figure \ref{fig: prelim experiment}.
To prepare for the construction, let $f$ be the Transformer of the latent model or sampler, and let $d$ be its feature dimension.
Given any input sequence $z$, denote the $|z|\times d$ feature matrix prior to the unembedding layer of $f$ by $f_{\text{pre}}(z)$, and the $|z|\times |\Omega|$ logit matrix after the unembedding layer by $f_{\text{post}}(z)$.
Recall that the standard routine of SFT is to compute the following $|x_r|\times |\Sigma|$ matrix to compute the loss
\begin{equation*}
f_{\text{post}}(xx_r)_{[-|x_r|-1,-1]}
\end{equation*}
where $[-|x_r|-1,-1]$ means taking the rows $|x|-1,\dots |x|+|x_r|-1$ of the matrix.
The $-1$ means that we are using the feature vector at the previous token to predict each token.
The gating trick is implemented as follows:
If $f$ comes from the latent model $P_{\theta}$, we modify its output to
\begin{equation}
\label{eq. gating output}
p_{\text{fast}} \odot f_{\text{post}}(xx_r)_{[-|x_r|-1,-1]} + p_{\text{slow}} \odot f_{\text{post}} \big(x\sot y \eot x_r \big)_{[-|x_r|-1,-1]}
\end{equation}
and the gatings $p_{\text{fast}},p_{\text{slow}}$ are $|x_r|\times 1$ matrices computed by
\begin{equation}
\label{eq. gating weights}
\begin{bmatrix}
p_{\text{fast}}\\
p_{\text{slow}}
\end{bmatrix}
= \text{softmax}\big(\text{MLP}\big(f_{\text{pre}}(xx_r)_{[-|x_r|-1,-1]} \oplus f_{\text{pre}}(x\sot y \eot x_r)_{[-|x_r|-1,-1]}\big)\big)
\end{equation}
where $\oplus$ means concatenating the matrices in the feature dimension.
The MLP is a 3-layer multilayer perceptron with SiLU activation, and with input dimension $2d$, hidden dimension $2d$, and output dimension $2$.
The softmax is applied to the 2 output entries for each row, so that $p_{\text{fast}}+p_{\text{slow}} \equiv 1$.
The MLP accounts for only $1.33\%$ of the total parameters in $f$, so its influence on the capacity of the slow thinking models should be negligible and thus is acceptable here.
It is trained from scratch, and its output layer is a PyTorch linear layer with weight initialized to all 0 and bias to $[2,-2]$.
Thus, at the beginning of training, $p_{\text{slow}} =\text{sigmoid}(-4) < 0.02$,
and the model mostly uses fast thinking.
Similarly, if $f$ comes from the explanatory sampler, then we modify its output to
\begin{align*}
&p_{\text{fast}} \odot f_{\text{post}}(x\sot y)_{[-|y|-1,-1]} + p_{\text{slow}} \odot f_{\text{post}} \big(\sor x_r \eor x\sot y\big)_{[-|y|-1,-1]} \\
&\begin{bmatrix}
p_{\text{fast}}\\
p_{\text{slow}}
\end{bmatrix}
= \text{softmax}\big(\text{MLP}\big(f_{\text{pre}}(x\sot y)_{[-|y|-1,-1]} \oplus f_{\text{pre}}(\sor x_r \eor x\sot y)_{[-|y|-1,-1]}\big)\big)
\end{align*}
The latent model and the explanatory sampler use separate MLPs.
The initialization is the same as above, so the explanatory sampler initially behaves like a predictive sampler.
This gating is also applied to the decoding stage (\ref{eq. experiment sampling}) of the explanatory sampler (after all, at the implementation level, the decoding of Transformer is repeated encoding).

One may worry about the doubled computation cost due to the gating trick.
Nevertheless, this trick is only used for finetuning.
If a slow thinking model with explanatory sampler(s) is trained from scratch, then the initial loss increase due to new formats would not happen.
Even for finetuned models, one may reduce the inference cost by distilling a finetuned model with gating to a base model without gating.

\begin{figure}[!ht]
\centering
\subfloat{\includegraphics[width=0.48\textwidth]{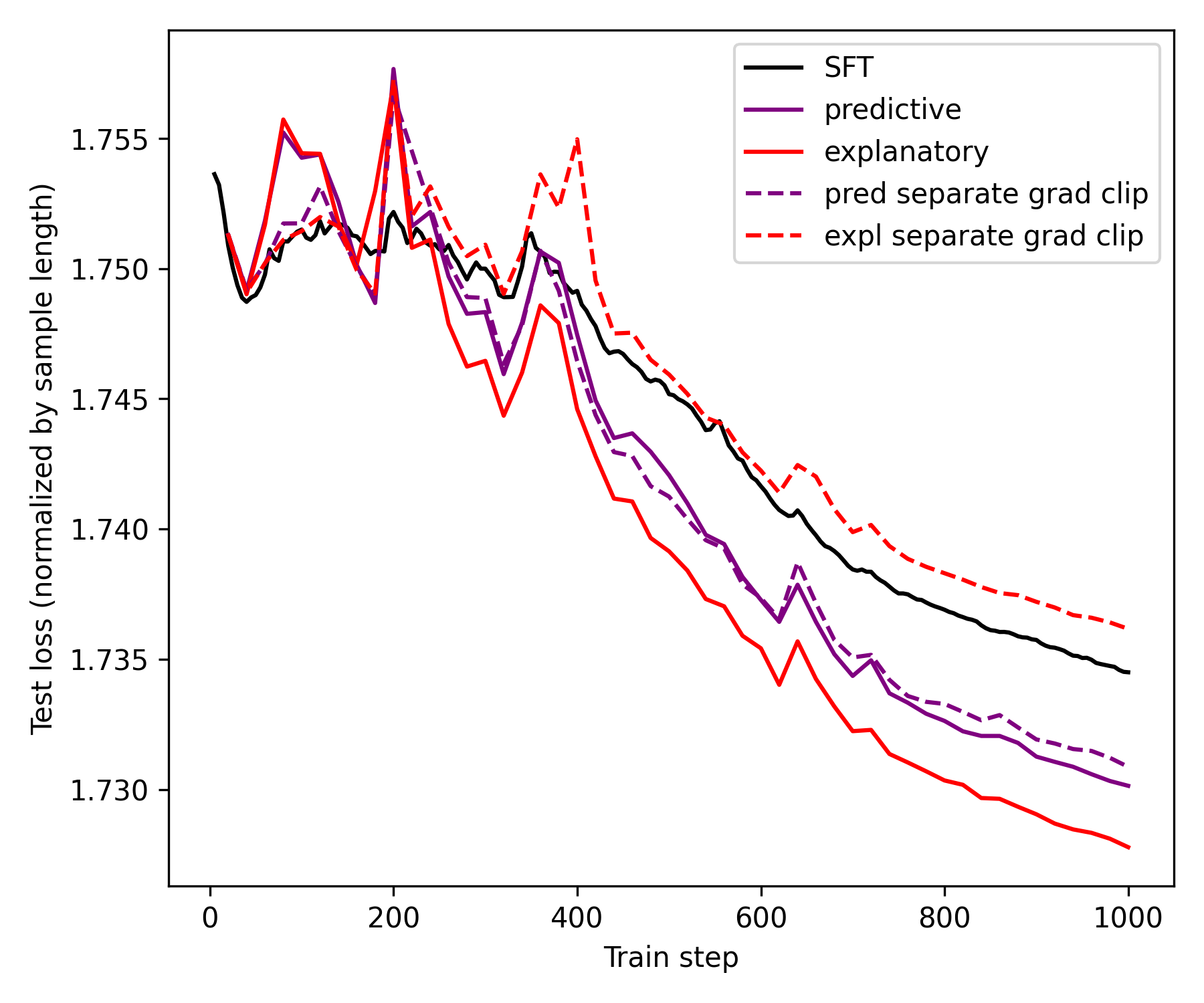}}
\quad
\subfloat{\includegraphics[width=0.48\textwidth]{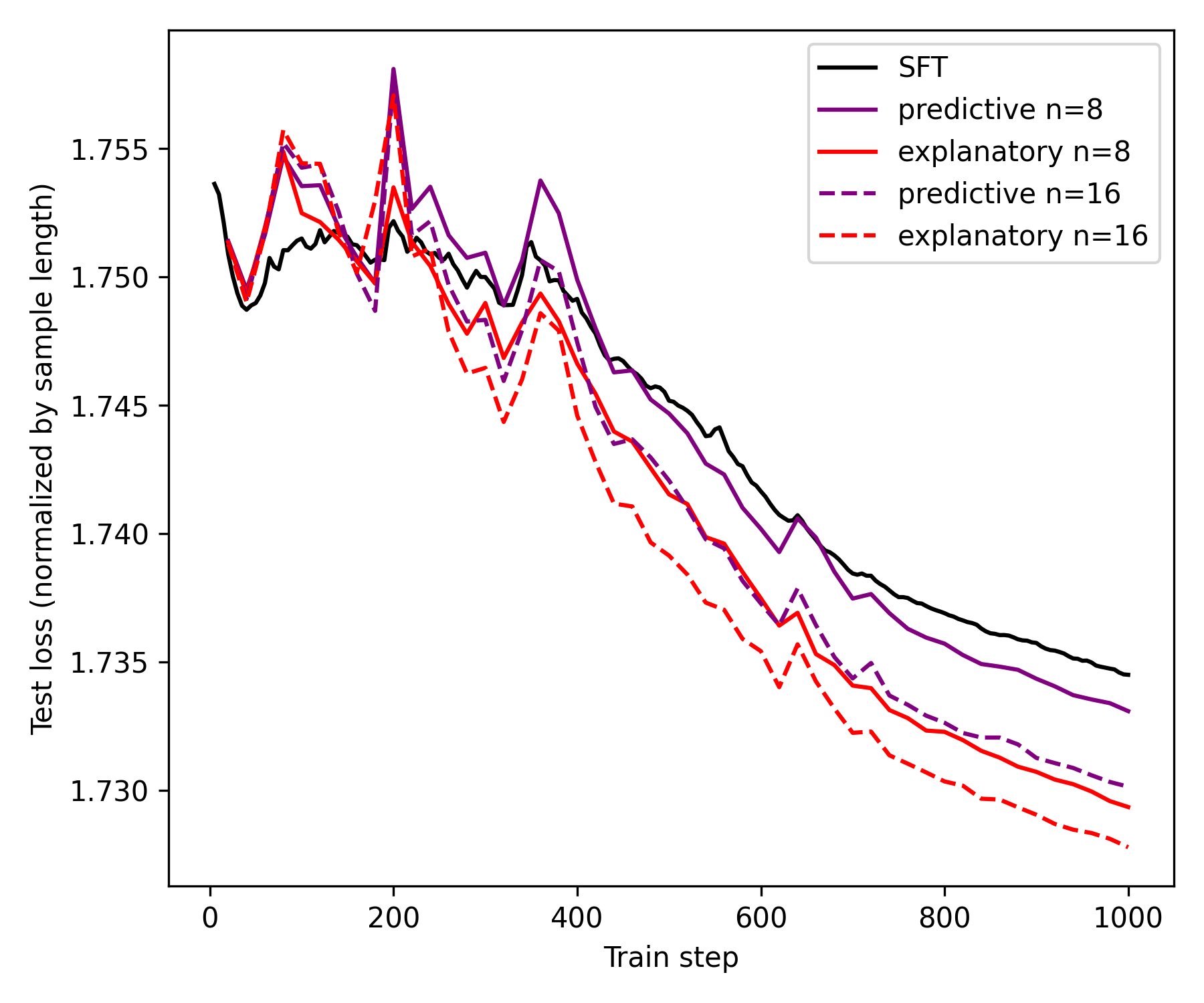}}
\caption{Left: Comparison of combined clipping and separate clipping.
The sampling size for this experiment is $n=16$.
Right: Comparison of training with sampling size $n=8$ and $n=16$.}
\label{fig: grad clip}
\end{figure}

Another technical detail is that gradient clipping treats the entire slow thinking model (the latent model and the sampler) as one model.
Specifically, while the gradient of the SFT model is clipped to $\vb_{\text{SFT}} / \max(1,\|\vb_{\text{SFT}}\|)$,
the gradient of each slow thinking is clipped to
\begin{equation*}
\frac{\vb_{\text{latent}} \oplus \vb_{\text{sampler}}}{\max\big( 1, \| \vb_{\text{latent}} \oplus \vb_{\text{sampler}} \|\big)}
\end{equation*}
instead of
\begin{equation*}
\frac{\vb_{\text{latent}}}{\max\big( 1, \| \vb_{\text{latent}} \|\big)} \oplus \frac{\vb_{\text{sampler}}}{\max\big( 1, \| \vb_{\text{sampler}} \|\big)}
\end{equation*}
In practice, one can simply wrap the parameters of the latent model and sampler into one optimizer, instead of two.
The consideration is that separate clipping makes the training worse, in particular for the explanatory sampler, as depicted in Figure \ref{fig: grad clip} (Left).
The mechanism of this failure has not been determined yet;
perhaps the model components need a smaller learning rate.

Next, an experiment is performed to study the influence of the sampling size $n$.
As plotted in Figure \ref{fig: grad clip} (Right), increasing $n$ from $8$ to $16$ improves the performance of both slow thinking models, but the loss gains tend to become saturated and the relative improvement of the explanatory sampler becomes smaller:
\begin{equation*}
\frac{L_{\text{fast}} - L_{\text{expl}}}{L_{\text{fast}} - L_{\text{pred}}} - 1 = \frac{1.7345-1.7278}{1.7345-1.7301} - 1 \approx 52\%
\end{equation*}
Such saturation is expected, since the samplers improve the training progress only in an indirect manner:
More expressive samplers such as the explanatory samplers can further reduce the divergences $\chi^2(Q_*(\cdot|x)\|Q(\cdot|x))$ and thus the estimation error of the training gradient (\ref{eq. grad mse one sample}) (also cf.\@ Figure \ref{fig: unified objective}), and then the latent model $P_{\theta}$ with more accurate gradients may converge faster.
No matter what sampler we use (as long as $\sprt Q_*(\cdot|x) \subseteq \sprt Q(\cdot|x)$), the gradient estimation error always converges to its minimum as $n\to\infty$, so the advantage of more expressive samplers diminishes.

\begin{figure}[!ht]
\centering
\includegraphics[width=0.8\textwidth]{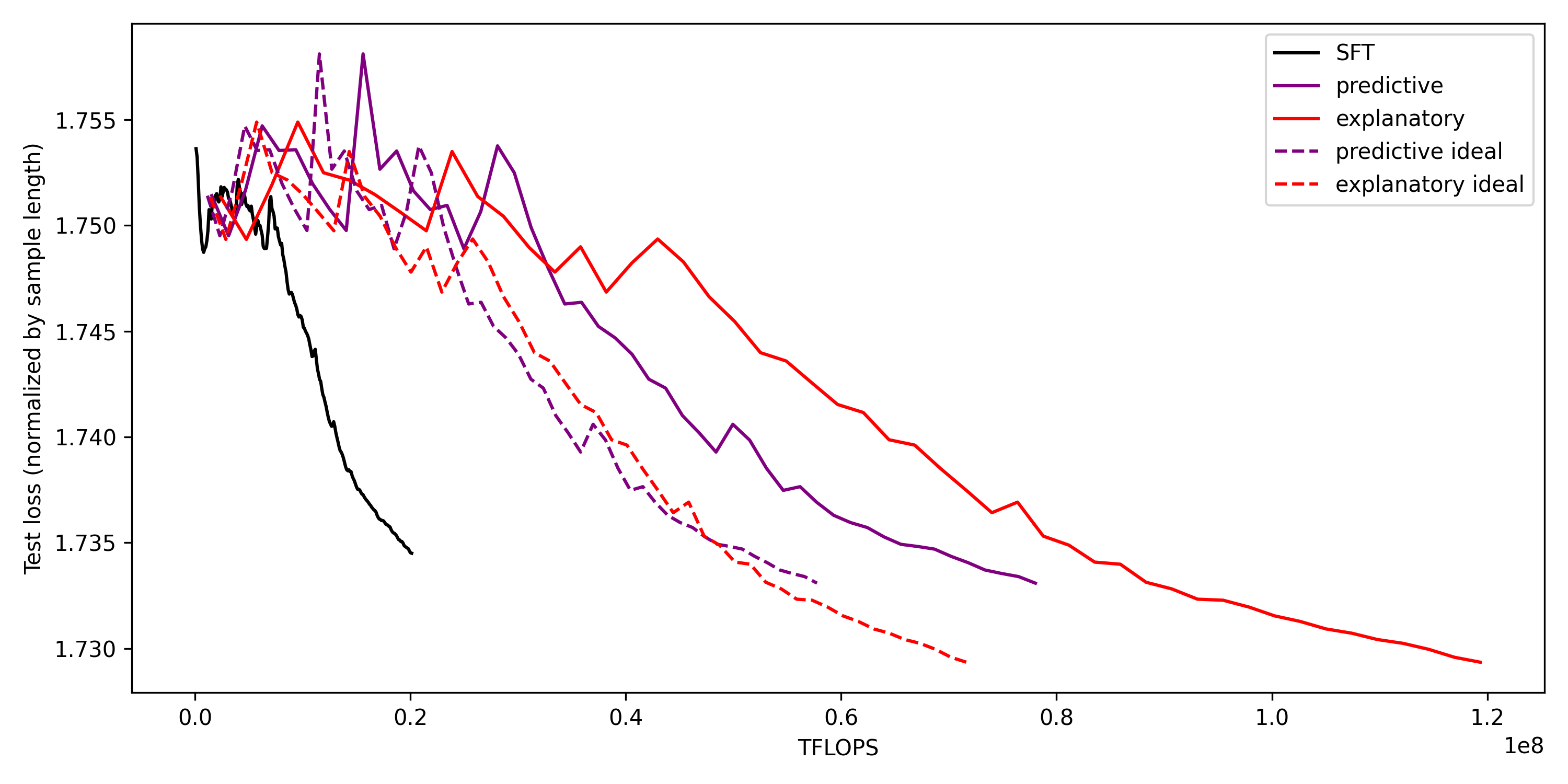}
\caption{Test loss vs.\@ TFLOPs.
The solid curves are the same as those in Figure \ref{fig: prelim experiment} (Left), except that their horizontal axes change from ``training step" to ``total TFLOPs consumed at the current step".
Similarly, the axes of the dotted curves are ``the ideal TFLOPs without gating".}
\label{fig: TFLOPs}
\end{figure}

Finally, we consider the compute cost of these experiments, measured by TFLOPs ($10^{12}$ floating-point operations).
It is not the most realistic measure, since the time cost consists of not only the compute time, but also the data transfer time.
The latter becomes dominant during LLM decoding (e.g.\@ the generation of chain-of-thoughts for slow thinking models) and is known as the memory wall \cite{gholami2024wall,wu2025combating}.
Nevertheless, since decoding (or more generally, sequential computation) seems to be an indispensable feature from the perspective of the representation hierarchy, it is reasonable to expect that technological advances will be made to alleviate the data transfer time, e.g.\@ through in-memory and near-memory computation \cite{wolters2024CIM,zeng2024flightllm}.
Thus, we adopt TFLOPs, and Figure \ref{fig: TFLOPs} plots the test loss curves of Figure \ref{fig: prelim experiment} (Left) with respect to the TFLOPs during training.

The TFLOPs are calculated as follows.
The number of floating-point operations of one forward pass of a Transformer LLM with gated MLP layers can be estimated by
\begin{equation*}
N_{\text{LLM}}(l_{\text{in}},l_{\text{out}}) = 2 \cdot \big[ L \cdot \big( l_{\text{in}} \cdot (2d^2 + 2dd_h H_{kv} + 3dW) + 2 \cdot \frac{l_{\text{in}}^2}{2} d \big) + l_{\text{out}} \cdot n_{\text{vocab}} d \big]
\end{equation*}
where $l_{\text{in}}$ is the input length and $l_{\text{out}}$ is the amount of tokens kept for logit computation.
The derivation of $N_{\text{LLM}}$ can be found in the appendices of \cite{yang2024memory,narayanan2021efficient}.
For Qwen2.5-7B base model \cite{qwen2025qwen25}, the constants are $L=28$ layers, $H=28$ heads, $H_{kv}=4$ key-value heads, hidden dimension $d=3584$, head dimension $d_h=128$, MLP width $W=18944$, vocabulary size $n_{\text{vocab}}=152064$.
For SFT training, the compute cost would be $3 \cdot N_{\text{LLM}}(l_q+l_r, l_r)$, where $l_q,l_r$ are the lengths of the query and response, and the coefficient $3=1+2$ means that the backward pass incurs twice the cost of the forward pass \cite{narayanan2021efficient}.
For decoding, the compute cost would be $N_{\text{LLM}}(l_q+l_r, l_r)$.
The extra cost due to the gating mechanism can be estimated by
\begin{equation*}
N_{\text{gate}}(l_{\text{out}}) = 2 \cdot l_{\text{out}} \cdot (L_{\text{MLP}}-1) (2d)^2 + 3 \cdot l_{\text{out}} \cdot  n_{\text{vocab}}
\end{equation*}
The first term is the cost of the forward pass of the MLP (\ref{eq. gating weights}), with hidden dimension $2d$ and depth $L_{\text{MLP}}=3$.
The $-1$ is due to the last linear layer, which can be omitted since its output dimension is just 2.
The second term is the cost of the weighted averaging (\ref{eq. gating output}).
The following constants
\begin{equation*}
l_{\text{chunk}}= l_{\text{thought}}=128, \quad l_{\sot}=l_{\sor}=4, \quad l_{\eot}=l_{\eor}=5
\end{equation*}
are the lengths of the text chunks (simulated queries and responses), thoughts, and aforementioned delimiters (with respect to the tokenizer of Qwen2.5-7B).
For SFT, the compute cost of each training sample is approximately
\begin{equation*}
N_{\text{SFT}} = 3 \cdot C_{\text{LM}}(2l_{\text{chunk}}, l_{\text{chunk}}) \approx 10.48 ~ \text{TFLOPs}
\end{equation*}
For the slow thinking model with predictive sampler, the per-sample cost is approximately
\begin{align*}
N_{\text{pred}} &= (1+3) \cdot N_{\text{LLM}}(l_{\text{chunk}} + l_{\sot}+l_{\text{thought}}, l_{\text{thought}}) \\
&\quad + 3 \cdot \big( N_{\text{LLM}}(2 l_{\text{chunk}} + l_{\sot} + l_{\text{thought}} + l_{\eot}, l_{\text{chunk}}) + N_{\text{LLM}}(2 l_{\text{chunk}}, l_{\text{chunk}}) + N_{\text{gate}}(l_{\text{chunk}}) \big) \\
&\approx 40.64 ~ \text{TFLOPs}
\end{align*}
where $1+3$ is for decoding and training.
For the slow thinking model with explanatory sampler, the per-sample cost is approximately
\begin{align*}
N_{\text{expl}} &= (1+3) \cdot \big( N_{\text{LLM}}(l_{\text{chunk}} + l_{\sot} + l_{\text{thought}}, l_{\text{thought}}) \\
&\quad \quad + N_{\text{LLM}}(2 l_{\text{chunk}} + l_{\sor} + l_{\eor} + l_{\sot} + l_{\text{thought}}, l_{\text{thought}}) \big) \\
&\quad + 3 \cdot \big( N_{\text{LLM}}(2 l_{\text{chunk}} + l_{\sot} + l_{\text{thought}} + l_{\eot}, l_{\text{chunk}}) + N_{\text{LLM}}(2 l_{\text{chunk}}, l_{\text{chunk}}) + N_{\text{gate}}(l_{\text{chunk}}) \big) \\
&\approx 62.16 ~ \text{TFLOPs}
\end{align*}
For reference, we also consider the ideal setting without the issue of loss increase at the beginning (e.g.\@ when training from scratch or given a better way to initialize finetuning) and thus without the need of gating.
The per-sample costs of the slow thinking models would become
\begin{align*}
N_{\text{pred-ideal}} &= (1+3) \cdot N_{\text{LLM}}(l_{\text{chunk}} + l_{\sot}+l_{\text{thought}}, l_{\text{thought}}) \\
&\quad + 3 \cdot N_{\text{LLM}}(2 l_{\text{chunk}} + l_{\sot} + l_{\text{thought}} + l_{\eot}, l_{\text{chunk}}) \\
&\approx 30.08 ~ \text{TFLOPs} \\
N_{\text{expl-ideal}} &= (1+3) \cdot N_{\text{LLM}}(2 l_{\text{chunk}} + l_{\sor} + l_{\eor} + l_{\sot} + l_{\text{thought}}, l_{\text{thought}}) \\
&\quad + 3 \cdot N_{\text{LLM}}(2 l_{\text{chunk}} + l_{\sot} + l_{\text{thought}} + l_{\eot}, l_{\text{chunk}}) \\
&\approx 37.31 ~ \text{TFLOPs}
\end{align*}
The cumulative cost at each training step $s$ would be the per-sample cost times the batch size $1920$ times $s$.
Thus, we obtain the five curves in Figure \ref{fig: TFLOPs}.

\end{document}